\documentclass[pdflatex,sn-mathphys]{sn-jnl}

\usepackage{hyperref}
\usepackage{graphicx}
\usepackage{acronym}
\usepackage{paralist}
\usepackage{url}
\usepackage{amsmath}
\usepackage{amsfonts}
\usepackage{scalerel}
\usepackage{listings}
\usepackage{caption}
\usepackage{gensymb}
\usepackage{enumitem}
\usepackage{amsthm}
\usepackage{subcaption}
\usepackage{float}
\usepackage[ruled, lined, commentsnumbered, longend, algo2e]{algorithm2e}

\SetCommentSty{mycommfont}
\usepackage [autostyle, english = american]{csquotes}

\acrodef{nsl}[FFNSL]{Feed-Forward Neural-Symbolic Learner}
\acrodef{asp}[ASP]{Answer Set Programming}
\acrodef{las}[LAS]{Learning from Answer Sets}
\acrodef{ilp}[ILP]{Inductive Logic Programming}
\acrodef{wcdpi}[WCDPI]{Weighted Context-Dependant Partial Interpretation}
\acrodef{cdpi}[CDPI]{Context-Dependent Partial Interpretation}
\acrodef{cnn}[CNN]{Convolutional Neural Network}
\acrodef{edl}[EDL]{Evidential Deep Learning}
\acrodef{ai}[AI]{Artificial Intelligence}
\acrodef{cnnlstm}[CNN-LSTM]{Convolutional Neural Network-Long Short Term Memory}
\acrodef{fcn}[FCN]{Fully Connected Network}
\acrodef{rf}[RF]{Random Forest}
\acrodef{d2k}[D2K]{Data-to-Knowledge}

\newcommand{\asp}[1]{\mbox{$\mathtt{#1}$}}
\DeclareMathOperator*{\argmax}{\arg\max}
\DeclareMathOperator{\codeif}{\mathtt{:-} }
\DeclareMathOperator{\naf}{\;\mathtt{not}\;}

\jyear{2022}%

\theoremstyle{definition}
\newtheorem{myexample}{Example}
\newtheorem{mydefinition}{Definition}

\begin{document}

\title[FFNSL: Feed-Forward Neural-Symbolic Learner]{FFNSL: Feed-Forward Neural-Symbolic Learner}


\author*[1,2]{\fnm{Daniel} \sur{Cunnington}}\email{dancunnington@uk.ibm.com}

\author[2,3]{\fnm{Mark} \sur{Law}}

\author[2]{\fnm{Alessandra} \sur{Russo}}
\author[4]{\fnm{Jorge} \sur{Lobo}}

\affil[1]{\orgname{IBM Research Europe}, \orgaddress{\city{Winchester}, \country{UK}}}

\affil[2]{\orgname{Imperial College London}, \orgaddress{\city{London}, \country{UK}}}

\affil[3]{\orgname{ILASP Limited},
\orgaddress{\city{Grantham}, \country{UK}}}

\affil[4]{\orgname{ICREA-Universitat Pompeu Fabra}, \orgaddress{\city{Barcelona}, \country{Spain}}}



\abstract{Logic-based machine learning aims to learn general, interpretable knowledge in a data-efficient manner. However, labelled data must be specified in a structured logical form. To address this limitation, we propose a neural-symbolic learning framework, called \textit{Feed-Forward Neural-Symbolic Learner (FFNSL)}, that integrates a logic-based machine learning system capable of learning from noisy examples, with neural networks, in order to learn interpretable knowledge from labelled unstructured data. We demonstrate the generality of FFNSL on four neural-symbolic classification problems, where different pre-trained neural network models and logic-based machine learning systems are integrated to learn interpretable knowledge from sequences of images. We evaluate the robustness of our framework by using images subject to distributional shifts, for which the pre-trained neural networks may predict incorrectly and with high confidence. We analyse the impact that these shifts have on the accuracy of the learned knowledge and run-time performance, comparing FFNSL to tree-based and pure neural approaches. Our experimental results show that FFNSL outperforms the baselines by learning more accurate and interpretable knowledge with fewer examples.}

\keywords{neural-symbolic learning, inductive logic programming, logic-based machine learning, distributional shift}



\maketitle

\section{Introduction}\label{sec:intro}
Logic-based machine learning~\cite{Muggleton1991, LawRB19} learns interpretable knowledge expressed in the form of a logic program, called a $hypothesis$, that explains labelled examples in the context of (optional) background knowledge. Recent logic-based machine learning systems have demonstrated the ability to learn highly complex and noise-tolerant hypotheses in a data efficient manner (e.g.,~\ac{las}~\cite{LawRB19}). However, they require labelled examples to be specified in a structured logical form, which limits their applicability to many real-world problems. On the other hand, differentiable learning systems, such as (deep) neural networks, are able to learn directly from unstructured data, but they require large amounts of training data and their learned models are difficult to interpret~\cite{8631448}. 

Within neural-symbolic artificial intelligence, many approaches aim to integrate neural and symbolic systems with the goal of preserving the benefits of both paradigms~\cite{nesy1,garcez2020neurosymbolic}. Most neural-symbolic integrations assume the existence of pre-defined knowledge expressed symbolically, or logically, and focus on training a neural network to extract symbolic features from raw unstructured data~\cite{manhaeve2018deepproblog,ijcai2020-243,serafini2016logic,cohen2016tensorlog,riegel2020logical}. In this paper, we introduce \emph{\ac{nsl}}, a neural-symbolic learning framework that assumes the opposite. Given a pre-trained neural network,~\ac{nsl} uses a logic-based machine learning system robust to noise to learn a logic-based hypothesis whose symbolic features are constructed from neural network predictions. The motivation is to enable logic-based machine learning systems to utilise \textit{pre-trained} neural networks\footnote{Examples include \url{https://modelzoo.co/} and those listed here: \url{https://github.com/collections/ai-model-zoos}} to learn symbolic features from unstructured data, and use these features to learn interpretable knowledge needed to solve a downstream classification task. \ac{nsl} preserves the benefits of both paradigms, increasing the scope of the tasks logic-based machine learning systems can be applied to. The challenge in performing such an integration, is that neural networks are vulnerable to \textit{distributional shifts}, where unstructured data belonging to a distribution different from that used for training often leads to incorrect predictions~\cite{ovadia2019can,sensoy2018evidential,Amodei2016ConcretePI}. By using a logic-based machine learning system that is robust to noise, such as a~\ac{las} system,~\ac{nsl} is capable of learning robust logic-based hypotheses from examples generated from labelled unstructured data, which may contain incorrect or noisy features as a result of incorrect neural network predictions.

The novel aspect of our~\ac{nsl} framework is the \textit{\ac{d2k} generator} that bridges the neural and symbolic learning components. The~\ac{d2k} generator automatically constructs a symbolic representation of the features predicted from the unstructured data, and weights such knowledge with a level of truthfulness that reflects the confidence score of the neural network predictions. The symbolic features can then be used by the symbolic learning component to automatically generate weighted examples from which to learn general and interpretable knowledge needed to solve the given downstream task.

\ac{nsl} is general enough to support the integration of any neural component capable of making discrete predictions from unstructured data (binary or multi-class classification), with any logic-based machine learning system capable of learning from noisy examples. In this paper, we present four instances of our framework, where the~\ac{las} systems, \textit{ILASP}~\cite{Law2018thesis} and \textit{FastLAS}~\cite{law2020fastlas}, are used as the symbolic learning component, and different neural network architectures are used as the neural component. The \ac{las} systems have been shown to learn optimal hypotheses from noisy examples~\cite{law2018inductive}, and to be suitable for different forms of symbolic learning tasks. In these systems, a noisy example includes a \emph{weight}, which defines the penalty paid by a hypothesis for not covering that example.~\ac{nsl} interprets this weight as a level of certainty of the example, and computes it using the confidence score of the related neural network predictions. In this way, the~\ac{las} systems become biased towards learning a hypothesis that has minimal penalty, i.e., a hypothesis that covers examples generated from high confidence neural network predictions (examples with high weights). For each proposed instance of our~\ac{nsl} framework, we investigate: \begin{inparaenum}[1)]\item whether~\ac{nsl} can learn an accurate and interpretable hypothesis from incorrect feature predictions of the neural component, \item how robust the learned hypothesis is in the presence of distributional shifts applied to an increasing percentage of the unstructured data, \item the impact of using an uncertainty-aware neural network component that provides more robust confidence estimates when distributional shifts are applied to the unstructured data, and \item how~\ac{nsl} performs in comparison to other hybrid systems where the same pre-trained neural networks, used for predicting features from the unstructured data, are integrated with a random forest and deep neural networks trained to learn the knowledge required to solve the downstream task.\end{inparaenum}

To evaluate our~\ac{nsl} framework, we use four neural-symbolic classification tasks, one for each proposed instance\footnote{See \url{https://github.com/DanCunnington/FFNSL} for code and data.}. Firstly, the \textit{Follow Suit Winner} task is a card game where 4 players each play a card and the goal is to predict the winning player. In order to solve the task, the neural network predicts the rank and suit of the playing card images and the rules of the game are learned as symbolic knowledge, where the winner is the player that plays the highest ranked card with the same suit as player 1. The second task is \textit{Sudoku Grid Validity} classification, which consists of observing a sequence of images of handwritten MNIST digits, corresponding to the digits in a Sudoku grid, and predicting if the grid is valid or not. The neural network classifies each digit and the symbolic knowledge required to be learned is the definition of valid (or invalid) Sudoku grids. The final tasks are \textit{Crop Yield Prediction} and \textit{Indoor Scene Classification}, which demonstrate the applicability of \ac{nsl} to real-world problems and datasets. The Crop Yield Prediction task requires predicting the quality of crop yield from an image containing potentially diseased crops, where the neural network predicts the crop's species and disease status, and the learned symbolic knowledge predicts the quality of yield. In the Indoor Scene Classification task, the neural network is pre-trained to predict the scene class from an image, and the learned symbolic knowledge maps scene classes to high-level super-classes. In the first task, the neural network is pre-trained on images of playing cards from a standard deck, but our~\ac{nsl} framework is applied on card images subject to distributional shifts, where a percentage of standard card images are replaced with images from alternative card decks. In the second task, the neural network is pre-trained on the standard MNIST dataset, 
and our~\ac{nsl} system is applied on an out-of-distribution MNIST dataset generated by rotating MNIST digits $90^\circ$ clockwise. In the Crop Yield Prediction task, we pre-train the neural network on the Plant Village dataset \cite{plants}, and apply distributional shifts using a hue filter. Finally, in the Indoor Scene Classification task, we adopt a neural network model pre-trained on the MIT Indoor Scene dataset \cite{quattoni2009recognizing}, and apply distributional shift using blur, hue, and rotation filters. 

Our evaluation demonstrates that~\ac{nsl} outperforms the baselines on all four tasks. The hypotheses learned from unstructured data, subject to distributional shifts, are more interpretable and more accurate than those learned by the random forest and deep neural networks even when these baselines are trained with significantly more data. We have also evaluated the robustness of the~\ac{nsl} instances when applied to a test set that is also subject to distributional shifts. The results show that~\ac{nsl} outperforms the baselines, trained with the same amount of unstructured data, when up to $\sim$80\% of the test set is subject to distributional shifts.

The paper is structured as follows. Section~\ref{sec:background} provides necessary background material on the~\ac{las} framework, alongside further discussion of the drawbacks of the standard neural network Softmax layer for providing robust confidence estimates, and details of the uncertainty-aware neural networks used in this paper. Section~\ref{sec:method} presents our general~\ac{nsl} framework followed by four instances discussed in detail in Section~\ref{sec:method:las}. We introduce our evaluation methodology in Section~\ref{sec:eval_method} and present the results of each~\ac{nsl} instance on the Follow Suit Winner and Sudoku Grid Validity tasks in Sections~\ref{sec:eval_fs_winner} and \ref{sec:eval_sud_grid_validity} respectively, followed by the Crop Yield Prediction and Indoor Scene Classification tasks in Section \ref{sec:real_world_app}. Related work is discussed in Section~\ref{sec:related_work} and Section~\ref{sec:conclusion} concludes the paper.




\section{Background}\label{sec:background}
This section provides an overview of the~\ac{las} framework and the neural network approaches used in~\ac{nsl}. We discuss the difference between confidence estimates of uncertainty-aware neural networks versus that of the standard Softmax layer, when applying these trained networks to out-of-distribution data. This is particularly relevant to our~\ac{nsl} framework, as~\ac{nsl} relies upon neural network predictions and their confidence scores to learn interpretable knowledge for solving a downstream task.

\subsection{Learning from Answer Sets}

\ac{las}~\cite{LawRB19} is a logic-based machine learning approach that extends the field of~\ac{ilp}~\cite{Muggleton1991} with systems ILASP~\cite{Law2018thesis} and FastLAS~\cite{law2020fastlas}. ILASP and FastLAS are capable of learning interpretable knowledge, expressed in the language of~\ac{asp}~\cite{gelfond_kahl_2014}, from noisy labelled examples in an effective and scalable manner. Typically, an~\ac{asp} program includes four types of rules: normal rules, choice rules, and hard and weak constraints. In this paper, we consider~\ac{asp} programs composed of \emph{normal rules} only\footnote{The reader is referred to~\cite{gelfond_kahl_2014} for a full overview of~\ac{asp}.}. A normal rule is of the form $\asp{h \codeif b_1,\ldots, b_n, \naf c_{1},\ldots,\naf c_{m}}$, where $\asp{h}, \asp{b_1},\ldots, \asp{b_n}, \asp{c_{1}},\ldots,\asp{c_{m}}$ are atoms, ``$\mathtt{not}$'' is negation as failure, $\asp{h}$ is the
\emph{head} of the rule and $\asp{b_1,\ldots, b_n, \naf c_{1},\ldots, \naf c_{m}}$ is
the \emph{body} of the rule. The Herbrand Base of an~\ac{asp} program $P$, denoted $HB_P$, is the set of ground (variable free) atoms that can be formed from predicates and constants in $P$. Subsets of $HB_P$ are called \emph{interpretations} of $P$. The semantics of an~\ac{asp} program $P$ is defined in terms of \emph{answer sets}, a subset, denoted as $AS(P)$, of all interpretations of $P$ that satisfy every rule in $P$.  Given an answer set $A$, a ground normal rule is satisfied if the head is satisfied by $A$ whenever all
positive atoms and none of the negated atoms of the body are in $A$, that is when the body is satisfied. A \textit{partial interpretation}, $e_{\mathrm{pi}}$, is a pair of sets of ground atoms $\left \langle e^{\mathrm{inc}}_{\mathrm{pi}}, e^{\mathrm{exc}}_{\mathrm{pi}} \right \rangle$, called the {\em inclusion} and {\em exclusion} sets respectively. An interpretation $I$ \emph{extends}
$e_{\mathrm{pi}}$ iff $e_{\mathrm{pi}}^{\mathrm{inc}} \subseteq I$ and $e_{\mathrm{pi}}^{\mathrm{exc}} \cap I = \emptyset$.

In the~\ac{las} framework, labelled examples are specified as \emph{\acp{cdpi}}. A~\ac{cdpi} example $e$ is a pair $\langle e_{\mathrm{pi}}, e_{\mathrm{ctx}}\rangle$, where
$e_{\mathrm{pi}}$ is a partial interpretation and $e_{\mathrm{ctx}}$ is an~\ac{asp} program called the context of $e$. An~\ac{asp} program $P$ is said to \emph{accept} $e$ if there is at least one answer set $A$ of $P \cup e_{\mathrm{ctx}}$ that extends $e_{\mathrm{pi}}$. Essentially, a~\ac{cdpi} states that a learned program $P$, together with $e_{\mathrm{ctx}}$, should bravely entail\footnote{A program $P$ \emph{bravely entails} an atom $\asp{a}$ if there is at least one answer set of $P$ that contains $a$.} all inclusion atoms and none of the exclusion atoms of $e$.
When a~\ac{cdpi} example is noisy, that is, the truthfulness of its context and/or partial interpretation is not guaranteed, it has a \emph{weight} or \emph{penalty} assigned to it, in the form of a positive integer. A \empty{\ac{wcdpi}} is therefore a~\ac{cdpi} weighted with a penalty. It is formally defined as a tuple $e=\langle e_{\mathrm{id}}, e_{\mathrm{pen}}, e_{\mathrm{pi}}, e_{\mathrm{ctx}} \rangle$ where $e_{\mathrm{id}}$ is a unique identifier of $e$, $e_{\mathrm{pen}}$ is the penalty of $e$, and $e_{\mathrm{pi}}$ and $e_{\mathrm{ctx}}$ represent a~\ac{cdpi}. A~\ac{las} system that is noise-tolerant learns an~\ac{asp} program $H$, called a \emph{hypothesis}, from~\ac{wcdpi} examples. If a hypothesis $H$ does not accept a~\ac{wcdpi} example, we say that it \emph{pays the penalty} of that example. Informally, penalties are used to calculate the cost associated with a hypothesis for not covering examples. The cost function of a hypothesis $H$ is the sum over the penalties of all of the examples that are not \emph{covered} by $H$, augmented with the length of the hypothesis. A~\ac{las} learning task with noisy examples, consists of an~\ac{asp} program denoting background knowledge, a hypothesis space defined by a language bias\footnote{For a detailed definition of a language bias see~\cite{Law2018thesis}.}, expressing the set of rules that can be used to construct a solution of the task, and a set of~\ac{wcdpi} examples. The goal of such a task is to find a hypothesis $H$ in the hypothesis space that minimises a cost function with respect to a given set of noisy examples. This is formally defined below, adapted from~\cite{Law2018thesis}. 

\begin{mydefinition}
\label{def:lnas}
  An $\mathrm{ILP}^{\mathrm{noise}}_{\mathrm{LAS}}$ task $T$ is a tuple $T=\langle B, S_M, E\rangle$, where $B$ is an~\ac{asp} program, $S_M$ is a hypothesis space, and $E$ is a set of~\acp{wcdpi}. Given a hypothesis $H \subseteq S_M$,
\begin{enumerate}
    \item $\mathrm{UNCOV}(H, T)$ is the set consisting of all examples $e \in E$ such that $B \cup H$ does not accept $e$.
    \item The penalty of $H$, denoted as $\mathrm{PEN}(H, T)$, is the sum $\sum_{e \in
      \mathrm{UNCOV}(H, T)} e_{\mathrm{pen}}$.
    \item The score of $H$, denoted as $\mathcal{S}(H, T)$, is calculated as $\vert H\vert + \mathrm{PEN}(H, T)$.
    \item $H$ is an \emph{optimal inductive solution} of $T$ if and only if $\nexists H' \subseteq S_M$ such that $\mathcal{S}(H', T) <
      \mathcal{S}(H, T)$.
  \end{enumerate}
\end{mydefinition}

\noindent
ILASP and FastLAS are two state-of-the-art systems capable of solving an $\mathrm{ILP}^{\mathrm{noise}}_{\mathrm{LAS}}$ task. The optimisation function used by both systems aims at learning a hypothesis $H$ that jointly minimises the total penalty paid for the uncovered examples and its length. In practice, this creates a bias towards shorter, and therefore more general solutions that cover examples with a high penalty value.

\subsection{Uncertainty-aware neural networks}

Our~\ac{nsl} framework relies on pre-trained neural networks to extract symbolic features from unstructured data. The neural network prediction and its confidence score may therefore affect the accuracy of a learned hypothesis. In this paper, we consider two different types of neural networks as \ac{nsl} neural components: a standard~\ac{cnn} that uses a Softmax layer, and an uncertainty-aware~\ac{cnn} that provides more robust confidence estimates when given data outside the training distribution.

Uncertainty can be formulated as either \textit{aleatoric} or \textit{epistemic} uncertainty~\cite{hullermeier2021aleatoric,pearce2021understanding}. In a machine learning classification task, aleatoric uncertainty can be thought of as the uncertainty along the class decision boundary, whereas epistemic uncertainty can be thought of as whether the sample falls into any of the classes at all. The confidence estimates output by a neural network Softmax layer in a classification task often only capture aleatoric uncertainty, as these outputs are based on a single probability distribution over a set of classes squashed into real values between 0 and 1. For example, given neural network output logits $\boldsymbol{l}$ and $k$ possible classes, the Softmax output $\sigma(\boldsymbol{l})$ for class $i$, where $1\leq i\leq k$ is calculated as: \[\sigma(\boldsymbol{l})_{i}=\frac{\boldsymbol{e}^{l_i}}{\sum_{j=1}^{k}\boldsymbol{e}^{l_j}}\] where $\boldsymbol{e}=2.71828...$ is the Euler number\footnote{We have used bold $\boldsymbol{e}$ to avoid confusion with a \ac{wcdpi} $e$.}. There are three challenges with this approach in terms of uncertainty quantification. Firstly, the exponent applied to neural network outputs inflates the confidence estimate. Secondly, as the Softmax output is a point-wise, multinomial distribution, it is only possible to compare the confidence of the predicted class among other classes, as opposed to estimating the predictive distribution variance~\cite{sensoy2018evidential}. Finally, when Softmax is paired with the commonly used cross-entropy loss, the network is only trained to minimise prediction error, as opposed to expressing uncertainty robustly. 

To address these challenges, many techniques have been proposed in the literature~\cite{rasmussen2003gaussian,bnns,blundell2015weight,abdar2021review}. In this paper we consider the \textit{EDL-GEN}~\cite{sensoy2020uncertainty} approach, which is a neural network based on generative models of~\ac{edl} systems~\cite{sensoy2018evidential} that have been shown to achieve state-of-the-art performance in handling epistemic uncertainty.
An~\ac{edl}~\cite{sensoy2018evidential} system replaces the Softmax layer in a neural network with a linear layer that represents the parameters of a Dirichlet distribution, a second-order distribution that inherently models the variance of a predictive distribution as opposed to the single point-wise output provided by Softmax. 
It then uses a new loss function that jointly minimises prediction error and the variance of the Dirichlet distribution, to reduce aleatoric uncertainty on the class decision boundary. EDL-GEN~\cite{sensoy2020uncertainty} extends this approach to also capture epistemic uncertainty by firstly treating the output of each class as a binary decision and secondly, using a variational auto-encoder to automatically generate out-of-distribution samples for training, in order to help the network discriminate between samples within and outside the training distribution.

To better understand how the uncertainty estimation of neural network predictions impacts the overall accuracy of our~\ac{nsl} framework, we analyse in the evaluation Sections~\ref{sec:eval_fs_winner} and~\ref{sec:eval_sud_grid_validity}, the predicted confidence scores generated by a standard~\ac{cnn} with a Softmax layer and an EDL-GEN neural network and evaluate how they affect the accuracy of~\ac{nsl} when increasing percentages of input training data are subject to distributional shifts.

\section{\ac{nsl} Framework}\label{sec:method}
\begin{figure*}[h]
  \centering
  \includegraphics[keepaspectratio, width=1\textwidth]{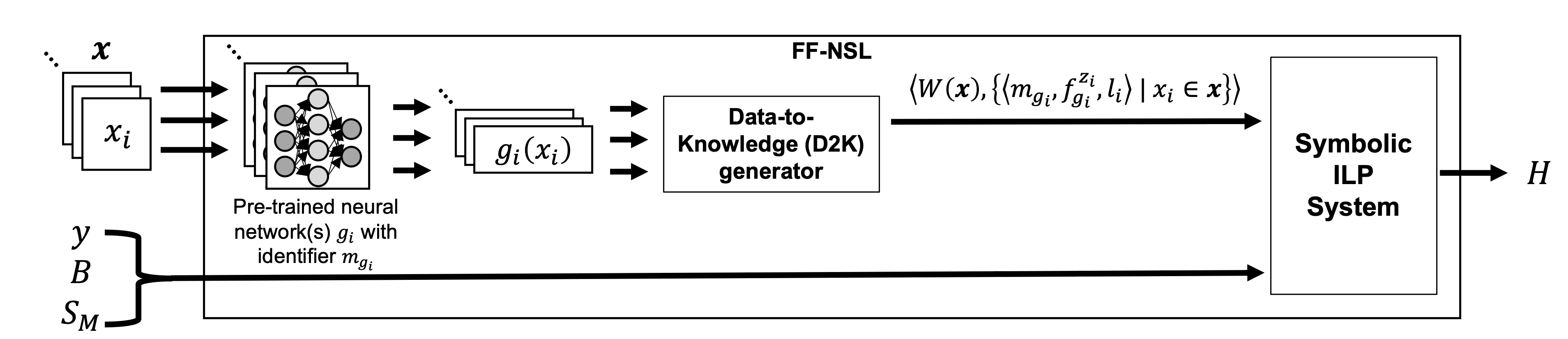}
  \caption{{\small \ac{nsl} architecture and data flow generated for a single data point $\langle \boldsymbol{x},y\rangle$, where $\boldsymbol{x}$ is a sequence of images and $y$ is a label for the sequence. $B$ is the background knowledge, $S_{M}$ is the hypothesis search space and $H$ is the learned hypothesis. In practice, the architecture is applied on a set of data points from which the~\ac{d2k} generator produces a set of symbolic examples passed in as input to the symbolic learner.}}
  \label{fig:arch}
\end{figure*}

In this section we present our general~\ac{nsl} framework. It consists of three components, a pre-trained neural network, a symbolic (logic-based) learning system and a~\ac{d2k} generator that bridges the neural and symbolic learning components. It takes as input a dataset $D$ of labelled (sequences of) unstructured data, alongside a background knowledge $B$ (if any) and a search space $S_{M}$. The output is a \emph{hypothesis} $H$ in the search space $S_{M}$ ($H\subseteq S_{M}$), that predicts the labels of (sequences of) unstructured data. An overview of the \ac{nsl} architecture is presented in Figure~\ref{fig:arch}. 


We now define each of the three components of our~\ac{nsl} architecture. Let us assume that the training dataset $D$ is given by a finite set $D=\{\langle \boldsymbol{x}_{w}, y_{w}\rangle \mid 1 \leq w\leq \vert D\vert\}$. The downstream task is a classification task where the objective is to predict the target label $y\!\in\! \mathcal{Y}$ given a sequence of unstructured data $\boldsymbol{x}\!\in\! \mathcal{X}_{1}\!\times\!\ldots\!\times\! \mathcal{X}_{n}$. Note that $\mathcal{X}_{i}$ could refer to different types of unstructured inputs and the sequence could also contain only a single input. The neural component of~\ac{nsl} contains up to $n$ \textit{pre-trained} neural network(s)\footnote{If the input sequence contains the same type of data, only one neural network is required.}. Each neural network \mbox{$g_{i}:\mathcal{X}_{i}\rightarrow [0,1]^{k_{i}}$} returns a vector denoting relative assignment to $k_{i}$ possible classes for an unstructured input $x_{i} \in \boldsymbol{x}$. Each possible class $z_{i}\in \left \{1,\ldots,k_{i}\right \}$ represents a set of symbolic feature and value pairs from a given set $F_{g_{i}}$ of symbolic feature mappings associated with the neural network $g_{i}$. For example, in the Follow Suit Winner task, $F_{g_{i}}$ contains all possible suit and rank values corresponding to the possible predictions of the neural network, when given an image of a playing card. 



The second component of~\ac{nsl} is the~\ac{d2k} generator that outputs a symbolic representation of the sequence of neural network predictions, together with an aggregated confidence value. Specifically, for a given sequence of unstructured data $\boldsymbol{x}=\langle x_{1},\ldots,x_{n}\rangle$, the~\ac{d2k} generator takes each neural network output $g_{i}(x_{i})$, and computes the corresponding prediction $z_{i}$. Each $z_{i}$, for $1\leq i\leq n$, is obtained by using the standard ``arg max" function, i.e., the class with the maximum confidence score:\[z_{i} = \argmax_{j\in\{1,\ldots,k_{i}\}} (g_{i}(x_{i})[j])\] The~\ac{d2k} generator then uses the set $F_{g_{i}}$, associated with $g_{i}$ and generates the set $f^{z_{i}}_{g_{i}}\subseteq F_{g_{i}}$ of symbolic feature and value pairs corresponding to the prediction $z_{i}$. As an example, in the Follow Suit Winner task, $z_{i}$ is an identifier for one of 52 playing cards, and $f^{z_{i}}_{g_{i}}$ contains two feature and value pairs, one for the suit, and one for the rank of the card $z_{i}$. The~\ac{d2k} generator also generates a set $l_{i}$ of pairs containing additional symbolic meta-data, associated with each input $x_{i}$. Again, each pair in $l_{i}$ contains a name and a value. In the Follow Suit Winner task, $l_{i}$ contains one pair indicating which player played the card $z_{i}$. The generated set of tuples $\{\langle m_{g_{i}}, f^{z_{i}}_{g_{i}}, l_{i}  \rangle \mid x_{i} \in \boldsymbol{x}\}$, where $m_{g_{i}}$ is a unique identifier for the neural network $g_{i}$, defines the symbolic features extracted from a sequence of unstructured data $\boldsymbol{x}$, based on the neural network predictions. Finally, the~\ac{d2k} generator computes an aggregated confidence value $W(\boldsymbol{x})$ for the generated symbolic features, representing the combined confidence scores of the neural network predictions:

\vspace{-2mm}
\begin{equation}\label{eqn:general_weight}
    W(\boldsymbol{x}) = \mathrm{min}(\left \{ g_{i}(x_{i})[z_{i}] \mid x_{i} \in \boldsymbol{x}\right \})
\end{equation}  

\vspace{2mm}
\noindent $W(\boldsymbol{x})$ is a generalisation of the binary Gödel \textit{t}-norm used in fuzzy logic to encode fuzzy conjunctions~\cite{metcalfe2008proof}. So, given a sequence of unstructured inputs $\boldsymbol{x}=\langle x_{1},\ldots,x_{n}\rangle$ and the predicted vector $\langle g_{1}(x_1)[z_1],\ldots,g_{n}(x_n)[z_{n}]\rangle$ from the neural network, the output of the~\ac{d2k} generator is formally defined as:

\vspace{-2mm}

\begin{equation}
     D2K(\boldsymbol{x}) = \langle W(\boldsymbol{x}), \left \{ \langle m_{g_{i}}, f^{z_{i}}_{g_{i}}, l_{i}  \rangle \mid x_{i} \in \boldsymbol{x} \} \right \rangle
\end{equation}

\noindent A pseudo-code implementation of the \ac{d2k} generator is presented in Algorithm \ref{alg:d2k}. Note that some aspects are task specific, such as the set of feature value pairs $f^{z_{i}}_{g_{i}}$, and meta-data $l_{i}$. These are left general in Algorithm \ref{alg:d2k}, and specified in more detail for each task in Section \ref{sec:method:las}.
\begin{algorithm}
    \caption{D2K generator}\label{alg:d2k}
    \KwIn{$\boldsymbol{x}$ \;\;\tcp{A sequence of images}}
    $SF = \emptyset$
    
    $CS = \emptyset$ 
    
    \For{$x_{i} \in \boldsymbol{x}$}{
        \tcp{Obtain the neural network prediction for each input $x_{i}$}
        $z_{i} = \argmax_{j\in\{1,\ldots,k_{i}\}} (g_{i}(x_{i})[j])$
        
        \tcp{Accumulate symbolic facts associated with the neural network prediction $z_{i}$, using the set of feature value pairs $f^{z_{i}}_{g_{i}} \subseteq F_{g}$, and meta-data $l_i$}
        $SF = SF \cup \{\langle m_{g_{i}}, f^{z_{i}}_{g_{i}}, l_{i} \rangle \}$
        
        \tcp{Accumulate the neural network confidence of prediction $z_{i}$}
        $CS = CS \cup \left \{ g_{i}(x_{i})[z_{i}]\right \}$
    }
    \tcp{Calculate the aggregated weight penalty}
    $W=\mathrm{min}(CS)$
    
    \tcp{Return the D2K output}
    \KwRet $\langle W, SF \rangle$

\end{algorithm}

\vspace{2mm}
The third component of our~\ac{nsl} framework is a symbolic logic-based machine learning system. For each labelled unstructured data $\langle \boldsymbol{x}, y\rangle \in D$, the symbolic learning system takes as input $D2K(\boldsymbol{x})$ and the label $y$, and generates a \emph{weighted symbolic labelled example} denoted as the tuple $\langle W^{\prime}(\boldsymbol{x}), e_{\langle \boldsymbol{x}, y\rangle}\rangle$ where $W^{\prime}(\boldsymbol{x})$ is a penalty for the example, calculated from the aggregated confidence score $W(\boldsymbol{x})$, and $e_{\langle \boldsymbol{x}, y\rangle}$ is a labelled example. The syntactic form of $e_{\langle \boldsymbol{x}, y\rangle}$ and the calculation of the penalty $W^{\prime}(\boldsymbol{x})$ depends on the specific symbolic learning system used in the instantiation of the framework. In Section~\ref{sec:method:las} we present two specific instances of~\ac{nsl} where the symbolic learning systems are~\ac{las} systems and we show how weighted symbolic labelled examples are defined as~\ac{wcdpi} examples. 
We denote with $E$ the set of weighted symbolic labelled examples defined by the symbolic learning system for all $\langle \boldsymbol{x}, y\rangle \in D$. A symbolic learning task $T=\langle B, S_{M}, E\rangle$ is then generated where $B$ and $S_{M}$ are respectively the background knowledge and a search space given as input to~\ac{nsl}. The symbolic learner then computes an optimal solution $H$ for this task $T$ as the output of~\ac{nsl}.

Formally, an~\ac{nsl} learning task is a tuple $T=\langle B, S_{M}, D\rangle$ where $D$ is a set of labelled unstructured data, $B$ is a set of optional background knowledge and $S_{M}$ is a search space of possible solutions for $T$. A \emph{hypothesis} $H\subseteq S_{M}$ is an inductive solution of $T$ if and only if $H$ is an \emph{optimal inductive solution} of the symbolic learning task $\langle B, S_{M}, E\rangle$, where $E$ is the set of weighted symbolic labelled examples automatically generated relative to the given set $D$. In the next section, we present four specific instances of our~\ac{nsl} framework and give specific examples of the components described here.  



\section{\ac{nsl} with~\ac{las} Systems}\label{sec:method:las}
The generality of our~\ac{nsl} framework allows it to be instantiated differently, using alternative neural and/or symbolic learning components, depending on the nature of the classification task in hand. We have considered four different classification tasks, called \emph{Follow Suit Winner}, \emph{Sudoku Grid Validity}, \textit{Crop Yield Prediction} and \textit{Indoor Scene Classification} respectively. The first requires the learning of concepts that are not directly observed in the labels, but linked to the label through the background knowledge, whereas the other tasks require the learning of concepts that define the classification label. Because of the different types of symbolic learning, we consider instantiations of our~\ac{nsl} framework with different~\ac{las} systems. In what follows we introduce these tasks, their datasets, define the respective~\ac{nsl} learning tasks and describe in more detail the~\ac{nsl} instances we have implemented to solve these tasks. Firstly, let us define the weighted symbolic labelled examples within a~\ac{las} system, based on the output from the~\ac{d2k} generator. Essentially, the predicted symbolic features and meta-data define the \textit{context} of a~\ac{las} example, represented as a conjunction of facts, and the aggregated confidence score $W(\boldsymbol{x})$ is used to calculate the associated \textit{weight penalty}:

\vspace{-2mm}
\begin{equation}\label{eqn:las_weight}
    W^{\prime}(\boldsymbol{x}) = \lfloor 100 \times W(\boldsymbol{x})\rfloor + 1
\end{equation}

\vspace{2mm}
\noindent which converts $W(\boldsymbol{x})$ to an integer $W^{\prime}(\boldsymbol{x})\!>\!0$ as required by the~\ac{las} systems. Given the output generated by~\ac{d2k}, a~\ac{las} system 
constructs a weighted symbolic labelled example $\langle W^{\prime}(\boldsymbol{x}), e_{\langle \boldsymbol{x}, y\rangle}\rangle$ as a~\ac{wcdpi} of the form $\langle e_{\mathrm{id}}, e_{\mathrm{pen}}(\boldsymbol{x}), e_{\mathrm{pi}}(y), e_{\mathrm{ctx}}(\boldsymbol{x})\rangle$, where  $e_{\mathrm{id}}$ is a unique identifier, $e_{\mathrm{pen}}(\boldsymbol{x}) = W^{\prime}(\boldsymbol{x})$, $e_{\mathrm{pi}}(y)$ is the partial interpretation $\langle \{y\}, \mathcal{Y}\setminus\{y\}\rangle$, defined in terms of the label $y$ and its domain $\mathcal{Y}$, and the context $e_{\mathrm{ctx}}(\boldsymbol{x})$ is a conjunction of facts created from the predicted symbolic features and meta-data. The components $e_{\mathrm{pi}}(y)$ and $e_{\mathrm{ctx}}(\boldsymbol{x})$ together constitute the labelled example $e_{\langle \boldsymbol{x}, y\rangle}$. 


Given a set $E^{\prime}$ of~\acp{wcdpi}, a background knowledge $B$, and a search space $S_{M}$, a hypothesis $H \subseteq S_{M}$ is learned such that $H$ is an optimal inductive solution of the task $T^{\mathrm{noise}}_{\mathrm{LAS}}=\langle B, S_{M}, E^{\prime}\rangle$. Let us now present the tasks used in our evaluation, alongside examples of each instantiated~\ac{nsl} component.

\subsection{Follow Suit Winner}
This is a classification task where 4 players each play 1 card and the goal is to predict the winning player. The symbolic knowledge required to solve the task defines the rules of the game, that is \emph{the winner is the player that plays the highest ranked card with the same suit as player 1}. Each $\langle \boldsymbol{x}, y\rangle\in D$ is composed of a sequence $\boldsymbol{x}$ of 4 card images corresponding to the cards played by players $1,\ldots,4$, and a label $y\in\{1,2,3,4\}$ denoting the player who wins the 4 card trick. 

Let us assume $\boldsymbol{x}= [ \text{\scalerel*{\includegraphics{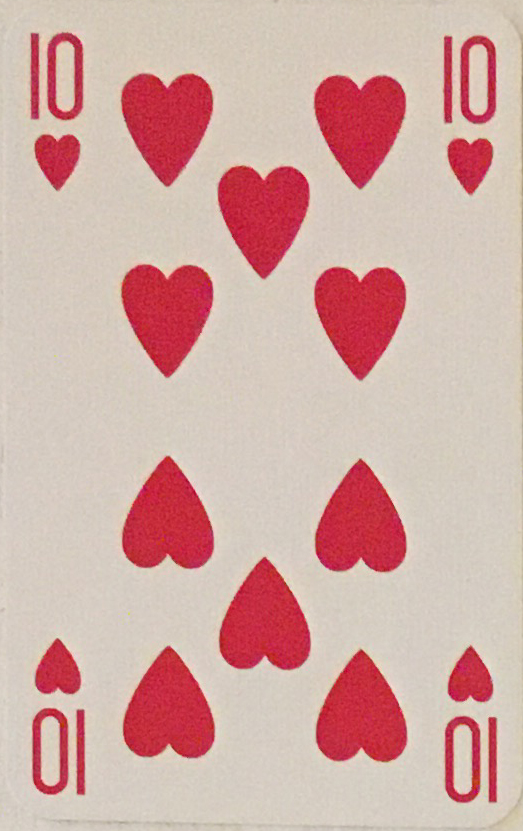}}{)}},
\text{\scalerel*{\includegraphics{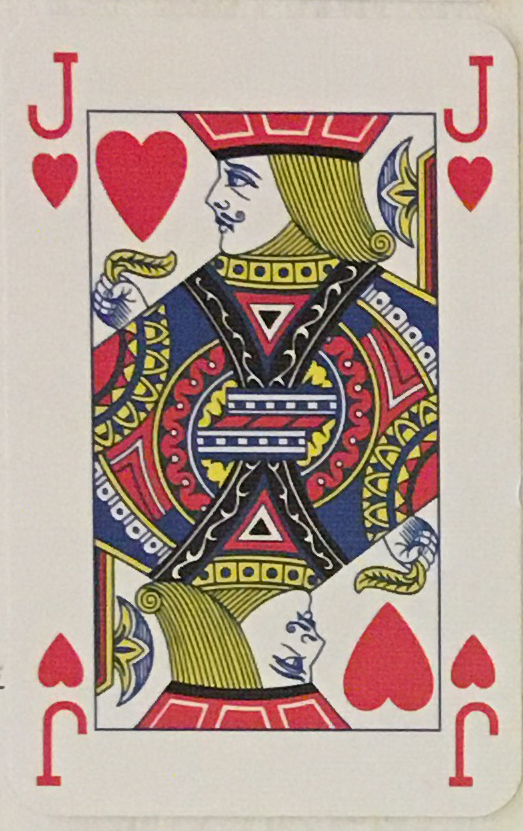}}{)}},
\text{\scalerel*{\includegraphics{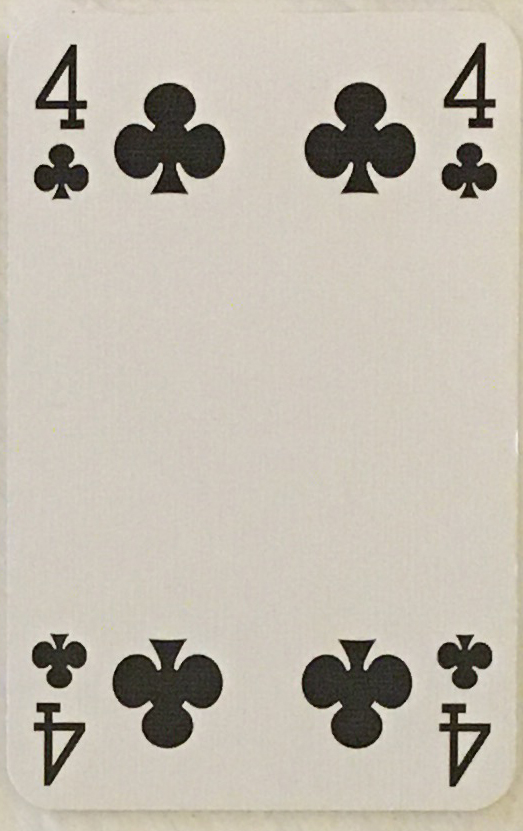}}{)}},
\text{\scalerel*{\includegraphics{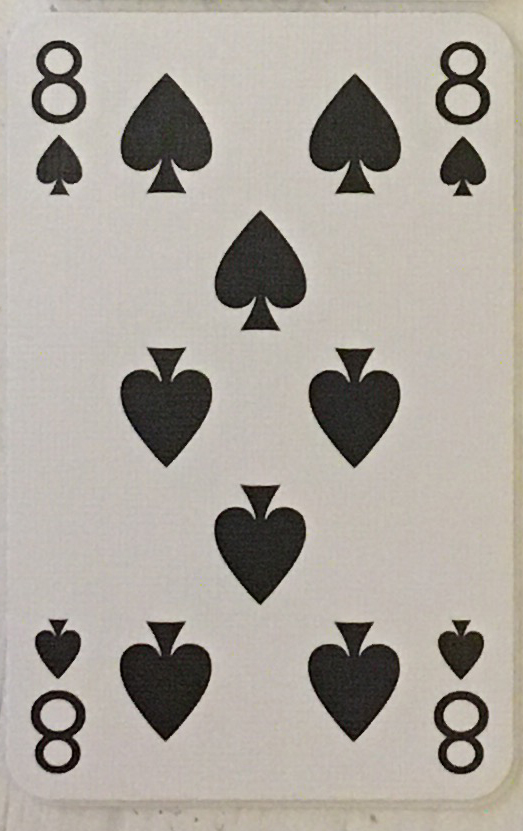}}{)}}
]$ which contains images of the cards \textit{10 of hearts}, \textit{jack of hearts}, \textit{4 of clubs} and \textit{8 of spades} played by player $1$, $2$, $3$ and $4$ respectively. For this trick, the ground truth label is $y=2$ indicating that player $2$ is the winner since player $2$ has played the highest ranked card with the same suit as player $1$. Since the unstructured inputs in the sequence $\boldsymbol{x}$ are of the same type (i.e., card images),~\ac{nsl} can simply use a single neural network $g$ pre-trained to predict the features of a card image, that is the rank and suit of each card. Therefore, $g$ has two associated symbolic features $rank$ and $suit$ each with values $\{2,\ldots,10, jack, queen, king, ace\}$ and $\{hearts, clubs, spades, diamonds\}$ respectively. For each input $x_{i}$, there are $52$ possible predictions, one for each combination of rank and suit, i.e., $g:\mathcal{X}\rightarrow[0,1]^{52}$, where $\mathcal{X}$ is the set of possible card images. $g$ has an associated feature value mapping $F_{g}$ which gives for each card prediction $z_{i}\in\{1,\ldots,52\}$, a unique set of two pairs, each containing a feature and value, i.e., $f^{z_{i}}_{g}=\{\langle rank, \nu_{rank}\rangle, \langle suit, \nu_{suit}\rangle\}$, where $\nu_{rank}$ is one of the $13$ rank values and $\nu_{suit}$ is one of the $4$ suit values. Furthermore, each input $x_{i}$ also has associated symbolic meta-data $l_{i}=\{\langle player, \nu_{player}\rangle\}$ where $\nu_{player}\in\{1,2,3,4\}$ indicates the player that has played card $x_{i}$.

We instantiate our~\ac{nsl} framework as follows. Given a sequence $\boldsymbol{x}$ of $4$ card images, the neural component of~\ac{nsl} generates $4$ vectors $g(x_{i})$, where $1\leq i\leq 4$. The~\ac{d2k} component generates for each $x_{i}$, the card prediction $z_{i}$ and its corresponding symbolic features and meta-data, thus computing the tuple $D2K(x_{i})=\langle \asp{card}, f^{z_{i}}_{g},l_{i}\rangle$, where $\asp{card}$ is the identifier for the network $g$ (i.e., $m_{g}=\asp{card}$). 

\begin{myexample}
\label{d2kFollowSuit}
Consider the sequence $\boldsymbol{x}= [ \text{\scalerel*{\includegraphics{Figures/cards/10h}}{)}},
\text{\scalerel*{\includegraphics{Figures/cards/jh}}{)}},
\text{\scalerel*{\includegraphics{Figures/cards/4c}}{)}},
\text{\scalerel*{\includegraphics{Figures/cards/8s}}{)}}]$ and $y=2$. Let us assume that the neural network $g$ computes the outputs $g(x_{1}),...,g(x_{4})$ from which the~\ac{d2k} generator generates the correct card predictions $z_{1}=10$, $z_{2}=11$, $z_{3}=17$, and $z_{4}=34$. Let us also assume the neural network confidence scores for these predictions are:

\noindent
\[
\begin{array}{llll}
g(x_{1})[z_{1}] = 0.95; &  g(x_{2})[z_{2}] = 0.92; & g(x_{3})[z_{3}] = 0.80; & g(x_{4})[z_{4}] = 0.94;
\end{array}
\] 
$D2K(\boldsymbol{x})$ is given by the following tuple:
\[
\begin{array}{lll}
D2K(\boldsymbol{x}) = & \langle 0.80, \{ &\langle \asp{card}, \{ \langle \asp{rank}, \asp{10} \rangle, \langle \asp{suit}, \asp{hearts}\rangle \}, \{ \langle \asp{player}, \asp{1} \rangle \}\rangle, \\
& & \langle \asp{card}, \{ \langle \asp{rank}, \asp{jack} \rangle, \langle \asp{suit}, \asp{hearts}\rangle \}, \{ \langle \asp{player}, \asp{2} \rangle \}\rangle,\\
& & \langle \asp{card}, \{ \langle \asp{rank},  \asp{4} \rangle, \langle \asp{suit}, \asp{clubs}\rangle \}, \{ \langle \asp{player}, \asp{3} \rangle \}\rangle, \\
& & \langle \asp{card}, \{ \langle \asp{rank}, \asp{8} \rangle, \langle \asp{suit}, \asp{spades}\rangle \}, \{ \langle \asp{player}, \asp{4} \rangle \}\rangle \;\;\;\}\rangle.
\end{array}
\]
\end{myexample}
\noindent
In this task,~\ac{nsl} uses the symbolic learner ILASP. The concept to be learned is not directly expressed as a label, but is related to it. The label is a single winning player for a trick, but the learned concept requires reasoning over the conditions of the suit and rank values of the other players' cards. We encode as background knowledge, possible suit and rank values, the four players, as well as the definition of a higher rank predicate. ILASP is particularly suited for solving such learning tasks, known as non-observational predicate learning. The full background knowledge $B$ and language bias used to construct the search space $S_{M}$ for this classification task are given in Appendix~\ref{sec:app:ilp}. To generate its learning task, ILASP has to generate its set $E^{'}$ of~\ac{wcdpi} examples based on the output of the~\ac{d2k} component. For example, the~\ac{wcdpi} generated from the~\ac{d2k} output and the corresponding label in Example~\ref{d2kFollowSuit} is: \[\langle e_{\mathrm{id}}, 81, \langle \{ 2 \}, \{ 1,3,4\}\rangle, e_{\mathrm{ctx}} \rangle\]
\noindent where $e_{\mathrm{id}}$ is a unique identifier and $e_{\mathrm{ctx}}$ is the set of facts 
{\small $\{\asp{card(1,10,hearts).}, \asp{card(2,jack,hearts).}, \asp{card(3,4,clubs).}, \asp{card(4,8,spades).}\}$}.

\subsection{Sudoku Grid Validity} 
Our second classification task is \emph{Sudoku Grid Validity}. This consists of observing a sequence of images of handwritten MNIST digits, corresponding to the digits in a Sudoku grid, and predicting if the grid is valid or not\footnote{We assume a Sudoku grid has been pre-processed to return images of digits in different cells and we do not process blank cells.}. The learned symbolic knowledge required to solve this task is the definition of a valid Sudoku grid. In this task, each $\langle\boldsymbol{x},y\rangle\in D$ contains a sequence of digit images $\boldsymbol{x}$ with a label $y\in\{0, 1\}$ for valid and invalid respectively. The length of the sequence depends on the size of the grid. We consider $4\times 4$ and $9\times 9$ Sudoku grids as two separate tasks, with respective datasets $D_{4\times 4}$ and $D_{9\times 9}$ where the maximum length of the sequence in input is given by $n=16$ and $n=81$ respectively. As the images are all MNIST digits,~\ac{nsl} uses two neural networks $g_{4\times 4}$ and $g_{9\times 9}$, depending on the grid size, pre-trained to predict the feature $digit$ of a single image $x_{i}$ in $\boldsymbol{x}$. So $g_{k\times k}:\mathcal{X}\rightarrow[0,1]^{k}$, where $\mathcal{X}$=MNIST. In the case of $D_{4\times 4}$, $n=16$ and $k=4$ whereas in the case of $D_{9\times 9}$, $n=81$ and $k=9$. The neural network $g_{k\times k}$ has associated a feature value mapping $F_{g_{k\times k}}$ which gives for each digit prediction $z_{i}\in\{1,\ldots,k\}$ a unique set of pairs $f^{z_{i}}_{g_{k\times k}} = \{\langle value, \nu \rangle\}$, where $\nu$ is one of the $k$ digits that can appear in a Sudoku grid of size $k\times k$. The meta-data related to each $x_{i}$ is a set of two feature value pairs denoting the row and column that the image $x_{i}$ has in the Sudoku grid, i.e., $l_{i}=\{\langle row, \nu_{row}\rangle, \langle col, \nu_{col}\rangle\}$, where $\nu_{row},\nu_{col}\in\{1,\ldots,k\}$.

The instantiated~\ac{nsl} framework for this classification task is defined as follows. Given a sequence, $\boldsymbol{x}$, of MNIST digit images, for each $x_i\in\boldsymbol{x}$, the pre-trained neural network $g_{k \times k}$ computes the vector $g(x_{i})$. The~\ac{d2k} component generates for each $x_{i}$ the tuple $D2K(x_{i})=\langle \asp{digit}, f^{z_{i}}_{g_{k\times k}}, l_{i}\rangle$ where $\asp{digit}$ is the network identifier, 
$f^{z_{i}}_{g_{k\times k}}$ is the set of symbolic feature values associated with the prediction $z_{i}$, and $l_{i}$ is the set of symbolic meta-data feature value pairs associated with $x_{i}$.

\begin{myexample}
\label{d2kSudoku}
Consider the task of predicting the validity of a $4\times4$ Sudoku grid. Let $\boldsymbol{x}= [ \text{\scalerel*{\includegraphics{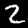}}{)}},
\text{\scalerel*{\includegraphics{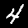}}{)}},
\text{\scalerel*{\includegraphics{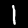}}{)}},
\text{\scalerel*{\includegraphics{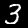}}{)}},
\text{\scalerel*{\includegraphics{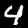}}{)}}
]$, with label $y=1$, and associated symbolic meta-data:  

\[
\begin{array}{ll}
l_{1} = & \{ \langle  \asp{row},\asp{1} \rangle, \langle \asp{col},\asp{1} \rangle\}\\
l_{2} = & \{ \langle \asp{row},\asp{1} \rangle, \langle \asp{col},\asp{3} \rangle\}\\
l_{3} = & \{ \langle \asp{row},\asp{1} \rangle, \langle \asp{col},\asp{4} \rangle\}\\
l_{4} = & \{\langle \asp{row},\asp{3} \rangle, \langle \asp{col},\asp{2} \rangle\}\\
l_{5} = & \{\langle \asp{row},\asp{4} \rangle, \langle \asp{col},\asp{3} \rangle\}
\end{array}
\]

\noindent
Let us assume the neural network $g = g_{4 \times 4}$ and $g$ computes the outputs
$g(x_{1}),...,g(x_{5})$ from which the~\ac{d2k} generator generates the correct digit predictions $z_{1}=2$, $z_{2}=4$, $z_{3}=1$, $z_{4}=3$, and $z_{5}=4$. Let us also assume the neural network confidence scores for these predictions are: $g(x_{1})[z_{1}] = 0.88$,  $ g(x_{2})[z_{2}] = 0.93$, $g(x_{3})[z_{3}] = 0.87$, $g(x_{4})[z_{4}] = 0.97$, and $g(x_{5})[z_{5}] = 0.99$. The aggregated confidence score $W(\boldsymbol{x}) = 0.87$. $D2K(\boldsymbol{x})$ is given by the following tuple:
\[
\begin{array}{lll}
D2K(\boldsymbol{x}) = & \langle 0.87, \{ & 
\langle \asp{digit}, \{ \langle \asp{value}, \asp{2} \rangle \}, \{ \langle \asp{row}, \asp{1} \rangle,\langle \asp{col}, \asp{1} \rangle \}\rangle,\\
& & \langle \asp{digit}, \{ \langle \asp{value}, \asp{4} \rangle \}, \{ \langle \asp{row}, \asp{1} \rangle,\langle \asp{col}, \asp{3} \rangle \}\rangle,\\
& & \langle \asp{digit}, \{ \langle \asp{value}, \asp{1} \rangle \}, \{ \langle \asp{row}, \asp{1} \rangle, \langle \asp{col}, \asp{4} \rangle \}\rangle,\\
& & \langle \asp{digit}, \{ \langle \asp{value}, \asp{3} \rangle \}, \{ \langle \asp{row}, \asp{3} \rangle,\langle \asp{col}, \asp{2} \rangle \}\rangle, \\
& & \langle \asp{digit}, \{ \langle \asp{value}, \asp{4} \rangle \}, \{ \langle \asp{row}, \asp{4} \rangle,\langle \asp{col}, \asp{3} \rangle \}\rangle\;\;\}\;\rangle.
\end{array}
\]
\end{myexample}

\noindent
In this task~\ac{nsl} uses the FastLAS symbolic learner because the task is to learn the definition of the classification label, and FastLAS has been shown, for these types of learning tasks, to be more scalable than ILASP~\cite{law2020fastlas}. For both $4\times4$ and $9\times9$ Sudoku grids, the knowledge of the grid is encoded as part of the background knowledge $B$, given in Appendix~\ref{sec:app:ilp} together with the language bias used to construct the search space $S_{M}$. For each $\langle \boldsymbol{x}, y\rangle$, FastLAS takes as input $D2K(\boldsymbol{x})$ and generates a~\ac{wcdpi} example. For instance, the~\ac{wcdpi} generated for the~\ac{d2k} output and the corresponding label in Example~\ref{d2kSudoku} is: \[\langle e_{\mathrm{id}}, 88, \langle \{ 1 \}, \{ 0\}\rangle, e_{\mathrm{ctx}} \rangle\]
\noindent where $e_{\mathrm{id}}$ is a unique identifier and $e_{\mathrm{ctx}}$ is given by the set of facts 
{\small $\{\asp{digit(1,1,2).},\;  \asp{digit(1,3,4).},\; \asp{digit(1,4,1).},\; \asp{digit(3,2,3).},\; \asp{digit(4,3,4).}\}$}.

\subsection{Crop Yield Prediction}\label{sec:method:crops}
To demonstrate the application of \ac{nsl} to a real-world problem and dataset, consider the \textit{Crop Yield Prediction} task. The goal is to classify the quality of yield, given an image and the location of a particular crop. The symbolic knowledge required to solve the task defines the quality of yield according to the crop's location, species, and any disease that may be present. Each $\langle \boldsymbol{x},y \rangle \in D$ is composed of a sequence $\boldsymbol{x}$ containing a single image, and a label $y\in\{0,1,2\}$ denoting the quality of yield as \textit{poor}, \textit{moderate}, and \textit{strong} respectively.

Let us assume $\boldsymbol{x}=[\text{\scalerel*{\includegraphics{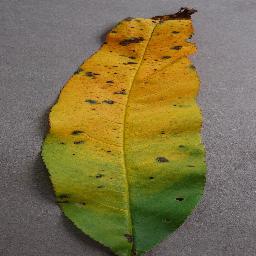}}{)}}]$ which contains an image of a peach crop with the bacterial spot disease. Given symbolic meta-data denoting the location of this crop, let us assume the label $y=0$, indicating poor yield. In this task, we use one neural network $g$ to predict the features of a crop image, which are the crop species and disease. In total, there are 38 possible combinations of crop species and diseases, and $g$ is trained to classify each combination. To assist with neural network training, the image dataset also contains a background class with unrelated images\footnote{In \ac{nsl}, if the neural network predicts the background class, no \ac{wcdpi} example is generated by the \ac{d2k} component.}. Therefore, $g : \mathcal{X} \rightarrow [0,1]^{39}$, where $\mathcal{X}$ is the set of possible crop and background images. $g$ has an associated feature value mapping $F_{g}$, which specifies for the crop prediction $z_{i} \in \{ 1,...,38\}$, a unique set of feature and value pairs $f_{g}^{z_{i}}= \{ \langle species, \nu_{species} \rangle, \langle disease, \nu_{disease} \rangle\}$, where $\nu_{species}$ and $\nu_{disease}$ are the crop species and disease values respectively. Also, each input $x_{i}$ has associated symbolic meta-data $l_{i} = \{ \langle location, \nu_{location}\rangle\}$ where $\nu_{location} \in \{ 1,...,19\}$ is the location of the crop\footnote{The dataset consists of two unique crops in each location, hence 19 possible location values.}.

We instantiate our \ac{nsl} framework as follows. Given a sequence $\boldsymbol{x}$ containing a single crop image, the neural component generates a single vector $g(x_{i})$. The \ac{d2k} component generates the prediction $z_{i}$ and its corresponding symbolic features and meta-data, thus computing the tuple $D2K(x_{i}) = \langle \asp{crop}, f_{g}^{z_{i}}, l_{i}\rangle$, where $\asp{crop}$ is the identifier for the network $g$ (i.e., $m_{g}=\asp{crop}$).

\begin{myexample}
\label{d2kcrop}
Consider the sequence $\boldsymbol{x}=[\text{\scalerel*{\includegraphics{Figures/plant_disease/49.JPG}}{)}}]$ and $y=0$. Let us assume the neural network $g$ computes the output $g(x_{1})$ from which the \ac{d2k} generator generates the correct crop prediction $z_{1}=17$. Let us also assume the neural network predicts with confidence $g(x_{1})[z_{1}]=0.98$, and this crop is in location 5. $D2K(\boldsymbol{x})$ is given by the following tuple:

\[
\begin{array}{llll}
D2K(\boldsymbol{x}) = & \langle 0.98, \{ & 
\langle \asp{crop}, & \{ \langle \asp{species}, \asp{peach} \rangle, \langle \asp{disease}, \asp{bacterial\_spot} \rangle \},\\& & & \{\langle \asp{location}, \asp{5} \rangle\}\rangle\;\;\}\;\rangle.
\end{array}
\]
\end{myexample}

In this task \ac{nsl} uses the FastLAS symbolic learner which is shown to be more scalable than ILASP. The background knowledge $B$ contains a rule that ensures a classification is performed, i.e., given a crop, disease, and a location, the learned hypothesis should output only one class of crop yield. This rule, alongside the language bias used to construct the search space $S_{M}$ is given in Appendix \ref{sec:app:ilp}. For each $\langle \boldsymbol{x},y\rangle$, FastLAS takes as input $D2K(\boldsymbol{x})$ and generates a \ac{wcdpi} example. For instance, the \ac{wcdpi} generated from the \ac{d2k} output and the corresponding label given in Example \ref{d2kcrop} is: \[\langle e_{\mathrm{id}}, 99, \langle \{ 0 \}, \{ 1,2\}\rangle, e_{\mathrm{ctx}} \rangle\]
\noindent where $e_{\mathrm{id}}$ is a unique identifier and $e_{\mathrm{ctx}}$ is given by the set of facts 
{\small $\{\asp{species(peach).},\; \asp{disease(bacterial\_spot).},\; \{\asp{location(5).}\}$}.

\subsection{Indoor Scene Classification}\label{sec:method:scene}
Our final instantiation of \ac{nsl} is with the \textit{Indoor Scene Classification} task, where both neural and symbolic components are trained with real data. The goal is to learn symbolic knowledge that maps indoor scene classes (e.g., bedroom, bathroom, kitchen) into higher level super-classes (e.g., home), given images of indoor scenes. Each $\langle \boldsymbol{x},y \rangle \in D$ is composed of a sequence $\boldsymbol{x}$ containing a single indoor scene image, and a label $y\in\{0,\ldots,4\}$ denoting the super-class as \textit{store}, \textit{home}, \textit{public space}, \textit{leisure}, and \textit{working place} respectively.

\begin{figure}[!htb]
    \centering
    \includegraphics[width=0.3\textwidth]{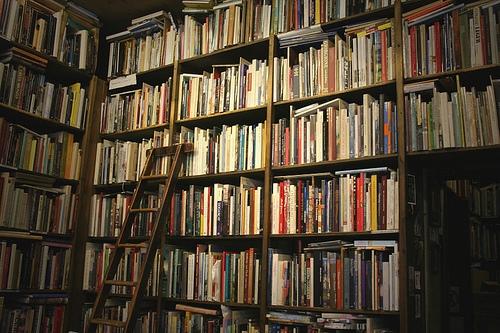}
    \caption{Example bookstore image from the MIT Indoor Scenes dataset. \cite{quattoni2009recognizing}}
    \label{fig:mit_im_example}
\end{figure}
Let us assume $\boldsymbol{x}=[\text{\scalerel*{\includegraphics{Figures/indoor_scenes/bookstore_32_10_flickr.jpg}}{)}}]$ which contains an image of a bookstore (also shown in Figure \ref{fig:mit_im_example}). The label for this example is $y=0$ (i.e., \textit{store}). We use one neural network $g$ to predict the scene class. In total, there are 67 different classes of various indoor scenes, and therefore $g:\mathcal{X}\rightarrow[0,1]^{67}$, where $\mathcal{X}$ is the set of possible images in the MIT Indoor Scene dataset. $g$ has an associated feature value mapping $F_{g}$, which for the scene prediction $z_{i} \in \{1,...,67\}$, gives a pair that denotes the symbolic scene name $\nu_{scene}$, i.e., $f^{z_{i}}_{g}=\{\langle scene, \nu_{scene} \rangle\}$. In this task there is no symbolic meta-data associated with each input $x \in \boldsymbol{x}$. Given a sequence $\boldsymbol{x}$ containing a single scene image, the neural component generates a single vector $g(x_{i})$. The \ac{d2k} component generates the prediction $z_{i}$ and its corresponding symbolic feature, thus computing the tuple $D2K(x_{i})=\langle \asp{image}, f^{z_{i}}_{g}, \{ \}\rangle$, where $\asp{image}$ is the identifier for the network $g$ (i.e., $m_{g}=\asp{image}$).

\begin{myexample}
\label{d2kscene}
Consider the sequence $\boldsymbol{x}=[\text{\scalerel*{\includegraphics{Figures/indoor_scenes/bookstore_32_10_flickr.jpg}}{)}}]$ and $y=0$. Let us assume the neural network $g$ computes the output $g(x_{1})$ from which the \ac{d2k} generator generates the correct scene prediction $z_{1}=8$. Let us also assume the neural network predicts with confidence $g(x_{1})[z_{1}]=0.96$. $D2K(\boldsymbol{x})$ gives as output the following tuple:

\[
\begin{array}{llll}
D2K(\boldsymbol{x}) = \langle 0.96, \{ & 
\langle \asp{image}, \{ \langle \asp{scene}, \asp{bookstore} \rangle \}, \{\} \rangle &\; \}\rangle.
\end{array}
\]
\end{myexample}

The FastLAS symbolic learner is also used in this task. No background knowledge is required, and the language bias is given in Appendix \ref{sec:app:ilp}. For each $\langle\boldsymbol{x}, y\rangle$, FastLAS takes as input $D2K(\boldsymbol{x})$ and generates a \ac{wcdpi} example. For instance, the \ac{wcdpi} generated for the D2K output in Example \ref{d2kscene} is:\[\langle e_{\mathrm{id}}, 96, \langle \{ 0 \}, \{ 1,2,3,4\}\rangle, e_{\mathrm{ctx}} \rangle\]
\noindent where $e_{\mathrm{id}}$ is a unique identifier and $e_{\mathrm{ctx}}$ is given by the set
{\small $\{\asp{scene(bookstore).}\}$}.



\section{Evaluation Methodology}\label{sec:eval_method}
In this section we describe the methodology used to evaluate the~\ac{nsl} framework. In the first two tasks, the focus is on learning complex first-order knowledge involving negation as failure and predicate invention, which are essential aspects of common-sense learning and reasoning. In the second two tasks, we demonstrate \acp{nsl} applicability to real-world problems and datasets. For each of the four classification tasks, we divide the evaluation into two types. Firstly, we evaluate the symbolic \textit{learning} capability of~\ac{nsl}, where the goal is to learn interpretable knowledge from symbolic features extracted from pre-trained neural network predictions. Secondly, we evaluate the \textit{inference} capability of~\ac{nsl}, where the pre-trained neural networks together with the learned knowledge are used to make a downstream classification of \textit{unseen} unstructured data. We refer to the first type of evaluation as the \textit{learned hypothesis evaluation} and the second type as the \textit{\ac{nsl} framework evaluation}, since this targets both neural and symbolic components. Let us now describe each evaluation type in more detail.

\subsection{Learned hypothesis evaluation}
We evaluate the learned hypothesis in terms of accuracy, interpretability and learning time. To measure accuracy, we use a symbolic test set containing ground truth symbolic features. This ensures that the evaluation only targets the accuracy of the learned hypothesis. For each example in the test set, the symbolic features are used with the learned hypothesis to make a prediction of the downstream label. This prediction is compared to the ground truth label in the test set and accuracy is computed using the standard measure. Since~\ac{nsl} learns knowledge from a pre-trained neural network, we consider the hypotheses that have been learned at each (increasing) percentage of distributional shift and evaluate the accuracy of the knowledge that~\ac{nsl} learns in the presence of incorrect neural network predictions. Note that the symbolic test set remains unchanged and is not affected by the distributional shifts, as we want to evaluate in this case just the accuracy of the learned hypotheses.

To perform a deeper analysis of the accuracy of the learned hypotheses, we take into consideration the following measures. Firstly, the accuracy and confidence score distribution of the pre-trained neural network(s) in classifying unstructured data in the training set $D$. Since the neural networks were pre-trained on a dataset different from $D$, this 
measure enables us to understand the reliability of the pre-trained neural network predictions over new unseen input data (For more dataset details, see Appendix~\ref{sec:app:datasets}.). Secondly, we measure the percentage of \ac{wcdpi} examples generated by the LAS system, that contains features in the context which are incorrect with respect to the label in the inclusion set. This enables us to understand the relationship between incorrect neural network predictions and the accuracy of the learned hypotheses, as well as analyse how many correct \ac{wcdpi} examples are needed to learn hypotheses with a certain level of accuracy. Thirdly, we calculate the weight penalty ratio $r$ over the generated \ac{wcdpi} examples, defined as \[r = \frac{\sum_{e \in E^{\prime}_{\mathrm{correct}}}{e_{\mathrm{pen}}}}{\sum_{e \in E^{\prime}}{e_{\mathrm{pen}}}}\] where $E^{\prime}_{\mathrm{correct}}$ is the set of correctly generated \ac{wcdpi} examples, (i.e.,~\ac{wcdpi} examples with features in the context that are consistent with the label in the inclusion set) and $E^{\prime}$ is the complete set of generated \ac{wcdpi} examples. This enables us to measure the bias given to the LAS system by the weights of the \ac{wcdpi} examples, which are based on the neural network confidence scores. Ideally,~\ac{nsl} should allocate a higher proportion of the total weight penalty to \ac{wcdpi} examples that contain correct neural network predictions. We compare the accuracy of the knowledge learned from these \ac{wcdpi} examples with that of knowledge learned from corresponding \ac{wcdpi} examples where we fix the penalty to be constant for all examples, as a baseline. To measure interpretability, we count the total number of atoms in a learned hypothesis: a hypothesis with a lower number of atoms is considered to be more interpretable~\cite{lakkaraju2016interpretable}. Finally, we measure the wall-clock time taken to learn a hypothesis at each percentage of distributional shift.

\subsection{\ac{nsl} framework evaluation}
When a hypothesis has been learned, the entire~\ac{nsl} framework can be evaluated using a test set containing unseen labelled unstructured data. In this case, the neural network component of~\ac{nsl} classifies each element of a sequence of unstructured data. The symbolic features predicted from the neural network classification are added to the background knowledge alongside the learned hypothesis. The symbolic component of~\ac{nsl} is used to compute the downstream prediction. This is compared to the ground-truth label associated with the sequence of unstructured data, and the accuracy is computed with the standard measure. To assist the evaluation, and to provide insight into where mistakes are being made, we evaluate the neural network accuracy in predicting the symbolic features from the unstructured data in the test set with respect to ground truth information. This enables us to identify whether any downstream classification error is due to neural network feature prediction, the learned hypothesis, or both. We also evaluate~\ac{nsl} under distributional shifts. We inject into the test data the same percentages of distributional shifts used during the learning of hypotheses, and evaluate the accuracy of~\ac{nsl}. This evaluates the performance of~\ac{nsl} in realistic scenarios where distributional shifts occur during learning and inference.

\subsection{Experimental setting}
In the next four sections, we present the results of the Follow Suit Winner, Sudoku Grid Validity, Crop Yield Prediction, and Indoor Scene Classification tasks, using the evaluation methodology outlined in this section. In the first three tasks we pre-train a Softmax~\ac{cnn} and an EDL-GEN neural network and when used in combination with a symbolic learning system in~\ac{nsl}, we refer to these as \textit{\ac{nsl} Softmax} and \textit{\ac{nsl} EDL-GEN} respectively. For the Indoor Scene Classification task we adopt a pre-trained network, called \textit{Semantic Aware Scene Recognition (SASR)}, tailored to the task of scene classification. We use a~\ac{rf} and neural network as baseline rule learning approaches in all tasks, and they both use the same pre-trained Softmax neural network for feature extraction as used in~\ac{nsl} Softmax, and are trained to learn the knowledge needed to predict the downstream label, given Softmax neural network predictions. In the Indoor Scene Classification task the SASR network is used. The \ac{rf} is chosen as a powerful decision tree approach, known for being a lightweight model that is quick to train and exhibits a certain level of interpretability. In the Follow Suit Winner and Crop Yield Prediction tasks, the neural network is a~\ac{fcn}, chosen to evaluate a deeper architecture, and in the Sudoku Grid Validity task, the neural network is a~\ac{cnnlstm} designed for sequence classification problems where the~\ac{cnn} component can learn spatial dependencies in the Sudoku grid. Full details of the baseline architectures are given in Appendix~\ref{sec:app:networks}. To measure interpretability of the \ac{rf} baseline, we used the first tree in the forest and extract a rule from each branch (from root to leaf) of this tree. For the neural network baselines, we fitted a surrogate decision tree model~\cite{molnar2020interpretable} to approximate black-box predictions, and applied the same rule extraction methods as that used for the \ac{rf}. Let us now present our results.



\section{Follow Suit Winner}\label{sec:eval_fs_winner}
\begin{figure}[!htb]
\centering
\begin{subfigure}{.16\columnwidth}
  \centering
  \includegraphics[width=0.85\linewidth]{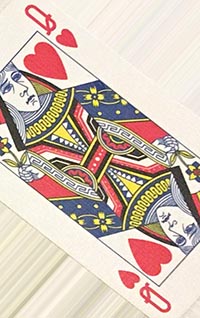}
  \caption{}
  \label{fig:standard_card}
\end{subfigure}%
\begin{subfigure}{.16\columnwidth}
  \centering
  \includegraphics[width=0.95\linewidth]{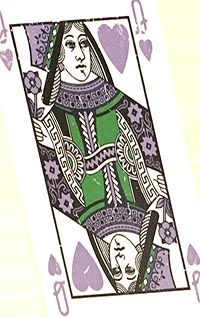}
  \caption{}
  \label{fig:bj_card}
\end{subfigure}
\begin{subfigure}{.16\columnwidth}
  \centering
  \includegraphics[width=0.95\linewidth]{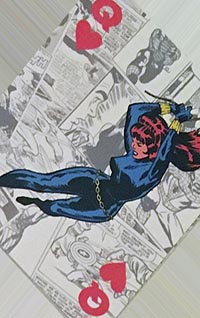}
  \caption{}
  \label{fig:ca_card}
\end{subfigure}
\hfill
\begin{subfigure}{.16\columnwidth}
  \centering
  \includegraphics[width=0.95\linewidth]{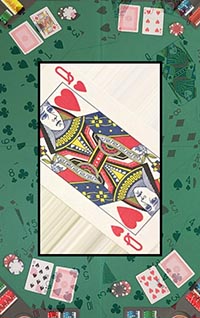}
  \caption{}
  \label{fig:advs_card}
\end{subfigure}
\begin{subfigure}{.16\columnwidth}
  \centering
  \includegraphics[width=0.95\linewidth]{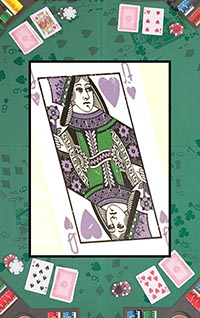}
  \caption{}
  \label{fig:advbj_card}
\end{subfigure}
\begin{subfigure}{.16\columnwidth}
  \centering
  \includegraphics[width=0.95\linewidth]{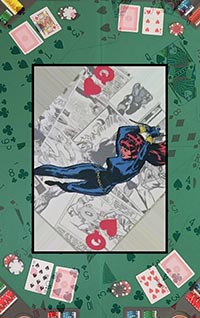}
  \caption{}
  \label{fig:advca_card}
\end{subfigure}
\caption{Example playing card images}
\label{fig:cards}
\end{figure}

In this section we present the results of the Follow Suit Winner task. We start with Softmax and EDL-GEN neural networks pre-trained on standard playing card images and apply \textit{minor} and \textit{major} distributional shifts by substituting standard playing card images with images from alternative decks. Example images are shown in Figure~\ref{fig:cards} for the queen of hearts card taken from the \textit{Standard} (\ref{fig:standard_card}), \textit{Batman Joker} (\ref{fig:bj_card}), \textit{Captain America} (\ref{fig:ca_card}), \textit{Adversarial Standard} (\ref{fig:advs_card}), \textit{Adversarial Batman Joker} (\ref{fig:advbj_card}) and \textit{Adversarial Captain America} (\ref{fig:advca_card}) decks. 

\begin{figure}[H]
\centering
\includegraphics[width=.7\columnwidth]{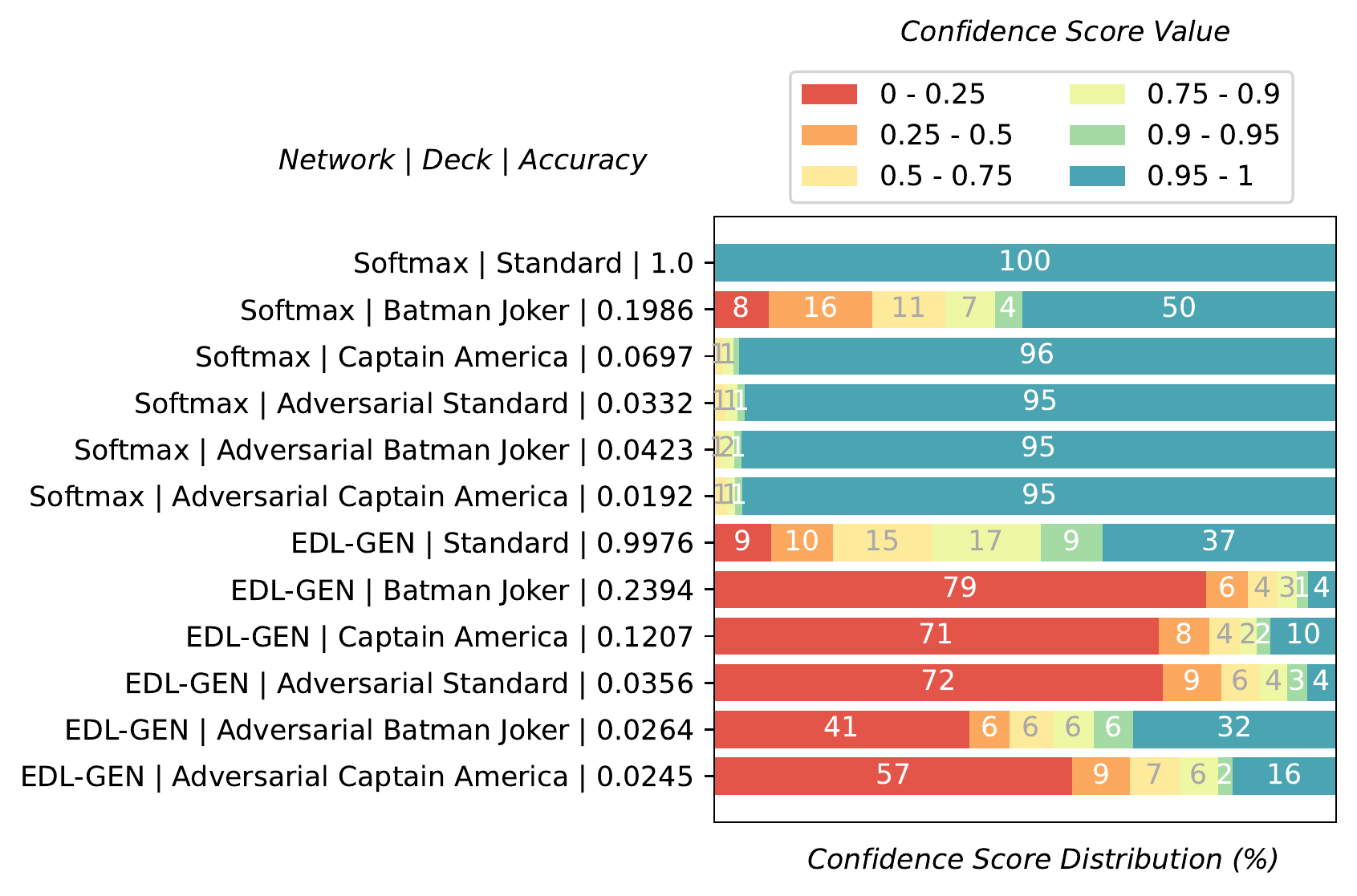}
\caption{Neural network performance under distributional shifts}
\label{fig:fs_nn_acc_conf}
\end{figure}

The Batman Joker and Captain America decks represent \textit{minor} distributional shifts; the adversarial decks represent instead \textit{major} distributional shifts where card images from each of the Standard, Batman Joker and Captain America decks are placed against a background containing additional card images from the Standard deck. These adversarial decks are designed to trick the neural networks into predicting incorrectly as images from the Standard deck are from the same distribution as the card images used during neural network pre-training. In order to understand the challenge faced by the~\ac{las} system when learning from neural network feature predictions in the presence of distributional shifts, Figure~\ref{fig:fs_nn_acc_conf} presents the accuracy and confidence score distribution of pre-trained neural networks when evaluated on different playing card decks than the one used for pre-training.

In Figure~\ref{fig:fs_nn_acc_conf}, each row shows the type of neural network, the playing card deck used for evaluation, the predictive accuracy, and the confidence score distribution. As one would expect, the accuracy was very high when classifying playing card images from the standard deck, as this was the deck used for pre-training. For the Softmax neural network the confidence score was also very high in this case, whereas EDL-GEN had more distributed confidence scores. When evaluating the pre-trained networks on decks different from the one used in training, the Softmax neural network still reported high confidence despite its overall low accuracy, whereas the EDL-GEN network reported comparable low accuracy but with much lower confidence. For example, evaluating the networks over the \textit{Captain America} deck (see 3rd and 9th rows), 96\% of Softmax predictions were made with confidence in the interval $[0.95,1]$, despite an accuracy of 0.0697, whereas only 10\% of EDL-GEN predictions were made within this same confidence interval. As for the overall accuracy, EDL-GEN performed slightly better than Softmax over decks representing minor distributional shifts, whereas both networks performed in a similar way when applied to decks representing major distributional shifts. This highlights the challenge for our \ac{nsl} framework in learning knowledge when presented with out-of-distribution data, as neural network predictions are likely to be incorrect, and may potentially be made with high confidence.

\subsection{Learned hypothesis evaluation}
\begin{figure}[H]
\centering
\begin{subfigure}{1\columnwidth}
  \centering
  \includegraphics[width=.75\linewidth]{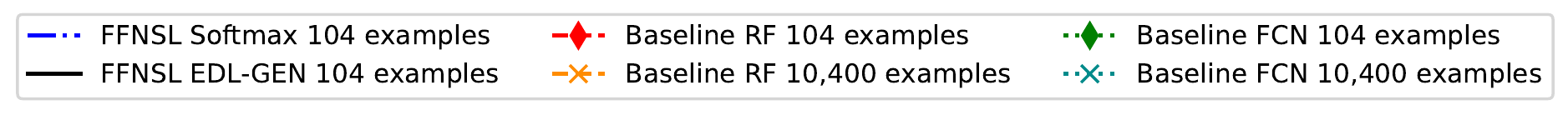}
  \label{fig:fs_legend}
\end{subfigure}
\begin{subfigure}{.3\columnwidth}
  \centering
  \includegraphics[width=1\linewidth]{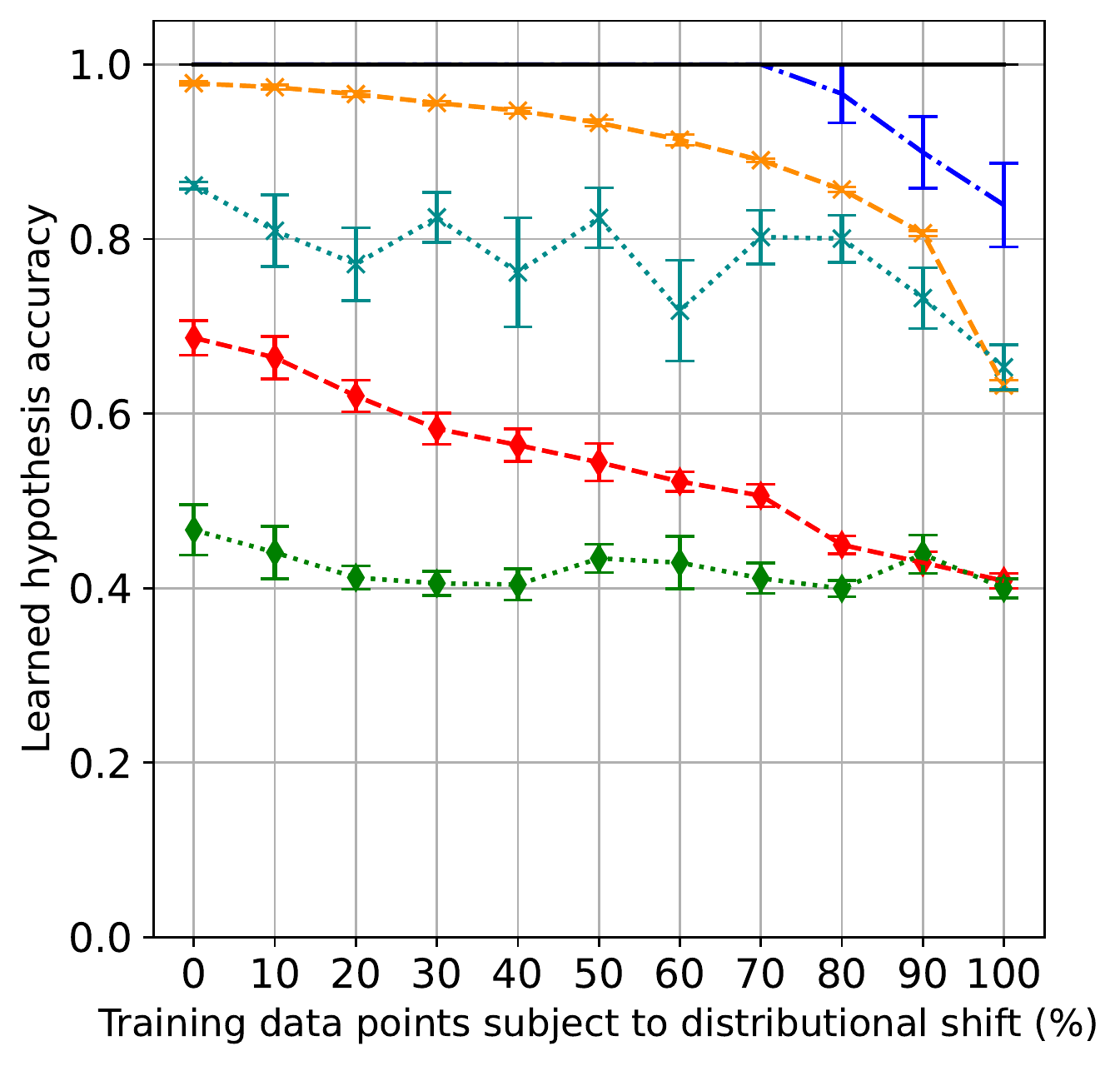}
  \caption{\centering Batman Joker}
  \label{fig:fs_bj_results_acc}
\end{subfigure}
\begin{subfigure}{.3\columnwidth}
  \centering
  \includegraphics[width=1\linewidth]{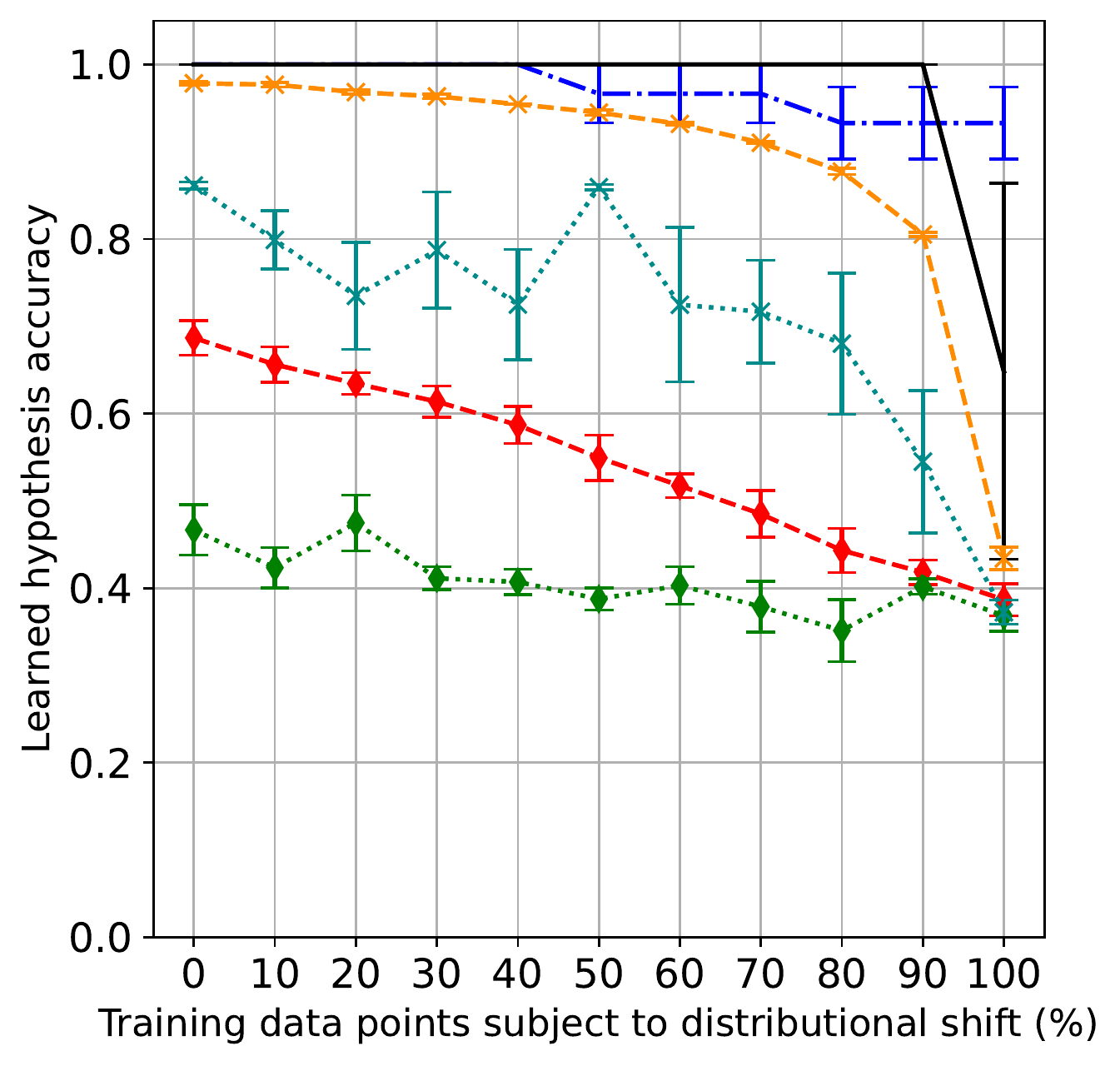}
  \caption{\centering Captain America}
  \label{fig:fs_ca_results_acc}
\end{subfigure}

\begin{subfigure}{.3\columnwidth}
  \centering
  \includegraphics[width=1\linewidth]{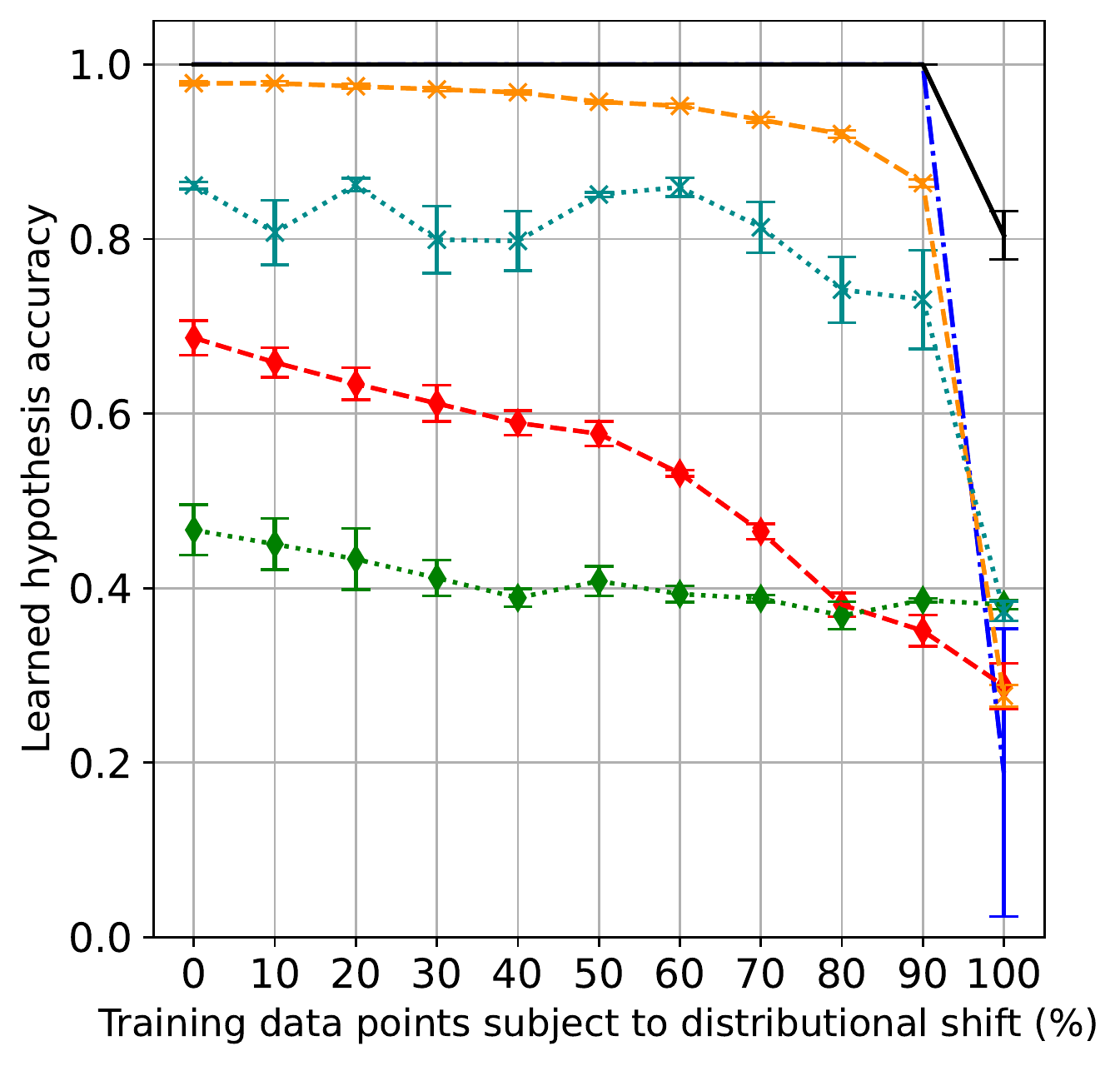}
  \caption{\centering Adversarial Standard}
  \label{fig:fs_as_results_acc}
\end{subfigure}
\begin{subfigure}{.3\columnwidth}
  \centering
  \includegraphics[width=1\linewidth]{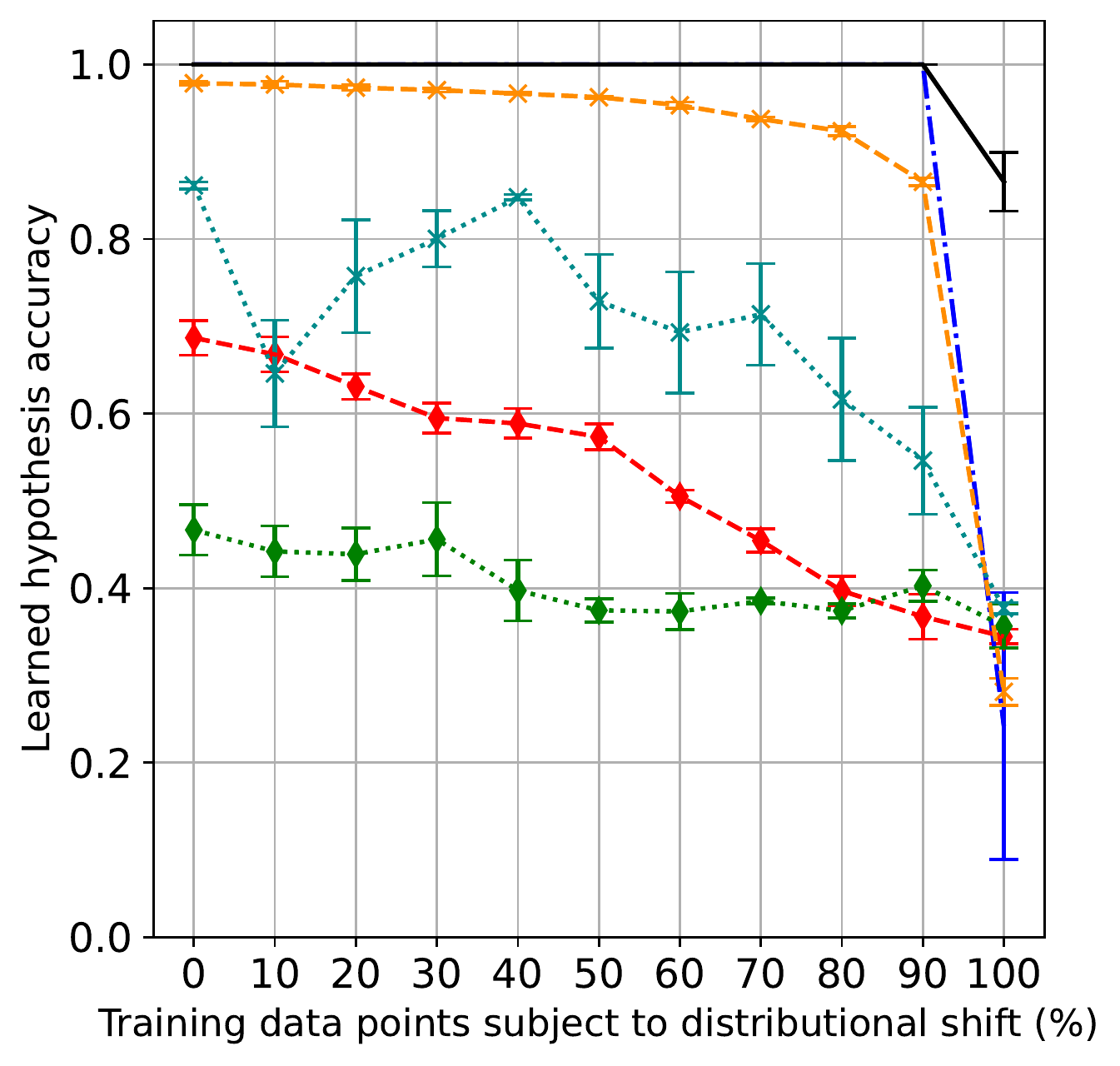}
  \caption{\centering Adversarial Batman Joker}
  \label{fig:fs_abj_results_acc}
\end{subfigure}
\begin{subfigure}{.3\columnwidth}
  \centering
  \includegraphics[width=1\linewidth]{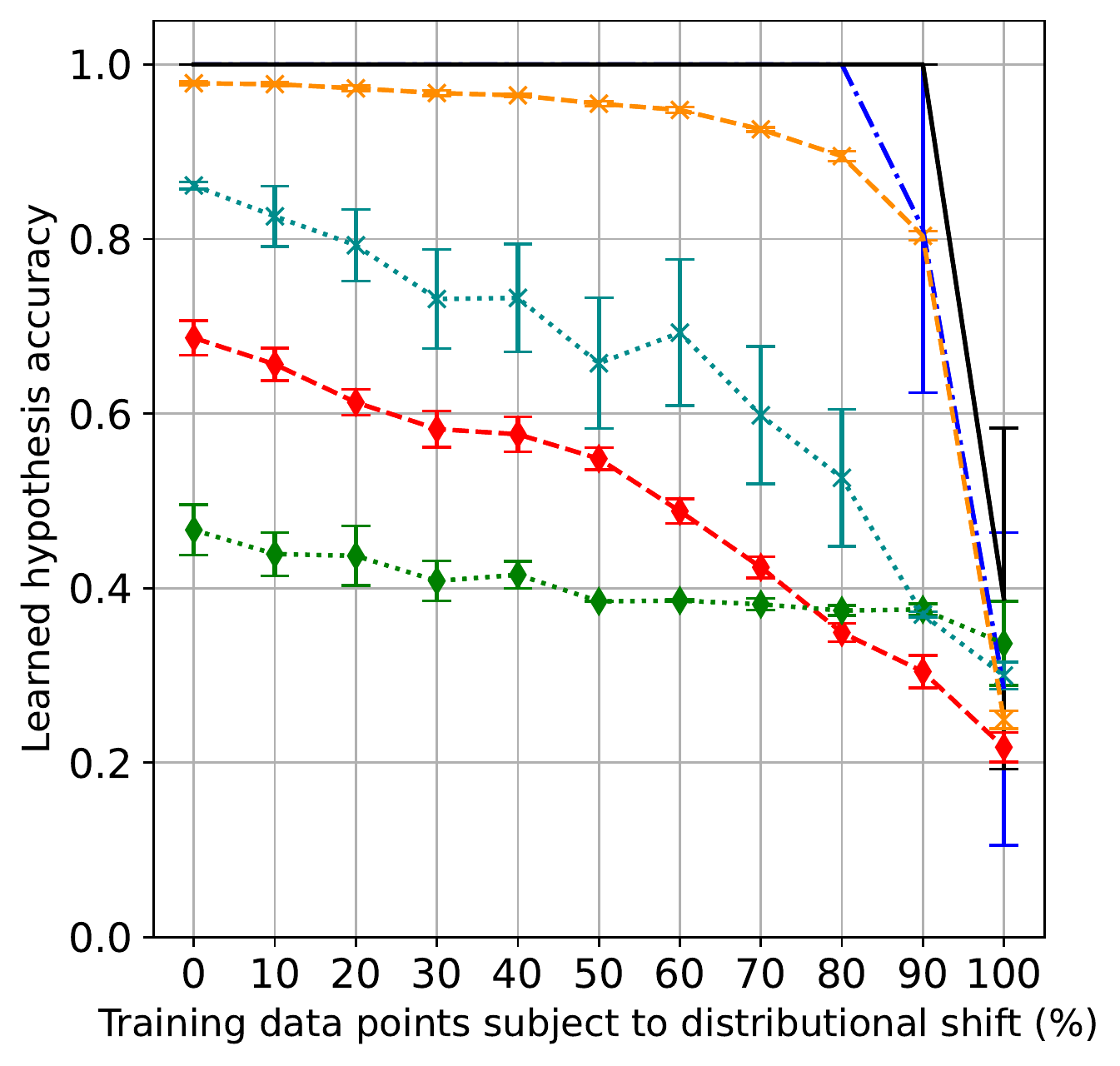}
  \caption{\centering Adversarial Captain America}
  \label{fig:fs_aca_results_acc}
\end{subfigure}
\caption{Accuracy of learned hypotheses with increasing percentages of data subject to distributional shifts, Follow Suit Winner task. 5 repeats.}
\label{fig:fs_winner_acc_results}
\end{figure}

Figure~\ref{fig:fs_winner_acc_results} presents the accuracy of the learned hypotheses when an increasing percentage of labelled unstructured data were subject to distributional shifts, applied with cards from the alternative decks. The reported accuracy is the mean accuracy over 5 repeats and the error bars indicate standard error.

\ac{nsl} outperformed the baselines and learned far superior hypotheses when up to 90\% of labelled unstructured data were subject to distributional shifts. This was the case for both instances of \ac{nsl}. The baselines required $100\times$ the number of examples in order to perform close to~\ac{nsl}, and despite the significant increase in the amount of data used by the baselines,~\ac{nsl} still learned more accurate hypotheses. Figures~\ref{fig:fs_bj_results_acc} and~\ref{fig:fs_ca_results_acc} refer to the injection of minor distributional shifts. In these two cases, when the percentage of distributional shift was very high (above $90\%$), the accuracy of the \ac{nsl} learned hypotheses decreased, but still remained between $\sim70-100\%$, whereas the accuracy of the baselines trained with the same amount of data reduced to $\sim40\%$. Figures~\ref{fig:fs_as_results_acc},~\ref{fig:fs_abj_results_acc} and~\ref{fig:fs_aca_results_acc} refer to the injection of major distribution shifts. \ac{nsl} Softmax had similar performance in Figures~\ref{fig:fs_as_results_acc} and \ref{fig:fs_abj_results_acc} but a much lower accuracy than that shown with minor distributional shifts when $90\%$ or more of the unstructured data were subject to distributional shifts. The \ac{nsl} EDL-GEN maintained instead a higher accuracy in these cases. 
We now perform a more in-depth analysis to explore the reasons for dropping accuracy in the presence of high percentages of distributional shifts. Given the two groups of similar behaviours we consider only two representative cases: \textit{Batman Joker}, as minor distributional shift, and \textit{Adversarial Batman Joker} as major distributional shift. A full set of analysis results, with respect to all the other card decks, is given in Appendix~\ref{sec:app:additional_fs_results}. 

\begin{figure}[ht]
\centering
\begin{subfigure}{1\columnwidth}
  \centering
  \includegraphics[width=.75\linewidth]{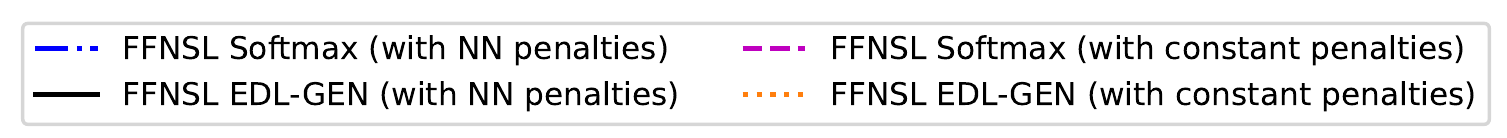}
  \label{fig:fs_50_repeats_legend}
\end{subfigure}
\begin{subfigure}{.3\columnwidth}
  \centering
  \includegraphics[width=1\linewidth]{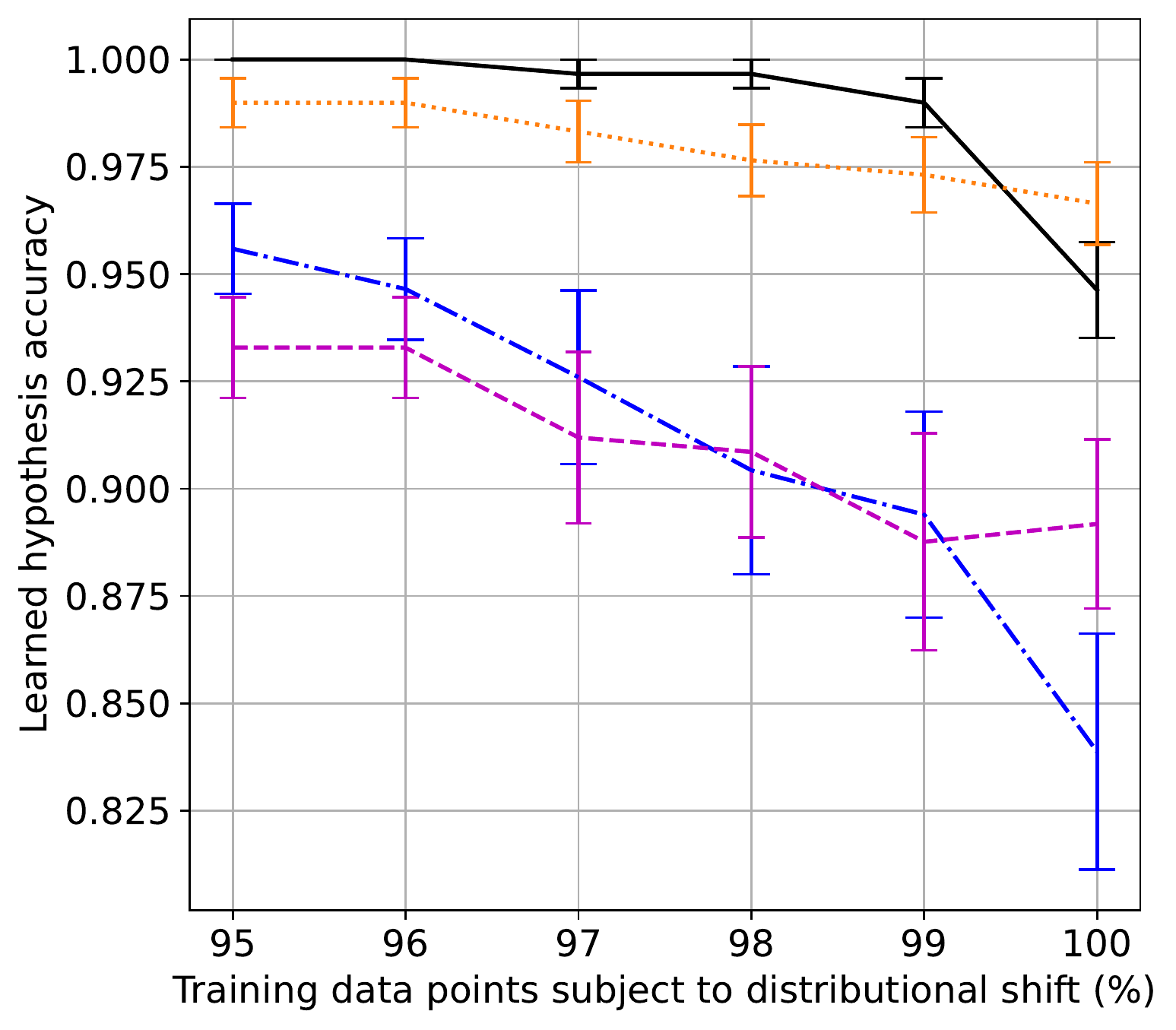}
  \caption{\centering Batman Joker}
  \label{fig:fs_bj_95_100_acc}
\end{subfigure}%
\begin{subfigure}{.27\columnwidth}
  \centering
  \includegraphics[width=1\linewidth]{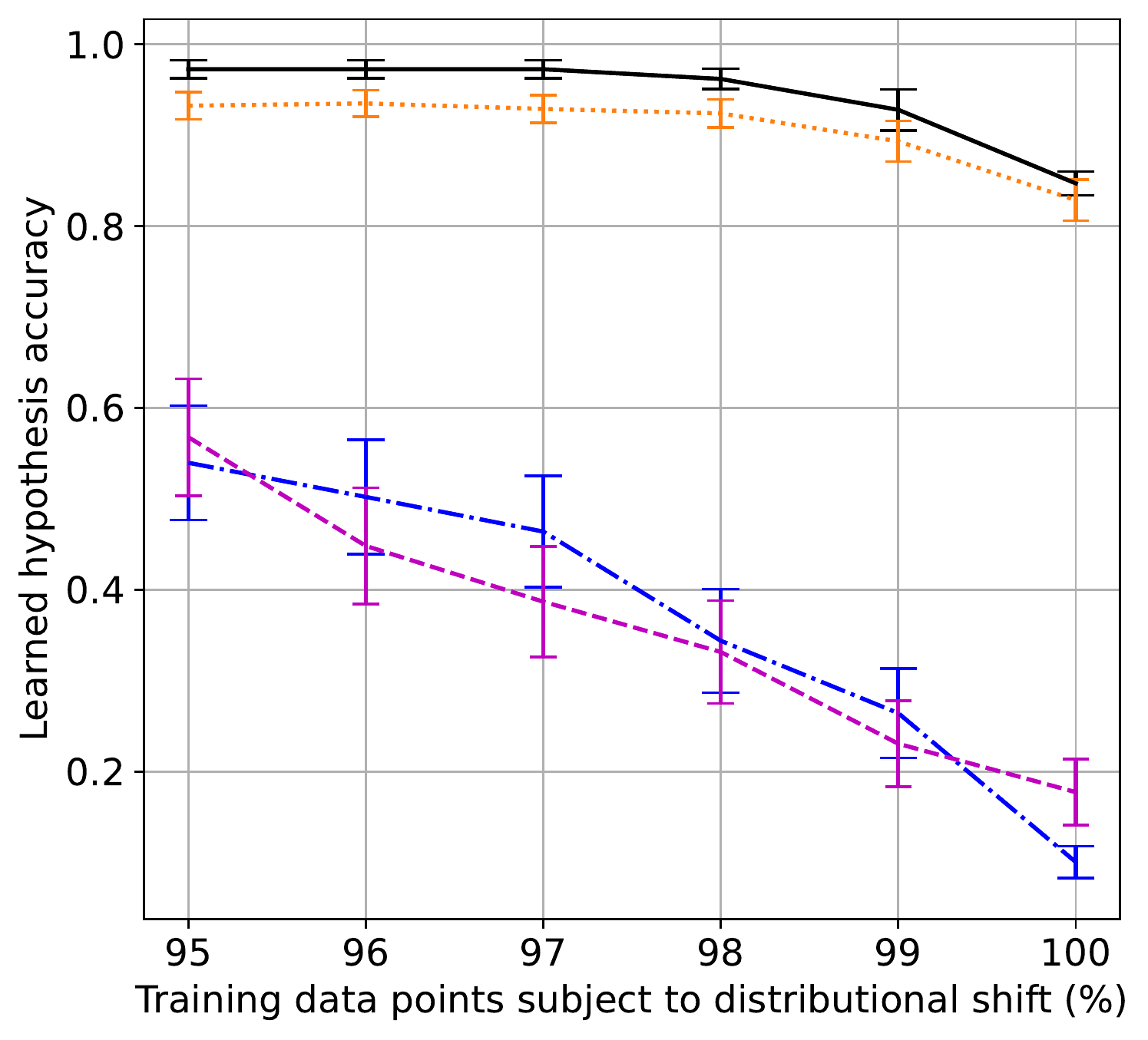}
  \caption{\centering Adversarial Batman Joker}
  \label{fig:fs_abj_95_100_acc}
\end{subfigure}
\begin{subfigure}{.27\columnwidth}
  \centering
  \includegraphics[width=1\linewidth]{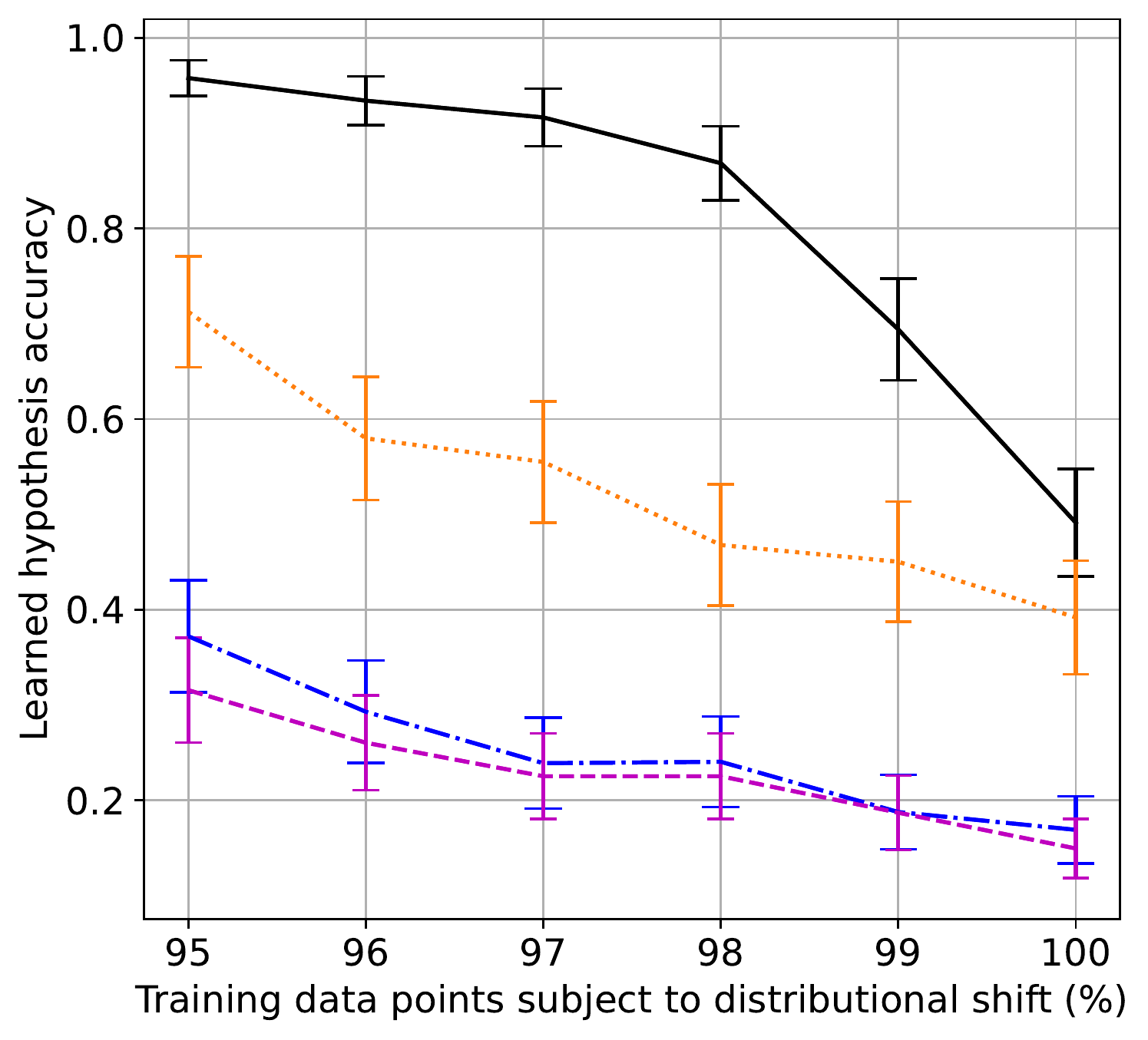}
  \caption{\centering Adversarial Captain America}
  \label{fig:fs_aca_95_100_acc_1}
\end{subfigure}
\caption{\ac{nsl} Softmax vs. \ac{nsl} EDL-GEN. Accuracy of learned hypotheses with 95-100\% distributional shifts using 50 repeats. Follow Suit Winner task. }
\label{fig:fs_95_100_results_bj_abj}
\end{figure}

In particular, we explore whether~\ac{nsl} EDL-GEN provides a performance benefit over~\ac{nsl} Softmax, and if so, what are the contributing factors. Specifically we analyse the accuracy performance in relation to either or both (i) better neural network predictive accuracy, when classifying out-of-distribution data, and (ii) more informative weight penalties of the generated~\ac{wcdpi} examples, calculated from the neural network confidence scores. For this analysis we focus on high percentages of distributional shifts, 95-100\%, as this was when~\ac{nsl} instances deteriorated in their learned hypothesis accuracy. We run 50 experimental repeats to generate statistically significant results. In order to isolate the effect of the example weight penalties, we also run two additional baseline~\ac{nsl} instances where the weight penalties of the generated~\ac{wcdpi} examples are all constant and equal to 10. The results are shown in Figure~\ref{fig:fs_95_100_results_bj_abj}. We have also included the performances with respect to distributional shifts given by the Adversarial Captain America deck (Figure~\ref{fig:fs_aca_95_100_acc_1}), since Figure~\ref{fig:fs_aca_results_acc} shows that in this case the accuracy of both \ac{nsl} instances decreased to around $40\%$ when nearly $100\%$ of the data were subject to distributional shifts. Full analysis of this deck is presented in Appendix~\ref{sec:app:additional_fs_results}.

For both Batman Joker and Adversarial Batman Joker, \ac{nsl} EDL-GEN outperformed~\ac{nsl} Softmax. Note, however, the difference in y-axis scale between Figure~\ref{fig:fs_bj_95_100_acc} and~\ref{fig:fs_abj_95_100_acc} and the difference in~\ac{nsl} performance. This was due to the fact that both Softmax and EDL-GEN neural networks predicted more accurately on the Batman Joker deck than the Adversarial Batman Joker deck, as presented in Figure~\ref{fig:fs_nn_acc_conf}. The improved performances of \ac{nsl} EDL-GEN versus \ac{nsl} Softmax did not seem to depend on the more informative weights of~\ac{wcdpi} examples with weights calculated from neural network confidence scores, versus constant weights, since the accuracy of \ac{nsl} instances (denoted ``...\textit{with NN penalties}") was similar to that of the respective baselines with constant weight penalties. However, Figure~\ref{fig:fs_aca_95_100_acc_1} shows that when the distributional shift was more severe\footnote{Recall that the accuracy of Softmax and EDL-GEN neural networks was the lowest for the Adversarial Captain America test set, as shown in Figure~\ref{fig:fs_nn_acc_conf}.}, the decrease in accuracy of \ac{nsl} EDL-GEN was less drastic than that of its corresponding baseline with constant penalty, whereas there was no difference in the case of \ac{nsl} Softmax. Even though the overall accuracy of the framework was lower than that reported for less drastic forms of distributional shifts, the more informative weight penalties of~\ac{wcdpi} examples, calculated from the EDL-GEN neural network confidence scores, provided a clear benefit compared to using constant weights, in particular when the percentage of distributional shift was very high. 

It still remains open the question as to why \ac{nsl} EDL-GEN performed better than~\ac{nsl} Softmax in Figures~\ref{fig:fs_bj_95_100_acc} and \ref{fig:fs_abj_95_100_acc}. For the percentages of distributional shifts between $95$-$100\%$, the pre-trained neural networks both reported low average accuracy. So a natural question to ask is whether EDL-GEN led to more consistent symbolic feature predictions than Softmax. This is important to investigate because~\ac{las} systems are capable of learning accurate hypotheses from few ``good" examples. So, we investigated the percentage of incorrect~\ac{wcdpi} examples generated when $95$-$100\%$ of the unstructured data were subject to distributional shifts. These were examples whose contextual symbolic features were inconsistent with the ground-truth label due to incorrect neural network predictions. 

\begin{figure}[H]
\centering

\begin{subfigure}{.4\columnwidth}
  \centering
  \includegraphics[width=1\linewidth]{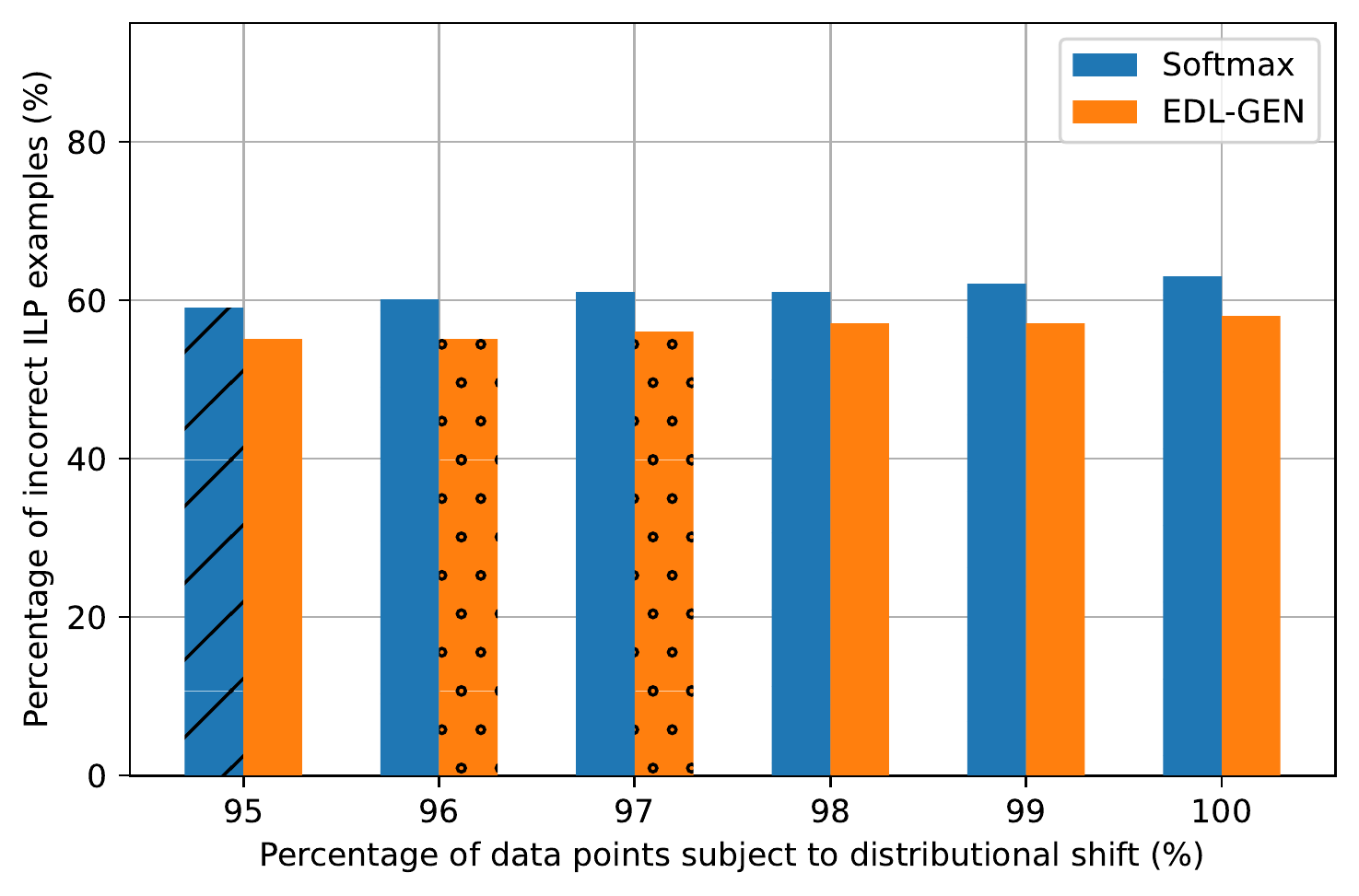}
  \caption{\centering Batman Joker}
  \label{fig:fs_bj_ilp_ex}
\end{subfigure}%
\begin{subfigure}{.37\columnwidth}
  \centering
  \includegraphics[width=1\linewidth]{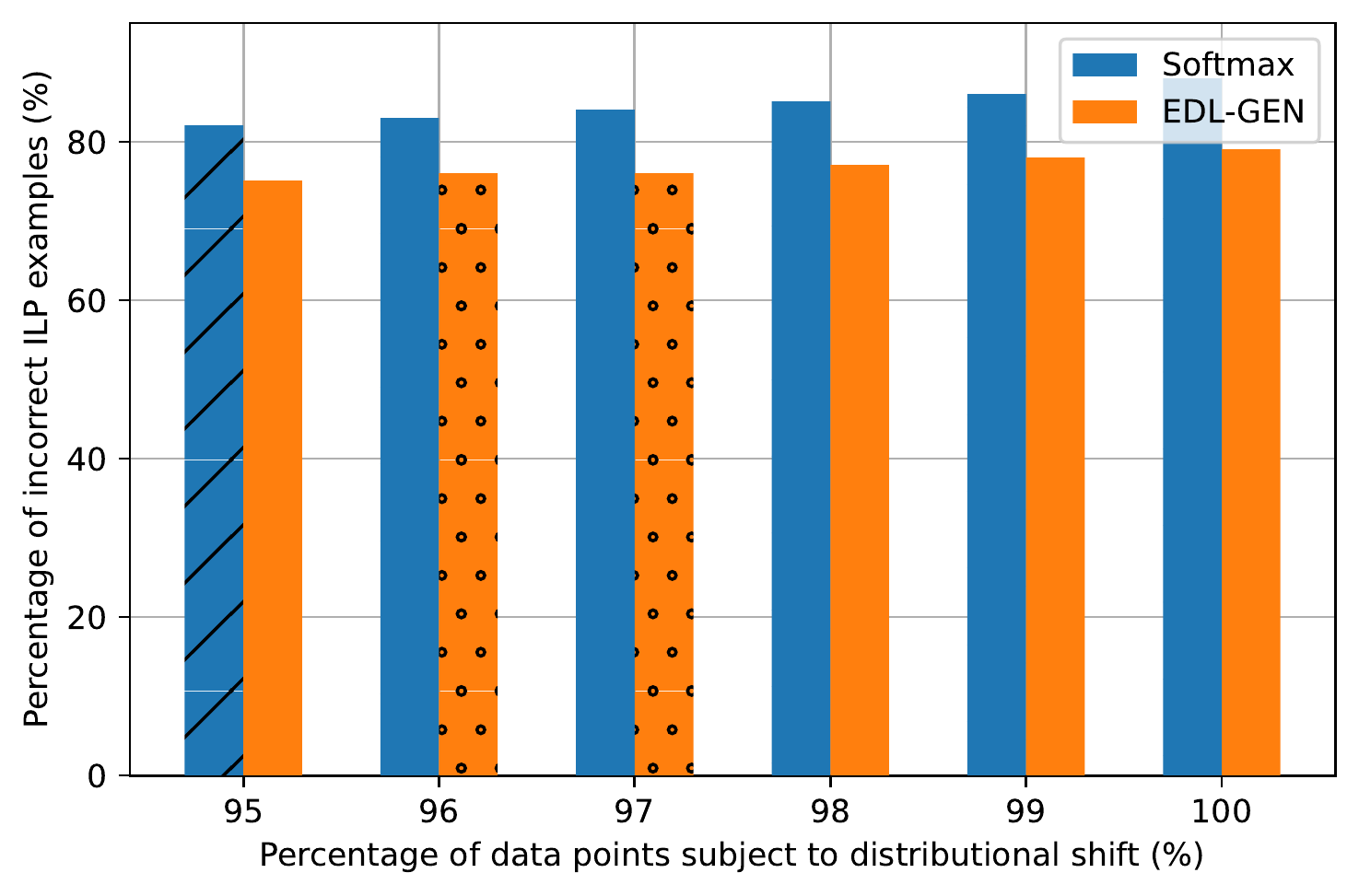}
  \caption{\centering Adversarial Batman Joker}
  \label{fig:fs_abj_ilp_ex}
\end{subfigure}
\caption{The effect of distributional shifts on percentage of incorrect~\ac{wcdpi} examples generated. Follow Suit Winner task.}
\label{fig:fs_ilp_ex_95_100}
\end{figure}

Figure~\ref{fig:fs_ilp_ex_95_100} shows, first of all, that the percentage of generated incorrect~\ac{wcdpi} examples was lower than the corresponding percentage of data subject to distributional shifts. This indicated that some correct~\ac{wcdpi} examples could be generated even when the neural networks made incorrect predictions. Incorrect predictions made over the 4 cards played could in combination lead to predicted symbolic features for the trick whose winning player would match the ground truth label. Secondly, more correct~\ac{wcdpi} examples were generated when the distributional shift was given by the Batman Joker deck, compared to that of the Adversarial Batman Joker deck. This was because, as indicated in Figure~\ref{fig:fs_nn_acc_conf}, the neural network accuracy for the former was better than that for the latter. Furthermore, EDL-GEN led to a lower number of incorrect~\ac{wcdpi} examples compared to that of Softmax in both forms of distributional shifts, and this difference was bigger in the case of the Adversarial Batman Joker deck. Given the relatively small number of~\ac{wcdpi} examples used by the~\ac{las} system (104), this difference contributed to the larger gap in accuracy between~\ac{nsl} Softmax and~\ac{nsl} EDL-GEN in Figure~\ref{fig:fs_abj_95_100_acc} than Figure~\ref{fig:fs_bj_95_100_acc}. 

\begin{figure}[ht]
\centering
\begin{subfigure}{1\columnwidth}
  \centering
  \includegraphics[width=.75\linewidth]{Figures/follow_suit_results/follow_suit_nn_penalties_legend.pdf}
  \label{fig:fs_50_repeats_legend_wpr}
\end{subfigure}
\begin{subfigure}{.4\columnwidth}
  \centering
  \includegraphics[width=1\linewidth]{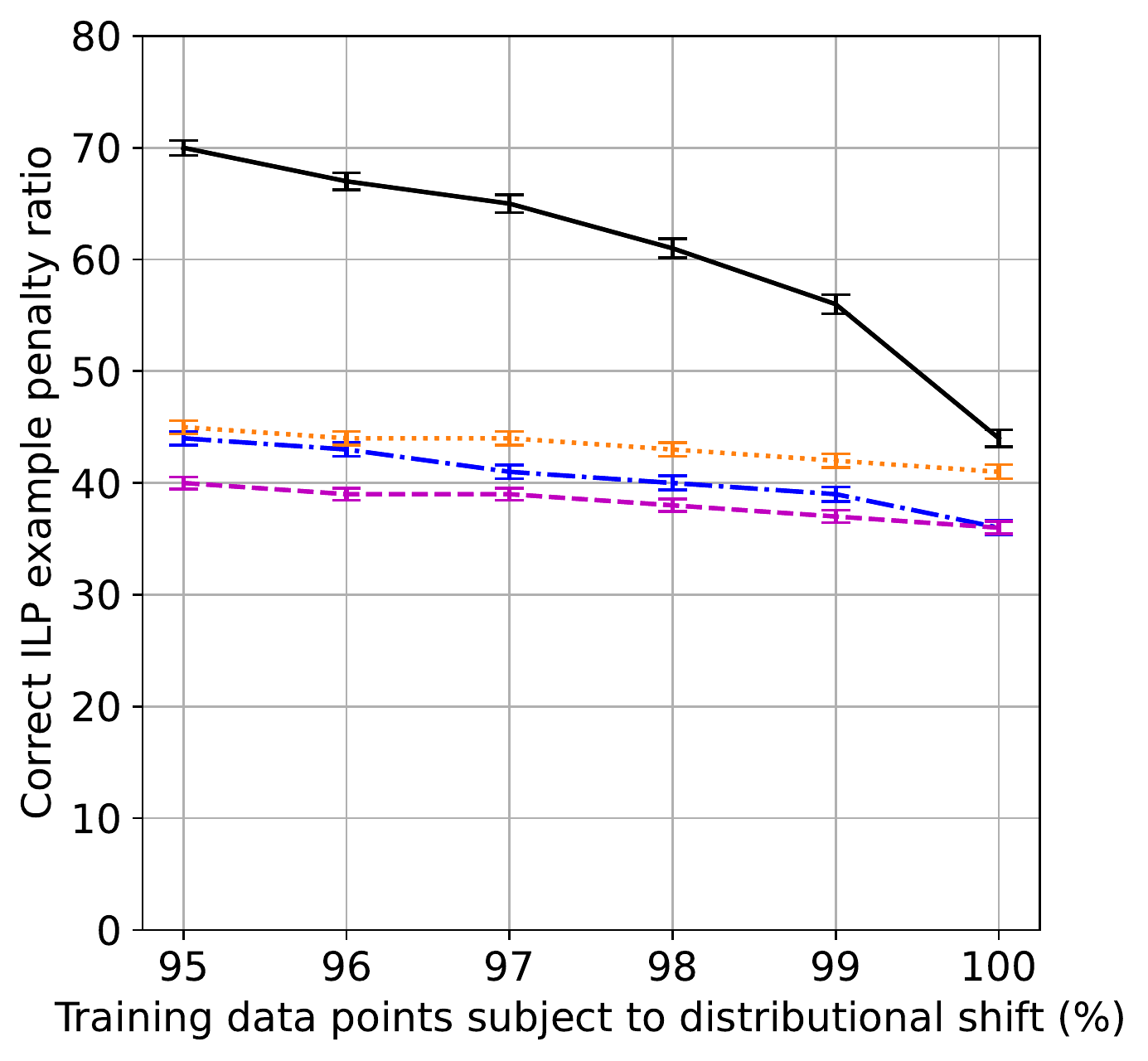}
  \caption{\centering Batman Joker}
  \label{fig:fs_bj_95_100_wpr}
\end{subfigure}
\begin{subfigure}{.4\columnwidth}
  \centering
  \includegraphics[width=1\linewidth]{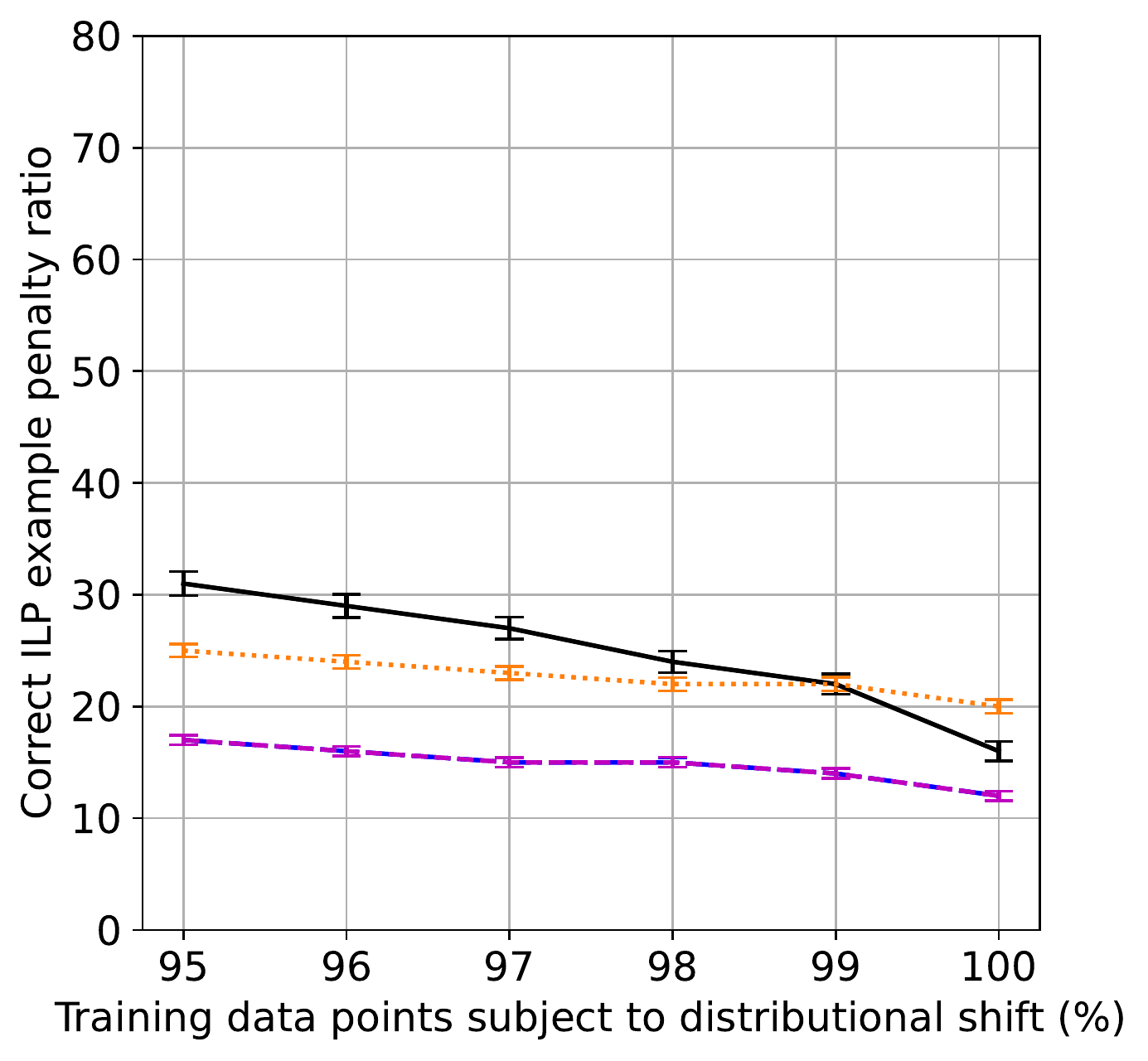}
  \caption{\centering Adversarial Batman Joker}
  \label{fig:fs_abj_95_100_wpr}
\end{subfigure}
\caption{\ac{wcdpi} example weight penalty ratio. Follow Suit Winner task.}
\label{fig:fs_nn_wpr}
\end{figure}

Now, how did the weight penalty, generated from the neural network confidence score, effect the accuracy of learned hypotheses? 
Clearly, EDL-GEN provided improved confidence scores than the Softmax neural network which in-turn, improved the accuracy of \ac{nsl}. 
This explains why the accuracy of each~\ac{nsl} approach was higher in Figure~\ref{fig:fs_bj_95_100_acc} than that shown in Figure~\ref{fig:fs_abj_95_100_acc}. However, Figure~\ref{fig:fs_95_100_results_bj_abj} shows that for \ac{nsl} Softmax, using~\ac{wcdpi} example weight penalties calculated from Softmax neural network confidence scores appears to have no benefit compared to using~\ac{wcdpi} examples with constant weight penalties. However, this was different in the case of \ac{nsl} EDL-GEN. To investigate this further, we calculated the weight penalty ratio for the~\ac{wcdpi} examples generated from both Softmax and EDL-GEN neural network confidence scores. The analysis is shown in Figure~\ref{fig:fs_nn_wpr} for each deck and $95$-$100\%$ distributional shifts.

\noindent
Figure~\ref{fig:fs_nn_wpr}
shows that the weight penalty ratio calculated from EDL-GEN confidence scores provided a clear benefit than that calculated from the Softmax neural network confidence scores, which was instead very similar to the weight penalty ratio given by constant penalties. At 100\% distributional shifts, the benefits of calculating~\ac{wcdpi} example weight penalties with the neural network confidence scores reduced, as there were very few correct examples. This explains why the gap between the accuracy of the \ac{nsl} EDL-GEN with neural network penalties and that of \ac{nsl} EDL-GEN with constant penalties, in both decks, reduced as distributional shifts increase towards 100\% (see Figure~\ref{fig:fs_95_100_results_bj_abj}). 
In summary, improved accuracy of the neural network predictions led to a higher percentage (even if small) of correct~\ac{wcdpi} examples and improved neural network confidence scores led to an improved penalty ratio of correct~\ac{wcdpi} examples. Together they provided an improved bias for the~\ac{las} system which even if it was reduced to learn from a small percentage of correct examples, these had improved penalty weight to guide the search for optimal solutions. 

\begin{figure}[ht]
\centering
\begin{subfigure}{1\columnwidth}
  \centering
  \includegraphics[width=.75\linewidth]{Figures/follow_suit_results/follow_suit_legend.pdf}
  \label{fig:fs_interp_legend}
\end{subfigure}
\begin{subfigure}{.4\columnwidth}
  \centering
  \includegraphics[width=1\linewidth]{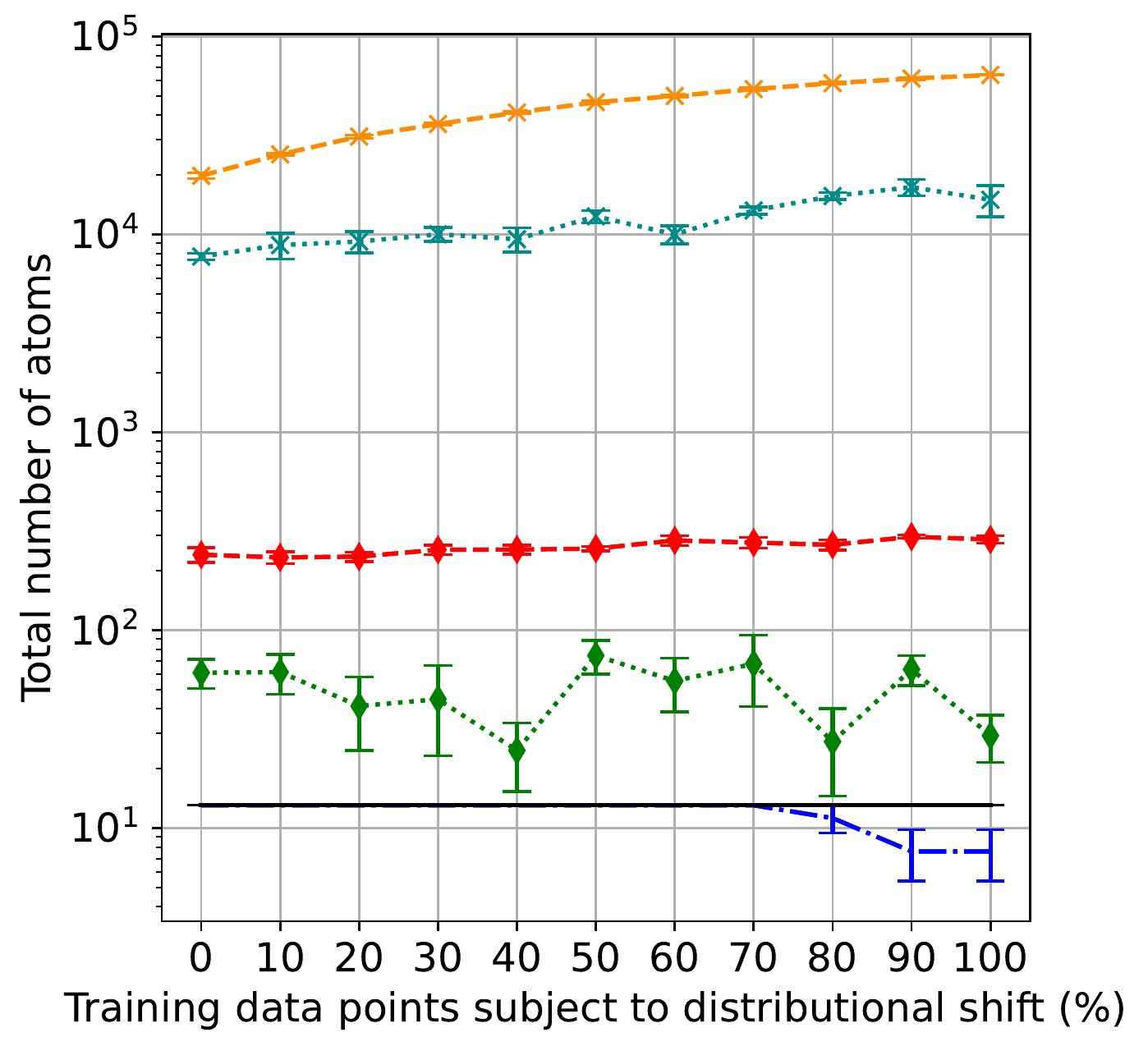}
  \caption{\centering Batman Joker}
  \label{fig:fs_bj_interp}
\end{subfigure}
\begin{subfigure}{.4\columnwidth}
  \centering
  \includegraphics[width=1\linewidth]{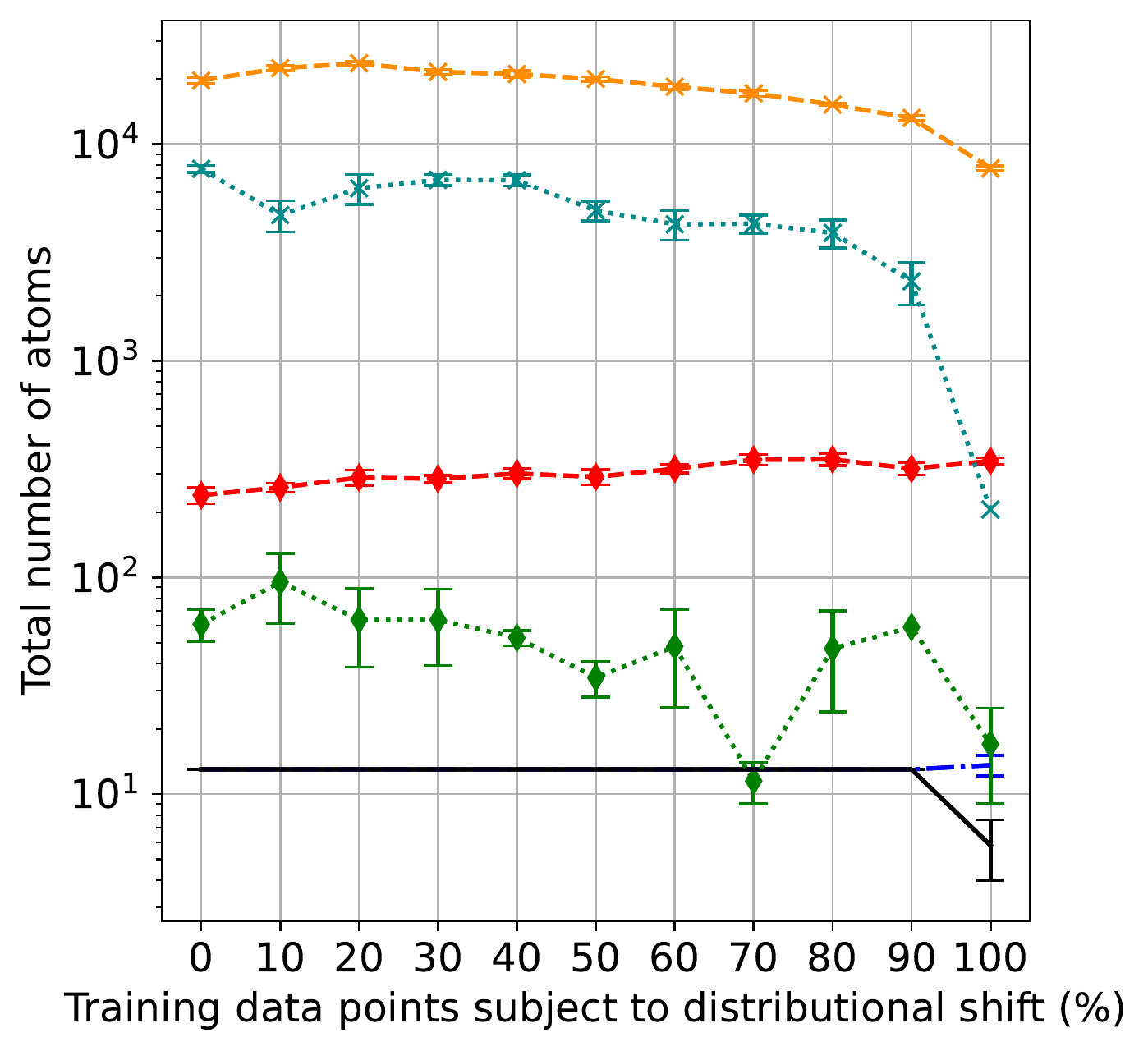}
  \caption{\centering Adversarial Batman Joker}
  \label{fig:fs_aca_interp}
\end{subfigure}
\caption{Interpretability of the learned hypotheses, Follow Suit Winner task.}
\label{fig:fs_interp_results}
\end{figure}

Let us now investigate the interpretability of the hypotheses learned using our \ac{nsl} framework compared to that of the baseline approaches. Figure~\ref{fig:fs_interp_results} shows the results, where interpretability was measured in terms of the number of atoms that formed the learned hypothesis. 

\noindent
\ac{nsl} learned significantly more interpretable knowledge than the baseline approaches (note the logarithmic scale on the y-axis). In the case of the minor form of distributional shift (see Figure~\ref{fig:fs_bj_interp}), the interpretability of the baseline models trained with $100\times$ the amount of data decreased as distributional shift increased. These models reached high accuracy by training over a much larger dataset (see Figure~\ref{fig:fs_bj_results_acc}), but they did so at the cost of much lower interpretability. This was because they learned a more complex mapping between input and output, instead of learning general rules, as was the case for our \ac{nsl} approach. The~\ac{fcn} trained with the same amount of data as \ac{nsl} had similar interpretability to that of~\ac{nsl}, because the model learned to largely predict the same class and the surrogate decision tree was very small. This was reflected in the poor performance of the \ac{fcn} shown in Figure~\ref{fig:fs_abj_results_acc} for the Adversarial Batman Joker deck. Examples of interpretable knowledge learned by our \ac{nsl} approaches are presented in Appendix~\ref{sec:app:learned_hyps}. 

\begin{figure}[H]
\centering
\begin{subfigure}{1\columnwidth}
  \centering
  \includegraphics[width=.75\linewidth]{Figures/follow_suit_results/follow_suit_legend.pdf}
  \label{fig:fs_lt_legend}
\end{subfigure}
\begin{subfigure}{.4\columnwidth}
  \centering
  \includegraphics[width=1\linewidth]{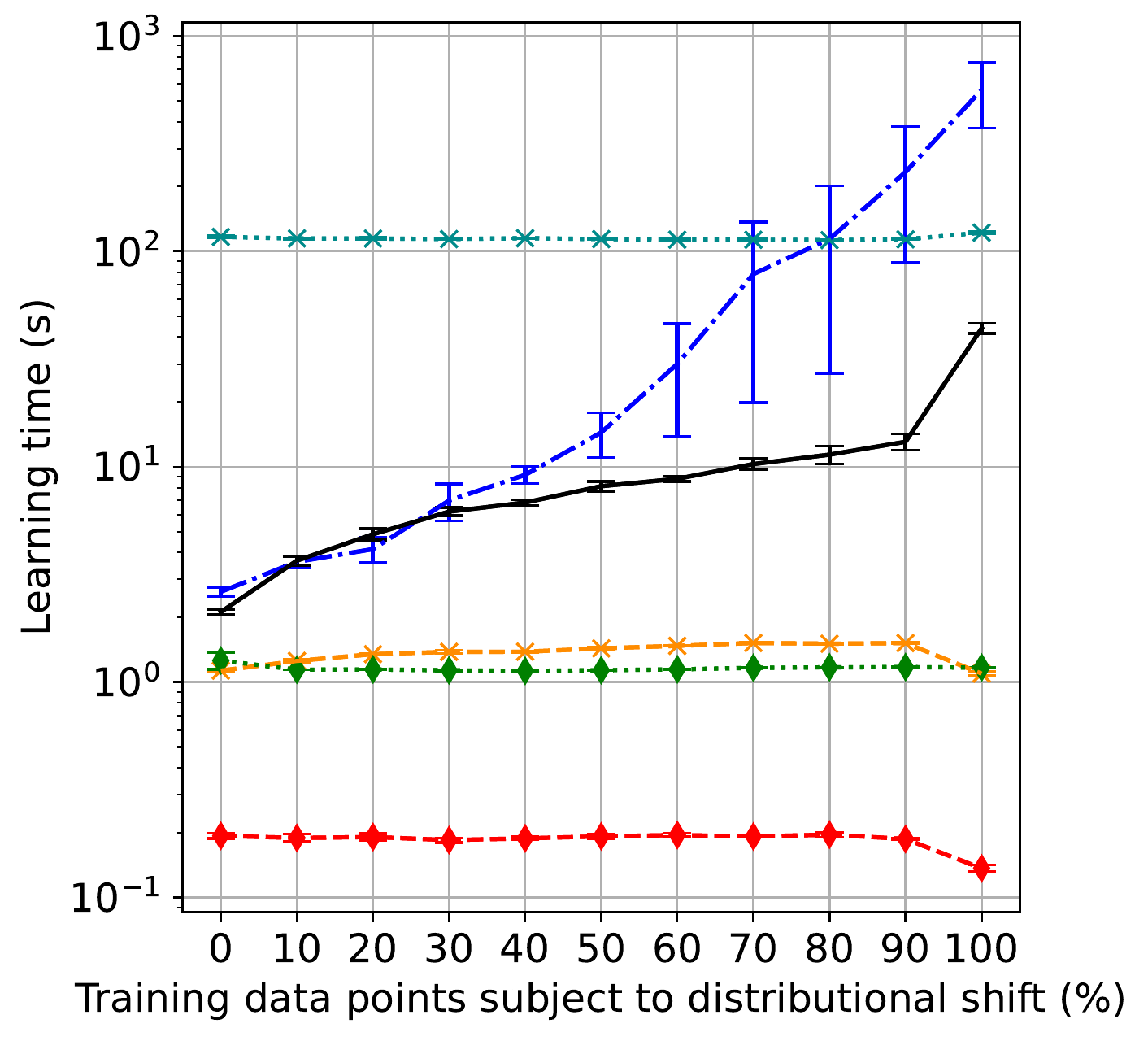}
  \caption{\centering Batman Joker}
  \label{fig:fs_bj_lt}
\end{subfigure}
\begin{subfigure}{.4\columnwidth}
  \centering
  \includegraphics[width=1\linewidth]{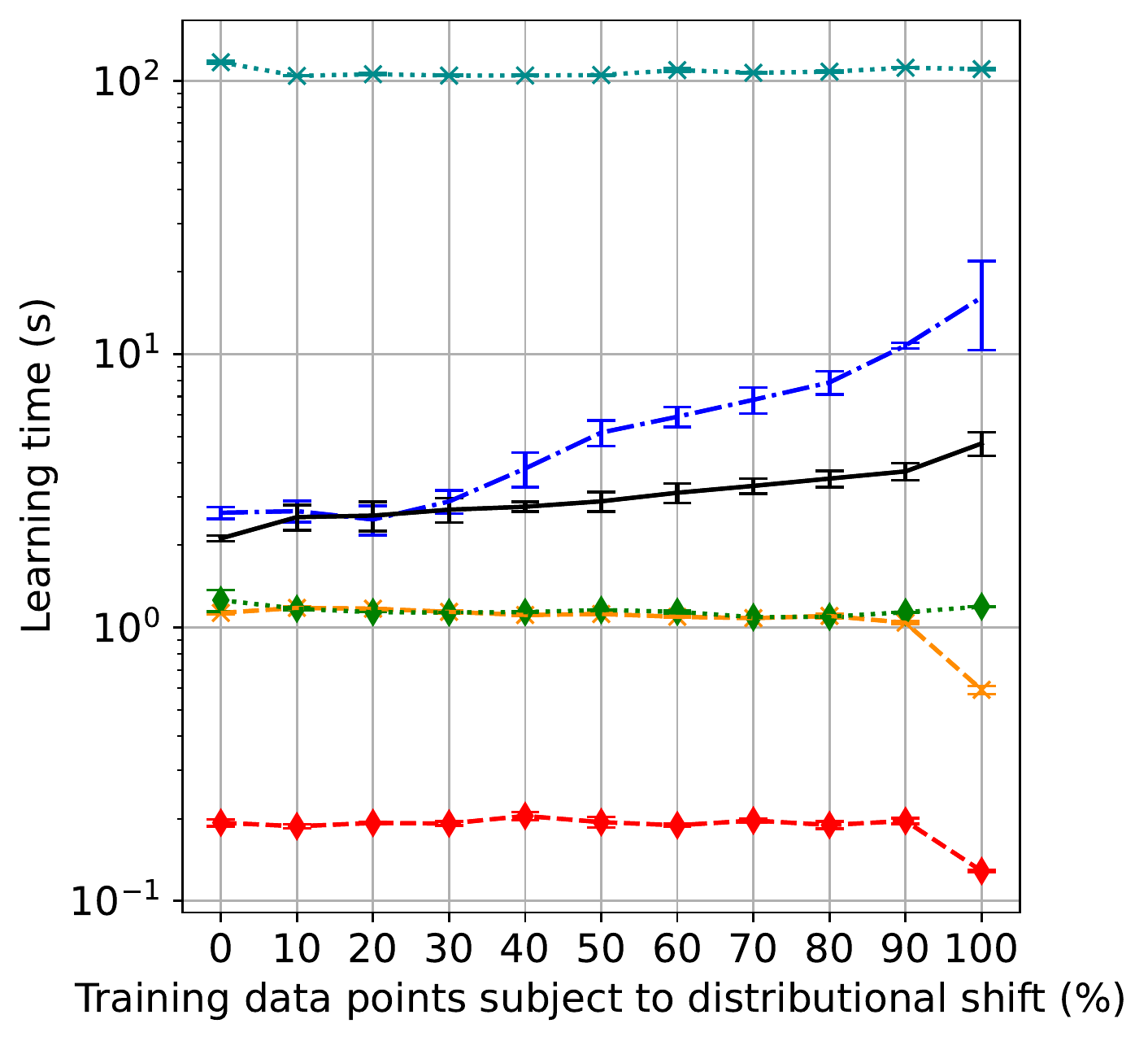}
  \caption{\centering Adversarial Batman Joker}
  \label{fig:fs_aca_lt}
\end{subfigure}
\caption{Learning time. Follow Suit Winner task.}
\label{fig:fs_lt_results}
\end{figure}

Finally, to investigate the scalability of~\ac{nsl}, we have also computed the time required to learn an interpretable hypothesis. The results are shown in Figure~\ref{fig:fs_lt_results}.

\noindent
Both~\ac{nsl} approaches learned with an order of magnitude of time similar to that of the \ac{fcn} trained with the same number of examples when no distributional shifts were applied to the data. As distributional shifts increased,~\ac{nsl} took longer because the ILASP system required more iterations to prove optimality with respect to minimising the total penalty on the examples. However, the learning time of~\ac{nsl} EDL-GEN did not increase as quickly, when compared to~\ac{nsl} Softmax. This was because the~\ac{wcdpi} example weight penalties were much more informative (see Figure~\ref{fig:fs_nn_wpr}) and the ILASP learning system required fewer iterations overall to prove optimality.

In conclusion, our analysis shows that~\ac{nsl} outperformed the baseline approaches in terms of accuracy and interpretability, even when the baselines were trained with $100\times$ the amount of data. \ac{nsl} EDL-GEN outperformed \ac{nsl} Softmax, in the accuracy of the learned hypotheses, as EDL-GEN neural network predictions were more accurate, and this influenced the downstream performance of the \ac{nsl} framework more than the neural network confidence scores. When major distributional shifts were applied, the EDL-GEN uncertainty-aware neural network led to significantly more informative~\ac{wcdpi} example weight penalties compared to the Softmax neural network, although this benefit diminished as the percentage of input data subject to distributional shifts approached 100\%. Finally, we have shown that more informative~\ac{wcdpi} example weight penalties resulted in faster hypothesis learning times, when the iterative ILASP system was used.

\subsection{\ac{nsl} framework evaluation}
\begin{figure}[H]
\centering
\begin{subfigure}{1\columnwidth}
  \centering
  \includegraphics[width=.75\linewidth]{Figures/follow_suit_results/follow_suit_legend.pdf}
  \label{fig:fs_50_repeats_legend_inf_eval}
\end{subfigure}
\begin{subfigure}{.4\columnwidth}
  \centering
  \includegraphics[width=1\linewidth]{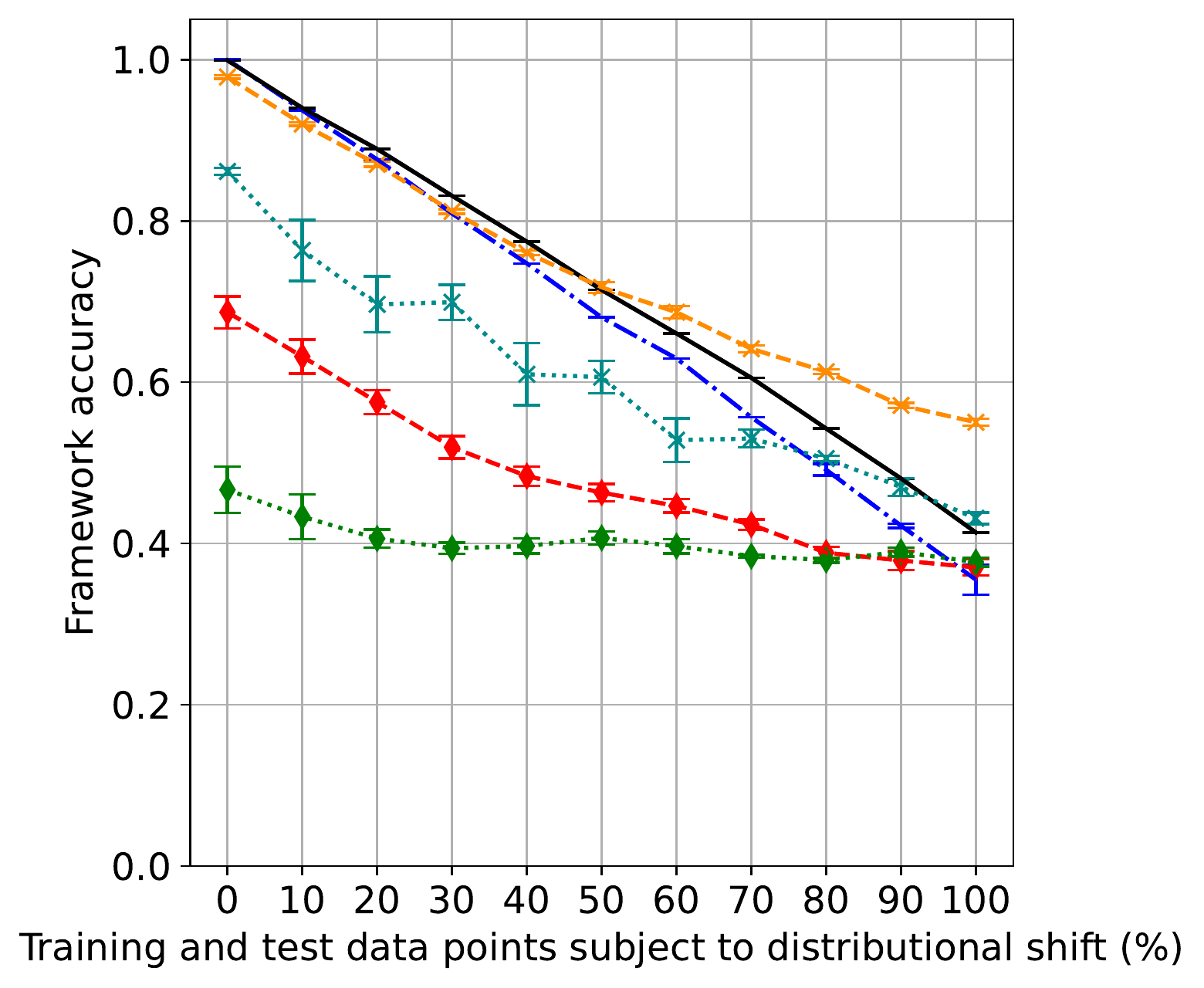}
  \caption{\centering Batman Joker}
  \label{fig:fs_bj_unstruc_acc}
\end{subfigure}%
\begin{subfigure}{.4\columnwidth}
  \centering
  \includegraphics[width=1\linewidth]{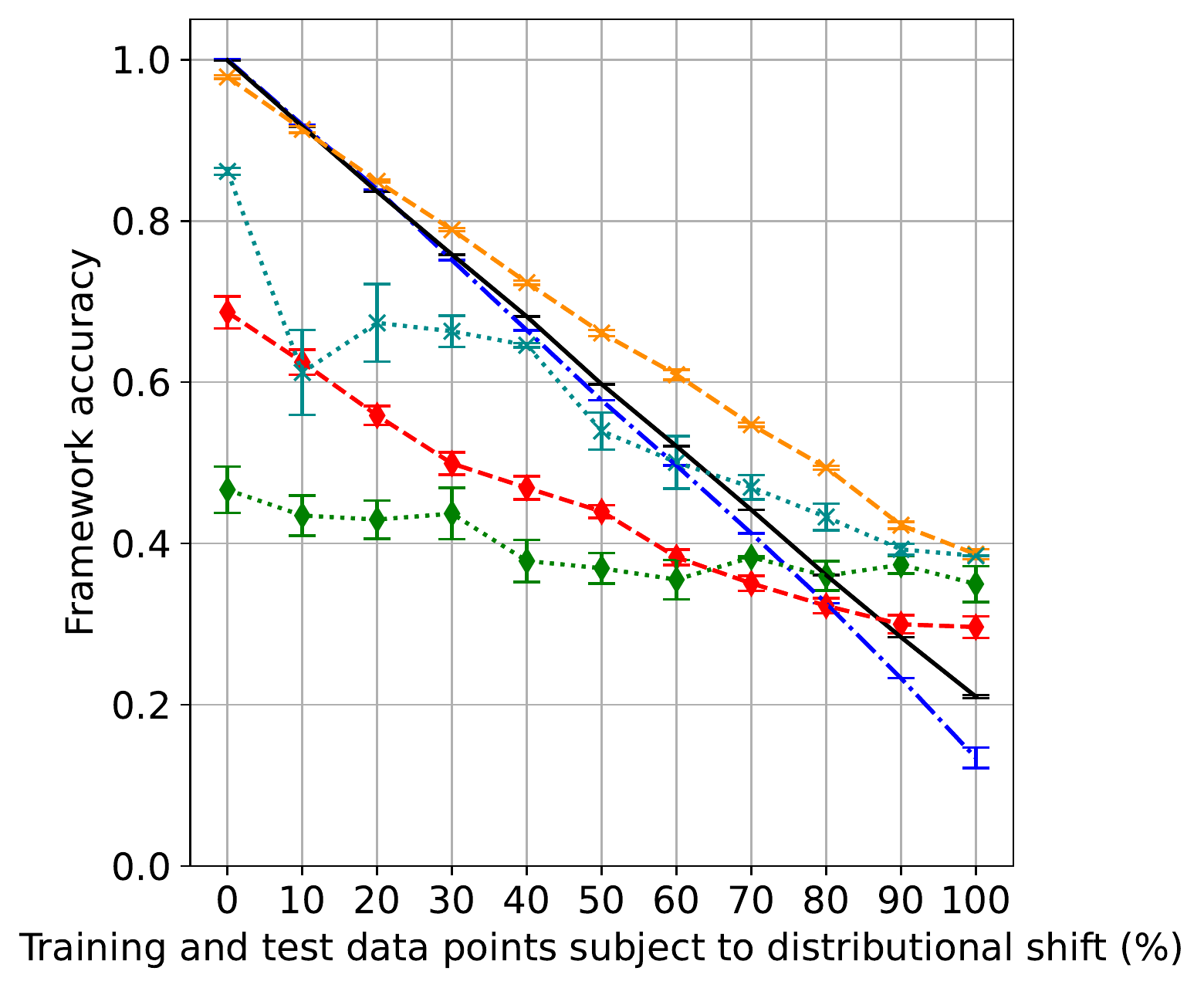}
  \caption{\centering Adversarial Batman Joker}
  \label{fig:fs_abj_acc_unstruc}
\end{subfigure}%
\caption{Accuracy of the~\ac{nsl} framework when training and test data were subject to distributional shifts. Follow Suit Winner.}
\label{fig:fs_unstruc_eval_bj_abj}
\end{figure}

Figure~\ref{fig:fs_unstruc_eval_bj_abj} presents the accuracy of the entire~\ac{nsl} framework when evaluated over a test data subject to the same types of distributional shifts used during the learning of interpretable knowledge. The mean accuracy is reported and the error bars denote standard error over 5 repeats.

\begin{figure}[H]
\centering
\begin{subfigure}{1\columnwidth}
  \centering
  \includegraphics[width=.3\linewidth]{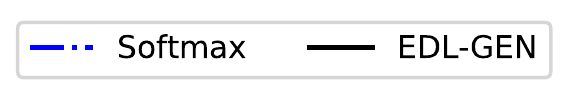}
  \label{fig:fs_50_repeats_legend_nn_card_acc}
\end{subfigure}
\begin{subfigure}{.3\columnwidth}
  \centering
  \includegraphics[width=1\linewidth]{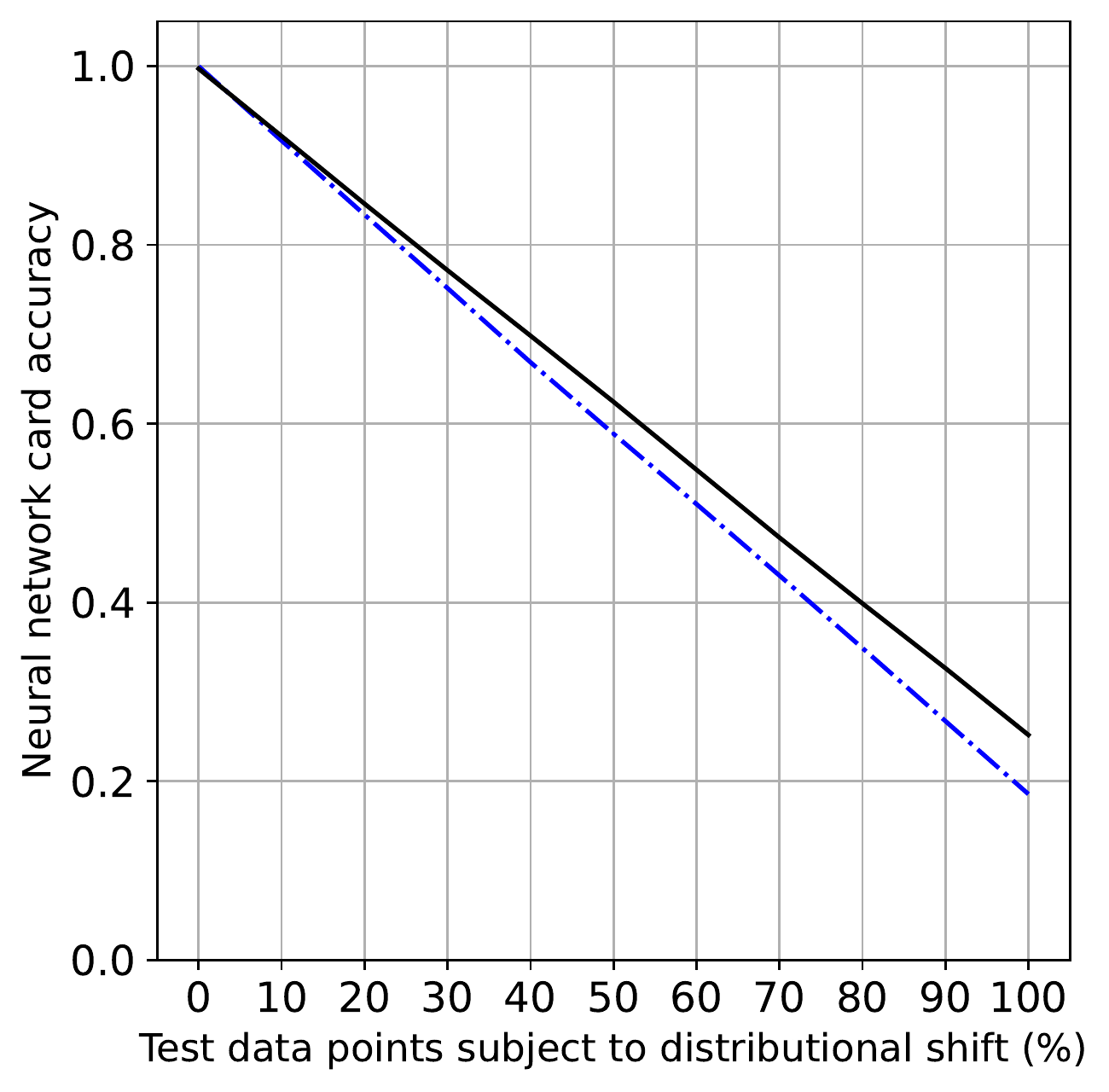}
  \caption{\centering Batman Joker}
  \label{fig:fs_bj_network_acc}
\end{subfigure}%
\begin{subfigure}{.28\columnwidth}
  \centering
  \includegraphics[width=1\linewidth]{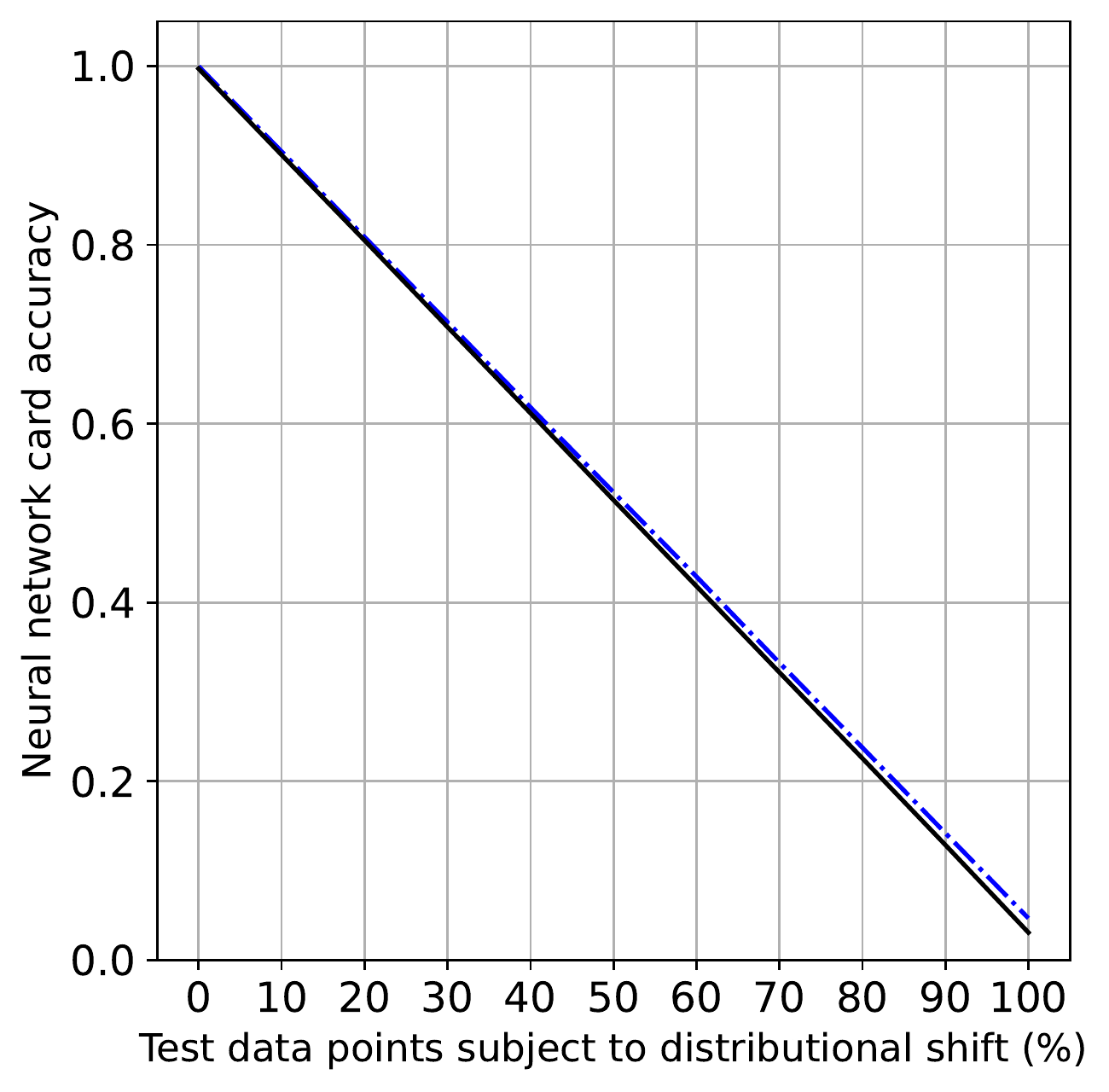}
  \caption{\centering Adversarial Batman Joker}
  \label{fig:fs_abj_network_acc}
\end{subfigure}%
\caption{Neural network card accuracy when test data points were subject to distributional shifts. Follow Suit Winner task.}
\label{fig:fs_network_acc_test_set}
\end{figure}

\noindent
\ac{nsl} outperformed the baselines trained with the same amount of data at each percentage of distributional shift on the Batman Joker deck, and until $\sim$80\% distributional shift on the Adversarial Batman Joker deck. The baselines required $100\times$ the amount of data in order to match or outperform~\ac{nsl}. On the Adversarial Batman Joker deck, the performance was lower for all approaches when the percentage of distributional shift was high, due to the neural networks predicting with lower accuracy (see Figure~\ref{fig:fs_nn_acc_conf}). The baselines trained with the same amount of data outperformed \ac{nsl} for $\gt 80\%$ distributional shifts. This was because they largely predicted player 1 and was sufficient to reach approximately $40\%$ of accuracy on the test set: the Follow Suit Winner task is biased towards player 1 because winning depends on playing the highest ranked card with the same suit as player 1. In the test set, 38.6\% of the data was indeed labelled with player 1 as the winner, which roughly corresponds to the performance of the baselines trained with the same number of examples at 100\% shifts in Figure~\ref{fig:fs_abj_acc_unstruc}. \ac{nsl} EDL-GEN outperformed \ac{nsl} Softmax because of two reasons. Firstly, the rules learned by \ac{nsl} EDL-GEN, in the presence of high percentage of distributional shifts, were more accurate (see Figures~\ref{fig:fs_bj_results_acc} and~\ref{fig:fs_abj_results_acc}) because of the lower number of incorrect~\ac{wcdpi} examples when the EDL-GEN neural network was used (Figure~\ref{fig:fs_ilp_ex_95_100}). In addition, the EDL-GEN neural network provided more informative bias to the~\ac{las} system through better~\ac{wcdpi} example weight penalties (Figure~\ref{fig:fs_nn_wpr}). Finally, the decrease in performance of the \ac{nsl} approaches over unseen data subject to distributional shift seemed to be linear in the percentage of applied distributional shift. This was primarily due to the accuracy of the neural network feature predictions. Figure~\ref{fig:fs_network_acc_test_set} shows that indeed the accuracy of neural network predictions over unseen card images decreased linearly with the increase of the percentage of distributional shifts. 

As shown in Figure~\ref{fig:fs_network_acc_test_set}, the EDL-GEN neural network was more accurate than Softmax in predicting unseen playing cards in the case of minor distributional shift given by the Batman Joker deck. However, the accuracy of the neural networks was the same in the case of major distributional shift given by the Adversarial Batman Joker deck. This was why in Figure~\ref{fig:fs_unstruc_eval_bj_abj},~\ac{nsl} EDL-GEN's showed better performance on the Batman Joker deck. For the Adversarial Batman Joker deck,~\ac{nsl} EDL-GEN's better performance than~\ac{nsl} Softmax was primarily due to more accurate hypotheses.



\section{Sudoku Grid Validity}\label{sec:eval_sud_grid_validity}
Having presented in detail the performance of our \ac{nsl} approaches on the Follow Suit Winner task, we now explore whether the approach can generalise to other tasks. We have applied our approach to a different classification task, the Sudoku Grid Validity task and we present the results in this section. We consider two cases: a $4\times4$ Sudoku grid size, for which the sequence $\boldsymbol{x}$ of unstructured data is much longer than that used for the Follow Suit Winner task. Therefore, each generated~\ac{wcdpi} example contains more contextual features that are likely to be predicted incorrectly, as a result of distributional shifts applied to input images. We then evaluate the scalability of the \ac{nsl} framework even further by considering $9\times9$ Sudoku grid sizes. For the Sudoku grid validiy tasks, the \ac{nsl} instance makes use of the FastLAS system, which has been shown to scale to handle large hypothesis spaces~\cite{law2020fastlas}.

\begin{figure}[H]
\centering
\begin{subfigure}{.5\columnwidth}
  \centering
  \includegraphics[width=1\linewidth]{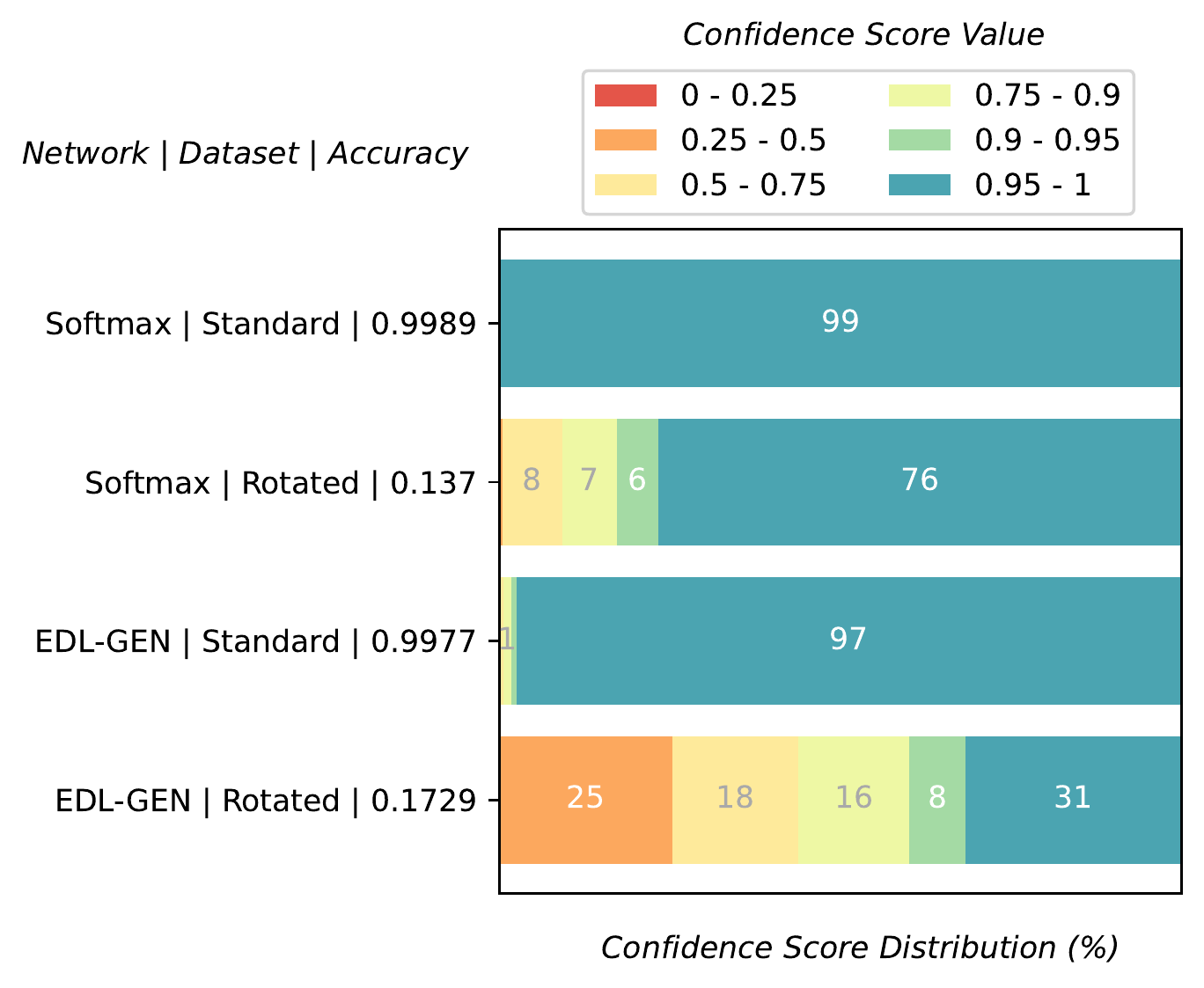}
  \caption{\centering $4\times4$ grids}
  \label{fig:4x4_sud_nn_acc_conf}
\end{subfigure}%
\begin{subfigure}{.5\columnwidth}
  \centering
  \includegraphics[width=1\linewidth]{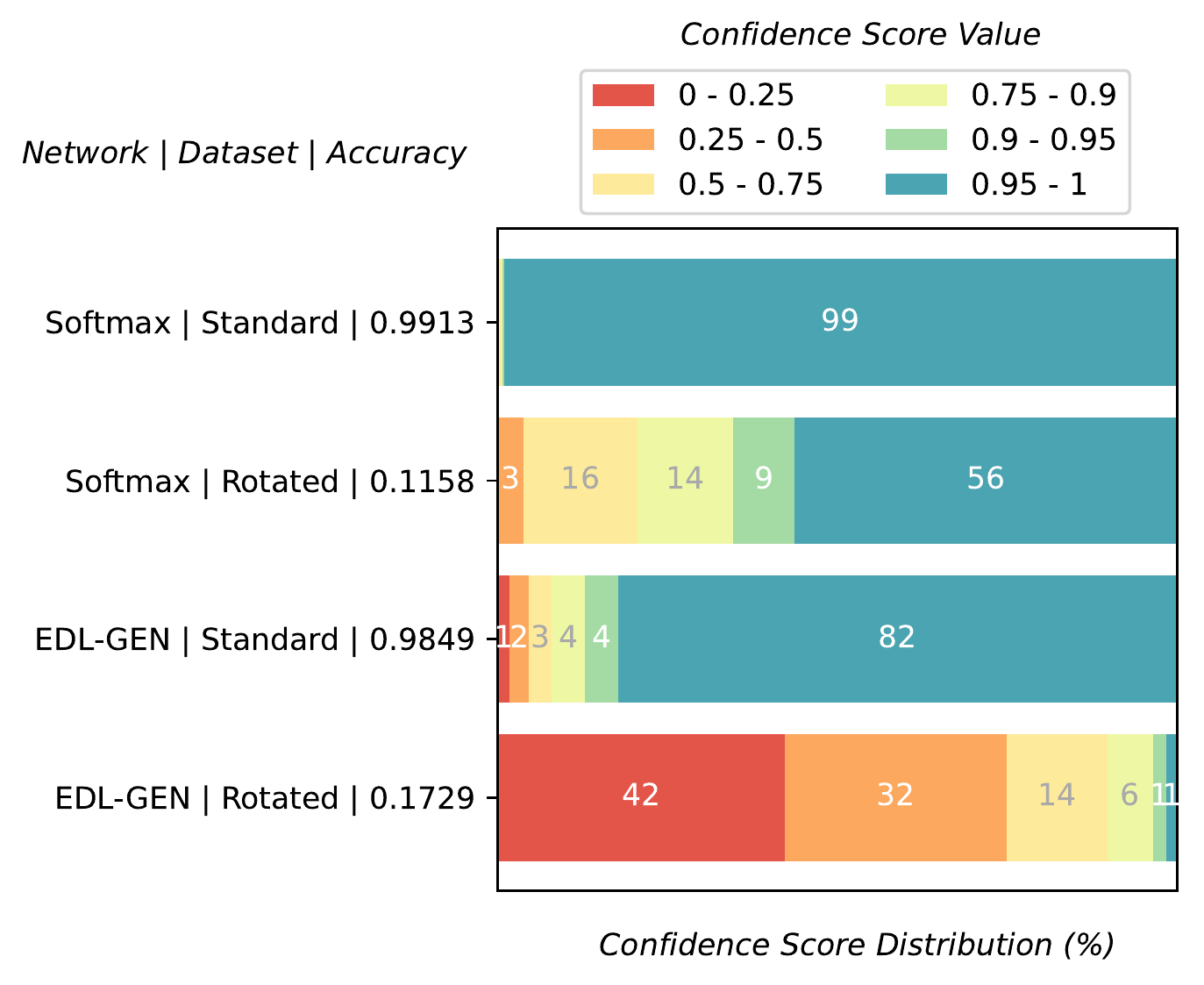}
  \caption{\centering $9\times9$ grids}
  \label{fig:9x9_sud_nn_acc_conf}
\end{subfigure}
\caption{Neural network performance under distributional shifts, Sudoku Grid Validity task}
\label{fig:sud_nn_perf}
\end{figure}

We first pre-train both Softmax and EDL-GEN neural networks on standard images from the MNIST training set. In all experiments, we used MNIST digits 1-4 and 1-9 for the respective Sudoku grid size tasks. Figure~\ref{fig:sud_nn_perf} shows the accuracy and confidence score distribution of the pre-trained neural networks for the $4\times4$ and $9\times9$ grid tasks, on two test sets: a \textit{standard} MNIST test set, and a test set where the MNIST digits have been \textit{rotated} 90$^{\circ}$ clockwise, representing a distributional shift. The test sets also contain MNIST digits 1-4 or 1-9, depending on the Sudoku grid size.

The results are similar to the Follow Suit Winner task. The Softmax neural network predicted with high confidence also over data subject to distributional shift, despite its low test set accuracy. The EDL-GEN neural network predicted more accurately than Softmax on data subject to distributional shift, but Softmax was slightly more accurate on the standard test sets.

\subsection{Learned hypothesis evaluation}
\begin{figure}[H]
\centering
\begin{subfigure}{1\columnwidth}
  \centering
  \includegraphics[width=.75\linewidth]{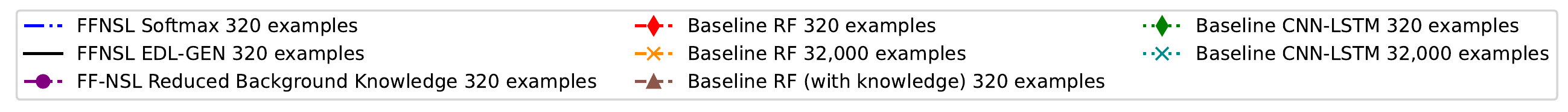}
  \label{fig:sudoku_legend}
\end{subfigure}
\begin{subfigure}{.4\columnwidth}
  \centering
  \includegraphics[width=1\linewidth]{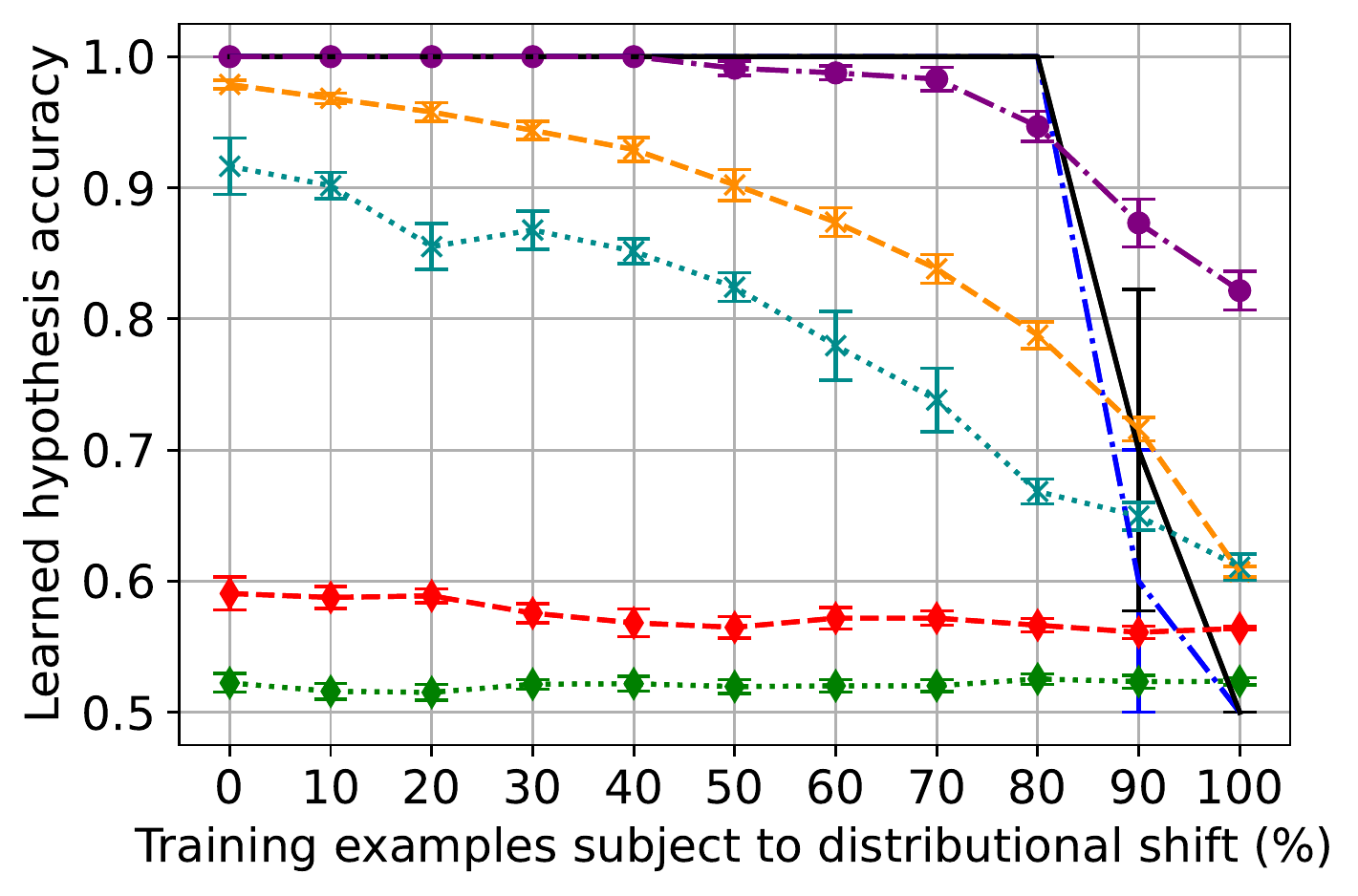}
  \caption{\centering $4\times4$ grids}
  \label{fig:4x4_sudoku_results_acc}
\end{subfigure}%
\begin{subfigure}{.4\columnwidth}
  \centering
  \includegraphics[width=1\linewidth]{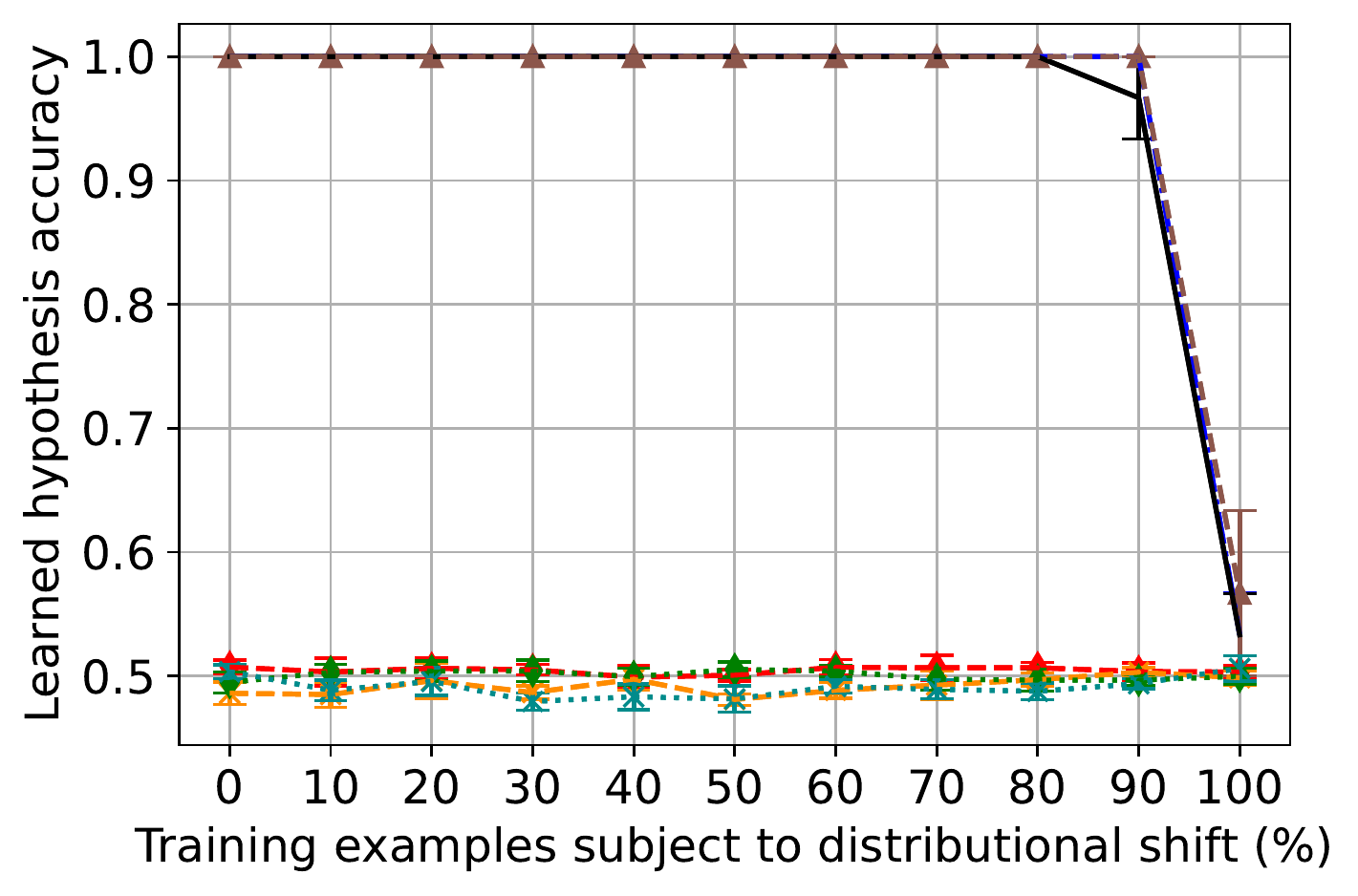}
  \caption{\centering $9\times9$ grids}
  \label{fig:9x9_sudoku_results_acc}
\end{subfigure}
\caption{Accuracy, over 5 repeats, of the learned hypotheses with increasing percentages of data subject to distributional shifts, Sudoku Grid Validity task.}
\label{fig:sud_acc_results}
\end{figure}

Figure~\ref{fig:sud_acc_results} presents the accuracy of the hypotheses learned from unstructured data with increasing percentages of distributional shift, given by rotating MNIST digit images. We plot the mean accuracy over 5 repeats and the error bars denote standard error. In both Sudoku Grid Validity tasks the \ac{nsl} approaches have as input a background knowledge that encodes the concept of a Sudoku grid (see Appendix~\ref{sec:app:ilp} for details). For the $4\times4$ task, we created an additional, more challenging task with a reduced background knowledge where facts about column, row and block were not given but implicitly inferred from a more general notion of division and cell coordinates (given as meta-data). For the $9\times9$ task, we also created an additional training task for the \ac{rf}, (which was the best performing baseline), where pre-trained neural network predictions were post-processed into 3 Boolean features: whether digits were in the same row, column or block, which was given as input to the \ac{rf}. This type of input effectively encoded the Sudoku grid knowledge into the \ac{rf} learning task, and constituted even more information than what was provided to our \ac{nsl} approaches. We demonstrate that~\ac{nsl} performed similarly to this baseline with additional background information. Full FastLAS task listings are given in Appendix~\ref{sec:app:ilp}. 

\noindent
\ac{nsl} approaches outperformed the baselines in both $4\times4$ and $9\times9$ tasks by learning far more accurate hypotheses. In the $4\times4$ task, the baselines required $100\times$ the amount of data to reach an accuracy closer to that of \ac{nsl}, whereas in the $9\times9$ task, the baselines failed completely. In Figure~\ref{fig:4x4_sudoku_results_acc} the purple line is~\ac{nsl} with the explicit background knowledge about the Sudoku grid removed. In this case, the \ac{nsl} approach used the EDL-GEN neural network, and it outperformed the baselines. It also outperformed the other two \ac{nsl} approaches, which used explicit background knowledge about the Sudoku grid, when 90\% and 100\% distributional shifts were applied to the data. This was because with less explicit facts about the grid, the symbolic learner FastLAS was less constrained and alternative hypotheses could be learned which better accommodated the (incorrect) predictions of the neural networks. With explicit facts about the Sudoku grid, the hypothesis space contained rules that performed either very well or very poorly. In Figure~\ref{fig:9x9_sudoku_results_acc}, the brown line shows the accuracy of the \ac{rf} with the 3 Boolean input features, post-processed from the pre-trained neural network predictions, indicating if digits were in the same row, column or block. \ac{nsl} approaches performed similarly to this baseline that used extra input knowledge. 

\begin{figure}[H]
\centering
\begin{subfigure}{1\columnwidth}
  \centering
  \includegraphics[width=.75\linewidth]{Figures/follow_suit_results/follow_suit_nn_penalties_legend.pdf}
  \label{fig:sud_pen_50_repeats_legend}
\end{subfigure}
\begin{subfigure}{.4\columnwidth}
  \centering
  \includegraphics[width=1\linewidth]{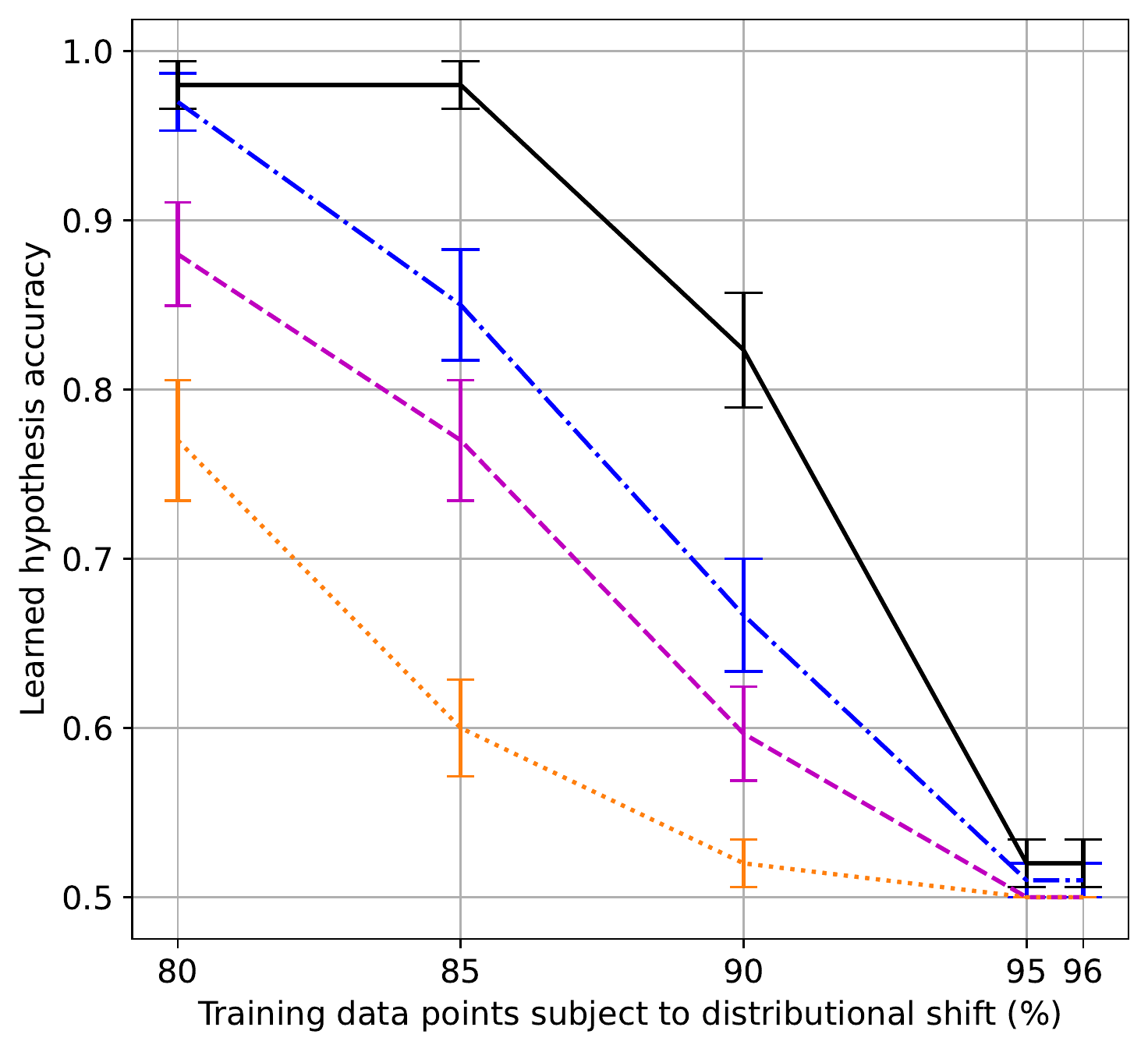}
  \caption{\centering $4\times4$ grids}
  \label{fig:4x4_sudoku_results_acc_50_repeats}
\end{subfigure}%
\begin{subfigure}{.4\columnwidth}
  \centering
  \includegraphics[width=1\linewidth]{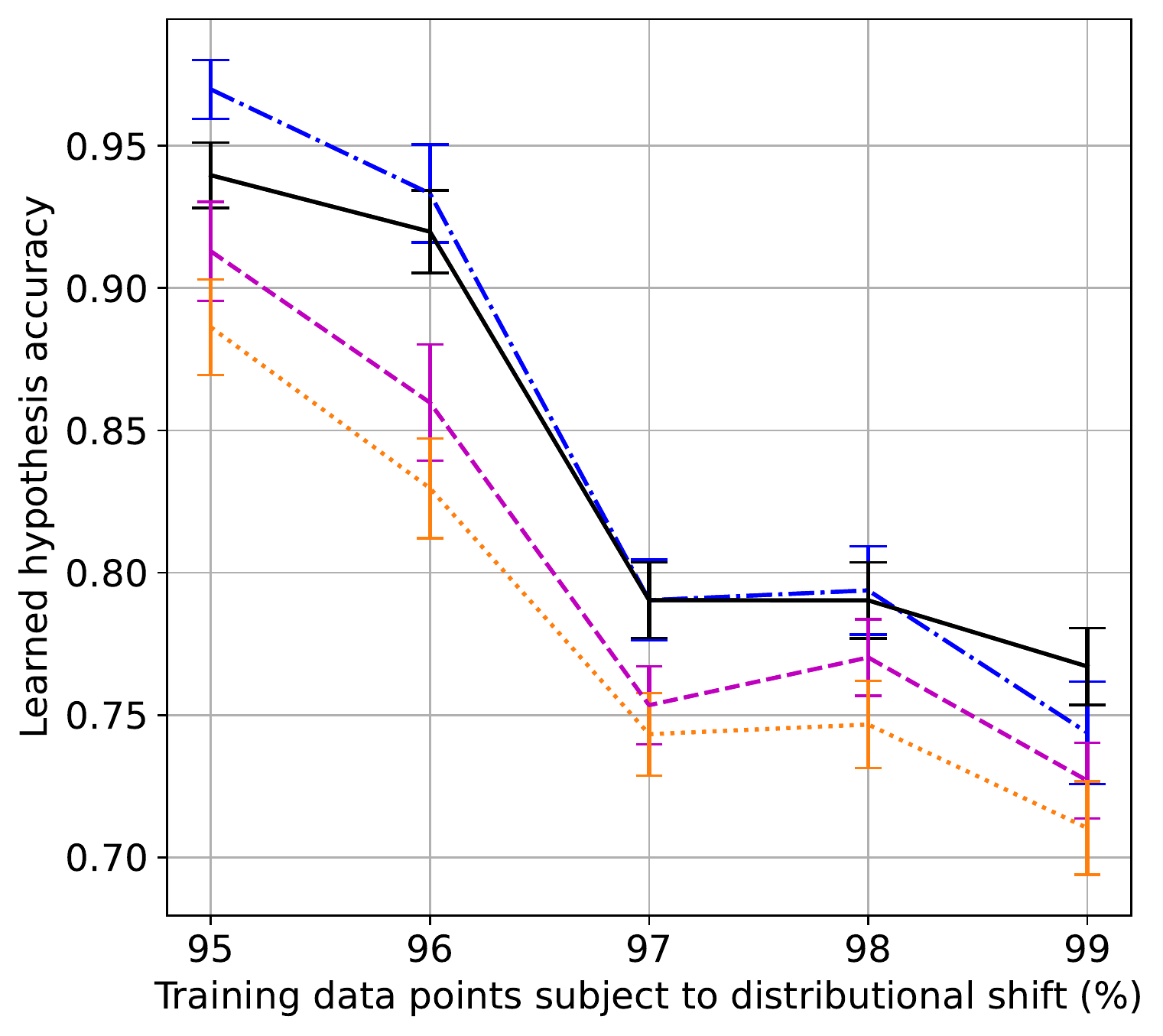}
  \caption{\centering $9\times9$ grids}
  \label{fig:9x9_sudoku_results_acc_50_repeats}
\end{subfigure}
\caption{\ac{nsl} Softmax vs.~\ac{nsl} EDL-GEN. Average accuracy of learned hypotheses over 50 repeats. Sudoku Grid Validity task.}
\label{fig:sud_acc_results_50_repeats}
\end{figure}

We investigate our results further to understand whether using the pre-trained EDL-GEN neural network provides a benefit over Softmax in the presence of high percentages of distributional shifts, also in this domain. We focused on 80-96\% distributional shifts for the $4\times4$ Sudoku Grid Validity task and 95-99\% distributional shifts for the $9\times9$ task, as this was where the performance of~\ac{nsl} deteriorated. Similarly to the Follow Suit Winner task, we run 50 experimental repeats and run two baseline~\ac{nsl} approaches with constant weight penalties. Figure~\ref{fig:sud_acc_results_50_repeats} shows our further experimental results.

Firstly, in both cases of $4\!\times\!4$ and $9\!\times\!9$ grids, the \ac{nsl} Softmax and~\ac{nsl} EDL-GEN that used~\ac{wcdpi} example weight penalties calculated from neural network confidence scores, outperformed the corresponding \ac{nsl} with constant weight penalties. To investigate this further, we explore the~\ac{wcdpi} example weight penalty ratio for the $4\times4$ and $9\times9$ tasks.  

\begin{figure}[ht]
\centering
\begin{subfigure}{1\columnwidth}
  \centering
  \includegraphics[width=.75\linewidth]{Figures/follow_suit_results/follow_suit_nn_penalties_legend.pdf}
  \label{fig:sud_nn_pens_legend}
\end{subfigure}
\begin{subfigure}{.4\columnwidth}
  \centering
  \includegraphics[width=1\linewidth]{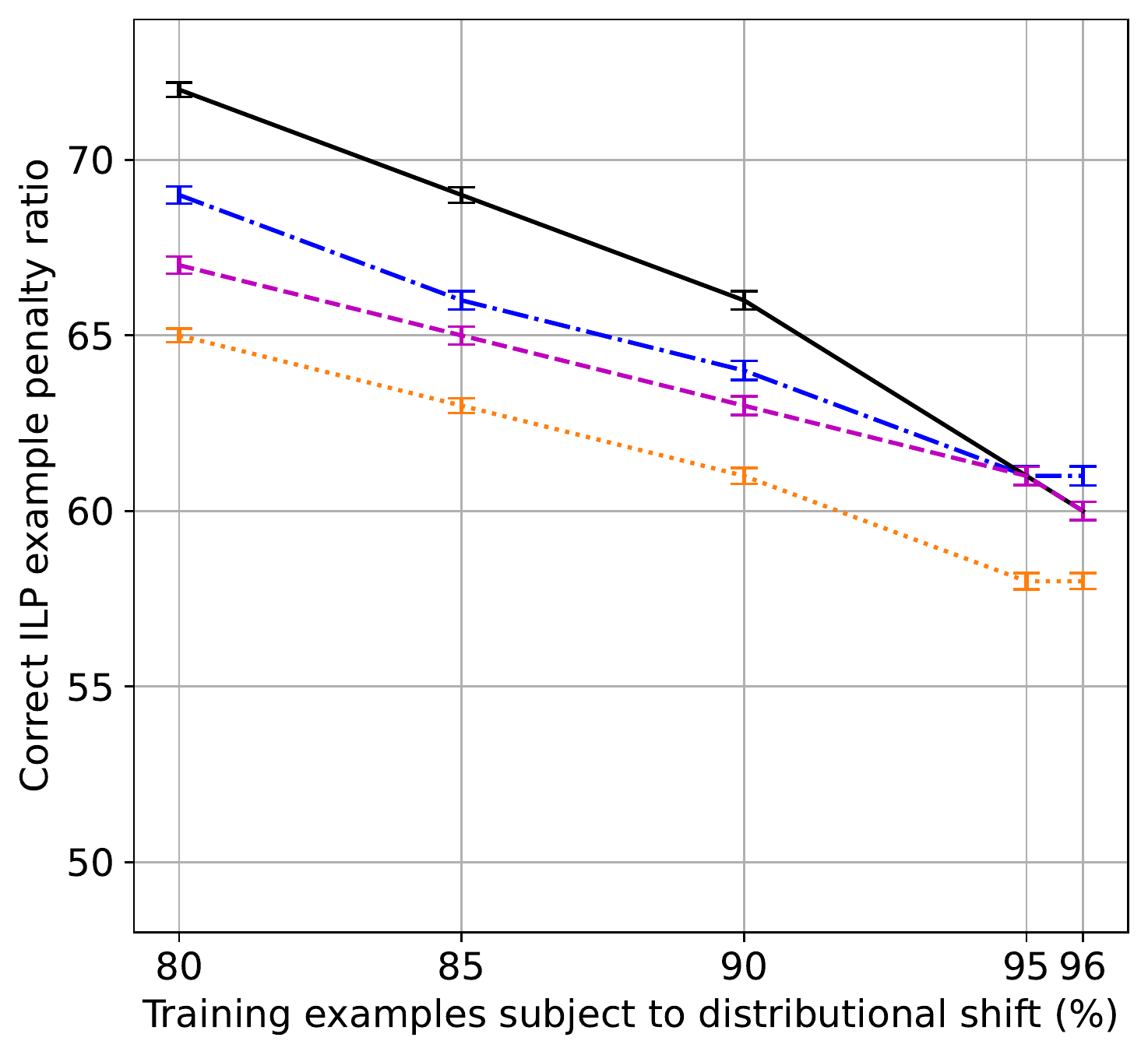}
  \caption{\centering $4\times4$ grids.}
  \label{fig:sud_4x4_pens_80_96}
\end{subfigure}
\begin{subfigure}{.4\columnwidth}
  \centering
  \includegraphics[width=1\linewidth]{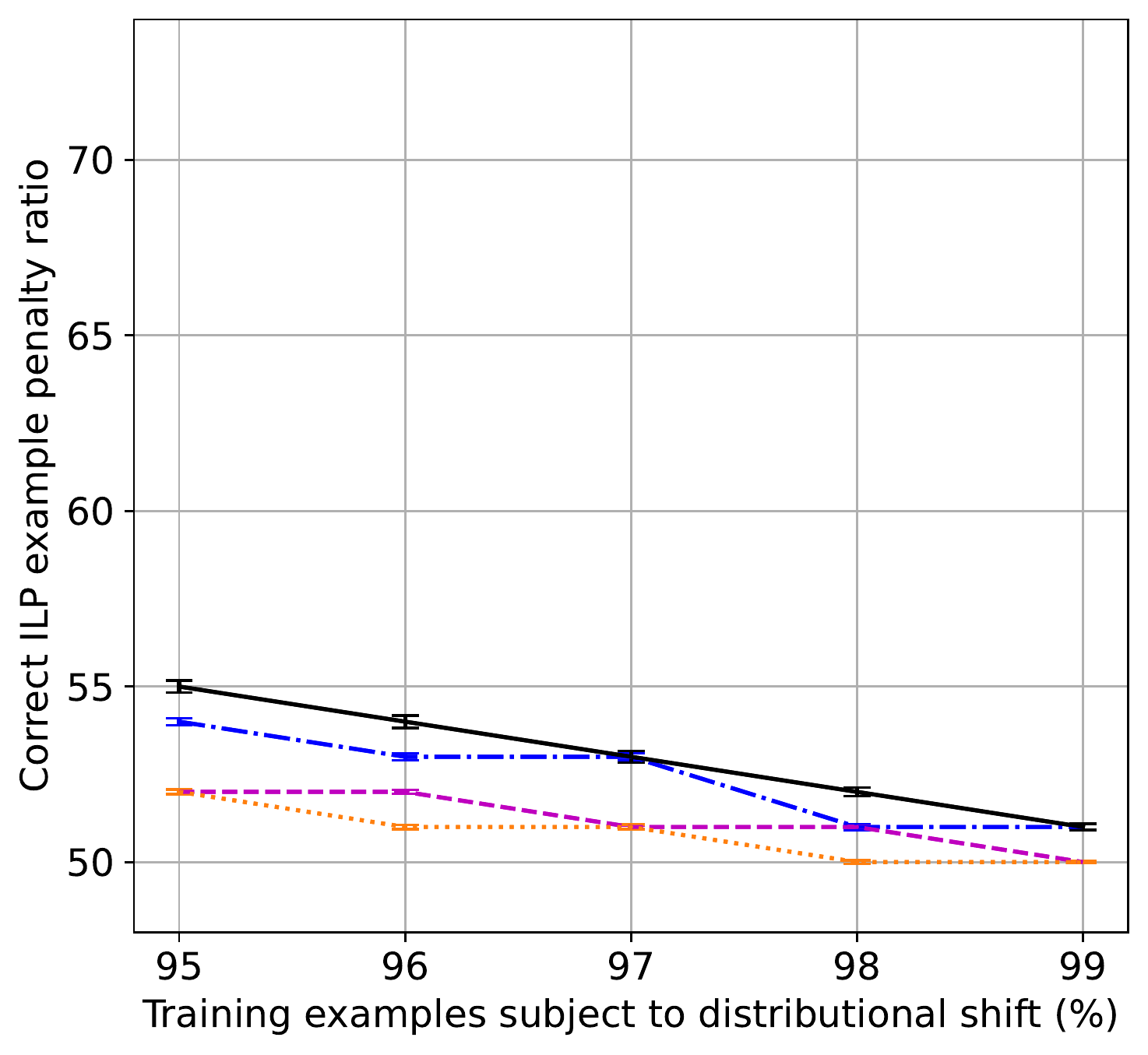}
  \caption{\centering $9\times9$ grids.}
  \label{fig:sud_9x9_pens_95_99}
\end{subfigure}
\caption{\ac{ilp} example weight penalty ratio, Sudoku Grid Validity task.}
\label{fig:sud_nn_wpr}
\end{figure}

Figure~\ref{fig:sud_nn_wpr} shows that both~\ac{nsl} Softmax and~\ac{nsl} EDL-GEN with neural network penalties had a larger weight penalty ratio than the corresponding \ac{nsl} with constant penalties. For \ac{nsl} EDL-GEN this was expected, and for \ac{nsl} Softmax, this is explained by the fact that, as shown in Figure~\ref{fig:sud_nn_perf}, the Softmax neural network had a more varied confidence score distribution over data subject to distributional shift (rotated digits). The difference between the~\ac{wcdpi} example weight penalty ratio with neural network penalties and constant penalties explains the performance gain of \ac{nsl} with neural network penalties versus \ac{nsl} with constant penalties in Figure~\ref{fig:sud_acc_results_50_repeats}. Similarly, the difference between the example weight penalty ratio of \ac{nsl} EDL-GEN and that of \ac{nsl} Softmax, with neural network penalties, also explains why \ac{nsl} EDL-GEN outperformed~\ac{nsl} Softmax in Figure~\ref{fig:4x4_sudoku_results_acc_50_repeats} for $4\!\times\!4$ grids. For the $9\!\times\!9$ task the difference between the example weight penalty ratio of \ac{nsl} EDL-GEN and that of \ac{nsl} Softmax, with neural network penalties, is very small and this explains why \ac{nsl} EDL-GEN and \ac{nsl} Softmax show a similar performance in Figure~\ref{fig:9x9_sudoku_results_acc_50_repeats}. 

\begin{figure}[H]
\centering
\begin{subfigure}{.4\columnwidth}
  \centering
  \includegraphics[width=1\linewidth]{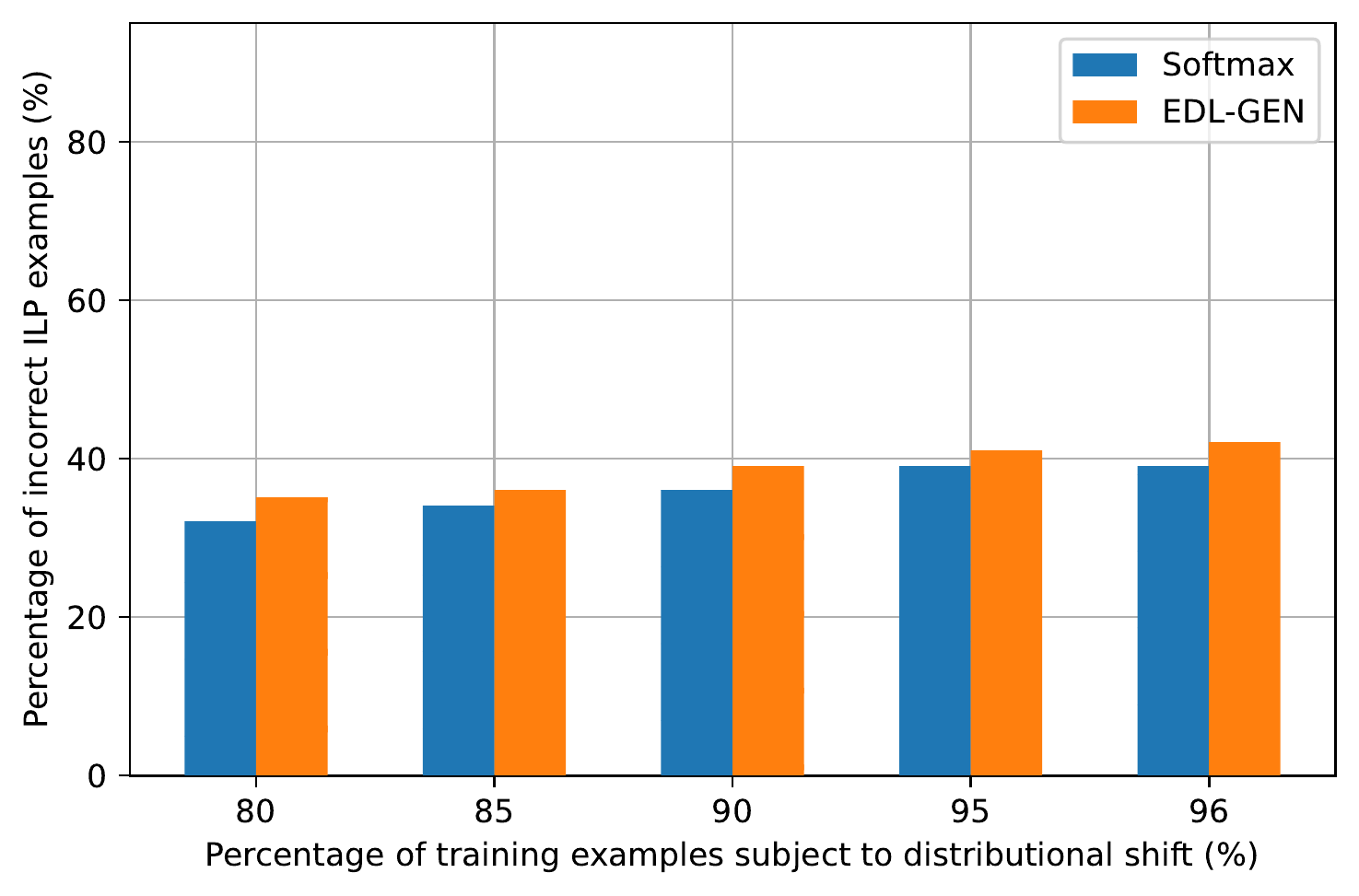}
  \caption{\centering $4\times4$ grids}
  \label{fig:fs_4x4_ilp_ex}
\end{subfigure}%
\begin{subfigure}{.4\columnwidth}
  \centering
  \includegraphics[width=1\linewidth]{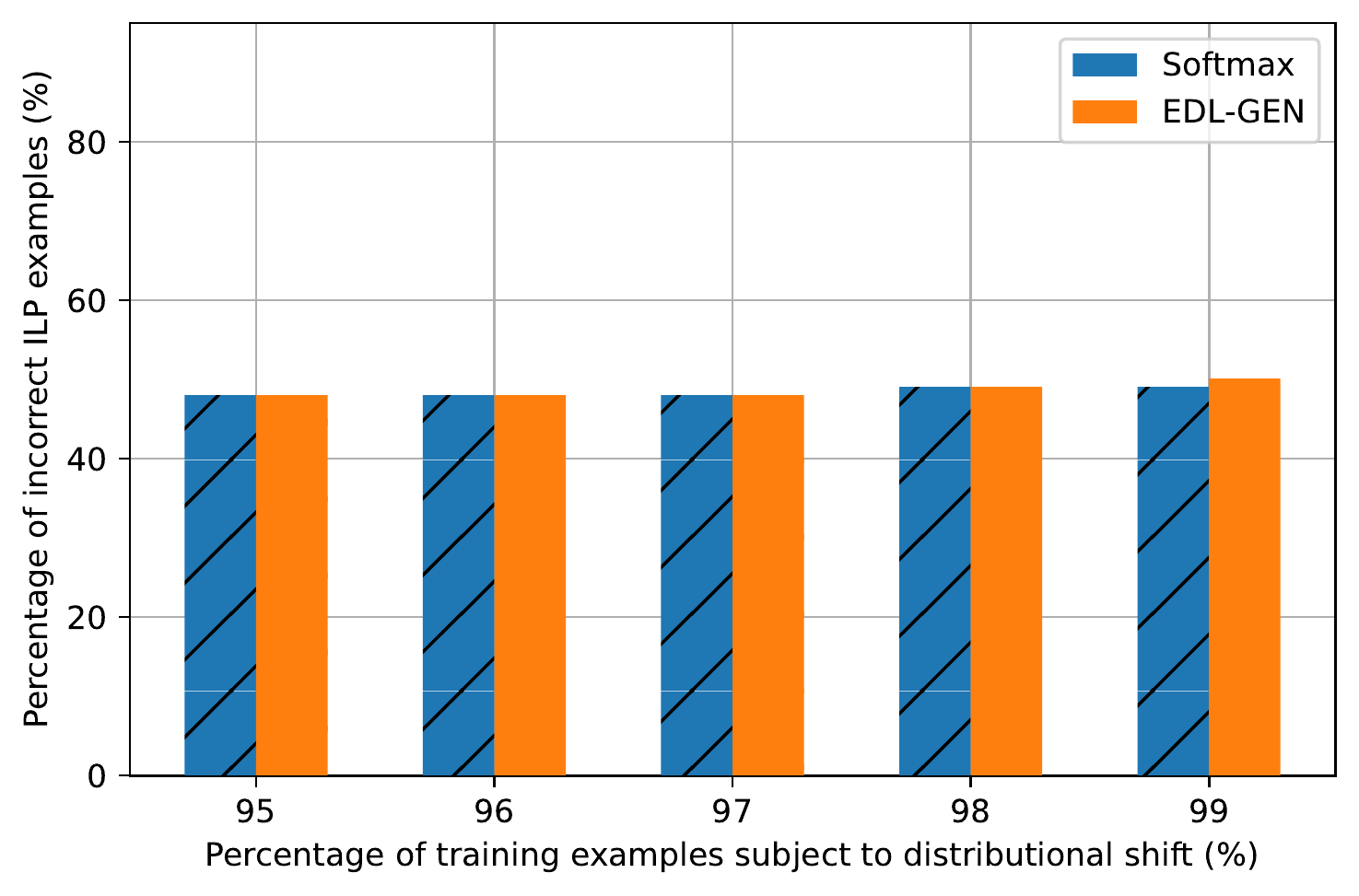}
  \caption{\centering $9\times9$ grids}
  \label{fig:fs_9x9_ilp_ex}
\end{subfigure}
\caption{The effect of applying distributional shifts on the percentage of incorrect~\ac{ilp} examples, Sudoku Grid Validity task.}
\label{fig:sud_ilp_ex_95_100}
\end{figure}

Now, in Figure~\ref{fig:4x4_sudoku_results_acc_50_repeats},~\ac{nsl} Softmax with constant weight penalties outperformed \ac{nsl} EDL-GEN with constant weight penalties, whereas in Figure~\ref{fig:9x9_sudoku_results_acc_50_repeats} these two approaches performed similarly. To investigate this further, we consider the percentage of incorrect~\ac{ilp} examples when distributional shifts were applied. The results are shown in Figure~\ref{fig:sud_ilp_ex_95_100}.

\vspace{-4mm}
For the $4\!\times\!4$ task, despite the pre-trained EDL-GEN neural network predicting on average more accurately (see Figure~\ref{fig:sud_nn_perf}), it led to a higher percentage of incorrect \ac{wcdpi} examples than the pre-trained Softmax neural network, as shown in Figure~\ref{fig:fs_4x4_ilp_ex}. This explains the lower performance in Figure~\ref{fig:4x4_sudoku_results_acc_50_repeats} of~\ac{nsl} EDL-GEN with constant penalties. Using \ac{wcdpi} example weight penalties calculated with EDL-GEN neural network confidence scores was able to rectify this and bias the~\ac{las} system to focus on learning a hypothesis from \ac{wcdpi} examples containing correct neural network predictions. For the $9\!\times\!9$ task both pre-trained neural networks led to a similar percentage of incorrect \ac{wcdpi} examples (see Figure~\ref{fig:fs_9x9_ilp_ex}), which explains the similar performance of \ac{nsl} EDL-GEN and \ac{nsl} Softmax with constant penalties (shown in Figure~\ref{fig:9x9_sudoku_results_acc_50_repeats}). 

\begin{figure}[H]
\centering
\begin{subfigure}{1\columnwidth}
  \centering
  \includegraphics[width=.75\linewidth]{Figures/sudoku_results/9x9/sudoku_9x9_legend.pdf}
  \label{fig:sudoku_legend_interp}
\end{subfigure}
\begin{subfigure}{.4\columnwidth}
  \centering
  \includegraphics[width=1\linewidth]{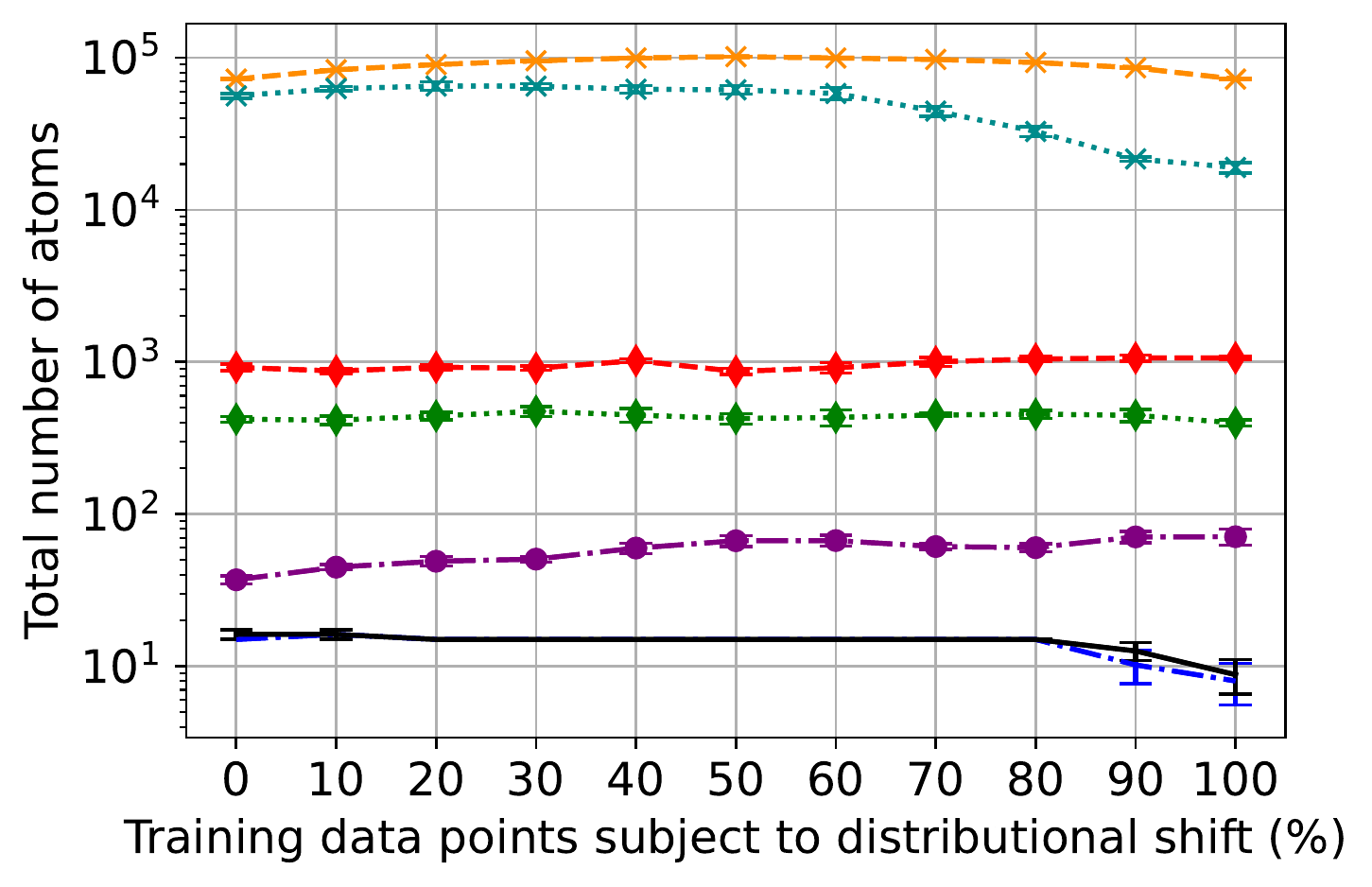}
  \caption{\centering $4\times4$ grids}
  \label{fig:4x4_sudoku_results_interp}
\end{subfigure}%
\begin{subfigure}{.32\columnwidth}
  \centering
  \includegraphics[width=1\linewidth]{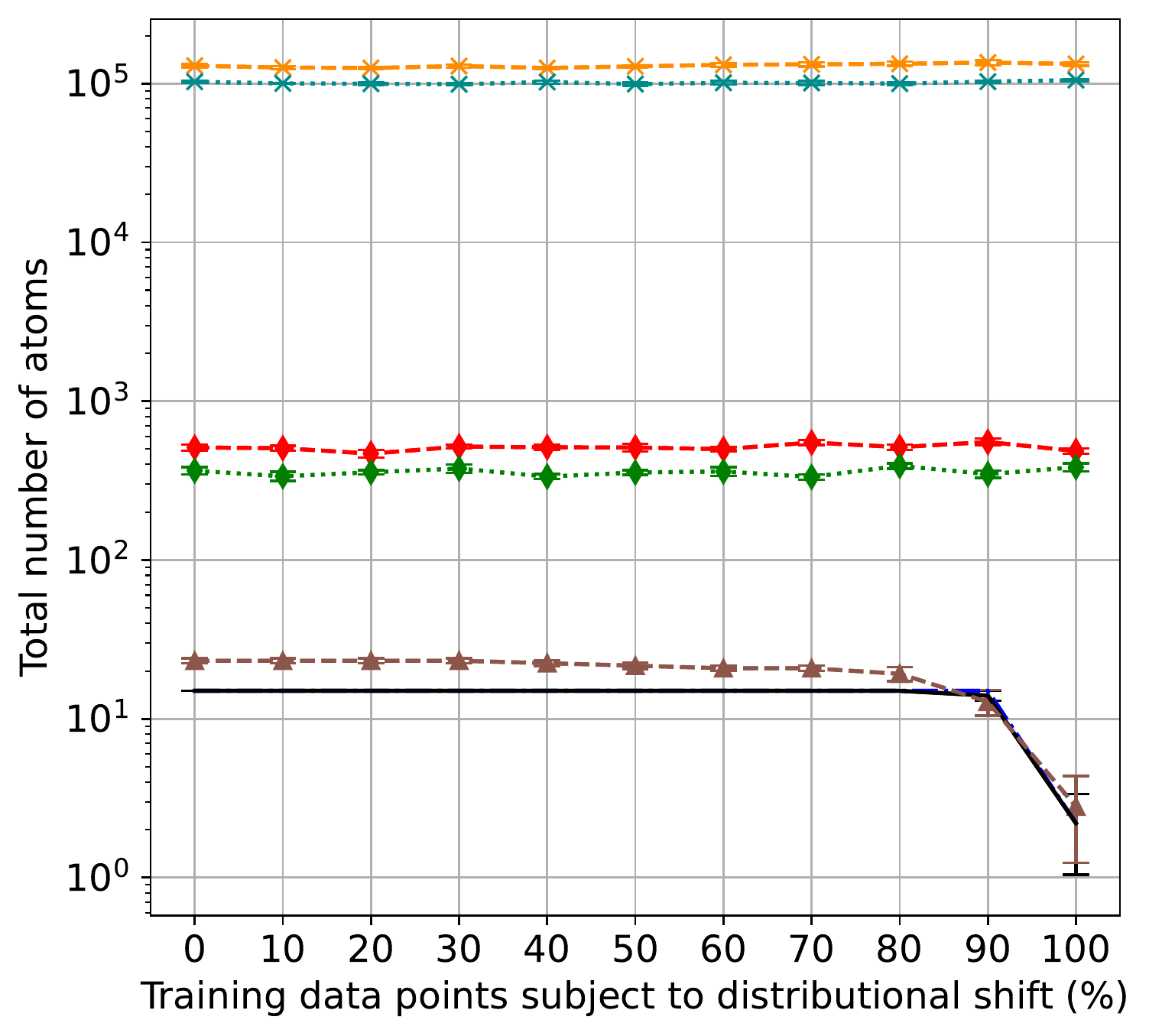}
  \caption{\centering $9\times9$ grids}
  \label{fig:9x9_sudoku_results_interp}
\end{subfigure}
\caption{Interpretability of the learned hypotheses, Sudoku Grid Validity task.}
\label{fig:sud_interp_results}
\end{figure}

Let us now investigate the interpretability of~\ac{nsl} compared to the baseline approaches. The results are shown in Figure~\ref{fig:sud_interp_results}.

\begin{figure}[H]
\centering
\begin{subfigure}{.75\columnwidth}
  \centering
  \includegraphics[width=1\linewidth]{Figures/sudoku_results/9x9/sudoku_9x9_legend.pdf}
  \label{fig:sudoku_legend_lt}
\end{subfigure}
\begin{subfigure}{.4\columnwidth}
  \centering
  \includegraphics[width=1\linewidth]{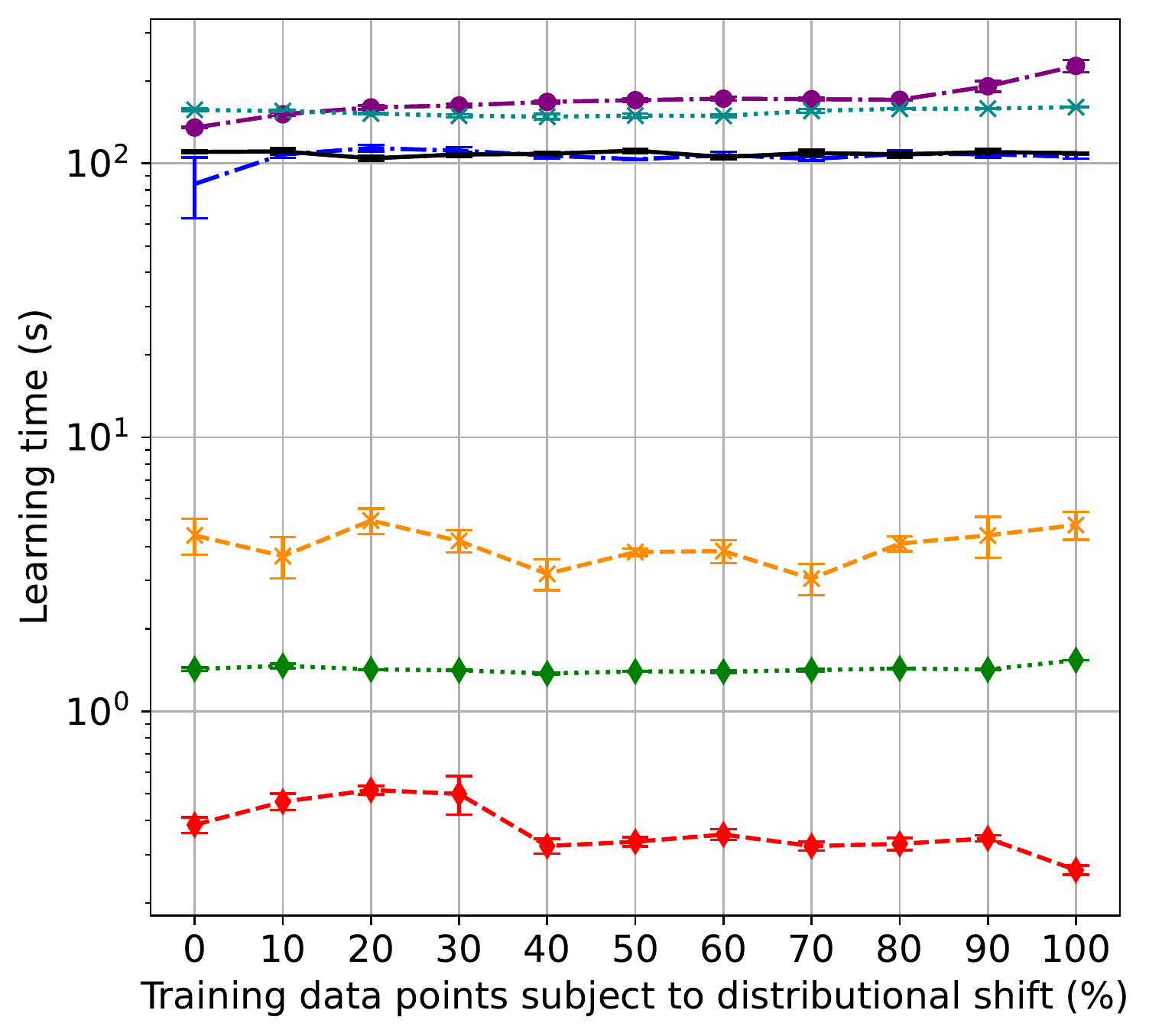}
  \caption{\centering $4\times4$ grids}
  \label{fig:4x4_sudoku_results_lt}
\end{subfigure}%
\begin{subfigure}{.4\columnwidth}
  \centering
  \includegraphics[width=1\linewidth]{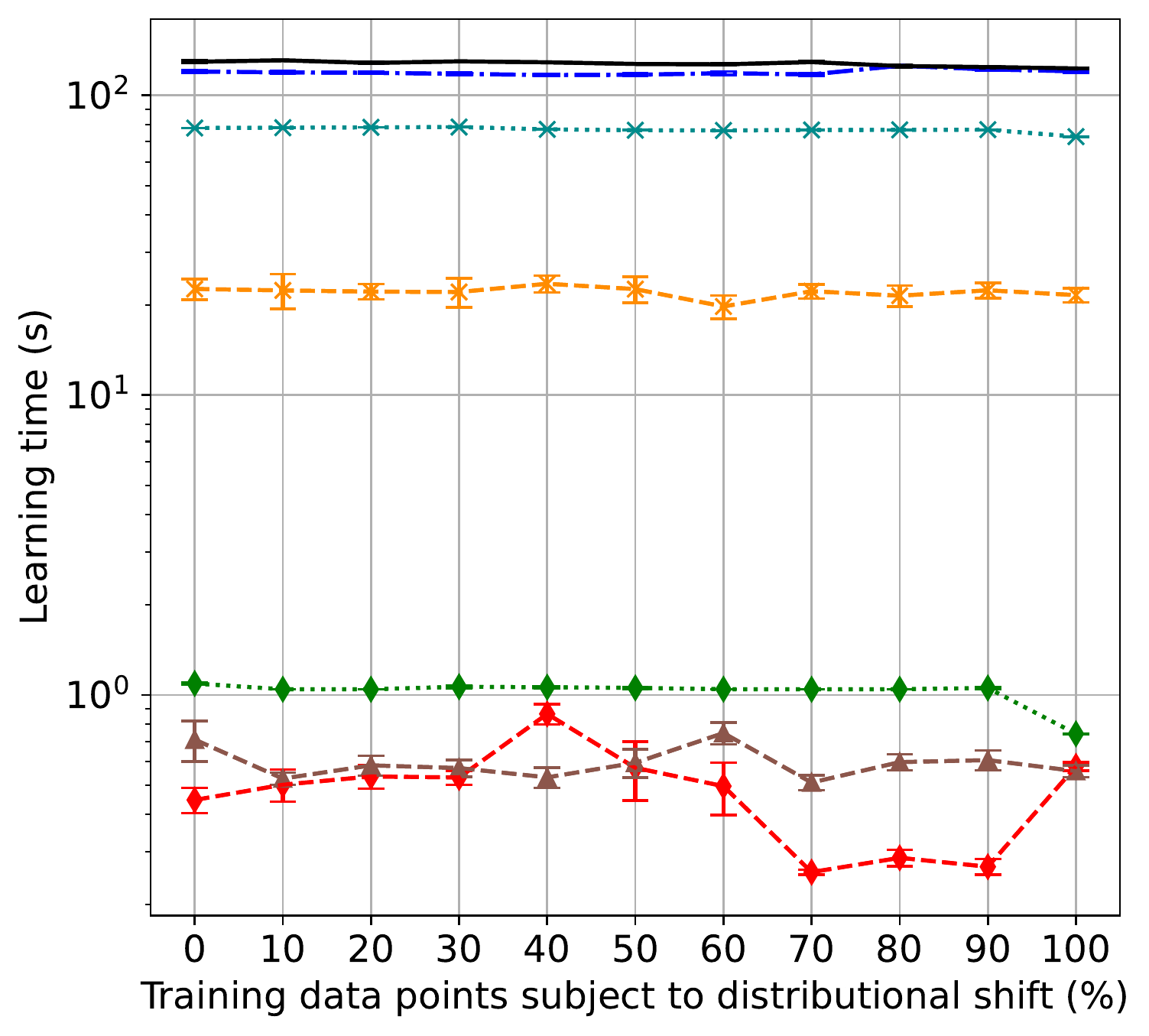}
  \caption{\centering $9\times9$ grids}
  \label{fig:9x9_sudoku_results_lt}
\end{subfigure}
\caption{Hypothesis learning time, Sudoku Grid Validity task.}
\label{fig:sud_learning_time}
\end{figure}

Again,~\ac{nsl} learns significantly more interpretable hypotheses than the baseline approaches. Example learned hypotheses are presented in Appendix~\ref{sec:app:learned_hyps}. As for the learning time, results are shown in Figure~\ref{fig:sud_learning_time}.

In both $4\!\times\!4$ and $9\!\times\!9$ tasks, the learning time for~\ac{nsl} did not increase as the percentage of input data subject to distributional shifts increased. This was because the FastLAS learning system used by \ac{nsl} learned a hypothesis by solving an optimisation problem with respect to all generated \ac{wcdpi} examples. This was not the case for the Follow Suit Winner task where the ILASP system learned an optimal hypothesis iteratively over the examples. It is interesting to note that in Figure~\ref{fig:4x4_sudoku_results_lt}, \ac{nsl}'s learning time had the same order of magnitude as that of the \ac{cnnlstm} trained with $100\times$ the amount of data, which had lower accuracy up to $90\%$ of distributional shifts. 

In conclusion, we have shown that for the learned hypotheses evaluation, \ac{nsl} outperformed the baseline approaches in terms of accuracy and interpretability, even when the baselines were trained with $100\times$ the amount of data. Furthermore, in this task, \ac{wcdpi} example weight penalties had a larger impact on the performance of \ac{nsl}. We have also shown that \ac{nsl} can scale to learning hypotheses where many more unstructured data points $x_{i}$ are observed per labelled input $\langle \boldsymbol{x},y\rangle$, and in these cases~\ac{nsl} learns in a timely manner.

\subsection{\ac{nsl} framework evaluation}
\begin{figure}[H]
\centering
\begin{subfigure}{1\columnwidth}
  \centering
  \includegraphics[width=.75\linewidth]{Figures/sudoku_results/9x9/sudoku_9x9_legend.pdf}
  \label{fig:sud_framework_eval_legend}
\end{subfigure}
\begin{subfigure}{.4\columnwidth}
  \centering
  \includegraphics[width=1\linewidth]{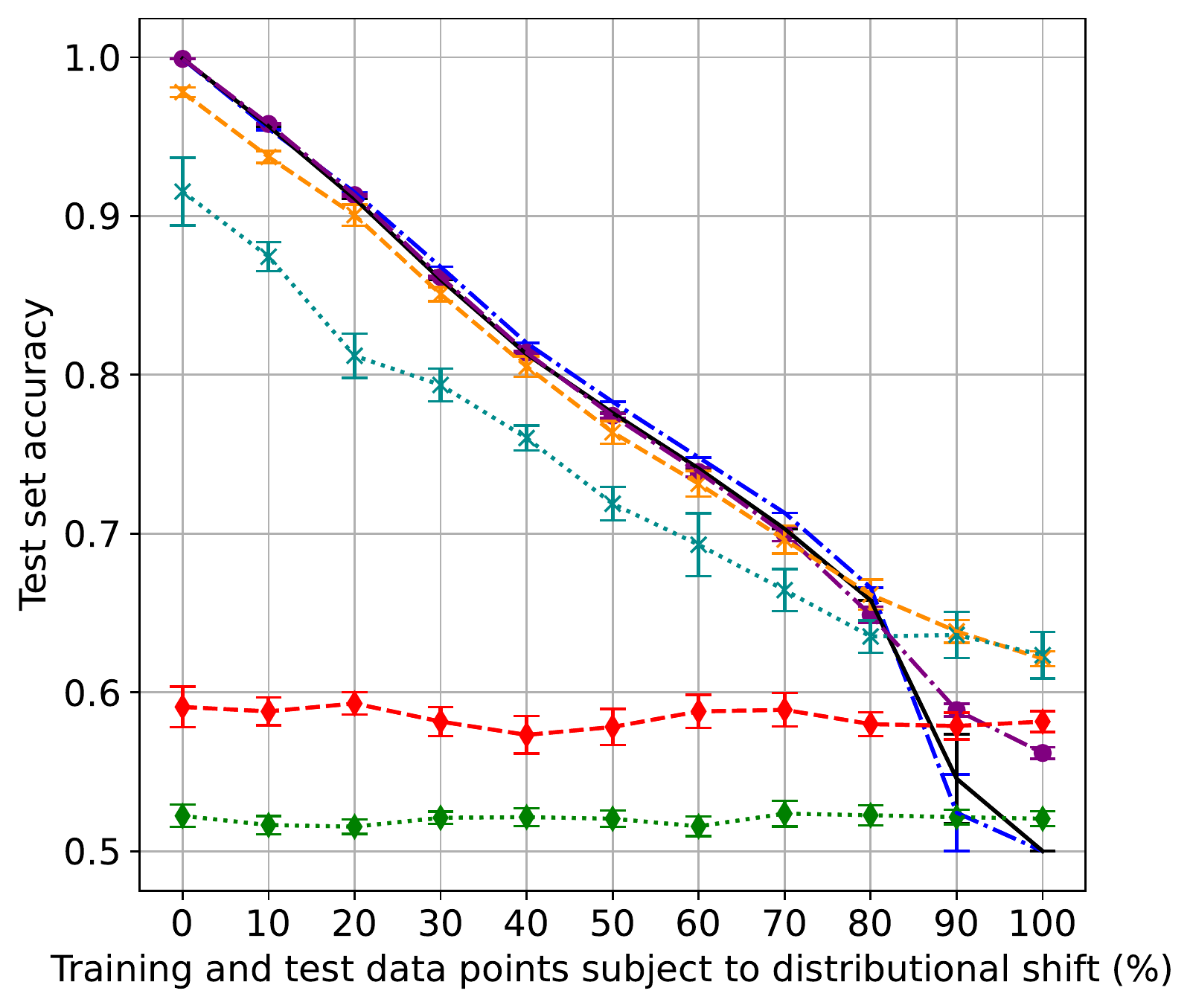}
  \caption{\centering $4\times4$ grids}
  \label{fig:sud_4x4_unstruc_acc}
\end{subfigure}%
\begin{subfigure}{.4\columnwidth}
  \centering
  \includegraphics[width=1\linewidth]{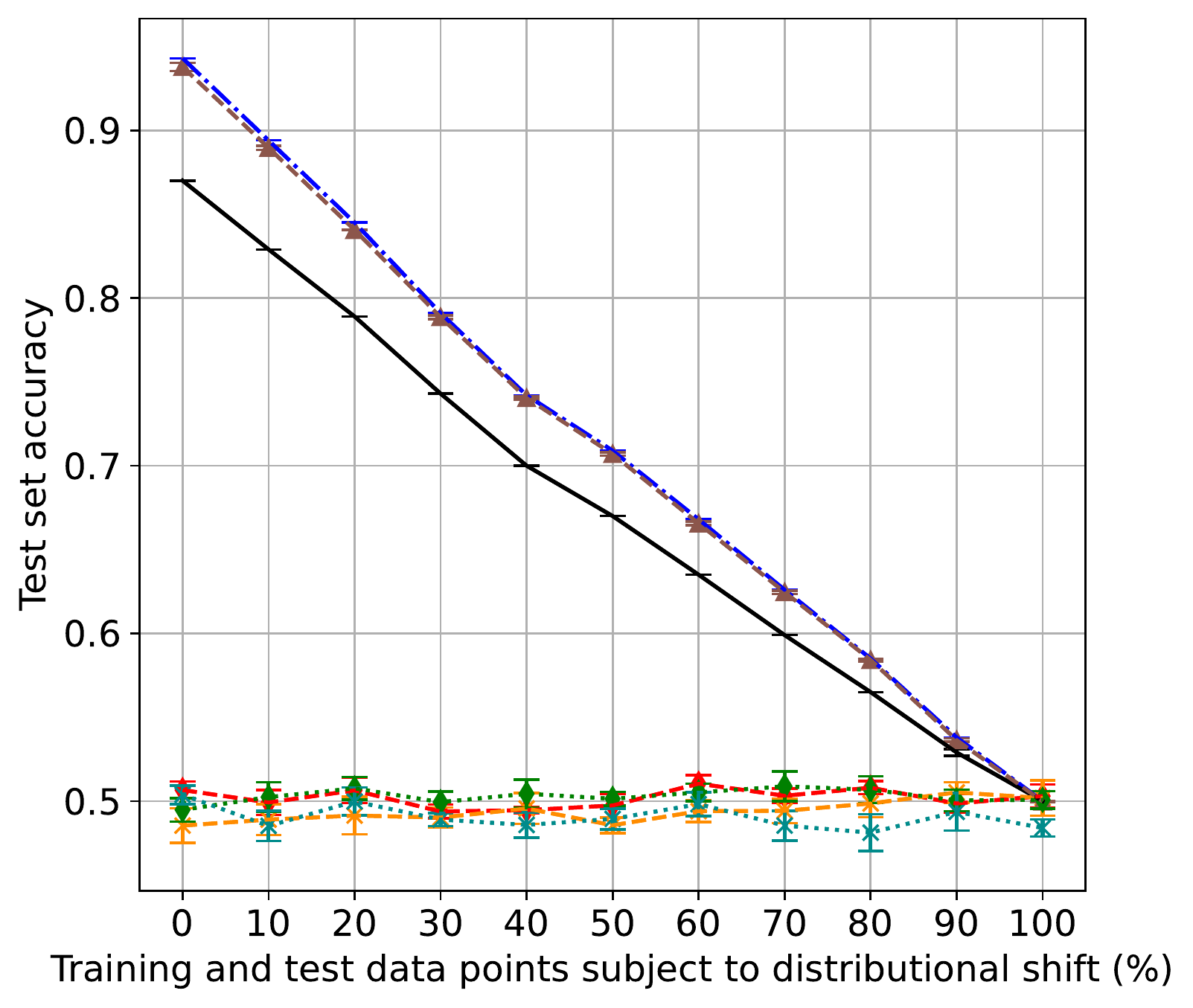}
  \caption{\centering $9\times9$ grids}
  \label{fig:sud_9x9_acc_unstruc}
\end{subfigure}%

\caption{Accuracy of the~\ac{nsl} framework when training and test data were subject to distributional shifts. Sudoku Grid Validity task.}
\label{fig:sud_framework_eval}
\end{figure}

The final evaluation is the accuracy of the overall \ac{nsl} framework when it is applied to a test set of unseen data also subject to distributional shifts. Figure~\ref{fig:sud_framework_eval} shows the mean accuracy over 5 repeats and the error bars denote standard error.

\noindent
In the $4\!\times\!4$ task,~\ac{nsl} outperformed the baselines until 80\% of the test data were subject to distributional shifts, even when the baselines were trained with $100\times$ the amount of data. In the $9\!\times\!9$ task, \ac{nsl} outperformed all baselines, with the exception of the \ac{rf} with additional background knowledge, which performed similarly to \ac{nsl} Softmax. It is indeed interesting to analyse why for the $9\!\times\!9$ task~\ac{nsl} Softmax outperformed \ac{nsl} EDL-GEN especially when low percentages of test data were subject to distributional shift.

Aside from the accuracy of the learned hypotheses, there were two contributing factors to the test set accuracy shown in Figure~\ref{fig:sud_9x9_acc_unstruc}. Firstly, the ability to correctly predict test examples when input data was subject to distributional shifts, and secondly, the ability to correctly predict test examples when no distributional shifts were applied. In the $9\!\times\!9$ task, there were many more digit images on the grid for the neural network to predict. For test examples that were not subject to distributional shift, just one single incorrect neural network prediction may have led to a miss-classified example. At $0\%$ shifts in Figure~\ref{fig:sud_9x9_acc_unstruc}, \ac{nsl} Softmax outperforms \ac{nsl} EDL-GEN. Now, the Softmax neural network accuracy over unseen and \textit{non-rotated} MNIST digits was $0.9927$, whereas that of EDL-GEN neural network was $0.9861$. This explains the drop in performance for~\ac{nsl} EDL-GEN at $0\%$ shifts. As distributional shifts were applied to the test set for percentages ranging between $10-80\%$, both~\ac{nsl} Softmax and~\ac{nsl} EDL-GEN failed to classify most examples subject to distributional shifts, but~\ac{nsl} EDL-GEN also failed to classify more examples that were not subject to distributional shifts, when compared to~\ac{nsl} Softmax. At distributional shifts $> 80\%$, the accuracy of the learned rules also became a factor and the performance of both approaches deteriorated towards $50\%$ accuracy.

\section{Real-world datasets}\label{sec:real_world_app}
In order to demonstrate \acp{nsl} applicability to real-world problems and datasets, in this section we present evaluations of two additional tasks: (1) \textit{Crop Yield Prediction}, and (2) \textit{Indoor Scene Classification}, introduced in Sections \ref{sec:method:crops} and \ref{sec:method:scene} respectively. Let us now summarise each task.

\begin{figure}[!htb]
\centering
\begin{subfigure}{.16\columnwidth}
  \centering
  \includegraphics[width=0.95\linewidth]{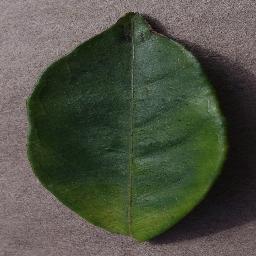}
  \caption{}
  \label{fig:grape_healthy_standard}
\end{subfigure}\hfill
\begin{subfigure}{.16\columnwidth}
  \centering
  \includegraphics[width=0.95\linewidth]{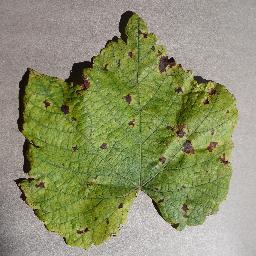}
  \caption{}
  \label{fig:grape_disease_standard}
\end{subfigure}\hfill
\begin{subfigure}{.16\columnwidth}
  \centering
  \includegraphics[width=0.95\linewidth]{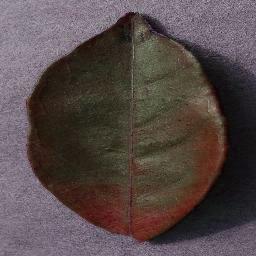}
  \caption{}
  \label{fig:grape_healthy_shift}
\end{subfigure}\hfill
\begin{subfigure}{.16\columnwidth}
  \centering
  \includegraphics[width=0.95\linewidth]{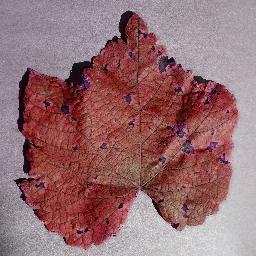}
  \caption{}
  \label{fig:grape_disease_shift}
\end{subfigure}
\caption{Example crop images from the Plant Village dataset with and without distributional shift.}
\label{fig:plants}
\end{figure}

\paragraph{Crop Yield Prediction}The goal is to classify the quality of yield given an image of a particular crop, and symbolic information denoting the crop's location. Softmax and EDL-GEN neural networks were trained to output species and disease information for each crop image, and the symbolic learner learned knowledge that identifies which predicted crop features correspond to different qualities of yield. We used the Plant Village dataset containing images of healthy and diseased crops \cite{plants}, and generated a synthetic symbolic dataset for yield prediction. A distributional shift was applied to crop images using a hue filter, after the neural networks were pre-trained on the unaltered images. Example images are shown in Figure \ref{fig:plants} for a grape crop. Figures \ref{fig:grape_healthy_standard} and \ref{fig:grape_disease_standard} show \textit{standard} images of a healthy and black measles image respectively, and Figures \ref{fig:grape_healthy_shift} and \ref{fig:grape_disease_shift} are the same as Figures \ref{fig:grape_healthy_standard} and \ref{fig:grape_disease_standard} respectively, with the distributional shift applied.

\begin{figure}[!htb]
\centering
\begin{subfigure}{.3\columnwidth}
  \centering
  \includegraphics[width=0.55\linewidth]{Figures/indoor_scenes/b28_standard.jpg}
  \caption{Standard image}
  \label{fig:bedroom_standard}
\end{subfigure}
\begin{subfigure}{.3\columnwidth}
  \centering
  \includegraphics[width=0.55\linewidth]{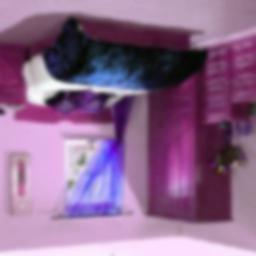}
  \caption{Dist. shift applied}
  \label{fig:bedroom_shift}
\end{subfigure}\hfill
\caption{Example \textit{bedroom} image from the MIT Indoor scene dataset with (b) and without (a) distributional shift applied.}
\label{fig:indoor_scenes}
\end{figure}

\paragraph{Indoor Scene Classification}The goal is to learn knowledge that maps scene level classifications (e.g., bedroom, bathroom, living room) into higher-level super-classes that correspond to a collection of scenes (e.g., home). In this task, we used a state-of-the art neural network called \textit{Semantic Aware Scene Recognition (SASR)} \cite{LOPEZCIFUENTES2020107256}. SASR is a dual-branch \ac{cnn} that is trained to output scene level classifications, utilising semantic segmentation information, and raw image RGB pixel data on each \ac{cnn} branch respectively. The symbolic learner then learned the super-class of each scene. Both neural and symbolic datasets are real and were constructed from the MIT Indoor Scenes dataset \cite{quattoni2009recognizing}. To apply a distributional shift, we transformed each image using a Gaussian blur, hue shift, and 180$^\circ$ rotation, after the neural network was trained on unaltered images. An example image for a \textit{bedroom} scene is shown in Figure \ref{fig:indoor_scenes}. In order to obtain results in a timely manner, in this task we implemented a timeout for FastLAS to return the most optimal hypothesis found after 10 minutes. Also, in contrast to the other tasks, all models were trained with the same dataset size, as the baselines performed strongly when no distributional shift was applied. Finally, only one experimental repeat was performed as the image train/test split was already defined in the dataset \cite{quattoni2009recognizing}.

\begin{figure}[H]
	\centering
	\begin{subfigure}{.49\columnwidth}
		\centering
		\includegraphics[width=0.9\linewidth]{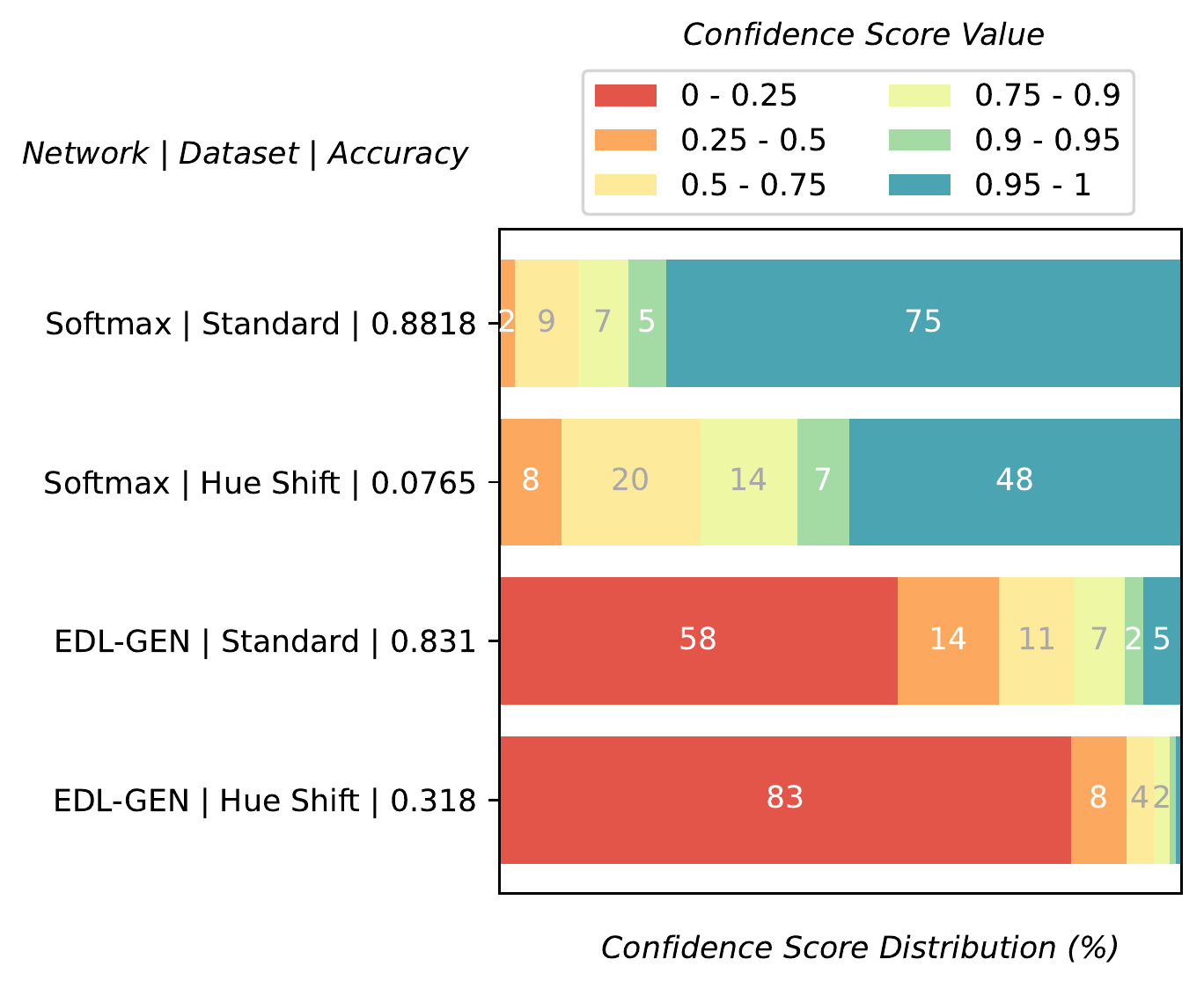}
		\subcaption{Crop images}
		\label{fig:crop_nn_perf}
	\end{subfigure}
	\begin{subfigure}{.49\columnwidth}
		\centering
		\includegraphics[width=0.85\linewidth]{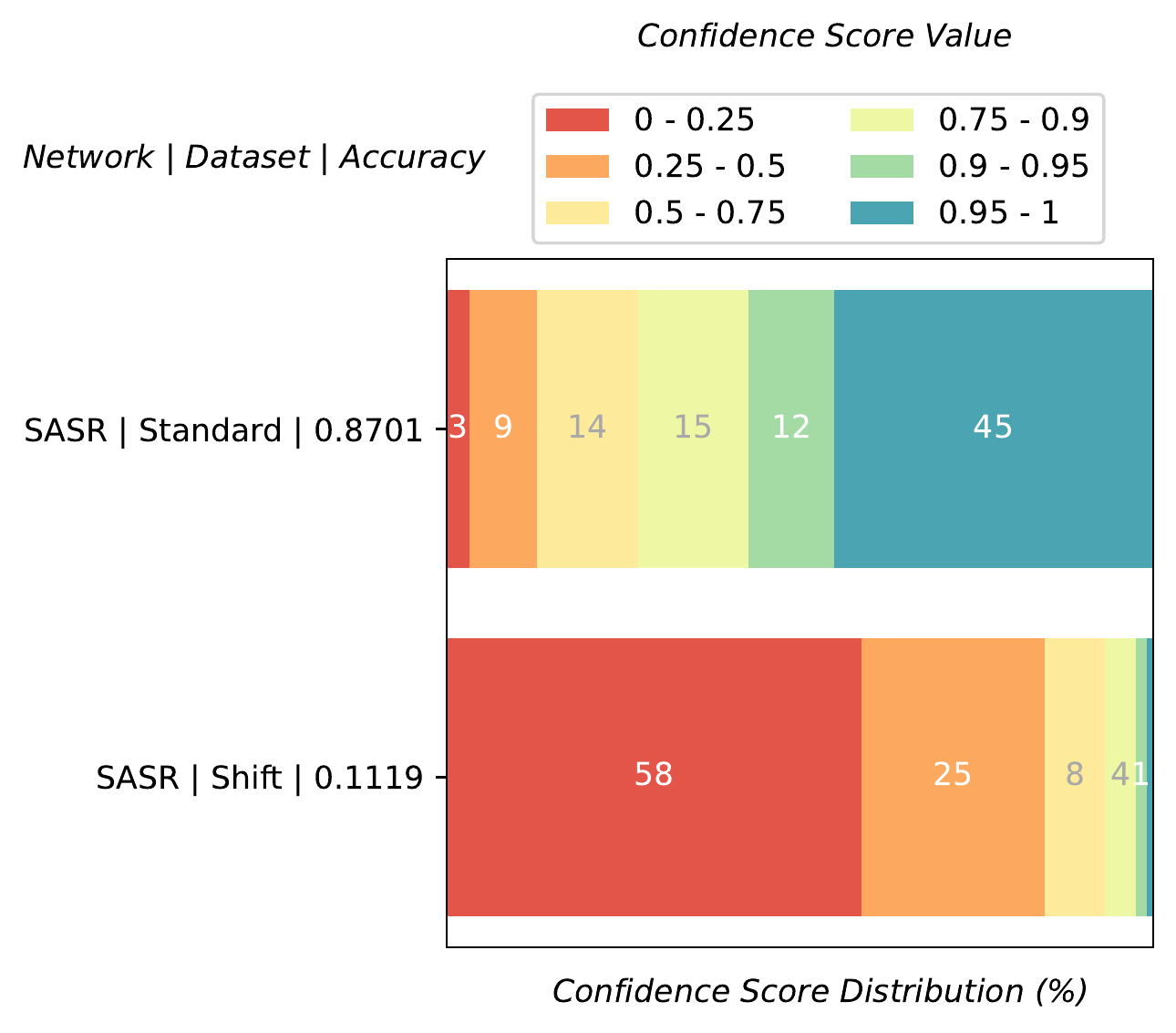}
		\subcaption{Indoor scene images}
		\label{fig:indoor_scene_nn_perf}
	\end{subfigure}
	\caption{Neural network performance under distributional shifts.}
	\label{fig:real_world_app_nn_perf}
\end{figure}

Figure \ref{fig:real_world_app_nn_perf} presents the neural network performance on both tasks, in terms of both accuracy and confidence score distribution when classifying unseen images with and without distributional shift. For the Crop Yield Prediction task in Figure \ref{fig:crop_nn_perf}, the Softmax neural network achieved 88.18\% accuracy for the standard images, and performed poorly when classifying hue shift images. In both datasets, predictions were made with very high confidence. The EDL-GEN neural network achieved 83.1\% accuracy on the standard images, and 31.8\% accuracy on the hue shift images which is much higher than Softmax. Crucially, the EDL-GEN neural network predicted with much lower confidence than Softmax on both datasets which better reflects the predictive accuracy. However, the confidence for standard images was somewhat lower than expected, as 58\% of predictions were made with less than 25\% confidence, despite 83\% accuracy. For the Indoor Scene Classification task in Figure \ref{fig:indoor_scene_nn_perf}, the distributional shift reduced the network accuracy from 87.01\% to 11.19\%, although the confidence scores from the SASR network appropriately reflected the reduced accuracy when distributional shift was applied. Although the SASR network does not have an uncertainty-aware architecture like the EDL-GEN networks used in the other tasks, SASR was able to predict with low confidence under our distributional shift. We suspect this was due to the shifted samples falling between the decision boundary of the 67 scene classes, rather than being completely out-of-distribution, enabling the network to better reflect its uncertainty amongst the possible classes. We now present our evaluation of the learned hypotheses in each task.

\subsection{Learned hypothesis evaluation}

\begin{figure}[H]
	\centering
	\begin{subfigure}{0.49\textwidth}
		\centering
		\includegraphics[width=0.8\linewidth]{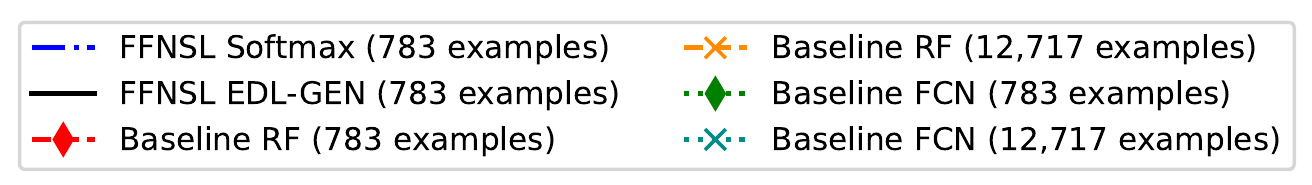}
	\end{subfigure}
	\begin{subfigure}{0.49\textwidth}
		\centering
		\includegraphics[width=0.8\linewidth]{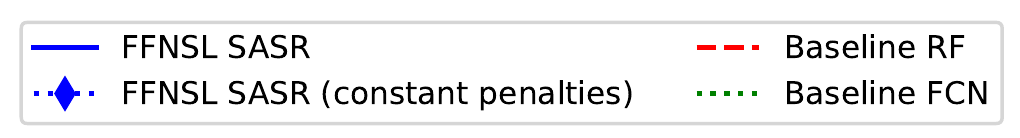}
	\end{subfigure}
	\begin{subfigure}{.49\columnwidth}
		\centering
		\includegraphics[width=0.8\linewidth]{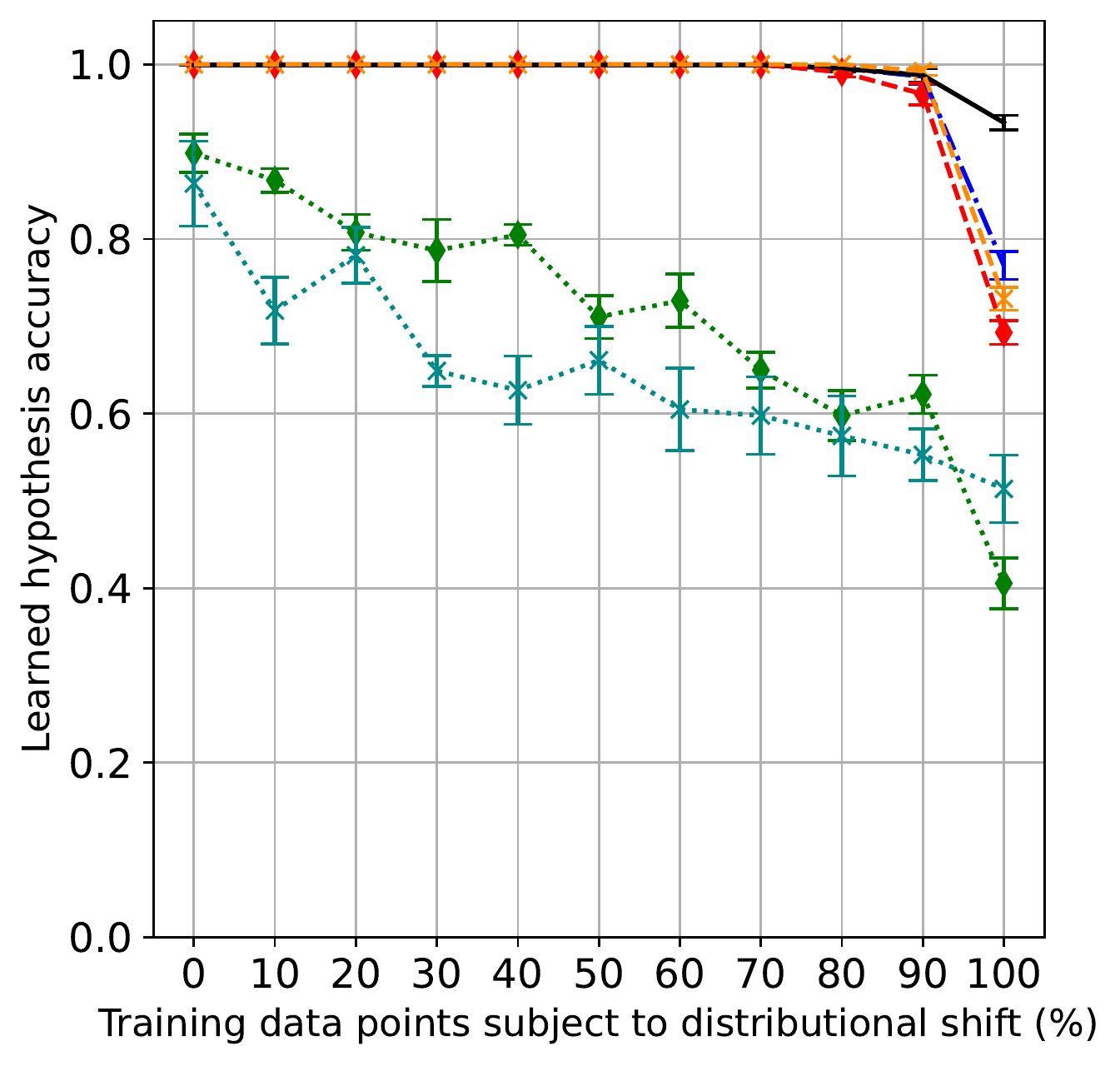}
		\subcaption{Crop Yield Prediction}
		\label{fig:crop_lh_acc}
	\end{subfigure}
	\begin{subfigure}{.49\columnwidth}
		\centering
		\includegraphics[width=0.8\linewidth]{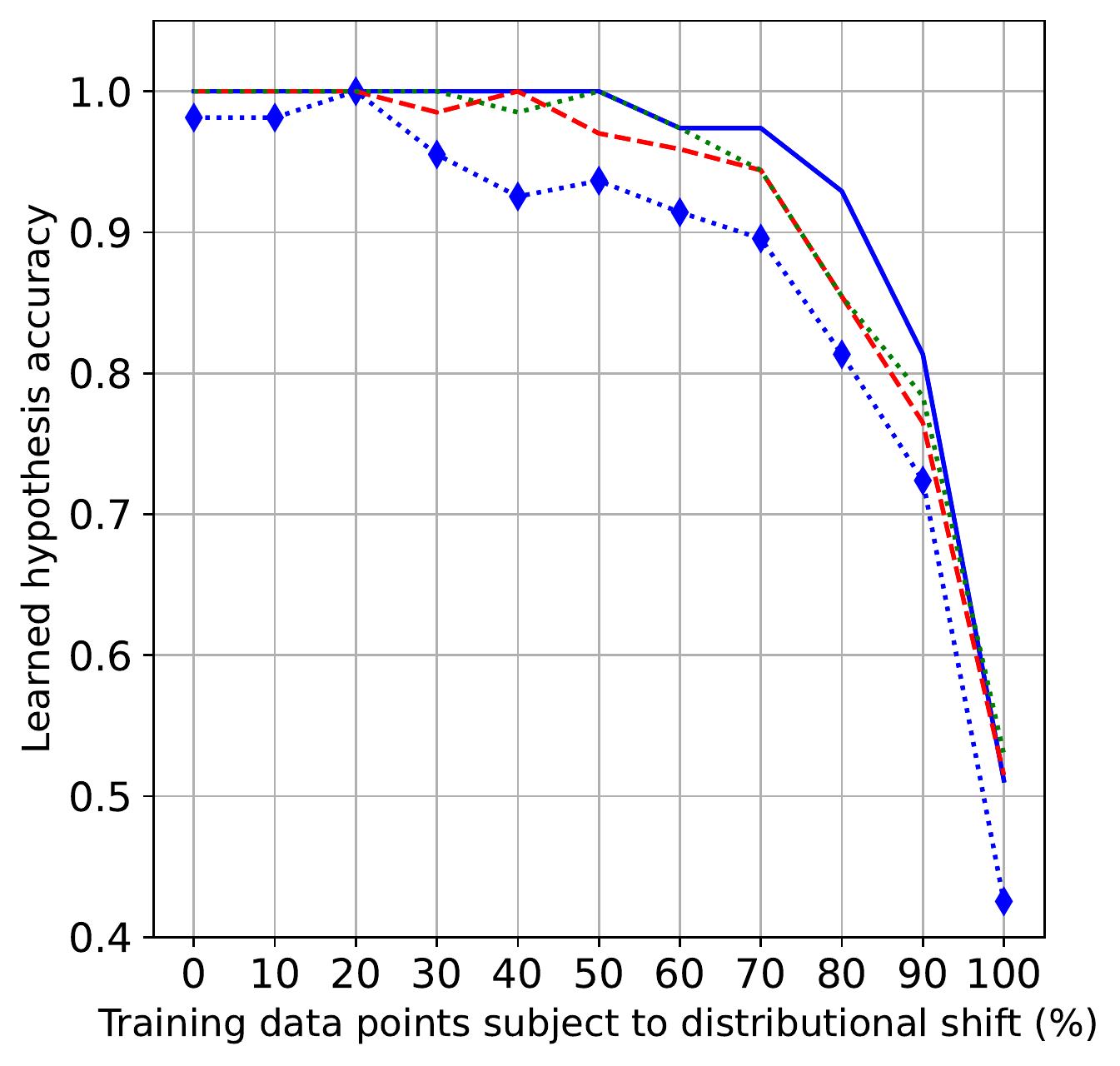}
		\subcaption{Indoor Scene Classification}
		\label{fig:scene_lh_acc}
	\end{subfigure}
	
	\caption{Learned hypothesis accuracy.}
	\label{fig:real_world_app_learned_hyp_acc}
\end{figure}

\begin{figure}[H]
    \centering
	
	\begin{subfigure}{0.49\textwidth}
		\centering
		\includegraphics[width=0.8\linewidth]{Figures/plant_disease/results/plant_disease_legend.pdf}
	\end{subfigure}
	\begin{subfigure}{0.49\textwidth}
		\centering
		\includegraphics[width=0.7\linewidth]{Figures/indoor_scenes/results/indoor_scene_legend.pdf}
	\end{subfigure}

	\begin{subfigure}{.49\columnwidth}
		\centering
		\includegraphics[width=0.8\linewidth]{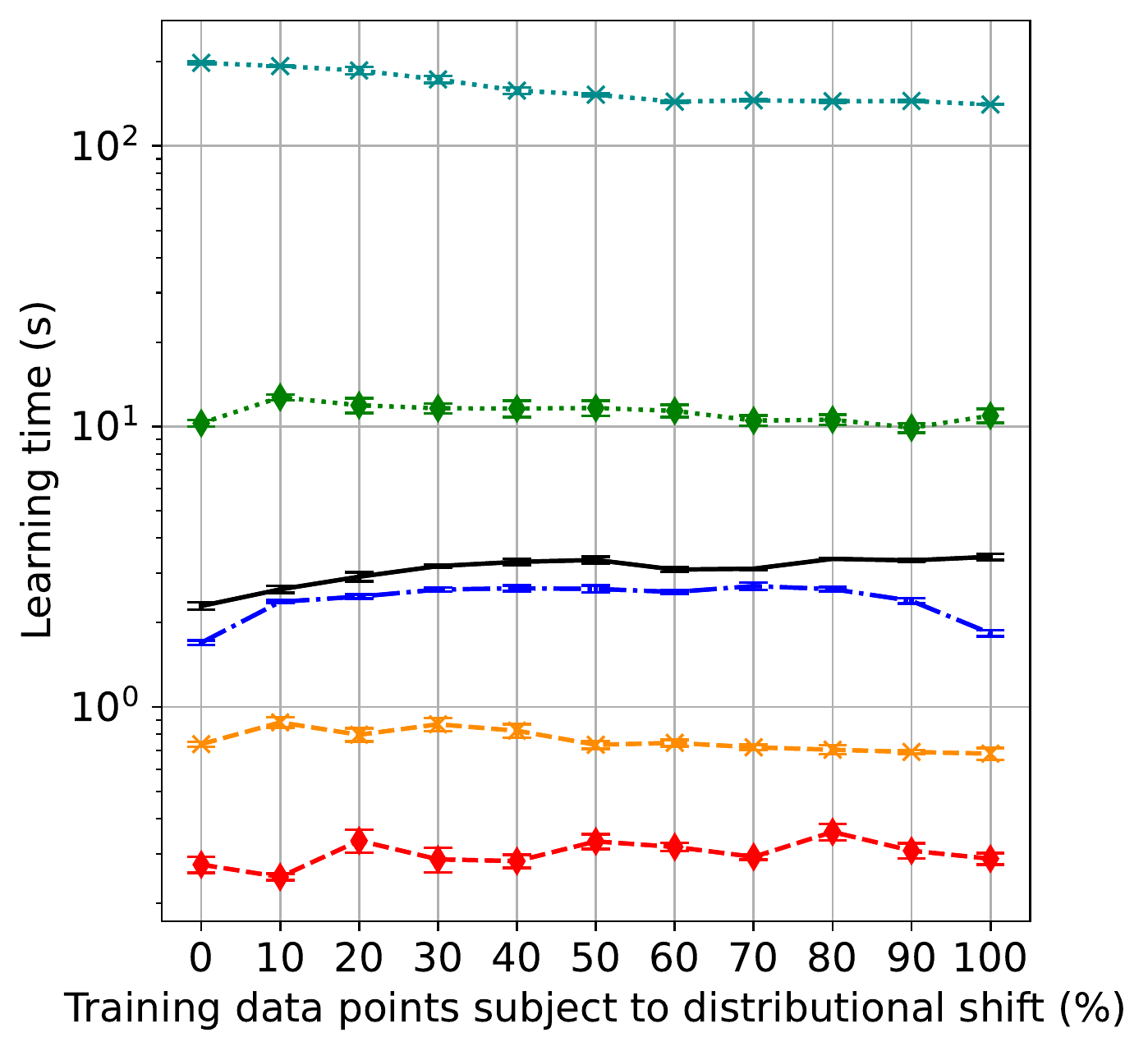}
		\subcaption{Crop Yield Prediction}
		\label{fig:crop_lt}
	\end{subfigure}
	\begin{subfigure}{.49\columnwidth}
		\centering
		\includegraphics[width=0.8\linewidth]{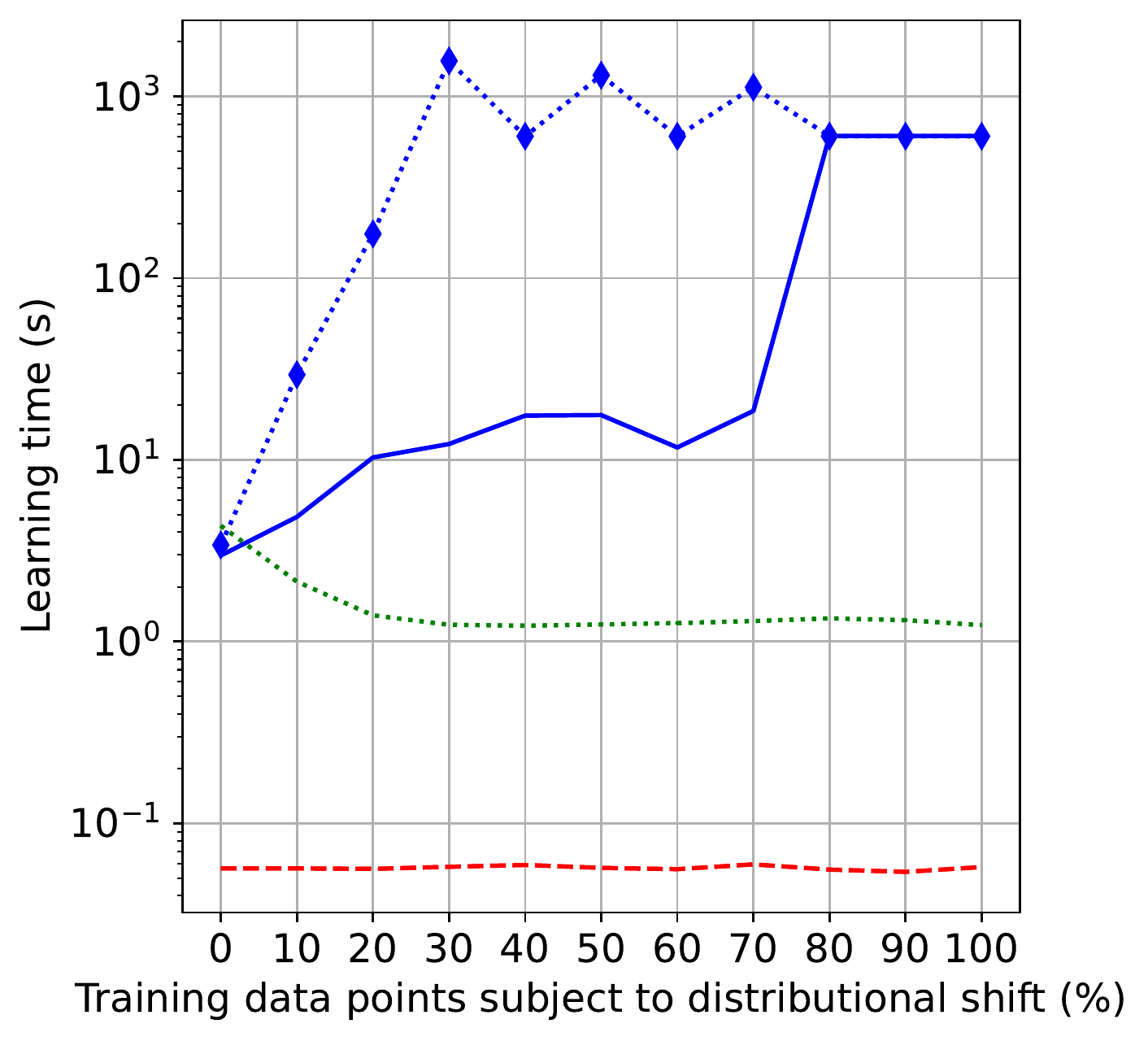}
		\subcaption{Indoor Scene Classification}
		\label{fig:scene_lt}
	\end{subfigure}
	
	\caption{Hypothesis learning time}
	\label{fig:real_world_lt}
\end{figure}

Figure \ref{fig:real_world_app_learned_hyp_acc} shows the accuracy of the learned hypotheses in each task, when an increasing percentage of labelled unstructured data were subject to distributional shift. In the Crop Yield Prediction task (Figure \ref{fig:crop_lh_acc}), the reported accuracy is the mean accuracy over 5 repeats, and the error bars indicate standard error. In this task, both instances of \ac{nsl} learned accurate hypotheses until 90\% of the data were subject to distributional shift, and outperformed all the baseline approaches, even when the baselines were trained with significantly more data. In the Indoor Scene Classification task (Figure \ref{fig:scene_lh_acc}), \ac{nsl} also outperformed the baseline approaches, learning the correct hypothesis up to 50\% of the data subject to distributional shift. Setting the weight penalties for the examples based on the neural network confidence scores led to more accurate hypotheses, compared to using constant penalties. This is because using the neural network-based weight penalties enabled FastLAS to find a better optimal solution within the 10 minute timeout, as the more informative weight penalties gave a clearer optimisation signal for the final solving stage. With constant weight penalties, the optimisation took significantly longer as the distributional shift increased (see Figure \ref{fig:scene_lt}).

\begin{figure}[H]
	\centering
	\begin{subfigure}{0.51\textwidth}
		\centering
		\includegraphics[width=1\linewidth]{Figures/follow_suit_results/follow_suit_nn_penalties_legend.pdf}
	\end{subfigure}
	
	\begin{subfigure}{.3\columnwidth}
		\centering
		\includegraphics[width=0.8\linewidth]{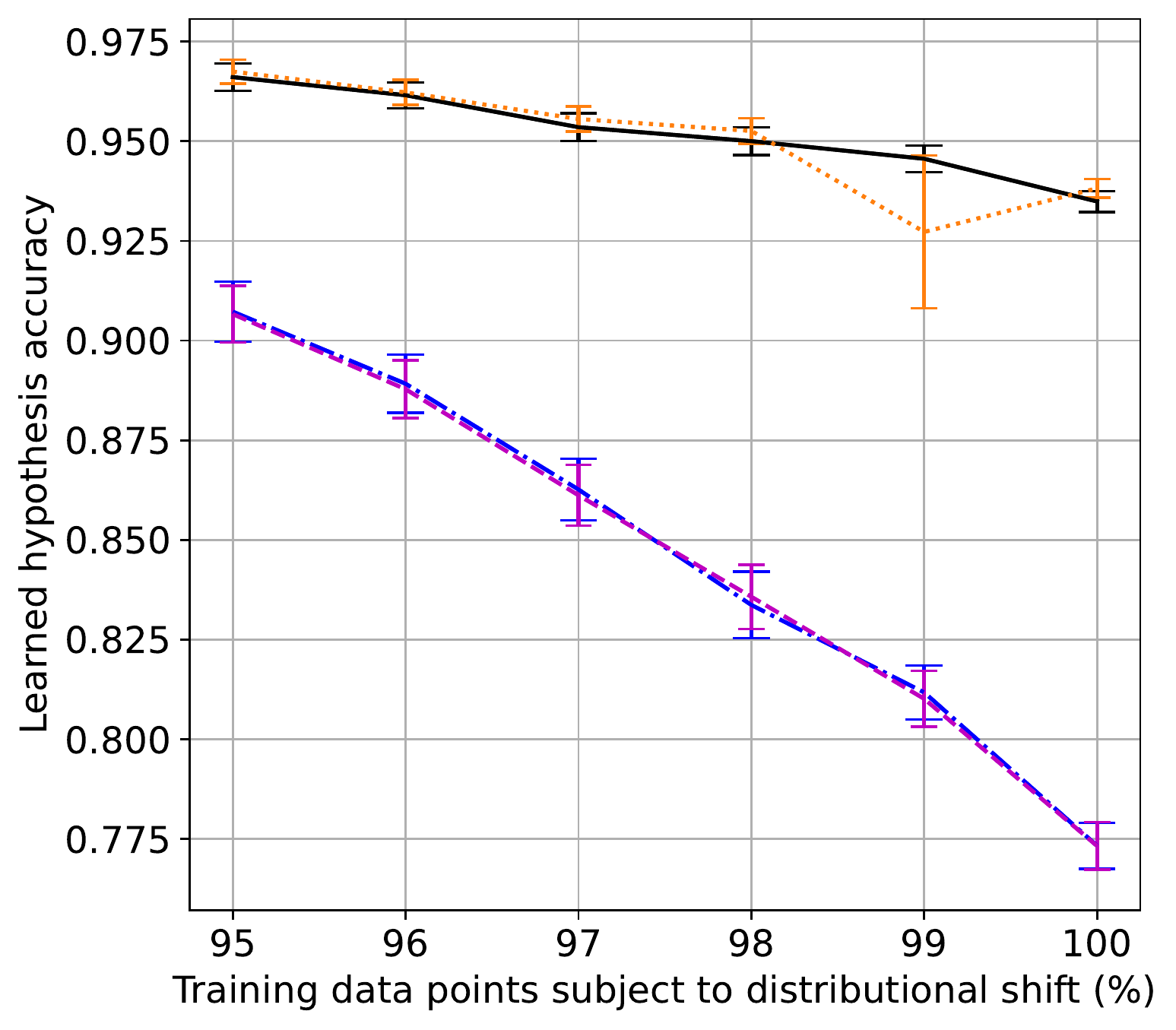}
		\subcaption{Hypothesis accuracy}
		\label{fig:crop_lh_acc_50_repeats}
	\end{subfigure}
	\begin{subfigure}{.3\columnwidth}
		\centering
		\includegraphics[width=0.8\linewidth]{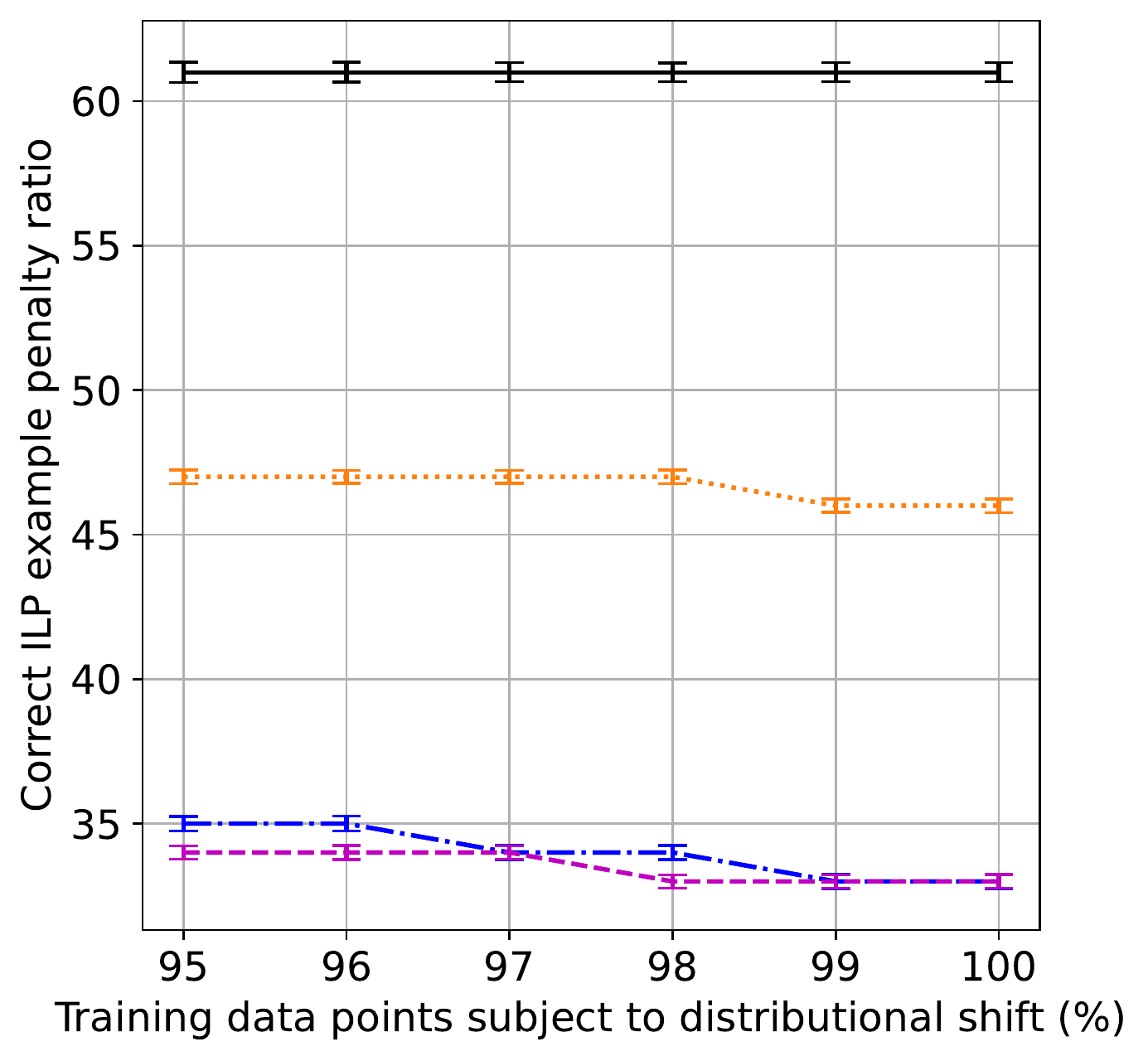}
		\subcaption{Weight penalty ratio}
		\label{fig:crop_wpr}
	\end{subfigure}
	\begin{subfigure}{.3\columnwidth}
		\centering
		\includegraphics[width=0.8\linewidth]{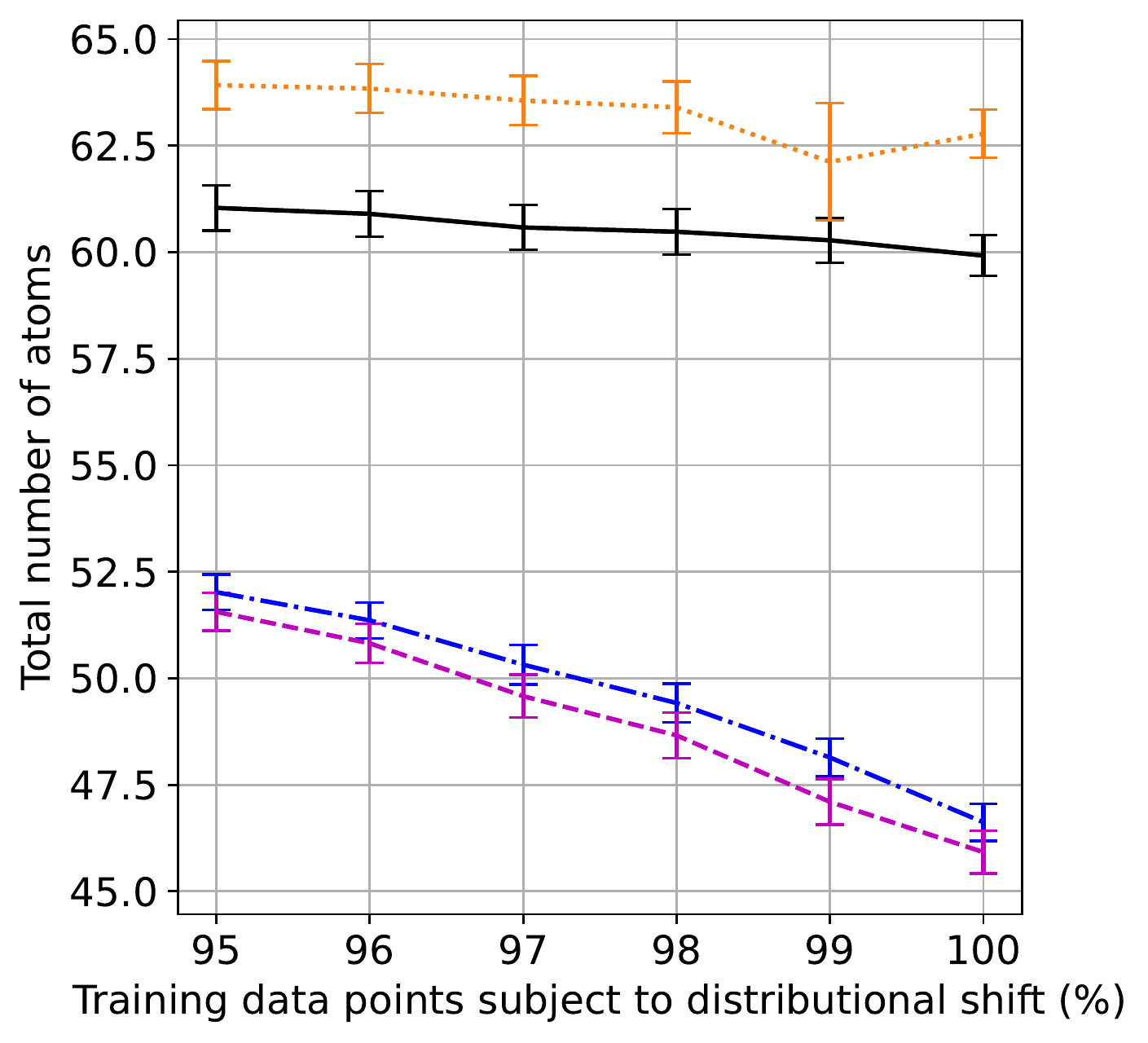}
		\subcaption{Hypothesis length}
		\label{fig:crop_hyp_length}
	\end{subfigure}
	
	\caption{The effect of setting ILP example weight penalties based on neural network confidence scores, compared to constant penalties, Crop Yield Prediction task. 95-100\% shifts, 50 repeats.}
	\label{fig:crop_penalty_comparison}
\end{figure}

Exploring deeper the effect of using example weight penalties set by neural network confidence scores, compared to using constant penalties, we ran 50 experimental repeats between 95-100\% shifts. Figure \ref{fig:crop_penalty_comparison} shows the accuracy, weight penalty ratio, and hypothesis length comparison for the Crop Yield Prediction task.

The EDL-GEN instances of \ac{nsl} clearly outperformed the Softmax \ac{nsl} instances (see Figure \ref{fig:crop_lh_acc_50_repeats}). This is because the EDL-GEN neural network predicted with greater accuracy than Softmax when distributional shifts were applied (see Figure \ref{fig:crop_nn_perf}). However, setting the weight penalties of the examples for the symbolic learner based on neural network confidence scores made very little difference in the \ac{nsl} EDL-GEN instances. Therefore, we investigated the weight penalty ratio (Figure \ref{fig:crop_wpr}). As expected, both \ac{nsl} Softmax instances had a similar weight penalty ratio, due to the Softmax neural network predicting with high confidence when distributional shift was applied (see Figure \ref{fig:crop_nn_perf}). The EDL-GEN instances do however show a difference, and the neural network-based weight penalties did provide a more informative signal. The question therefore, is why did this not translate into an improvement in learned hypothesis accuracy? It turns out that the benefit was realised in the length of the learned hypothesis (Figure \ref{fig:crop_hyp_length}), as \ac{nsl} EDL-GEN with neural network weight penalties learned a shorter hypothesis than when constant penalties were used. Comparing Figure \ref{fig:crop_hyp_length} to Figure \ref{fig:crop_lh_acc_50_repeats}, you can see that at 99\% shifts, when the accuracy of \ac{nsl} EDL-GEN with constant penalties decreased, the length of the learned hypotheses also decreased, whilst \ac{nsl} EDL-GEN with neural network weight penalties achieved a higher accuracy with a shorter hypothesis. With constant penalties, to account for the level of noise, the symbolic learner had to learn more rules that map additional values of location type, plant species and disease to crop yield, in order to maintain the same level of accuracy as when neural network-based penalties were used.

\begin{figure}[H]
	\centering
	\begin{subfigure}{0.49\textwidth}
		\centering
		\includegraphics[width=0.8\linewidth]{Figures/plant_disease/results/plant_disease_legend.pdf}
	\end{subfigure}
	\begin{subfigure}{0.49\textwidth}
		\centering
		\includegraphics[width=0.7\linewidth]{Figures/indoor_scenes/results/indoor_scene_legend.pdf}
	\end{subfigure}
	
	\begin{subfigure}{.49\columnwidth}
		\centering
		\includegraphics[width=0.8\linewidth]{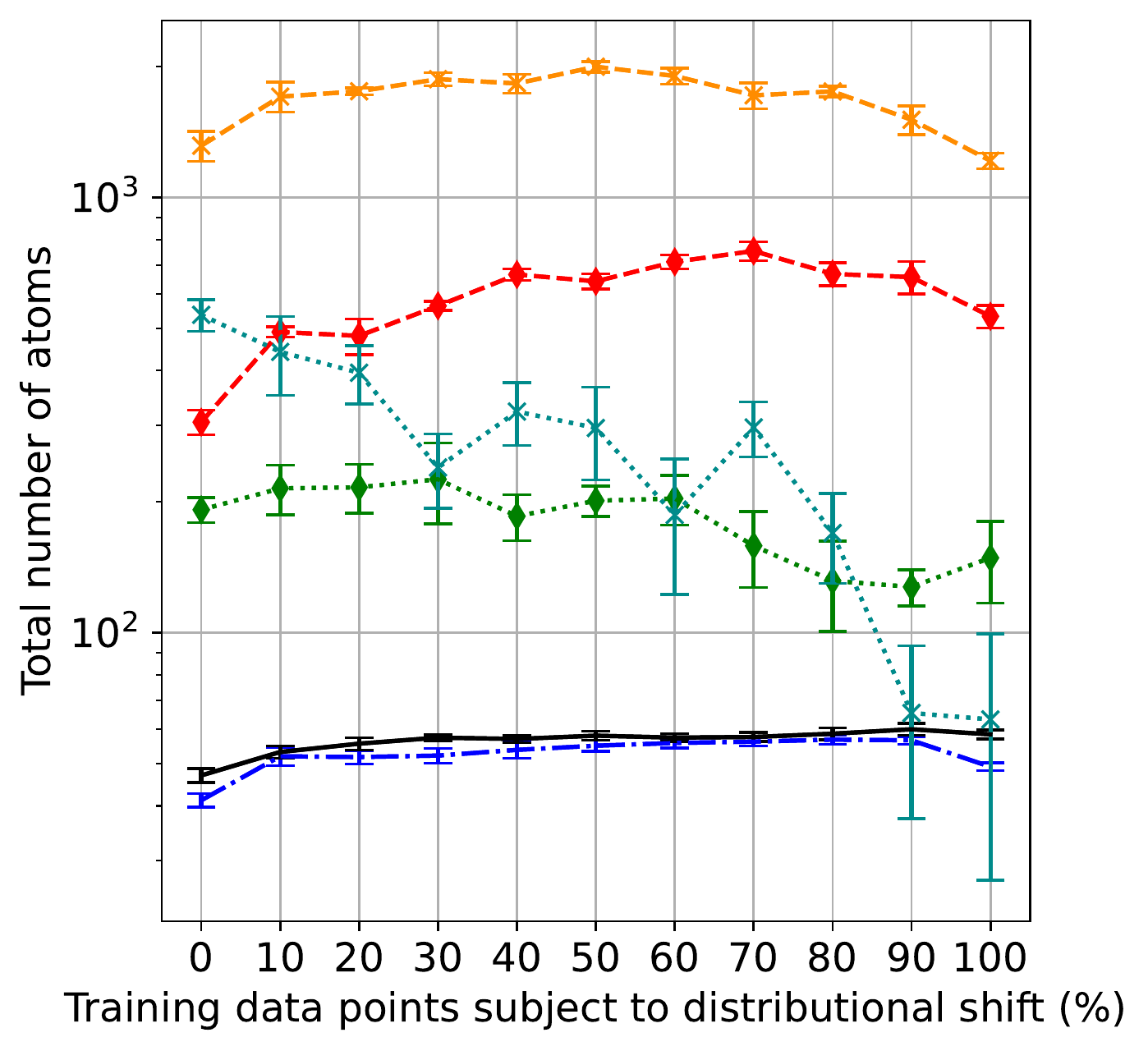}
		\subcaption{Crop Yield Prediction}
		\label{fig:crop_interp}
	\end{subfigure}
	\begin{subfigure}{.49\columnwidth}
		\centering
		\includegraphics[width=0.8\linewidth]{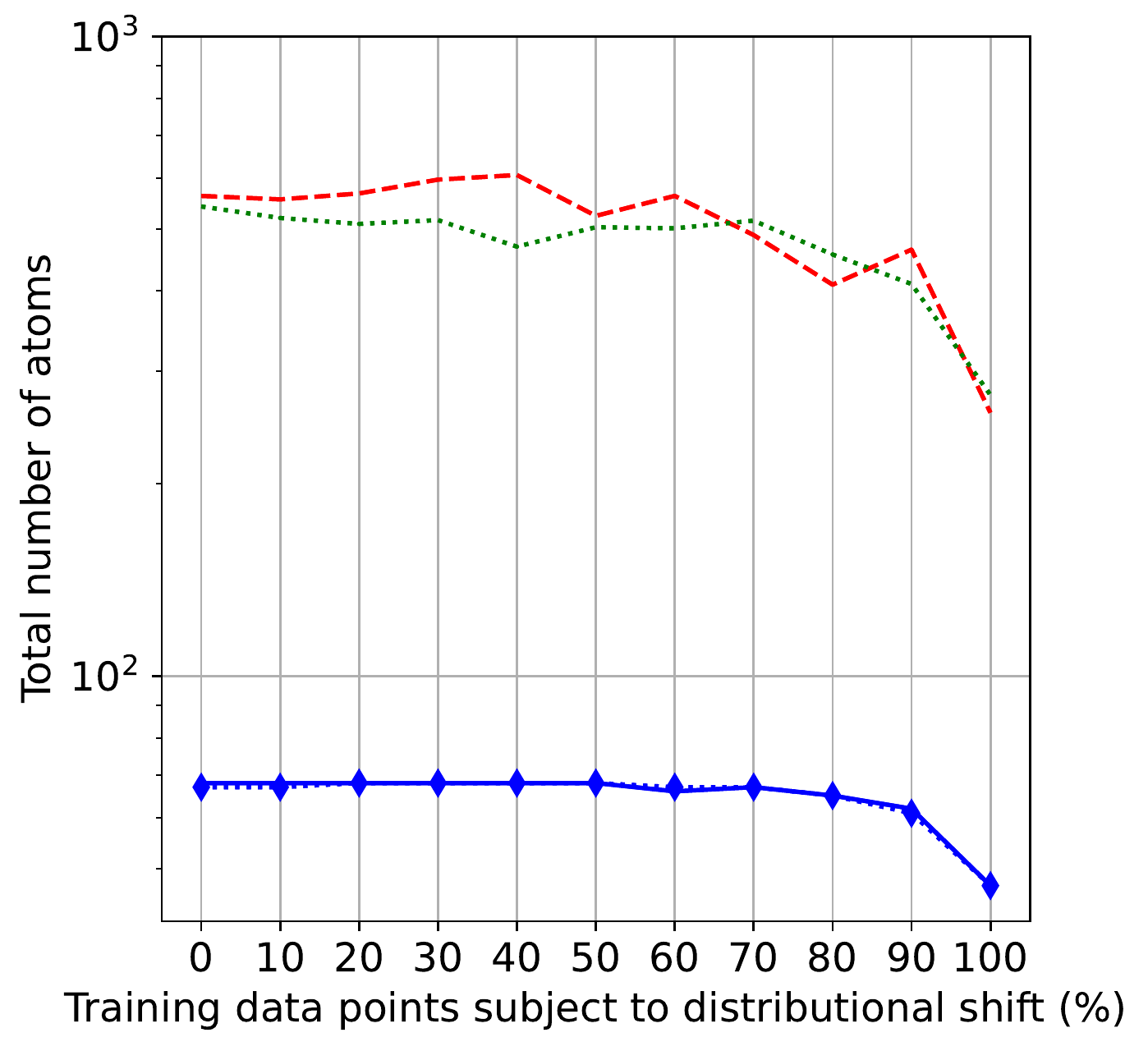}
		\subcaption{Indoor Scene Classification}
		\label{fig:scene_interp}
	\end{subfigure}

	\caption{Interpretability of the learned hypotheses.}
	\label{fig:real_world_interp}
\end{figure}

Finally, Figures \ref{fig:real_world_interp} and \ref{fig:real_world_lt} present the interpretability and learning time results for both tasks. \ac{nsl} learned significantly more interpretable hypotheses than the baseline approaches in both tasks. In terms of learning time, \ac{nsl} learned a hypothesis faster than the baselines trained with more examples in the Crop Yield Prediction task, and was slower than the baselines in the Indoor Scene Classification task. Figure \ref{fig:scene_lt} clearly shows the computational benefit of setting \ac{ilp} example weight penalties based on neural network confidence scores, as a hypothesis was learned significantly faster than when constant penalties were used. The near constant learning times at 80-100\% shifts for \ac{nsl} SASR with neural network-based penalties, and 30-100\% shifts with constant penalties, was due to the 10 minute timeout imposed on each FastLAS learning task.

\subsection{\ac{nsl} framework evaluation}
\begin{figure}[H]
    \centering
	
	\begin{subfigure}{0.49\textwidth}
		\centering
		\includegraphics[width=0.8\linewidth]{Figures/plant_disease/results/plant_disease_legend.pdf}
	\end{subfigure}
	\begin{subfigure}{0.49\textwidth}
		\centering
		\includegraphics[width=0.7\linewidth]{Figures/indoor_scenes/results/indoor_scene_legend.pdf}
	\end{subfigure}

	\begin{subfigure}{.49\columnwidth}
		\centering
		\includegraphics[width=0.8\linewidth]{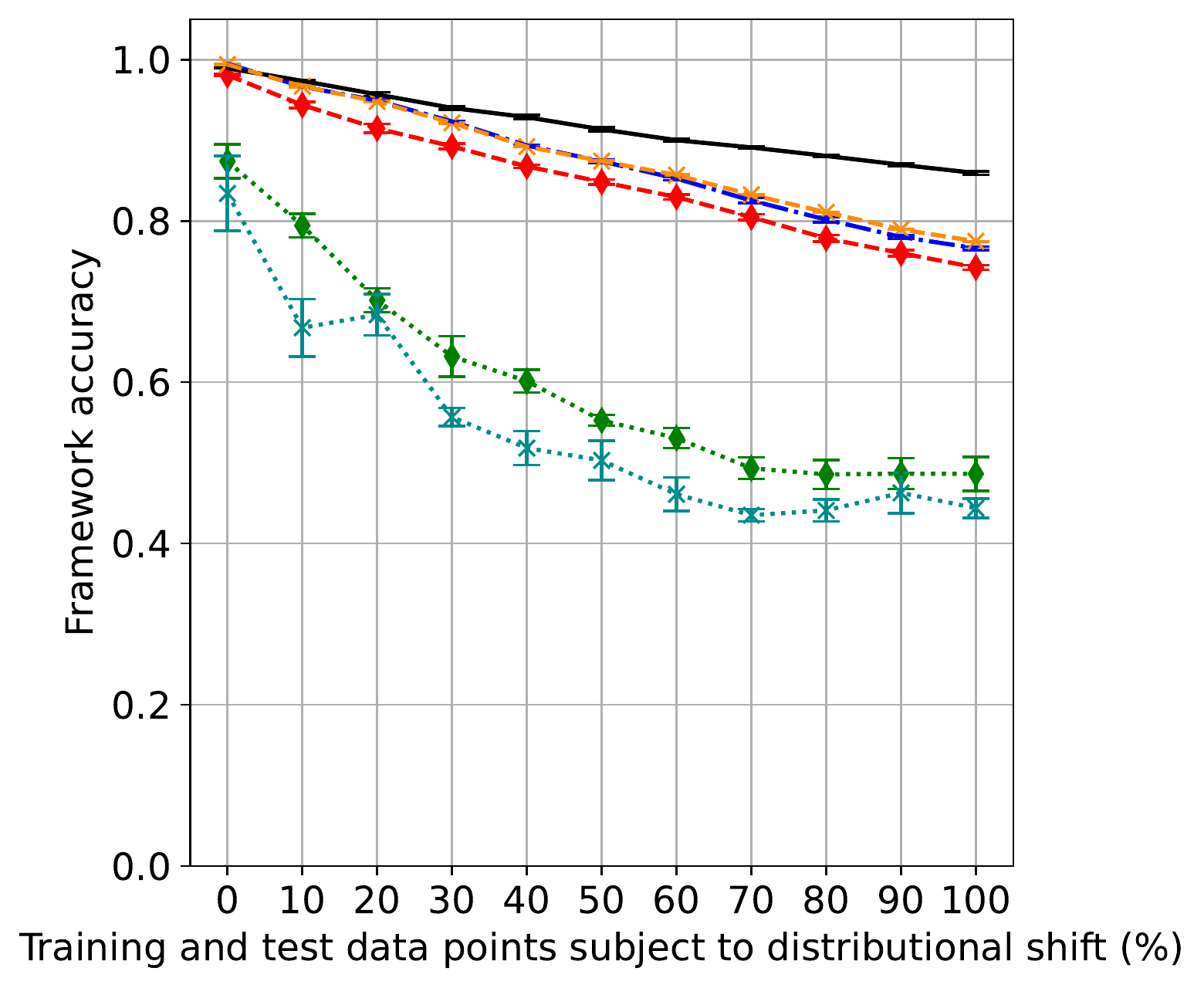}
		\subcaption{Crop Yield Prediction}
		\label{fig:crop_framework_eval}
	\end{subfigure}
	\begin{subfigure}{.49\columnwidth}
		\centering
		\includegraphics[width=0.8\linewidth]{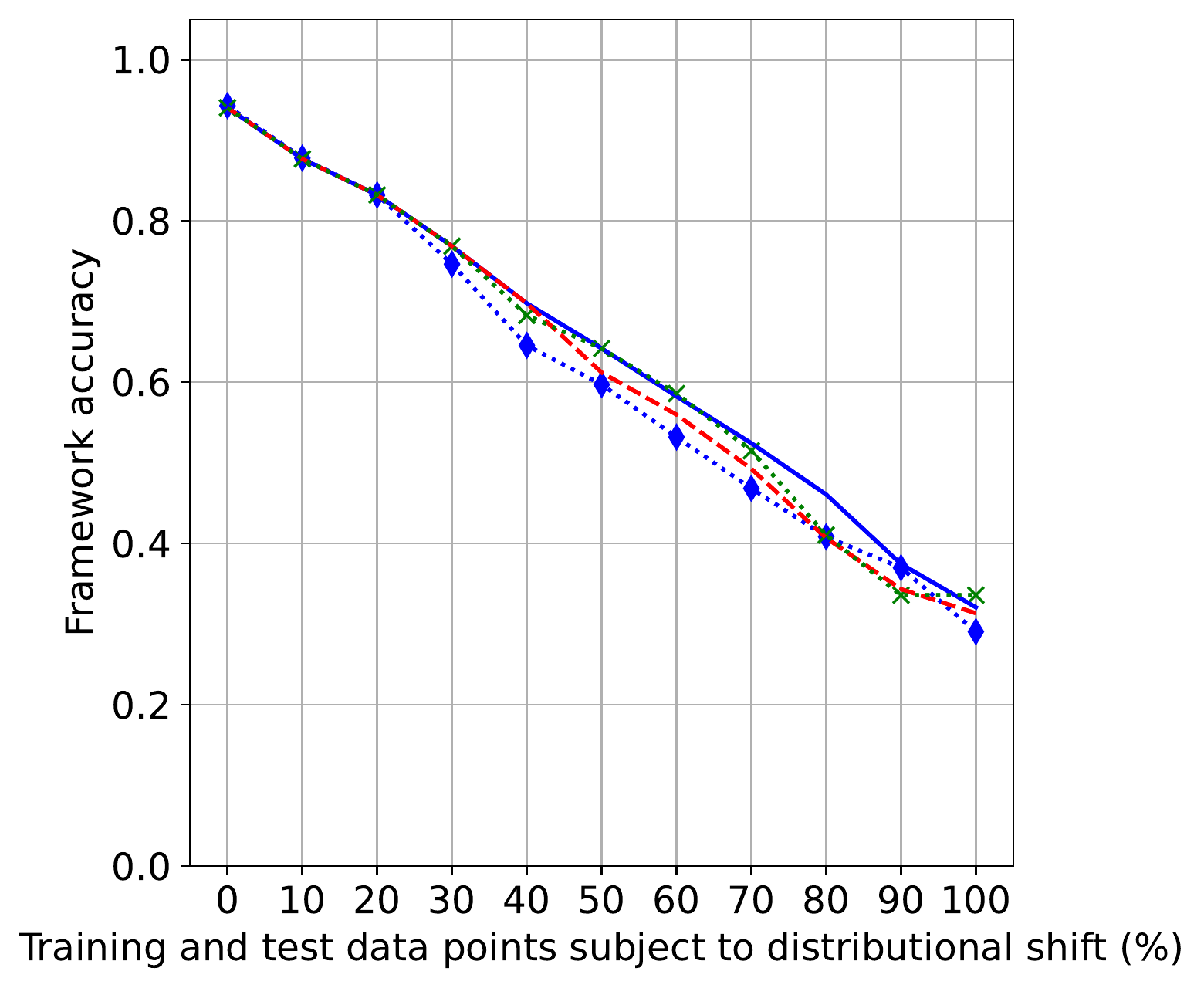}
		\subcaption{Indoor Scene Classification}
		\label{fig:scene_framework_eval}
	\end{subfigure}
	
	\caption{Accuracy of the \ac{nsl} framework when training and test data were subject to distributional shift.}
	\label{fig:real_world_framework_results}
\end{figure}

Figure \ref{fig:real_world_framework_results} presents the accuracy of the entire \ac{nsl} framework when evaluated over test data also subject to the same percentage of distributional shift as used during learning.

In the Crop Yield Prediction task (Figure \ref{fig:crop_framework_eval}), \ac{nsl} EDL-GEN outperformed all other methods, and \ac{nsl} Softmax outperformed all other methods trained with the same amount of data. The next best approach, the random forest, required significantly more data to match the performance of \ac{nsl} Softmax. The superior performance of \ac{nsl} EDL-GEN compared to \ac{nsl} Softmax was due to the EDL-GEN neural network predicting more accurately for images subject to distributional shift (see Figure \ref{fig:crop_nn_perf}). In the Indoor Scene Classification task, \ac{nsl} performed similarly to the best baseline approach, and all approaches degraded gracefully as the percentage of data points subject to distributional shift increased. 

To conclude, this evaluation of \ac{nsl} to real-world datasets shows that the framework can support a wide range of neural modules, and the \ac{d2k} component is flexible enough to support the interface between different neural and symbolic modules. When taking into account the Follow Suit Winner and Sudoku Grid Validity results, we have also shown that \ac{nsl} can learn complex, first-order symbolic knowledge, using essential aspects of common-sense learning and reasoning such as negation as failure and predicate invention. In the next section, we discuss related work before concluding the paper.



\section{Related Work}\label{sec:related_work}
Our proposed \ac{nsl} approach is a specific form of a neural-symbolic learning and reasoning system that, differently from other neural-symbolic methods, uses pre-trained neural networks and logic-based machine learning systems to learn interpretable, logic-based knowledge from unstructured data that can be used to solve a given task.  Most of the recently proposed neural-symbolic approaches focus on ways in which a given fixed knowledge can be used to improve the training of a neural network~\cite{serafini2016logic,donadello2017logic, riegel2020logical,manhaeve2018deepproblog,ijcai2020-243,tsamoura2021neural}. These approaches leverage the notions of Real Logic~\cite{serafini2016logic,donadello2017logic} or t-norm functions~\cite{FM2006} to enable the injection of logical reasoning in data-driven relational machine learning. This is the case, for instance, of the Logic Tensor Network approaches proposed in \cite{serafini2016logic, donadello2017logic}. Our \ac{nsl} approach also uses a similar notion of t-norms but not to embed logic into the differentiable setting, rather to ``combine'' neural network predictive approximations with logic-based learning optimisation so enabling the composition of these two different machine learning paradigms. 

Neural-symbolic approaches that preserve the composition of neural and symbolic inference include DeepProbLog~\cite{manhaeve2018deepproblog}, NeurASP~\cite{ijcai2020-243} and NeuroLog \cite{tsamoura2021neural}. They compose deep learning architectures with symbolic reasoning in order to use existing background knowledge, expressed as logic programs, to train deep learning models. DeepProbLog~\cite{manhaeve2018deepproblog} uses ProbLog~\cite{de2007problog} to interpret network outputs as probabilistic atoms, and symbolic knowledge compiled into an arithmetic circuit, to train the network. NeurASP extends \ac{asp} with neural predicates, expressed as choice rules, to symbolically capture possible network outputs. The probability of each model of the \ac{asp} program is computed based on the network predictions, which is in turn used to optimise a semantic loss function for training the network~\cite{Xu18}. NeuroLog also trains the neural network using a semantic loss function, although uses abduction to prune the space of possible pseudo-label revisions for the latent concepts, instead of considering all possibilities as in NeurASP. Although compositional in their architectural solution, and novel in their end-to-end approach for differentiable training of the neural networks, these methods require the logic-based knowledge to be manually engineered. Our \ac{nsl} approach, on the other hand, enables the learning of logic-based knowledge from unstructured data exploiting pre-trained neural models. The semantics of the underlying logic-based learning algorithm in \ac{nsl} is the Answer Set semantics, as it is the case for the symbolic component of the NeurASP system, but with the advantage in \ac{nsl} that knowledge expressed in~\ac{asp} programs is learned instead of being fully encoded as input. 

Contrary to the end-to-end feature of DeepProbLog and NeurASP neural-symbolic systems, our \ac{nsl} adopts a pipeline approach. It is therefore somewhat related to the Concept Bottleneck Model architecture proposed in \cite{koh2020concept}, which advocates the idea of training first a model to predict ``primary'' concepts and then using these concepts to train a downstream model for predicting the labels. These models are however differentiable and even though they can be trained in an end-to-end fashion to improve the overall accuracy~\cite{koh2020concept}, the trained downstream model is not interpretable. Their interpretability is limited to extracting correlations between the primary concepts and the final label. In our \ac{nsl} approach, the use of LAS logic-based machine learning systems allow the learning of knowledge that is fully interpretable and that is more robust to distributional shifts and noise in the data. In fact, the~\ac{cnnlstm} and~\ac{fcn} baselines used in our tasks could be considered as independent concept bottleneck models, and \ac{nsl} outperformed both of these models in our evaluation. 

The compositional aspect of our framework could, in principle, make it amenable to instantiations where the symbolic component is a probabilistic rule learning system. Different probabilistic rule learning and statistical relational learning systems have been proposed, such as ProbFOIL \cite{DeRaedtLuc2015Iprr}, SLIPCOVER \cite{slipcover}, Markov Logic Networks \cite{richardson2006markov} and Credal-FOIL \cite{tuckeytowards}. They adopt a probabilistic notion of uncertainty which is different from the notion of~\ac{wcdpi} example weight penalties used in our \ac{nsl} approach. Such systems would, however, make \ac{nsl} not applicable to tasks where non-observational predicate learning with negation as failure is required, like our Follow Suit Winner task, and limit its scalability. This is because it still remains to be shown whether current probabilistic rule learning systems are scalable to a large number of probabilistic facts and large hypothesis search spaces.  

Related approaches that support the learning of interpretable knowledge from (unstructured) data in a neural-symbolic manner include $\delta$ILP~\cite{evans2018learning}, and NeuralLP~\cite{YangYC17}. They make use of rule templates and differentiable reasoning to approximate the inference process and learn instances of the rule templates that cover given labelled examples or to answer given queries. Such approaches, preserve the symbolic, logic-based representation of the knowledge, but replace the logic-based inference process with a purely differentiable one. Our \ac{nsl} approach uses instead a pure symbolic inference process to learn interpretable knowledge, leveraging on state-of-the-art logic-based machine learning systems such as ILASP and FastLAS. The composition of these systems with differentiable feature extraction from unstructured data enables \ac{nsl} to learn knowledge that is more expressive than the definite clausal form supported by $\delta$ILP and NeuralLP, broadening the applicability of \ac{nsl} to real-world problems where non-monotonicity and preference learning are required. Results in \cite{law2018inductive} have already demonstrated that, in the case of structured data, the ILASP system used by our \ac{nsl} framework outperforms $\delta$ILP when learning interpretable knowledge from noisy examples. 

Neural-symbolic systems such as Neural-Theorem Prover \cite{Rocktaschel017} and its extensions, adopt instead a counterpart approach whereby knowledge is expressed as dense vector embedding representations that are learned in a differentiable manner by using a symbolically inspired backward chaining algorithm and (soft) unification. In these systems, the knowledge is represented in a high-dimensional differentiable space and the inference is symbolically inspired. More recently, a fully differentiable rule induction approach based on Logical Neural Networks has been proposed \cite{grey2021} that uses differentiable operators from fuzzy and real logic to learn rules from structured data within a very controlled search space expressed using templates. Although some of these systems have recently shown to be somewhat scalable over large knowledge bases~\cite{Minervini0SGR20, MinerviniBR0G20}, they are all limited in the expressivity of the knowledge that they can learn and they are not guaranteed to learn (mathematically provable) optimal solutions. These are two main properties that our \ac{nsl} framework instead benefits from, making our approach particularly suited for safe and trusted AI applications where data are unstructured, complex, and interpretable knowledge is required to be learned to solve complex tasks.

Recent approaches train a neural network to extract primary concepts from raw data, whilst learning interpretable symbolic knowledge in an end-to-end fashion \cite{dai2019bridging,ijcai2021-254}. These methods don't require labels for the primary concepts, and train a neural network from scratch whilst simultaneously learning knowledge. The Abductive Learning framework (ABL) \cite{dai2019bridging} learns ground operation facts that complete a symbolic knowledge base, to map neural network outputs to downstream labels. This knowledge is then used to abduce revised pseudo-labels to improve the training of the neural network. Crucially, \cite{dai2019bridging} cannot perform program induction, and assumes monotonicity of the background knowledge, as ground operation facts are abduced and accumulated during an iterative sampling process over the training data. In contrast, our approach learns first-order rule-based programs, which contain universally quantified variables, and are therefore applicable to a range of input sizes greater than the sizes used for training. We can also handle non-monotonicity, thus enabling the learning of more complex knowledge. The $Meta_{Abd}$ approach \cite{ijcai2021-254} extends \cite{dai2019bridging} to perform rule induction using the Metagol symbolic learner \cite{Muggleton13}. 
The key drawback of $Meta_{Abd}$ is that Metagol can only learn symbolic knowledge expressed as definite logic programs without function symbols, which can compute only polynomial functions \cite{dantsin2001complexity}. $Meta_{Abd}$ cannot learn more expressive knowledge involving defaults, exceptions, constraints and choice, which are essential aspects of common-sense learning and reasoning. In \ac{nsl}, we learn first-order complex knowledge expressed in the language of \ac{asp}, which is more general than symbolic learning of definite clauses~\cite{Law2018thesis,law2020fastlas,LawAAAI}, and can solve computationally harder problems \cite{karp1972reducibility}. Also, due to the high level of difficulty of such an end-to-end neuro-symbolic task, $Meta_{Abd}$ has only been applied to very simple classification problems. Our architecture is motivated by a completely different requirement, that of using already trained and therefore possibly much more complex neural components for extracting features from challenging raw data. 

\section{Conclusion}\label{sec:conclusion}
This paper introduces a neural-symbolic learning framework, \textit{\ac{nsl}}, that learns interpretable knowledge from unstructured data that is robust to distributional shifts. Three main instantiations of this framework have been presented, which use the ILASP and FastLAS logic-based machine learning systems, according to the type of symbolic learning task required. In each instantiation, pre-trained neural networks have been used for extracting symbolic features from the unstructured data. The novel component of \ac{nsl} is the~\ac{d2k} generator, which generates symbolic features, weighted by neural network confidence scores, that together with a label, form the input to the logic-based machine learning system which then learns interpretable knowledge needed to solve the given downstream task.

Our evaluation on four neural-symbolic classification tasks, Follow Suit Winner, Sudoku Grid Validity, Crop Yield Prediction and Indoor Scene Classification, demonstrates that \ac{nsl} is robust to distributional shifts in the input data, outperforming random forest and deep neural network baselines. \ac{nsl} learns more accurate and interpretable knowledge than the baselines even when the latter are trained with significantly more data. The application of \ac{nsl} learned knowledge to unseen data also subject to similar proportions of distributional shifts shows that \ac{nsl} is again capable of outperforming the baseline approaches trained with the same amount of data up to $\sim$80\% of data subject to distributional shifts. A detailed analysis of the performance in accuracy of our \ac{nsl} framework shows that using an uncertainty-aware neural network provides an improved bias to the logic-based machine learning system compared to Softmax neural networks, with a greater proportion of the total weight penalty allocated to~\ac{wcdpi} examples containing correct contextual information extracted from the unstructured data.  

\section*{Acknowledgements}
This research was sponsored by the U.S. Army Research
Laboratory and the U.K. Ministry of Defence under Agreement
Number W911NF-16-3-0001. The views and conclusions
contained in this document are those of the authors and should
not be interpreted as representing the official policies, either
expressed or implied, of the U.S. Army Research Laboratory,
the U.S. Government, the U.K. Ministry of Defence or the
U.K. Government. The U.S. and U.K. Governments are
authorized to reproduce and distribute reprints for Government
purposes notwithstanding any copyright notation hereon.

\section*{Statements and Declarations}
\textbf{Funding}. This research was sponsored by the U.S. Army Research
Laboratory and the U.K. Ministry of Defence under Agreement
Number W911NF-16-3-0001. \textbf{Conflicts of interest / competing interests}. Daniel Cunnington, Alessandra Russo and Jorge Lobo have no relevant financial or non-financial interests to disclose. Mark Law is the director of ILASP Limited, which owns the intellectual property of the ILASP system used in this paper. \textbf{Ethics approval}. Not applicable to this paper. \textbf{Consent to participate}. Not applicable to this paper as no humans were used to conduct the experimental evaluations. \textbf{Consent for publication}. Not applicable, all data, figures and tables are original and are generated synthetically, with the exception of the MNIST dataset~\cite{lecun1998gradient}. \textbf{Availability of data and material}. The Sudoku and Follow Suit Winner datasets introduced in this paper are available at the following GitHub repository: \url{https://github.com/DanCunnington/FFNSL}. \textbf{Code availability}. All the experimental code is also available at the GitHub repository. \textbf{Authors' contributions}. Daniel Cunnington defined the~\ac{nsl} method, performed the experimental evaluation and wrote the initial version of the paper. Mark Law provided support with running the ILASP and FastLAS systems and helped define the correct encoding for each task. Mark Law also suggested the Follow Suit Winner task and provided feedback on the final paper. Alessandra Russo and Jorge Lobo both equally contributed to the papers positioning, the generalised~\ac{nsl} method and gave suggestions for the experimental approach. Alessandra Russo and Jorge Lobo also contributed to the writing of the paper.

\bibliography{sn-bibliography}


\clearpage
\appendix



\section{Additional Follow Suit Winner Results}\label{sec:app:additional_fs_results}
In this section we present additional Follow Suit Winner results and analysis when the Captain America, Adversarial Standard and Adversarial Captain America decks were used to apply distributional shifts to the input data points. These results supplement the results and analysis presented in Section~\ref{sec:eval_fs_winner} for the Batman Joker and Adversarial Batman Joker decks. 

\subsection{Learned hypothesis evaluation}

\begin{figure}[ht]
\centering
\begin{subfigure}{1\columnwidth}
  \centering
  \includegraphics[width=.75\linewidth]{Figures/follow_suit_results/follow_suit_nn_penalties_legend.pdf}
  \label{fig:fs_50_repeats_legend_other_decks}
\end{subfigure}
\begin{subfigure}{.3\columnwidth}
  \centering
  \includegraphics[width=1\linewidth]{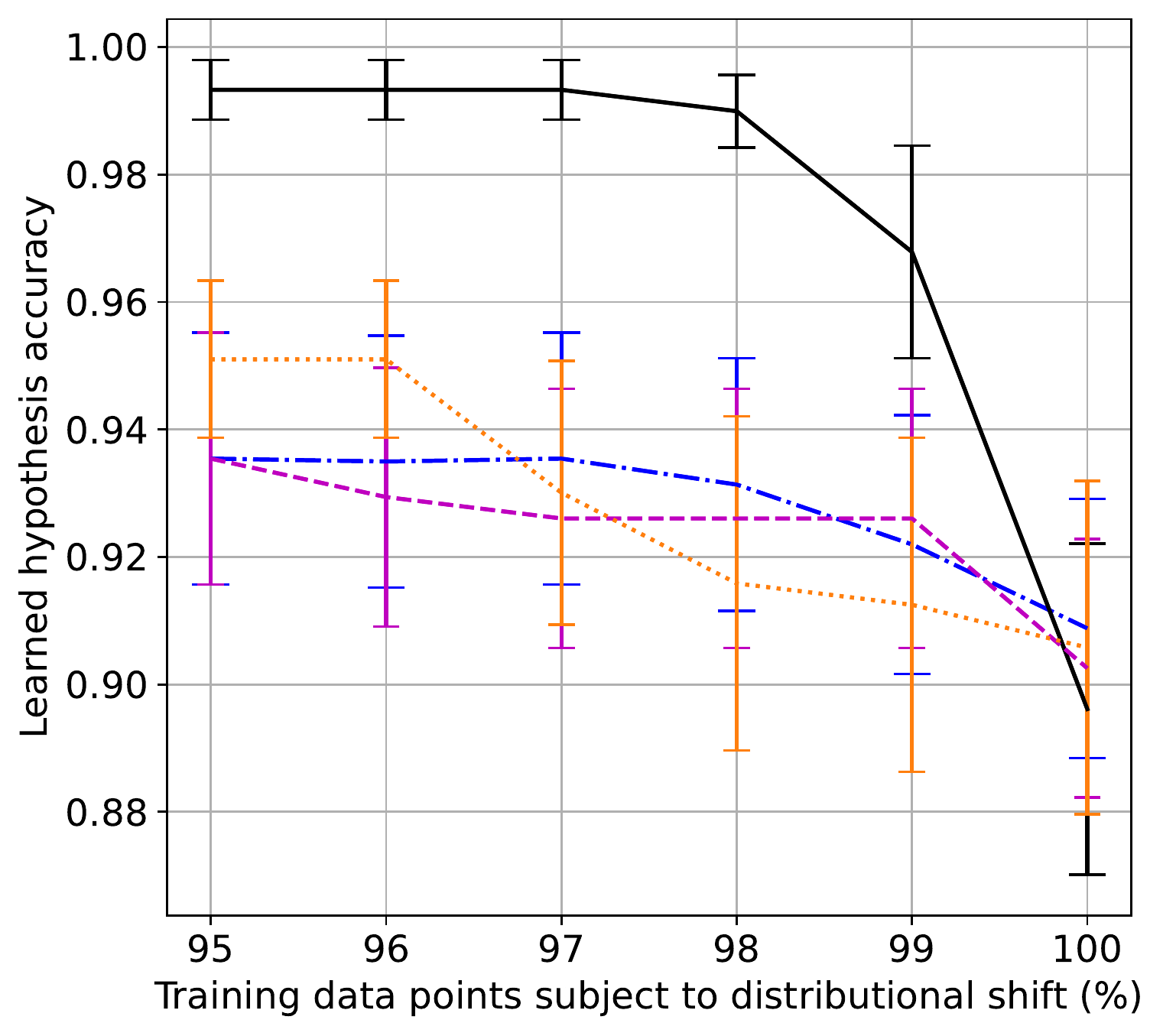}
  \caption{\centering Captain America}
  \label{fig:fs_ca_95_100}
\end{subfigure}%
\begin{subfigure}{.27\columnwidth}
  \centering
  \includegraphics[width=1\linewidth]{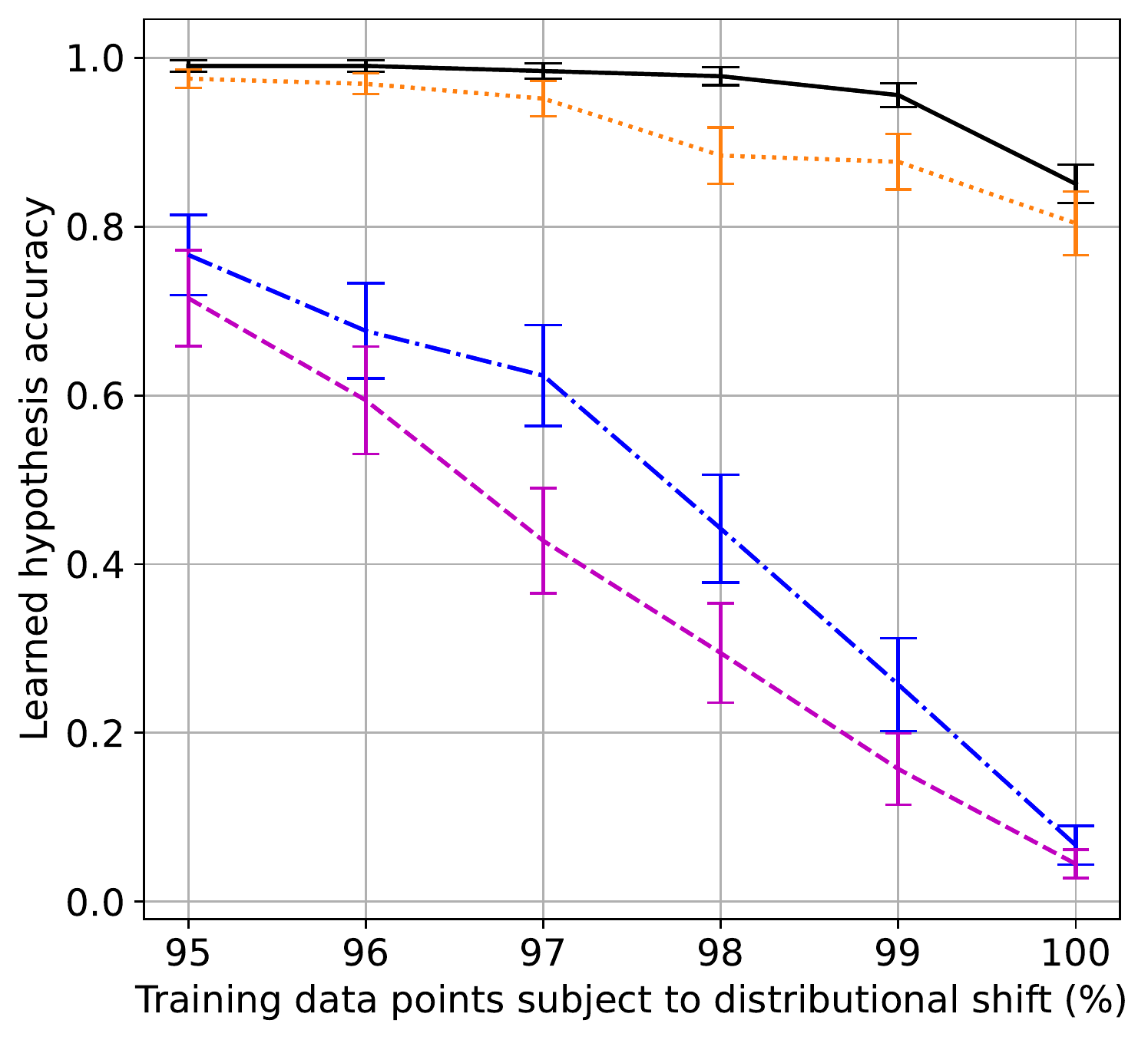}
  \caption{\centering Adversarial Standard}
  \label{fig:fs_as_95_100_acc}
\end{subfigure}
\begin{subfigure}{.27\columnwidth}
  \centering
  \includegraphics[width=1\linewidth]{Figures/follow_suit_results/adversarial_captain_america_follow_suit_structured_test_data_95_100_results.pdf}
  \caption{\centering Adversarial Captain America}
  \label{fig:fs_aca_95_100_acc}
\end{subfigure}
\caption{\ac{nsl} Softmax vs. \ac{nsl} EDL-GEN. Accuracy of learned hypotheses, 95-100\% distributional shifts, 50 repeats.}
\label{fig:fs_95_100_results_other_decks}
\end{figure}

Firstly, let us present the comparison of~\ac{nsl} EDL-GEN vs.~\ac{nsl} Softmax, for 95-100\% distributional shifts and 50 experimental repeats. The results are presented in Figure~\ref{fig:fs_95_100_results_other_decks}, which extend the results presented in Figure~\ref{fig:fs_95_100_results_bj_abj}.

\begin{figure}[H]
\centering

\begin{subfigure}{.3\columnwidth}
  \centering
  \includegraphics[width=1\linewidth]{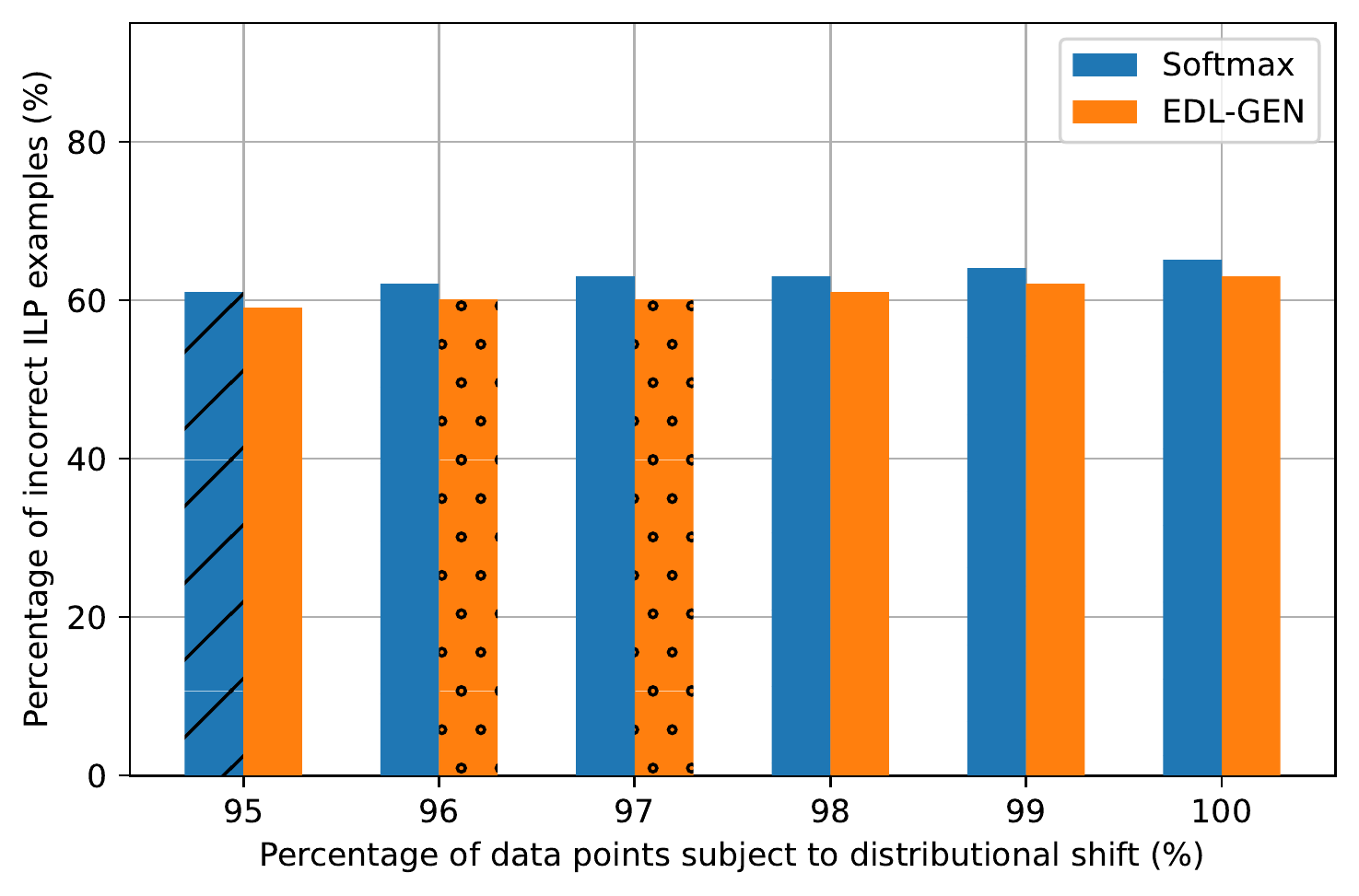}
  \caption{\centering Captain America}
  \label{fig:fs_ca_ilp_ex}
\end{subfigure}%
\begin{subfigure}{.27\columnwidth}
  \centering
  \includegraphics[width=1\linewidth]{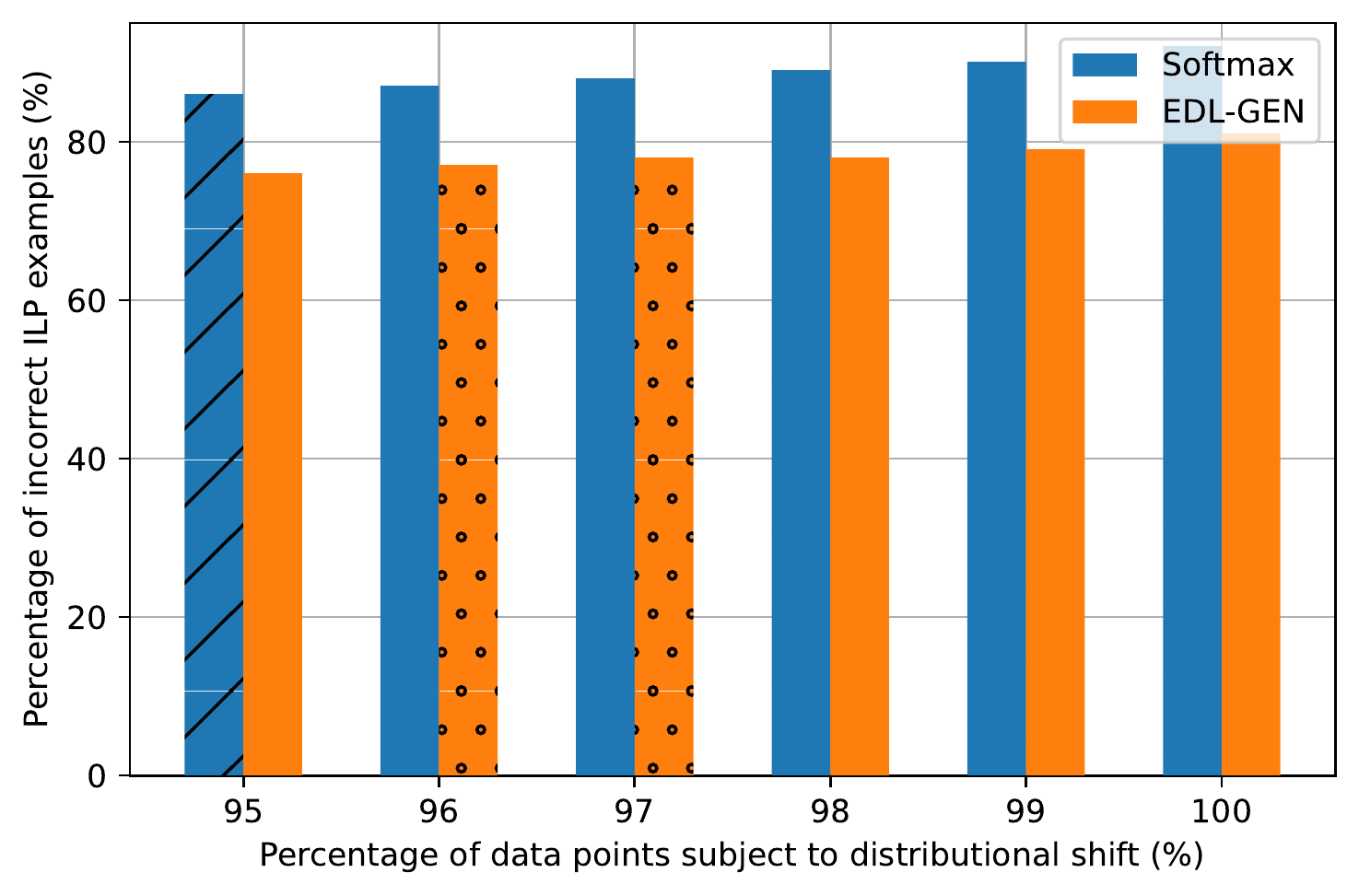}
  \caption{\centering Adversarial Standard}
  \label{fig:fs_as_ilp_ex}
\end{subfigure}
\begin{subfigure}{.27\columnwidth}
  \centering
  \includegraphics[width=1\linewidth]{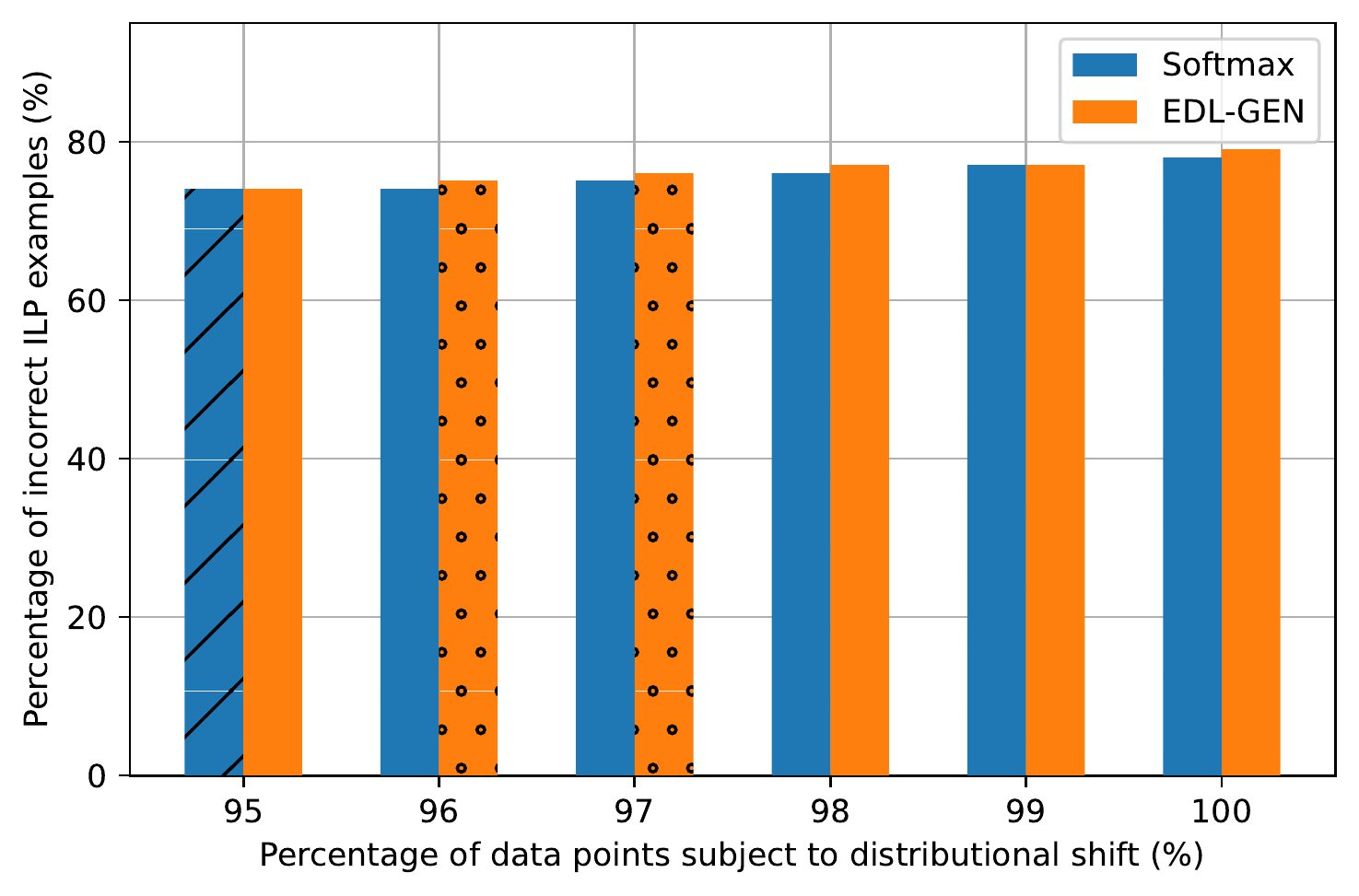}
  \caption{\centering Adversarial Captain America}
  \label{fig:fs_aca_ilp_ex}
\end{subfigure}
\caption{The effect of applying distributional shifts on the percentage of incorrect~\ac{ilp} examples, Follow Suit Winner task}
\label{fig:ilp_ex_anlaysis_other_decks}
\end{figure}

The results for the Captain America deck in Figure~\ref{fig:fs_ca_95_100} are very similar to the Batman Joker deck presented in Figure~\ref{fig:fs_bj_95_100_acc}, with the exception of~\ac{nsl} EDL-GEN with constant penalties that performed similarly to the~\ac{nsl} Softmax approaches. The Adversarial Standard deck results in Figure~\ref{fig:fs_as_95_100_acc} are very similar to the Adversarial Batman Joker results in Figure~\ref{fig:fs_abj_95_100_acc}, however, the Adversarial Captain America results in Figure~\ref{fig:fs_aca_95_100_acc} are different to the other decks. The two~\ac{nsl} Softmax approaches had much lower accuracy at 95\% shifts compared to the other decks, and there was a significant gap between~\ac{nsl} EDL-GEN with neural network penalties and~\ac{nsl} EDL-GEN with constant penalties. Let us now investigate each of these decks w.r.t. the percentage of incorrect~\ac{ilp} examples and the~\ac{ilp} example weight penalty ratios calculated from neural network confidence scores. The incorrect~\ac{ilp} example analysis is presented in Figure~\ref{fig:ilp_ex_anlaysis_other_decks}.

\begin{figure}[H]
\centering
\begin{subfigure}{.3\columnwidth}
  \centering
  \includegraphics[width=1\linewidth]{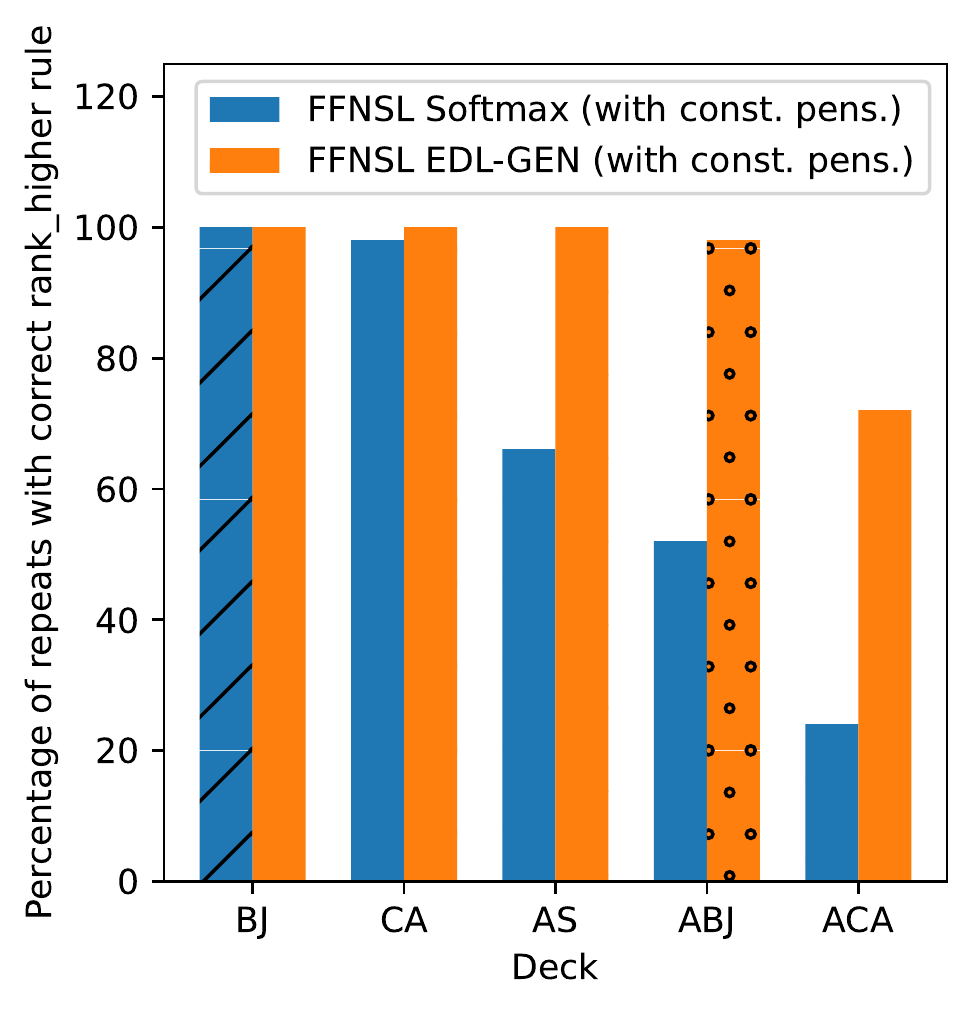}
   \caption{$\asp{rank\_higher}$ rule}
   \label{fig:correct_rank_higher_rule}
\end{subfigure}
\begin{subfigure}{.3\columnwidth}
  \centering
  \includegraphics[width=1\linewidth]{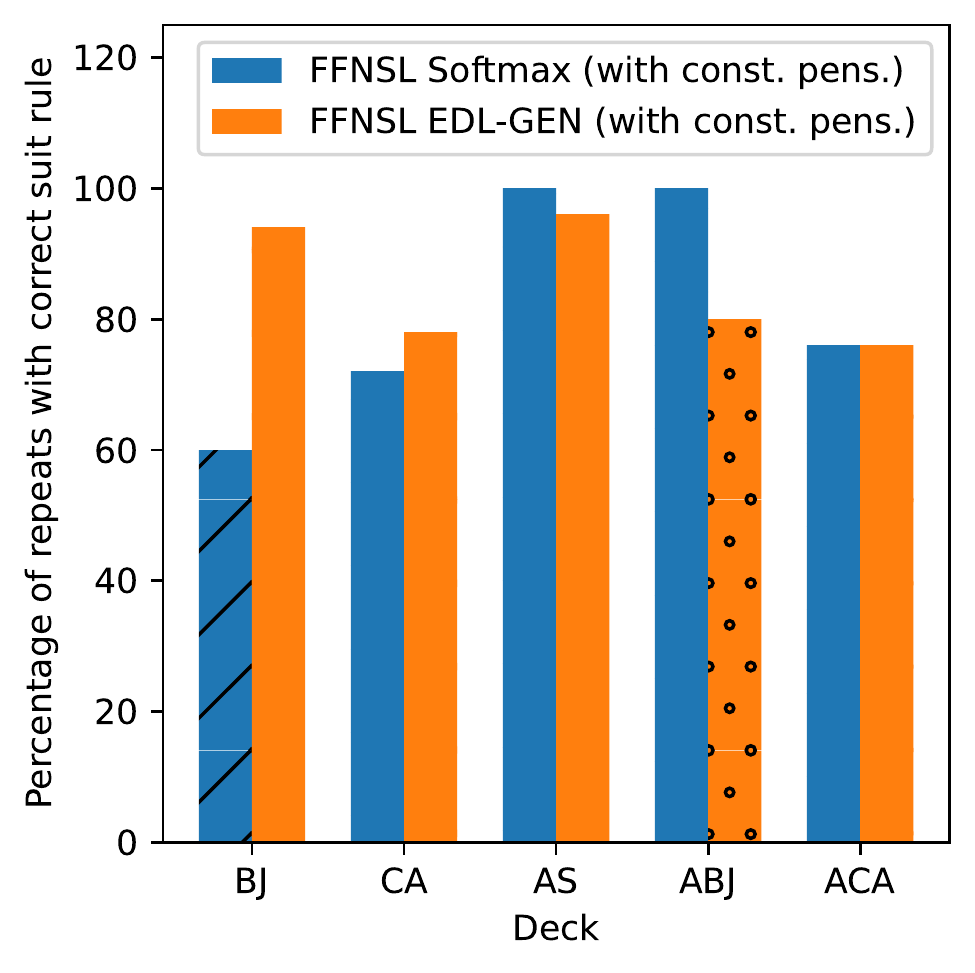}
   \caption{$\asp{suit}$ rule}
   \label{fig:correct_suit_rule}
\end{subfigure}
\caption{The percentage of experimental repeats that learned the correct follow suit rules, when 95\% of input data points were subject to distributional shifts.}
\label{fig:correct_rank_higher_and_suit_rules}
\end{figure}

Firstly, with the Captain America deck in Figure~\ref{fig:fs_ca_ilp_ex}, both Softmax and EDL-GEN had a very similar percentage of incorrect~\ac{ilp} examples, more similar than the other decks with the exception of Adversarial Captain America. This explains why~\ac{nsl} EDL-GEN with constant penalties performs similarly to~\ac{nsl} Softmax approaches in Figure~\ref{fig:fs_ca_95_100}. For the Adversarial Standard deck in Figure~\ref{fig:fs_as_ilp_ex}, the Softmax neural network resulted in a significantly higher percentage of incorrect~\ac{ilp} examples compared to EDL-GEN, which explains the large performance gap between~\ac{nsl} EDL-GEN and~\ac{nsl} Softmax approaches in Figure~\ref{fig:fs_as_95_100_acc}. For the Adversarial Captain America deck, using the EDL-GEN neural network resulted in a higher percentage of incorrect~\ac{ilp} examples in Figure~\ref{fig:fs_aca_ilp_ex}, yet~\ac{nsl} EDL-GEN clearly outperformed~\ac{nsl} Softmax in Figure~\ref{fig:fs_aca_95_100_acc}. To investigate this further, we look at the percentage of experimental repeats for both~\ac{nsl} approaches with constant penalties that learned the correct $\asp{rank\_higher}$ and $\asp{suit}$ rules (respectively; the winning player had a higher ranked card than other players, and the winning player also had the same suit as player 1), both of which are key to solving the task successfully. The analysis is presented in Figure~\ref{fig:correct_rank_higher_and_suit_rules} when 95\% of the input data points were subject to distributional shifts. The x-axis labels are the abbreviated names for each deck.

\begin{figure}[H]
\centering

\begin{subfigure}{.25\columnwidth}
  \centering
  \includegraphics[width=1\linewidth]{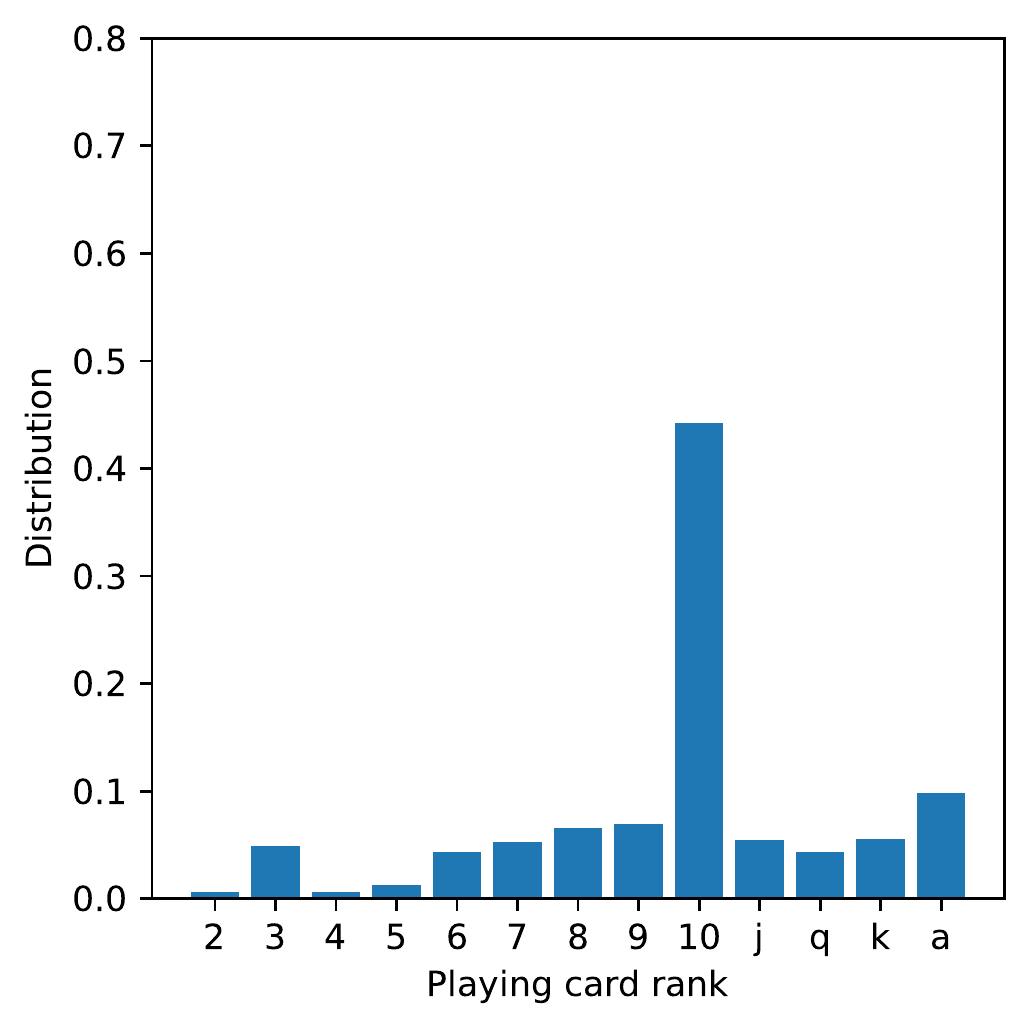}
  \caption{\centering Captain America, Softmax}
  \label{fig:Softmax_rank_ca}
\end{subfigure}%
\begin{subfigure}{.25\columnwidth}
  \centering
  \includegraphics[width=1\linewidth]{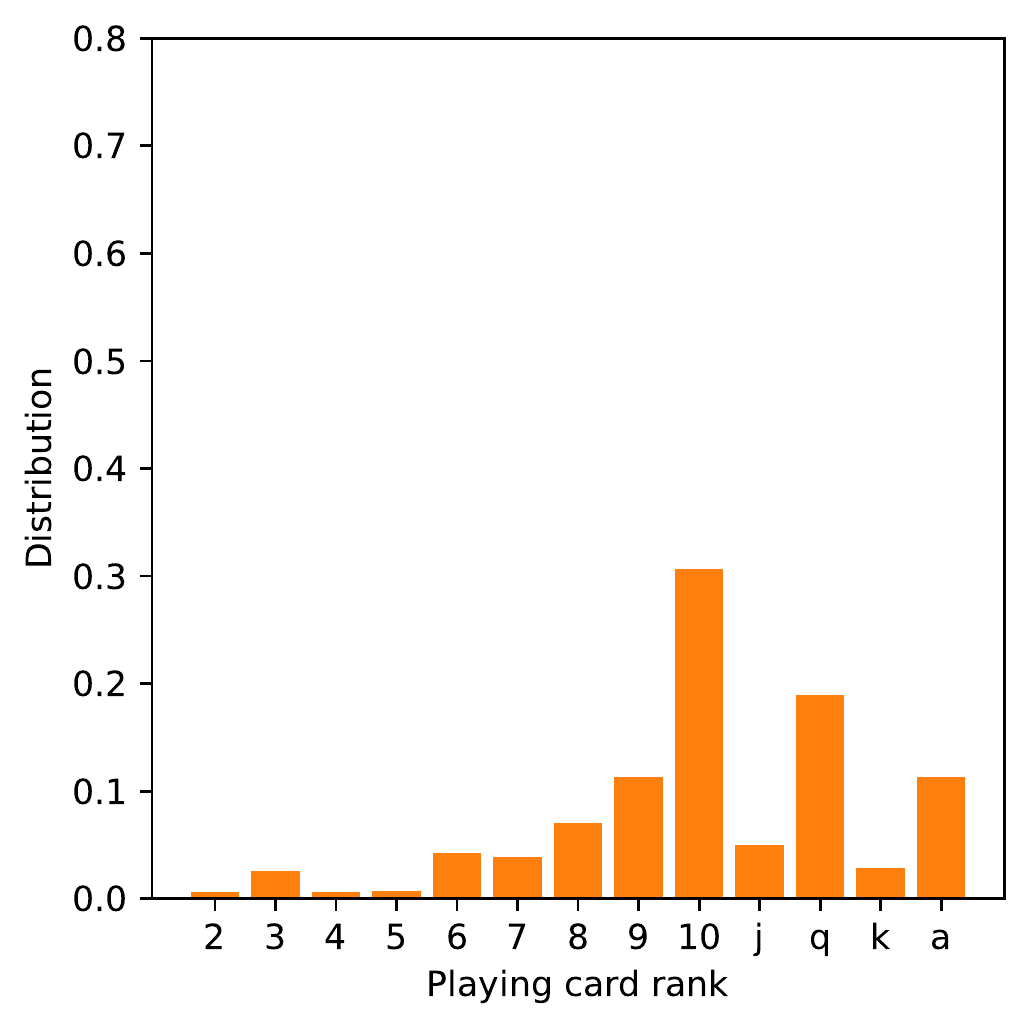}
  \caption{\centering Captain America, EDL-GEN}
  \label{fig:edl_gen_rank_ca}
\end{subfigure}
\begin{subfigure}{.23\columnwidth}
  \centering
  \includegraphics[width=1\linewidth]{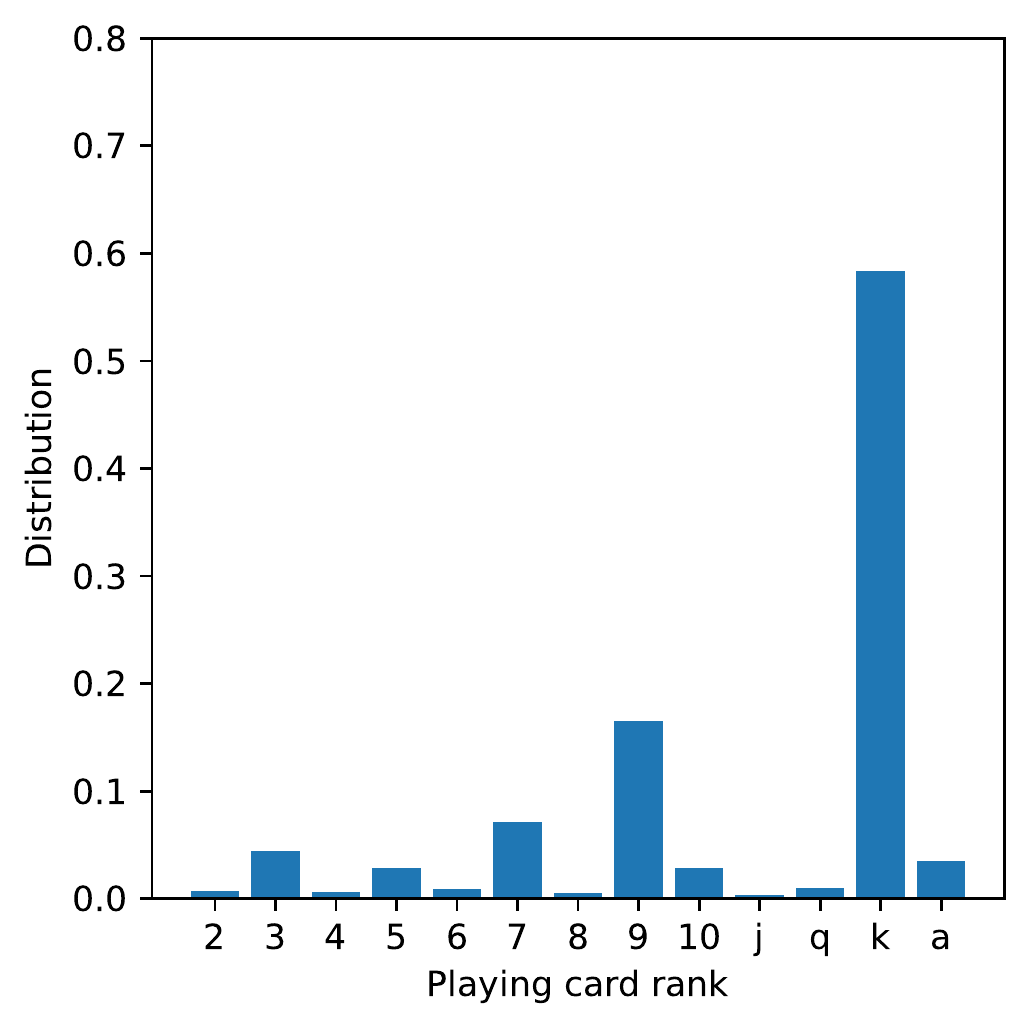}
  \caption{\centering Adversarial Captain America, Softmax}
  \label{fig:Softmax_rank_aca}
\end{subfigure}%
\begin{subfigure}{.23\columnwidth}
  \centering
  \includegraphics[width=1\linewidth]{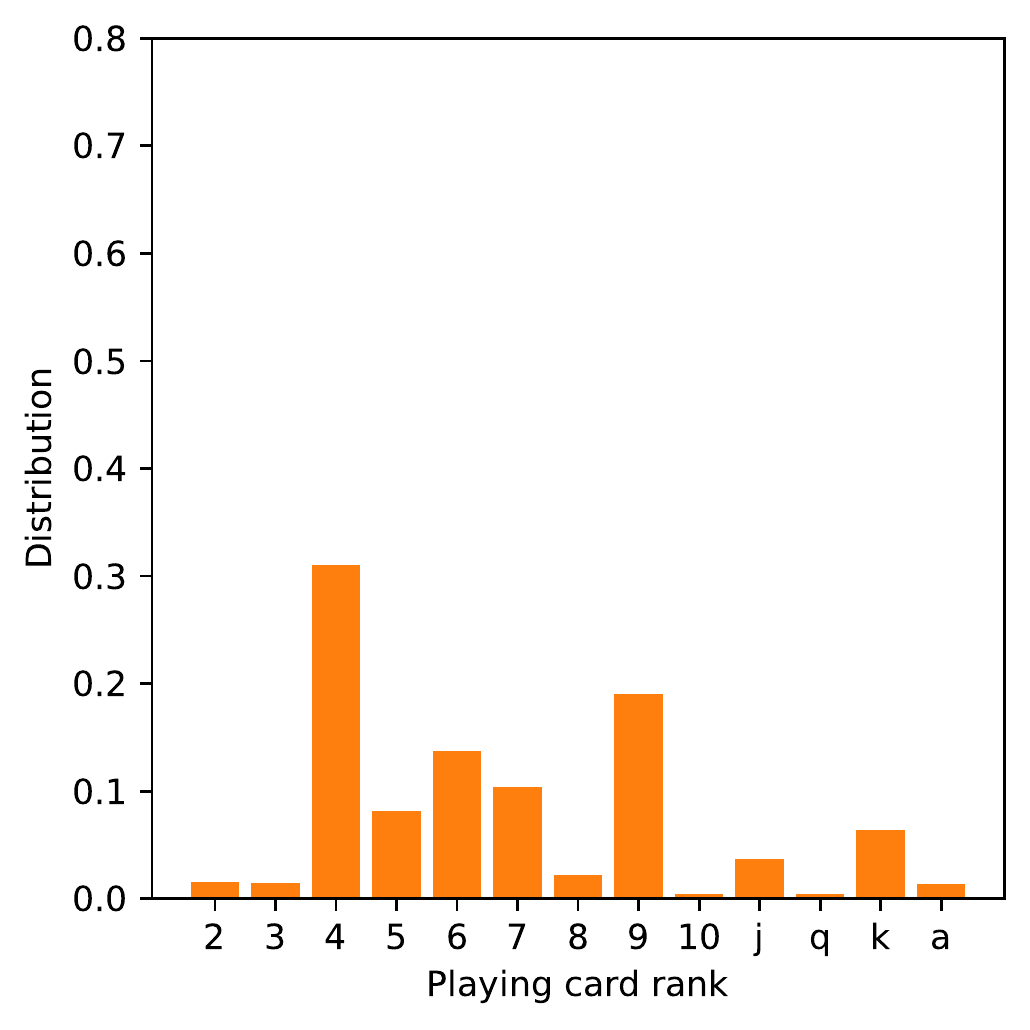}
  \caption{\centering Adversarial Captain America, EDL-GEN}
  \label{fig:edl_gen_rank_aca}
\end{subfigure}

\caption{Distribution of playing card rank predictions for Softmax and EDL-GEN neural networks when the Captain America and Adversarial Captain America decks were used to apply distributional shift for 95\% of input data points.}
\label{fig:network_dist_aca_95}
\end{figure}

With the Adversarial Captain America deck (ACA in Figure~\ref{fig:correct_rank_higher_and_suit_rules}), only 24\% of experimental repeats with~\ac{nsl} Softmax learned the correct $\asp{rank\_higher}$ rule, compared to 72\% with~\ac{nsl} EDL-GEN (both with constant penalties). There isn't much difference between~\ac{nsl} Softmax and~\ac{nsl} EDL-GEN in learning the correct suit rule (Figure~\ref{fig:correct_suit_rule}). So, as~\ac{nsl} Softmax fails to learn the correct $\asp{rank\_higher}$ rule, this explains why in Figure~\ref{fig:fs_aca_95_100_acc}, the~\ac{nsl} Softmax approaches performed worse than~\ac{nsl} EDL-GEN approaches. The question now becomes, why does ILASP fail to learn the correct $\asp{rank\_higher}$ rule for 76\% of the experimental repeats with~\ac{nsl} Softmax, despite~\ac{nsl} Softmax having a similar or lower percentage of incorrect~\ac{ilp} examples compared to~\ac{nsl} EDL-GEN? To answer this question, Figure~\ref{fig:network_dist_aca_95} shows the distribution of playing card rank predictions for both Softmax and EDL-GEN neural networks when the Captain America and Adversarial Captain America decks were used to apply distributional shift for 95\% of input data points, as these two decks have a similar percentage of incorrect~\ac{ilp} examples between Softmax and EDL-GEN (see Figures~\ref{fig:fs_ca_ilp_ex} and~\ref{fig:fs_aca_ilp_ex}).

The Softmax neural network predicted the same playing card rank more often than the EDL-GEN neural network. With the Captain America deck, 45\% of playing cards were predicted with rank $10$ (Figure~\ref{fig:Softmax_rank_ca}), and with the Adversarial Captain America deck, nearly 60\% of playing cards were predicted with rank $King$ (Figure~\ref{fig:Softmax_rank_aca}). The EDL-GEN neural network predicted with a more even distribution. Now, with the Softmax neural network and the Adversarial Captain America deck, ILASP didn't learn the correct $\asp{rank\_higher}$ rule very often. Investigating the neural network card predictions within the generated~\ac{ilp} examples when 95\% of distributional shifts were applied, we calculate the percentage of examples where the ground-truth winner has a predicted card with a higher rank than the other players. For~\ac{nsl} Softmax, only 9\% of the examples contained a higher ranked card for the ground-truth winning player, compared to 19\% with~\ac{nsl} EDL-GEN. Looking at the Captain America deck, 35\% of the examples for~\ac{nsl} Softmax contained a higher ranked card for the ground-truth winning player, compared to 37\% for~\ac{nsl} EDL-GEN.

This explains why, in the Adversarial Captain America deck~\ac{nsl} Softmax struggled to learn the $\asp{rank\_higher}$ rule and therefore, why there was a drop in performance in Figure~\ref{fig:fs_aca_95_100_acc} for~\ac{nsl} Softmax. As the Softmax neural network failed to predict playing card ranks correctly, the~\ac{ilp} examples didn't contain a higher ranked card for the ground-truth winner, and therefore 76\% of the experimental repeats failed to learn the correct $\asp{rank\_higher}$ rule. Comparing with the Captain America deck in Figure~\ref{fig:fs_ca_95_100}, the~\ac{nsl} Softmax approaches performed much better, because there was a higher number of generated~\ac{ilp} examples that contained higher ranked card predictions for the ground-truth winning player.

\begin{figure}[H]

\centering
\begin{subfigure}{1\columnwidth}
  \centering
  \includegraphics[width=.75\linewidth]{Figures/follow_suit_results/follow_suit_nn_penalties_legend.pdf}
  \label{fig:fs_50_repeats_legend_3}
\end{subfigure}
\begin{subfigure}{0.3\columnwidth}
  \centering
  \includegraphics[width=1\textwidth]{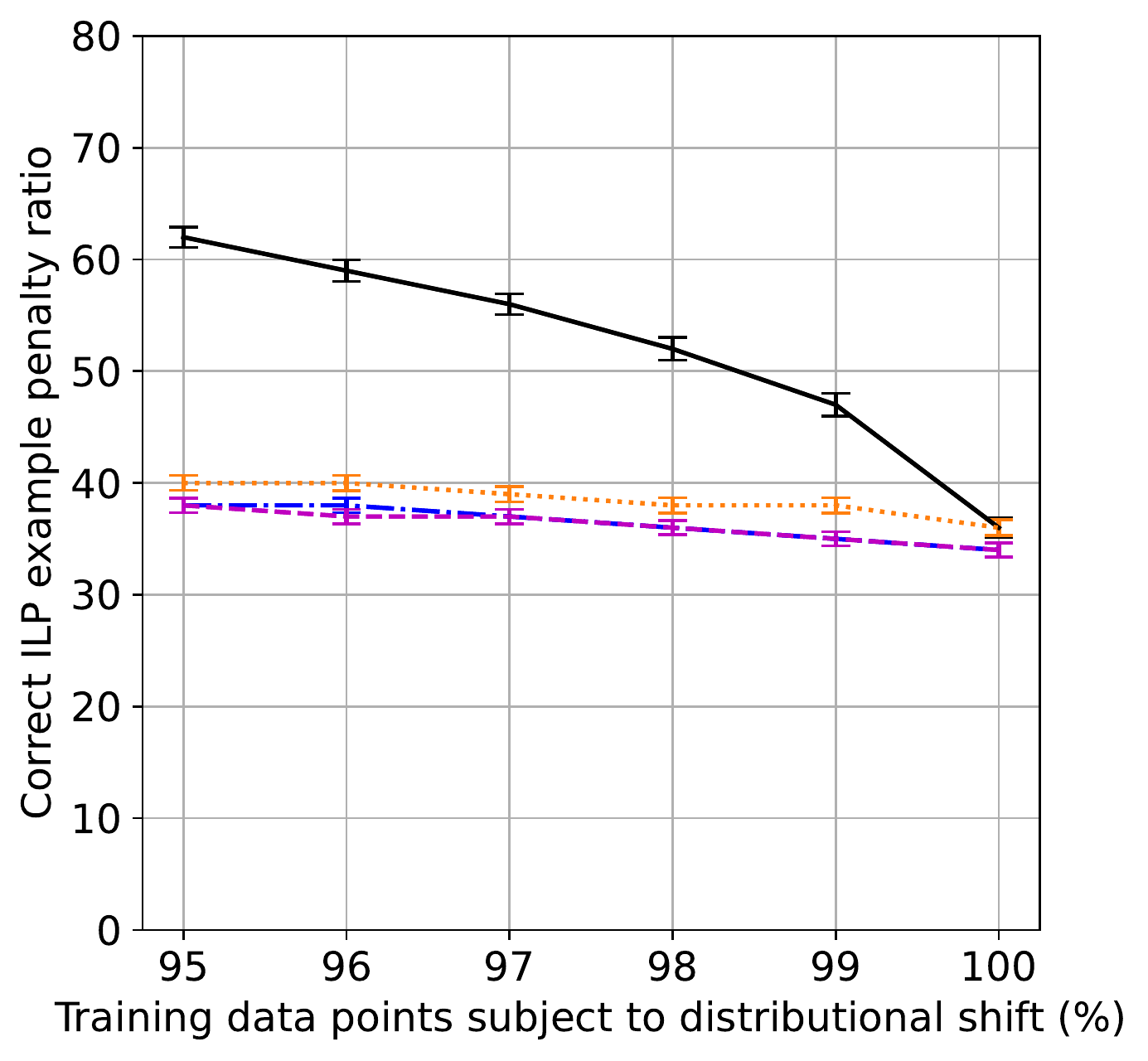}
  \caption{Captain America}
  \label{fig:ca_wpr}
\end{subfigure}
\begin{subfigure}{0.3\columnwidth}
  \centering
  \includegraphics[width=1\textwidth]{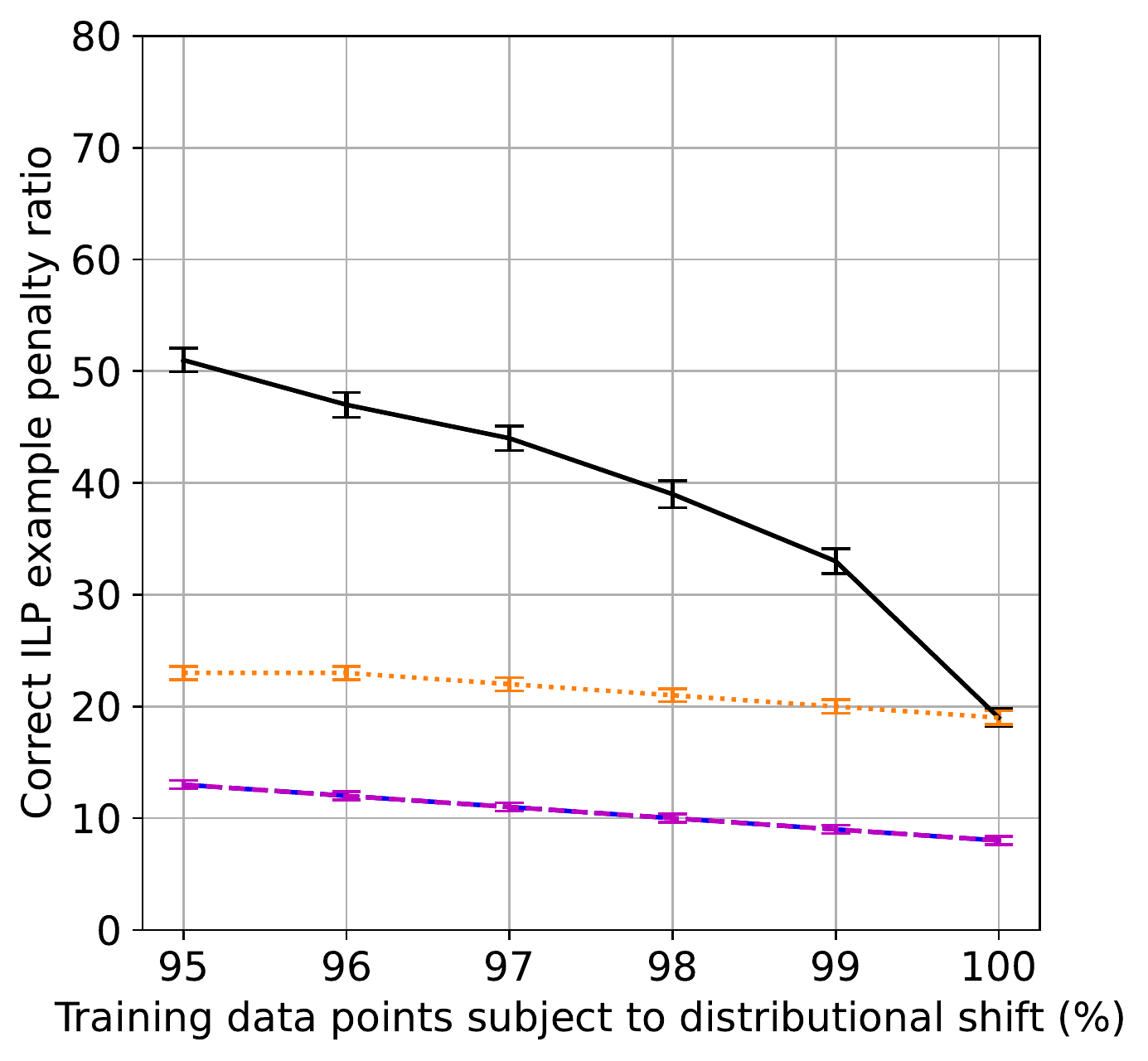}
  \caption{Adversarial Standard}
  \label{fig:as_wpr}
\end{subfigure}
\begin{subfigure}{0.3\columnwidth}
  \centering
  \includegraphics[width=1\textwidth]{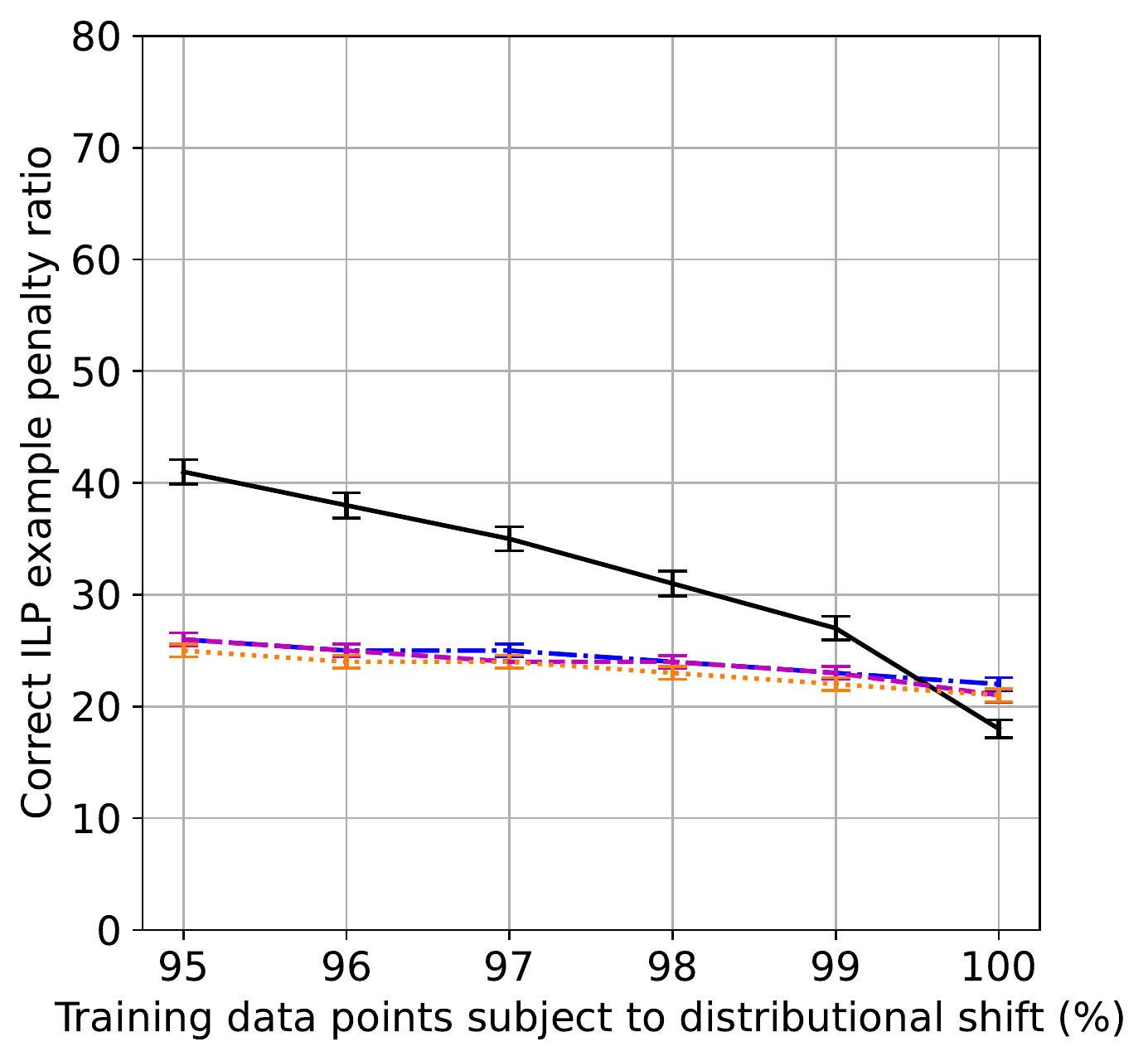}
  \caption{Adversarial Captain America}
  \label{fig:aca_wpr}
\end{subfigure}

\caption{\ac{ilp} example weight penalty ratio, 95-100\% shifts.}
\label{fig:extra_wpr}
\end{figure}

Finally, in Figure~\ref{fig:aca_wpr} we now investigate the~\ac{ilp} example weight penalty ratio to explain why there was a significant gap between the two~\ac{nsl} EDL-GEN approaches in Figure~\ref{fig:fs_aca_95_100_acc} for the Adversarial Captain America deck.

The weight penalty ratio of~\ac{nsl} EDL-GEN with penalties calculated from neural network confidence scores outperformed~\ac{nsl} EDL-GEN with constant penalties (Figure~\ref{fig:aca_wpr}). As the Adversarial Captain America deck was more challenging for ILASP in terms of the predictions from the neural networks, the~\ac{ilp} example weight penalties had more impact on the accuracy of the learned hypotheses, as ILASP was able to focus on covering the~\ac{ilp} examples that contained the correct neural network predictions.

\subsection{\ac{nsl} framework evaluation}
\begin{figure}[H]

\centering
\begin{subfigure}{1\columnwidth}
  \centering
  \includegraphics[width=.75\linewidth]{Figures/follow_suit_results/follow_suit_legend.pdf}
  \label{fig:fs_50_repeats_legend_4}
\end{subfigure}
\begin{subfigure}{0.3\columnwidth}
  \centering
  \includegraphics[width=1\textwidth]{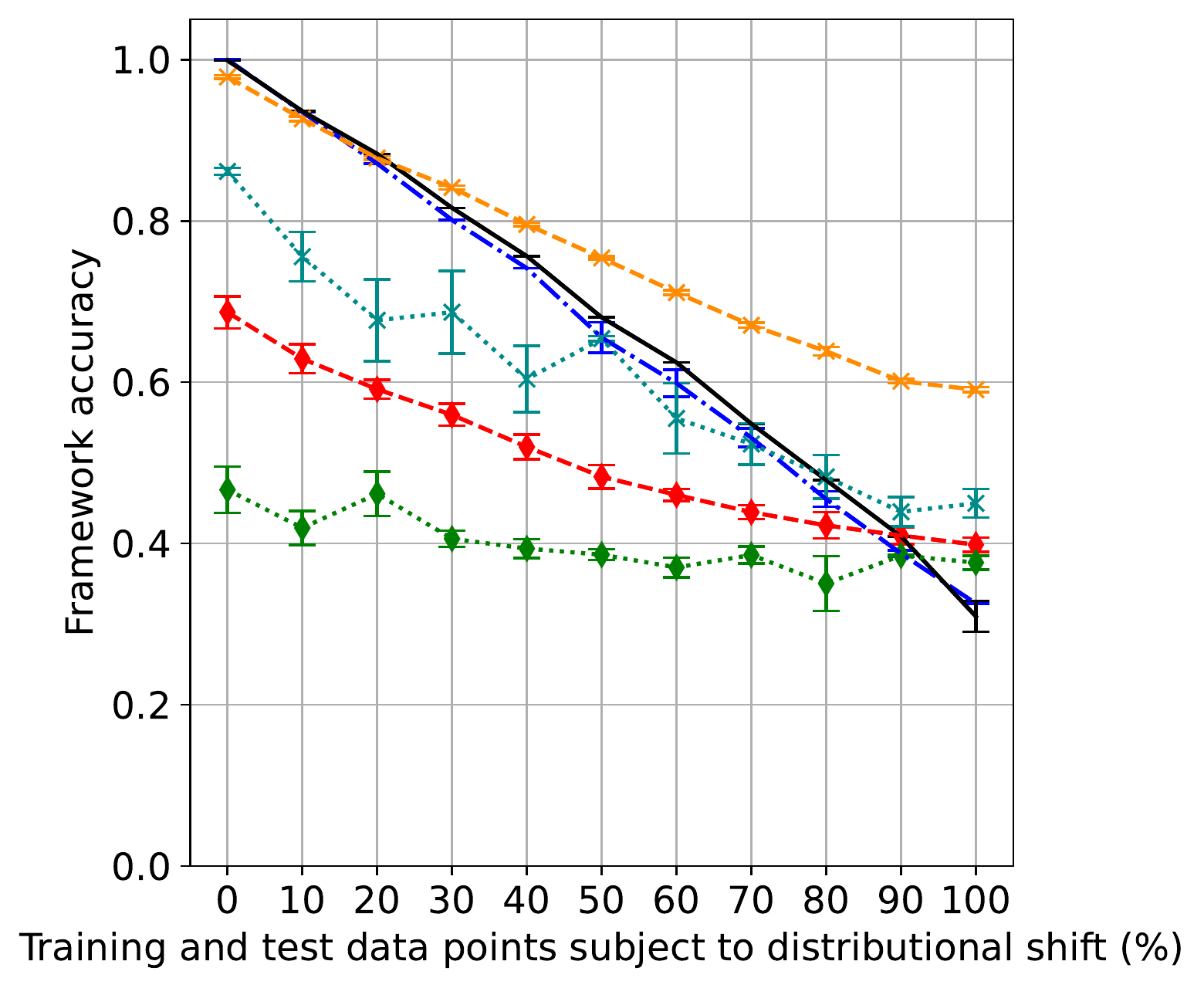}
  \caption{Captain America}
  \label{fig:ca_frameork_eval}
\end{subfigure}
\begin{subfigure}{0.3\columnwidth}
  \centering
  \includegraphics[width=1\textwidth]{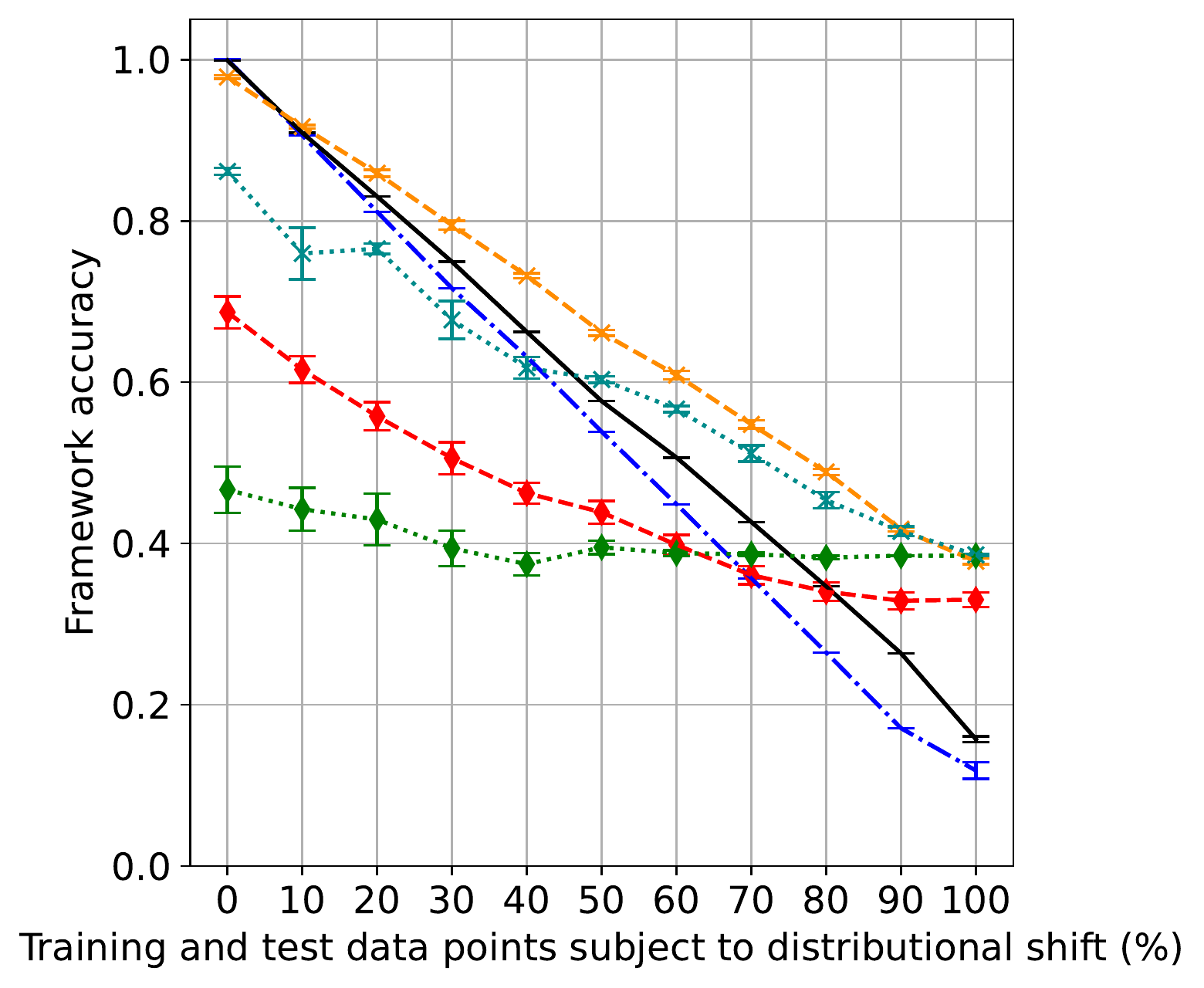}
  \caption{Adversarial Standard}
  \label{fig:as_frameork_eval}
\end{subfigure}
\begin{subfigure}{0.3\columnwidth}
  \centering
  \includegraphics[width=1\textwidth]{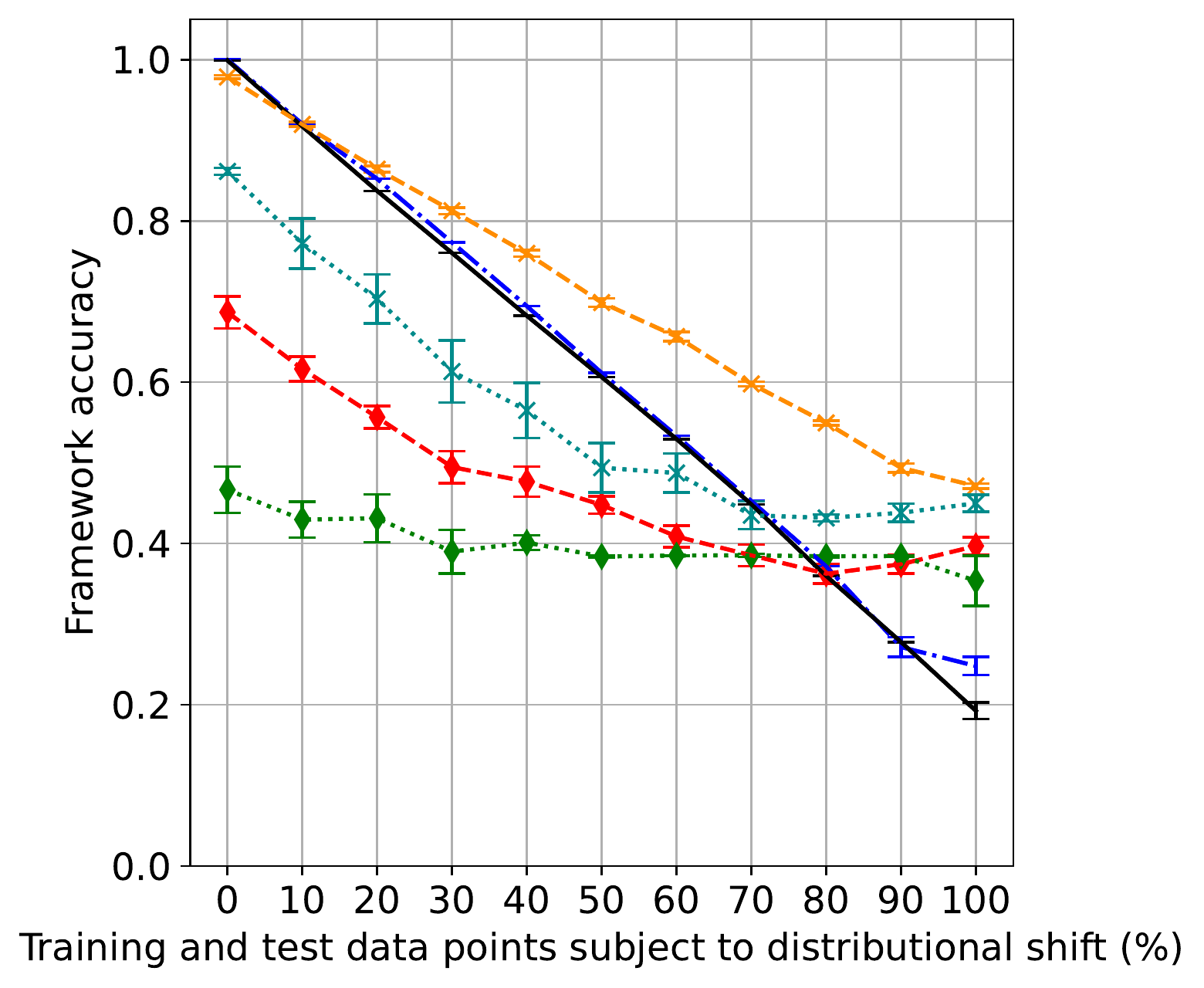}
  \caption{Adversarial Captain America}
  \label{fig:aca_frameork_eval}
\end{subfigure}

\caption{Accuracy of the~\ac{nsl} framework when training and test data points were subject to distributional shifts.}
\label{fig:extra_framework_eval}
\end{figure}

Figure~\ref{fig:extra_framework_eval} presents the accuracy of the entire~\ac{nsl} framework when both training and test data points were subject to distributional shifts. The results for the Captain America deck in Figure~\ref{fig:ca_frameork_eval} are very similar to the results for the Batman Joker deck presented in Figure~\ref{fig:fs_bj_unstruc_acc}, and the results for the Adversarial Standard and Adversarial Captain America decks in Figures~\ref{fig:as_frameork_eval} and~\ref{fig:aca_frameork_eval} are very similar to the results presented in Figure~\ref{fig:fs_abj_acc_unstruc} for the Adversarial Batman Joker deck.

\section{Learned Hypotheses}\label{sec:app:learned_hyps}
In this section we present a sample of the hypotheses learned by~\ac{nsl} when distributional shifts were applied to input data points.

\subsection{Follow Suit Winner}
When no distributional shifts were applied, i.e., at points $0\%$ on the x-axes in Figure~\ref{fig:fs_winner_acc_results}, the following hypothesis was learned by~\ac{nsl}:

\begin{lstlisting}
winner(X) :- not p1(X), player(X).
p1(V1) :- V2 != V3; suit(1,V2); suit(V1,V3); player(V1); suit(V2); suit(V3).
p1(V1) :- rank_higher(V2,V1); suit(1,V3); suit(V2,V3); player(V1); player(V2); suit(V3).
\end{lstlisting}

\noindent The first rule states that player $\asp{X}$ is a winner if \textit{neither} of the bottom two rules hold. The second rule holds if the suit of player $\asp{X}$ is different to the suit of player 1, and the final rule holds if there is another player with a higher ranked card with the same suit as player 1. The $\asp{player}$, $\asp{suit}$ and $\asp{rank\_higher}$ predicates were defined in the background knowledge (for details, see Appendix~\ref{sec:app:ilp}). As an example, in the case of 100\% distributional shifts using the Batman Joker deck in Figure~\ref{fig:fs_bj_results_acc}, the following rules were learned on one experimental repeat:

\begin{lstlisting}
winner(X) :- not p1(X), player(X).
p1(V1) :- rank_higher(V2,V1); player(V1); player(V2).
\end{lstlisting}

\noindent In this hypothesis, a player is the winner if they have played the highest ranked card. In this case, the rule denoting the suit having to match the suit of player 1 was missed.

\subsection{Sudoku Grid Validity}
For the $4\times4$ and $9\times9$ grid Sudoku tasks, when no distributional shifts were applied (i.e., at points $0\%$ on the x-axes in Figure~\ref{fig:sud_acc_results}), the following hypothesis was learned by~\ac{nsl}, which states that a Sudoku grid is invalid if there are two of the same digits in a block, column or row:

\begin{lstlisting}
invalid :- neq(V2,V1), digit(V1,V3), block(V2,V0), block(V1,V0), digit(V2,V3).
invalid :- neq(V1,V0), digit(V0,V2), digit(V1,V2), row(V0,V3), row(V1,V3).
invalid :- neq(V1,V0), digit(V0,V3), digit(V1,V3), col(V0,V2), col(V1,V2).
\end{lstlisting}

The $\asp{neq}$, $\asp{digit}$, $\asp{block}$, $\asp{row}$ and $\asp{col}$ predicates were defined in the background knowledge (for details, see Appendix~\ref{sec:app:ilp}). For the $4\times4$ grid task, when 100\% of the training data points were subject to distributional shifts in Figure~\ref{fig:4x4_sudoku_results_acc}, the following hypothesis was learned:

\begin{lstlisting}
invalid :- neq(V2,V1), neq(V3,V1), neq(V3,V2), block(V2,V0), block(V3,V0), block(V1,V0).
invalid :- not block(V1,V0), block(V2,V0), col(V2,V3), col(V1,V3).
invalid :- not block(V2,V0), block(V1,V0), row(V1,V3), row(V2,V3).
\end{lstlisting}

The first argument in the $\asp{block}$, $\asp{row}$ and $\asp{col}$ predicates is a string representing cell coordinates and the second argument is an identifier (e.g., block 1, block 2, etc...). The first rule states that a grid is invalid if there are three cells in the same block that have different coordinates. The second rule states that a grid is invalid if there are two cells within the same column that are in different blocks and the third rule states that a grid is invalid if there are two cells within the same row that are in different blocks. Therefore, this hypothesis always returned invalid at test-time.

\subsection{Crop Yield Prediction}
When no distributional shift was applied, the following hypothesis was learned by \ac{nsl}. Note that we trim the number of rules for compactness and the full listing is available in the experiment code\footnote{\url{https://github.com/DanCunnington/FFNSL}}:

\begin{lstlisting}
yield(0) :- disease(bacterial_spot), location(18).
yield(0) :- disease(black_rot), location(18).
yield(2) :- disease(late_blight), location(6).
yield(2) :- disease(leaf_scorch).
yield(2) :- location(16).
yield(2) :- disease(healthy), location(7).
yield(2) :- disease(healthy), location(17).
yield(2) :- location(11).
...
\end{lstlisting}

\noindent At 100\% shifts, the following hypothesis was learned:
\begin{lstlisting}
yield(0) :- location(19), disease(early_blight).
yield(1) :- location(7), disease(bacterial_spot).
yield(2) :- location(16).
yield(2) :- location(11).
yield(2) :- species(corn), disease(healthy).
yield(2) :- disease(powdery_mildew), location(7).
yield(0) :- location(19), species(potato).
\end{lstlisting}

Here you can see the yield quality has changed for the bacterial spot disease, and this also depends on a different location. This is due to incorrect neural network predictions for the bacterial spot disease.

\subsection{Indoor Scene Classification}
When no distributional shift was applied, the following hypothesis was learned by \ac{nsl}. Note that we also trim the number of rules for compactness and the full listing is available in the experiment code:

\begin{lstlisting}
label(4) :- image(meeting_room).
label(2) :- image(inside_subway).
label(2) :- image(elevator).
label(3) :- image(bowling).
label(0) :- image(shoeshop).
label(4) :- image(classroom).
...
\end{lstlisting}

\noindent At 100\% shifts, the following hypothesis was learned:

\begin{lstlisting}
label(4) :- image(meeting_room).
label(2) :- image(inside_subway).
label(2) :- image(elevator).
label(3) :- image(bowling).
label(0) :- image(shoeshop).
label(2) :- image(airport_inside).
...
\end{lstlisting}

Analysing these rules further, it appears that at 100\% shifts, whilst some super-class rules are correct, others were not learned at all. For example, the $\asp{classroom}$ rule is missing. In total, there were 46 rules learned at 100\% shifts, compared to the full set of 67 rules at 0\% shifts. In Figure \ref{fig:scene_interp} you can see the number of rules in the learned hypothesis decreases at 100\% shifts.



\section{Dataset Details}\label{sec:app:datasets}

\paragraph{Follow Suit Winner}
The Follow Suit Winner dataset was generated by simulating multiple games, where each game began with a randomly shuffled deck of playing cards split between the four players. Each game consisted of 13 tricks and the card played by each player along with the winner of each trick was stored. The \textit{small} training datasets contained 104 example tricks from 8 games and the \textit{large} training datasets contained 10,400 example tricks from 800 games. A test set was created containing 1001 example tricks from 77 games. For the neural network, an image was taken of every playing card in a standard deck. The \textit{ImageDataGenerator} class from the Keras image pre-processing library\footnote{\href{https://keras.io/api/preprocessing/image/}{https://keras.io/api/preprocessing/image/}} was used to apply transformations to each playing card image, generating 750 variations of each image. We set the \textit{rotation range} to 55, \textit{brightness range} to 0.5-1.5, \textit{shear range} to 15, \textit{channel shift range} to 2.5, \textit{zoom range} to 0.1 and enable \textit{horizontal flip}. From a total of 39,000 images, we created a training set of 27,300 images and a test set of 11,700 images (70\%/30\% split), maintaining an equal representation of each playing card. Similarly to the Sudoku Grid Validity task, the test set was further split into two datasets ($\sim$70\%/30\%), maintaining an equal representation of each playing card, as follows. The first, denoted CARDS\_TEST\_A contains 8164 images and was used to create~\ac{nsl} training sets for learning a hypothesis. Playing cards in the Follow Suit Winner training sets were replaced with a random image of the corresponding playing card from CARDS\_TEST\_A. The second split, denoted CARDS\_TEST\_B contained 3536 images and was used to create a hold out test set such that~\ac{nsl} can be evaluated on unseen data once a hypothesis has been learned.

Distributional shifts were applied by replacing playing card images from the standard deck with playing card images from alternative decks in an increasing percentage of data points in the Follow Suit Winner training sets. We used playing card images from \textit{Batman Joker} and \textit{Captain America} decks and also created \textit{adversarial} data points from each deck, placing the candidate playing card image on a background containing playing card images from the standard deck. We applied the same image transformations to the alternative decks such that standard playing card images can be directly swapped with a corresponding card image from an alternative deck. Figure~\ref{fig:cards} shows an example queen of hearts playing card image from each deck: \textit{Standard} (\ref{fig:standard_card}), \textit{Batman Joker} (\ref{fig:bj_card}), \textit{Captain America} (\ref{fig:ca_card}),\\\textit{Adversarial} \textit{Standard} (\ref{fig:advs_card}), \textit{Adversarial Batman Joker} (\ref{fig:advbj_card}) and \textit{Adversarial Captain America} (\ref{fig:advca_card}).

\paragraph{Sudoku Grid Validity}
The Sudoku Grid Validity datasets were generated using valid $4\times4$ and $9\times9$ Sudoku starting configurations obtained from Hanssen's Sudoku puzzle generator\footnote{\href{https://www.menneske.no/sudoku/2}{https://www.menneske.no/sudoku/2}}. Invalid starting configurations were obtained by taking a valid example (that didn't exist in the set of valid data points) and changing one digit at random in a row, column or block to match another digit in the same row, column or block. All sets of invalid data points contained an equal distribution of data points containing two of the same digit in a row, column or block. The \textit{small} training datasets contained 320 data points, each consisting of 160 \textit{valid} starting configurations and 160 \textit{invalid} starting configurations. The \textit{large} training datasets contained 32,000 data points, with 16,000 valid and 16,000 invalid data points. Finally, separate test sets were created for $4\times4$ and $9\times9$ grids, which contained 1000 data points: 500 valid and 500 invalid.

For the neural network used in the $4\times4$ grids, we used digit classes 1-4 from the standard MNIST dataset~\cite{lecun1998gradient} and created a training set of 24,674 data points and a test set of 4,160 data points. The MNIST test set was further split ($\sim$70\%/30\%), maintaining an equal representation of digits, into two datasets as follows. The first, denoted \textit{MNIST\_TEST\_A} contained 2910 images and was used to create~\ac{nsl} training sets for learning a hypothesis. Digits in the Sudoku training sets were replaced with a random image of the corresponding digit from MNIST\_TEST\_A. The second split, denoted \textit{MNIST\_TEST\_B} contained 1249 images and was used to create a hold out test set such that~\ac{nsl} could be evaluated on unseen data once a hypothesis was learned. Digits in the Sudoku test set were replaced with a random image of the corresponding image from MNIST\_TEST\_B. 

For the neural network used in the $9\times9$ grids, we used digit classes 1-9 from the standard MNIST dataset~\cite{lecun1998gradient} and created a training set of 54,078 data points and a test set of 9,021 data points. The MNIST test set was further split ($\sim$70\%/30\%), maintaining an equal representation of digits, into two datasets as follows. The first, denoted \textit{MNIST\_TEST\_C} contained 6310 images and was used to create~\ac{nsl} training sets for learning a hypothesis. Digits in the Sudoku training sets were replaced with a random image of the corresponding digit from MNIST\_TEST\_C. The second split, denoted \textit{MNIST\_TEST\_D} contained 2710 images and was used to create a hold out test set such that~\ac{nsl} could be evaluated on unseen data once a hypothesis was learned. Digits in the Sudoku test set were replaced with a random image of the corresponding image from MNIST\_TEST\_D. Note that data observed by~\ac{nsl} at learning time was completely unseen by the neural network and was therefore vulnerable to distributional shifts. Also, data observed by~\ac{nsl} at evaluation time was completely unseen by the neural network and also~\ac{nsl} itself during learning.

Distributional shifts were applied by rotating MNIST digit images 90$^{\circ}$ clockwise in an increasing percentage of data points in the Sudoku training sets. When we evaluated with unstructured test data, the same procedure applied to the Sudoku test set, i.e., when we evaluated a hypothesis learned from a training set with 20\% of the data points containing rotated images, 20\% of the test set data points also contained rotated images.




\section{Neural Network and Baseline Details}\label{sec:app:networks}
\paragraph{Follow Suit Winner}
Firstly, for~\ac{nsl} Softmax, we trained a Softmax-based~\ac{cnn} with 4 2D convolutional layers and 2 fully connected layers for 20 epochs in PyTorch. The network accepts 3-channel RGB input with images of size 274x174 pixels and outputs a 52 dimensional Softmax vector to predict each playing card. Secondly, for~\ac{nsl} EDL-GEN, we trained an uncertainty-aware neural network based on evidential deep learning~\cite{sensoy2020uncertainty}. We used the available architecture and implementation in TensorFlow,\footnote{\label{edl_gen_tensorflow}\href{https://muratsensoy.github.io/gen.html}{https://muratsensoy.github.io/gen.html}} and modified $k$, the number of outputs to 52 and the layer dimensions to accept 274x174 RGB card images. We also trained this neural network for 20 epochs.

The baseline random forest model was implemented with scikit-learn 0.23.2 and tuned on the first \textit{small} dataset with 0 data points subject to distributional shift. The number of estimators was tuned across: $\{10,20,50,100,200\}$. The best performing parameter value of 100 estimators was chosen and used for all Follow Suit Winner experiments. The random seed was set to 0 to enable reproducability.

The baseline~\ac{fcn} consists of 3 fully connected layers with the ReLU activation function applied to each layer. Dropout was also applied after the first and second layers. Finally, a Softmax layer squashed the final logits into 4 classes, representing each possible winner. The input consisted of one-hot encoded suit values and the rank value of the playing card for each player. Therefore, the input size to the first fully connected layer was 20. We implemented the architecture in PyTorch v1.7.0.

To tune the~\ac{fcn}, we sampled the number of output units in the first and second layers, i.e., $l1 \in \left \{ 20, 32, 46, 52 \right \}$ and $l2 \in \left \{ 52, 64, 74, 80 \right \}$ respectively, along with the dropout probability in both dropout layers $dr \in \left \{0.1, 0.2, 0.5 \right \}$. We sampled all possible parameter combinations and tuned on the first \textit{small} dataset, with no data points subject to distributional shift, trained for 50 epochs. The best performing parameter values of $l1 = 20$, $l2 = 74$ and $dr = 0.1$ were chosen. These parameters were then fixed for all models trained and following tuning, each model was trained for 50 epochs. Finally, the random seed was set to 0 to enable reproducability.

\paragraph{Sudoku Grid Validity}
Within~\ac{nsl}, we trained two types of neural networks. Firstly, for~\ac{nsl} Softmax, we adopted the~\ac{cnn} architecture available in the MNIST PyTorch tutorial\footnote{\href{https://github.com/pytorch/examples/tree/master/mnist}{https://github.com/pytorch/examples/tree/master/mnist}} and replaced the \textit{LogSoftmax} layer with a \textit{Softmax} layer and the \textit{Negative Log Likelihood} loss function with \textit{Cross-Entropy Loss}. This is to satisfy the neural network definition in Section~\ref{sec:method} such that a confidence score $c\in[0,1]^{k}$ is returned for $k$ possible feature values. For the two grid sizes, $4\times4$ and 9x9, we train two separate networks. For $4\times4$ grids, we set $k=4$ and train on digits 1-4 inclusive, whilst for $9\times9$ grids we set $k=9$ and train on digits 1-9 inclusive. We adopted all existing hyper-parameter values and trained for 20 epochs.

Secondly, for~\ac{nsl} EDL-GEN, we trained two uncertainty-aware neural networks \cite{sensoy2020uncertainty} using the available architecture and implementation in TensorFlow\textsuperscript{\ref{edl_gen_tensorflow}}, and set $k$, the number of outputs, to 4 and 9, for $4\times4$ and $9\times9$ grids respectively. We used existing hyper-parameter values and trained for 20 epochs.

The baseline random forest model was implemented with scikit-learn 0.23.2 and tuned on the first \textit{small} dataset with no data points subject to distributional shift. The number of estimators was tuned across: $\{10,20,50,100,200\}$. The best performing parameter value of 100 estimators was chosen and used for all Sudoku Grid Validity experiments. The random seed was set to 0 to enable reproducability.

The baseline~\ac{cnnlstm} consisted of an embedding layer, followed by a 1D convolutional layer with a kernel size of 3 and the ReLU activation function. Then, a 1D max pooling layer with pool size 2 was used, followed by a dropout layer, an LSTM layer and a second dropout layer. Finally, a dense fully connected layer with the sigmoid activation function was used to produce a binary classification of the input digit sequence. The input sequence length to the embedding layer was 16 for $4\times4$ grids and 81 for $9\times9$ grids, representing each cell on the Sudoku grid. We implemented the architecture in PyTorch v1.7.0.

To tune the~\ac{cnnlstm}, we sampled the learning rate $lr \in \left \{ 0.1, 0.001, 0.0001\right \}$, the embedding dimension of the embedding layer $ed \in \left \{ 32, 96, 256\right \}$, the number of output channels of the 1D convolution layer $oc \in \left \{ 64, 96 \right \}$, the number of hidden features in the LSTM layer $lh \in \left \{ 32, 96, 128 \right \}$ and the dropout probability $dr \in \left \{ 0.01, 0.05, 0.1\right \}$ in both dropout layers. We performed 10 samples and evaluated the model on the first \textit{large} dataset with 0 data points subject to distributional shift, trained for 2 epochs. The best performing parameter values of $lr = 0.0001$, $ed = 96$, $oc = 64$, $lh = 96$ and $dr = 0.01$ were chosen. These parameters were then fixed for all models trained and following tuning, each model was trained for 5 epochs. Finally, the random seed was set to 0 to enable reproducability.



\section{System Details}\label{sec:app:system}
All experiments in this paper (with the exception of the deep neural network baselines) were run on the same machine with the following specifications:

\noindent \textbf{Hardware:} QEMU KVM virtual machine standard PC (i440FX + PIIX 1996) with 10 nodes of 8-core AMD EPYC Zen 2 CPUs (80 cores total), 16GB RAM.

\noindent \textbf{Operating System:} Ubuntu 18.04.4 LTS.

\noindent \textbf{Software:} FastLAS 1.1 (FastLAS 3 for $4\times4$ Sudoku Grid Validity with reduced background knowledge), ILASP 4, Python 3.7.3, PyTorch 1.7.0, TensorFlow 1.14.0, Keras 2.4.0, scikit-learn 0.23.2, numpy 1.19.1, problog 2.1.0.42.
\noindent The neural network baselines were run on a machine with the following specifications:

\noindent \textbf{Hardware:} x86 compute node with 24 cores (CPU) and an NVIDIA Tesla K80 GPU, 512GB RAM.

\noindent \textbf{Operating System:} Red Hat Enterprise Linux 7.6.

\noindent \textbf{Software:} Same as above.



\section{ILP task listings}\label{sec:app:ilp}

\subsection{Follow Suit Winner} For the Follow Suit Winner task, we used the ILASP~\cite{Law2018thesis}~\ac{ilp} system as ILASP supports \textit{predicate invention}~\cite{stahl1993predicate}. Predicate invention was required for this task to link the winning player to the suit and rank of other players cards. We encoded as background knowledge possible suit and rank values, the four players, as well as the definition of the \asp{rank\_higher} predicate. The set of body mode declarations included a \asp{suit} predicate, which linked a player's card to a suit, alongside the \asp{rank\_higher} predicate. The set of head mode declarations included a player variable, specified to support predicate invention. The hypothesis space for this task contained 96 possible rules (therefore $2^{96}$ potential hypotheses, computed as the power set).

\paragraph{Background Knowledge}
\begin{lstlisting}
% Suits
suit(h).
suit(s).
suit(d).
suit(c).

% Ranks
rank(a).
rank(2).
rank(3).
rank(4).
rank(5).
rank(6).
rank(7).
rank(8).
rank(9).
rank(10).
rank(j).
rank(q).
rank(k).

% Rank Value
rank_value(2, 2).
rank_value(3, 3).
rank_value(4, 4).
rank_value(5, 5).
rank_value(6, 6).
rank_value(7, 7).
rank_value(8, 8).
rank_value(9, 9).
rank_value(10, 10).
rank_value(j, 11).
rank_value(q, 12).
rank_value(k, 13).
rank_value(a, 14).

% 4 Players
player(1..4).

% Definition of higher rank
rank_higher(P1, P2) :- card(P1, R1, _), card(P2, R2, _), rank_value(R1, V1), rank_value(R2, V2), V1 > V2.

% Link player's card to suit
suit(P1, S) :- card(P1, _, S).
\end{lstlisting}

\paragraph{Mode Declarations}
\begin{lstlisting}
P(X) :- Q(X), identity(P, Q).
P(X) :- player(X), not Q(X), inverse(P, Q).
#modem(2, inverse(target/1, invented/1)).
#modem(2, identity(target/1, invented/1)).
#predicate(target, winner/1).
#predicate(invented, p1/1).

#constant(player, 1).
#constant(player, 2).
#constant(player, 3).
#constant(player, 4).
#modeh(p1(var(player))).
#modeb(1, var(suit) != var(suit)).
#modeb(1, suit(var(player), var(suit)), (positive)).
#modeb(1, suit(const(player), var(suit)), (positive)).
#modeb(1, rank_higher(var(player),var(player)),(positive)).
\end{lstlisting}

\subsection{Sudoku Grid Validity}
There are two variations of~\ac{ilp} tasks presented in this paper, where knowledge of the Sudoku grid was specified, and where grid knowledge was removed and replaced with a division predicate, which enabled FastLAS to learn column, row and block identifiers, based on the cell coordinates given in the example contexts. Both of these variations are presented below, with an example for $9\times9$ grids with the grid knowledge, and $4\times4$ grids without the grid knowledge. For each variation, we present the background knowledge specified and the mode declarations used. The argument in quotes for each column, row and block fact is a unique identifier for each cell. The subset of the hypothesis space computed by FastLAS for both $4\times4$ and $9\times9$ grids contained 2350 possible rules (therefore $2^{2350}$ potential hypotheses, computed as the power set).

\paragraph{Encoding the Sudoku grid: Background knowledge} For $9\times9$ Sudoku grids:
\begin{lstlisting}
col("1, 1", 1).
col("1, 2", 2).
col("1, 3", 3).
col("1, 4", 4).
col("1, 5", 5).
col("1, 6", 6).
col("1, 7", 7).
col("1, 8", 8).
col("1, 9", 9).
...

row("1, 1", 1).
row("1, 2", 1).
row("1, 3", 1).
row("1, 4", 1).
row("1, 5", 1).
row("1, 6", 1).
row("1, 7", 1).
row("1, 8", 1).
row("1, 9", 1).
...

block("1, 1", 1).
block("1, 2", 1).
block("1, 3", 1).
block("2, 1", 1).
block("2, 2", 1).
block("2, 3", 1).
block("3, 1", 1).
block("3, 2", 1).
block("3, 3", 1).
...
\end{lstlisting}

\paragraph{Encoding the Sudoku grid: Mode Declarations} For $9\times9$ Sudoku grids:
\begin{lstlisting}
#modeh(invalid).
#modeb(digit(var(cell), var(num))).
#modeb(row(var(cell), var(row))).
#modeb(col(var(cell), var(col))).
#modeb(block(var(cell), var(block))).
#modeb(neq(var(cell), var(cell))).
#maxv(4).
num(1..9).
row(1..9).
col(1..9).
block(1..9).
cell(C) :- digit(C, _).
neq(X, Y) :- cell(X), cell(Y), X != Y.
\end{lstlisting}

\paragraph{Without encoding the Sudoku grid: Background knowledge} For $4\times4$ Sudoku grids:
\begin{lstlisting}
div_same1(X,Y,C) :- (X - 1) / C = (Y - 1) / C, idx1(X), idx1(Y), X < Y, quotient(C).
div_same2(X,Y,C) :- (X - 1) / C = (Y - 1) / C, idx2(X), idx2(Y), X < Y, quotient(C).

quotient(1..3).
idx1(1..4).
idx2(1..4).
\end{lstlisting}

\paragraph{Without encoding the Sudoku grid: Mode Declarations} For $4\times4$ Sudoku grids:
\begin{lstlisting}
#modeh(invalid).
#modeb(digit(var(idx1), var(idx2), var(num))).
#modeb(div_same1(var(idx1), var(idx1), const(quotient))).
#modeb(div_same2(var(idx2), var(idx2), const(quotient))).

#maxv(5).
num(1..4).

#bias("penalty(1, head).").
#bias("penalty(1, body(X)) :- in_body(X).").
#ground_without_replacement.
\end{lstlisting}

\subsection{Crop Yield Prediction}

\paragraph{Background Knowledge and Mode Declarations}

\begin{lstlisting}
:- yield(X), yield(Y), X < Y.

yield_type(0).
yield_type(1).
yield_type(2).

#modeh(yield(const(yield_type))).
#modeb(1, location(const(location))).
#modeb(1, species(const(species))).
#modeb(1, disease(const(disease))).
\end{lstlisting}

\subsection{Indoor Scene Classification}
\paragraph{Mode Declarations}
\begin{lstlisting}
label_type(0).
label_type(1).
label_type(2).
label_type(3).
label_type(4).

#modeh(label(const(label_type))).
#modeb(1,image(const(image))).
\end{lstlisting}

\end{document}